%% file: main.tex
\documentclass[]{worldbench}
\input{preamble}

\title{\ours: Full-Spectrum Evaluations of Driving World Models in Real World}

\author[]{Ao~Liang~\raisebox{0.2em}{\includegraphics[width=0.019\linewidth]{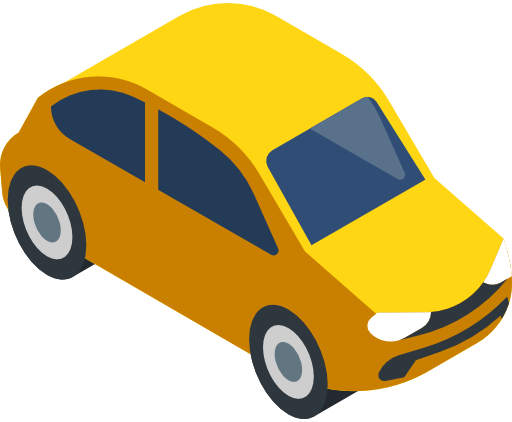}}}
\author[]{Lingdong~Kong~\raisebox{0.2em}{\includegraphics[width=0.019\linewidth]{figures/icons/car1.png}}~\raisebox{0.2em}{\includegraphics[width=0.019\linewidth]{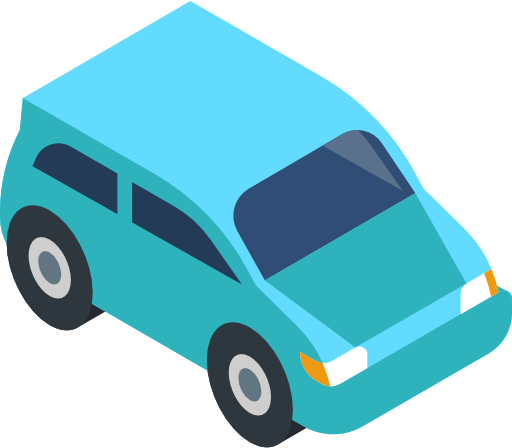}}}
\author[]{Tianyi~Yan~\raisebox{0.2em}{\includegraphics[width=0.019\linewidth]{figures/icons/car1.png}}}
\author[]{Hongsi~Liu~\raisebox{0.2em}{\includegraphics[width=0.019\linewidth]{figures/icons/car1.png}}}
\author[]{Wesley~Yang~\raisebox{0.2em}{\includegraphics[width=0.019\linewidth]{figures/icons/car1.png}}}
\author[]{Ziqi~Huang}
\author[]{Wei~Yin}
\author[]{Jialong~Zuo}
\author[]{Yixuan~Hu}
\author[]{Dekai~Zhu}
\author[]{Dongyue~Lu}
\author[]{Youquan~Liu}
\author[]{Guangfeng~Jiang}
\author[]{Linfeng~Li}
\author[]{Xiangtai~Li}
\author[]{Long~Zhuo}
\author[]{Lai~Xing~Ng}
\author[]{Benoit~R.~Cottereau}
\author[]{Changxin~Gao}
\author[]{Liang~Pan}
\author[]{Wei~Tsang~Ooi}
\author[]{Ziwei~Liu~\raisebox{0.15em}{\includegraphics[width=0.017\linewidth]{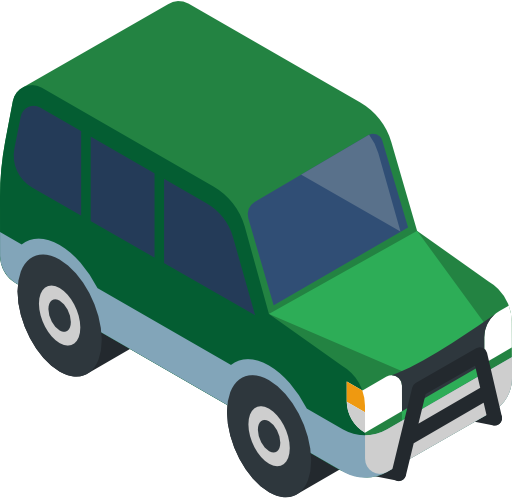}}}

\affiliation[]{
\raisebox{-0.1em}{\includegraphics[width=0.029\linewidth]{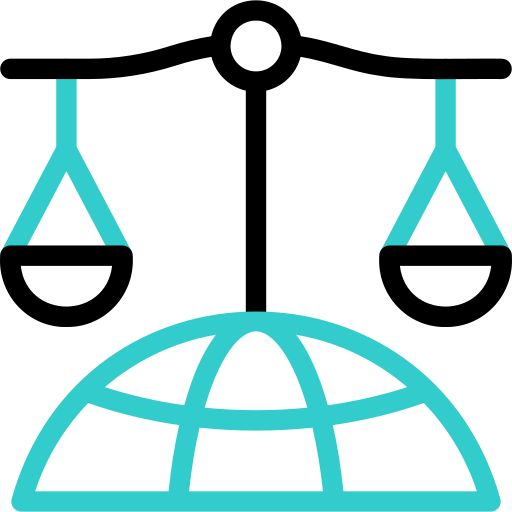}}~WorldBench Team
\\[1.2ex]
~\raisebox{-0.2em}{\includegraphics[width=0.032\linewidth]{figures/icons/car1.png}}~{\small \textbf{Equal Contributions}}
\quad
\raisebox{-0.2em}{\includegraphics[width=0.031\linewidth]{figures/icons/car2.png}}~{\small \textbf{Project Lead}}
\quad
\raisebox{-0.2em}{\includegraphics[width=0.028\linewidth]{figures/icons/car4.png}}~{\small \textbf{Corresponding Author}}
}

\abstract{
Generative world models are reshaping embodied AI, enabling agents to synthesize realistic 4D driving environments that look convincing but often fail physically or behaviorally. Despite rapid progress, the field still lacks a unified way to assess whether generated worlds preserve geometry, obey physics, or support reliable control. We introduce \textbf{\ours}, a full-spectrum benchmark evaluating how well a model builds, understands, and behaves within its generated world. It spans five aspects -- Generation, Reconstruction, Action-Following, Downstream Task, and Human Preference -- jointly covering visual realism, geometric consistency, physical plausibility, and functional reliability. Across these dimensions, no existing world model excels universally: those with strong textures often violate physics, while geometry-stable ones lack behavioral fidelity. To align objective metrics with human judgment, we further construct \textbf{WorldLens-26K}, a large-scale dataset of human-annotated videos with numerical scores and textual rationales, and develop \textbf{WorldLens-Agent}, an evaluation model distilled from these annotations to enable scalable, explainable scoring. Together, the benchmark, dataset, and agent form a unified ecosystem for measuring world fidelity -- standardizing how future models are judged not only by how real they look, but by how real they behave.
}

\metadata[
\raisebox{-0.2em}{\includegraphics[width=0.025\linewidth]{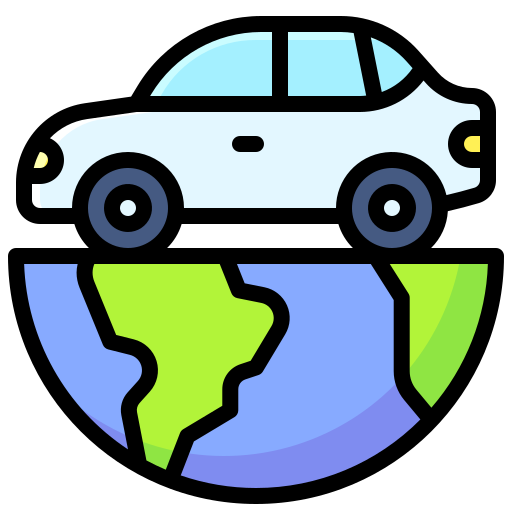}}~~Project Page]{\href{https://worldbench.github.io/worldlens}{\texttt{https://worldbench.github.io/worldlens}}
\\[-1.5ex]}

\metadata[
\raisebox{-0.2em}{\includegraphics[width=0.025\linewidth]{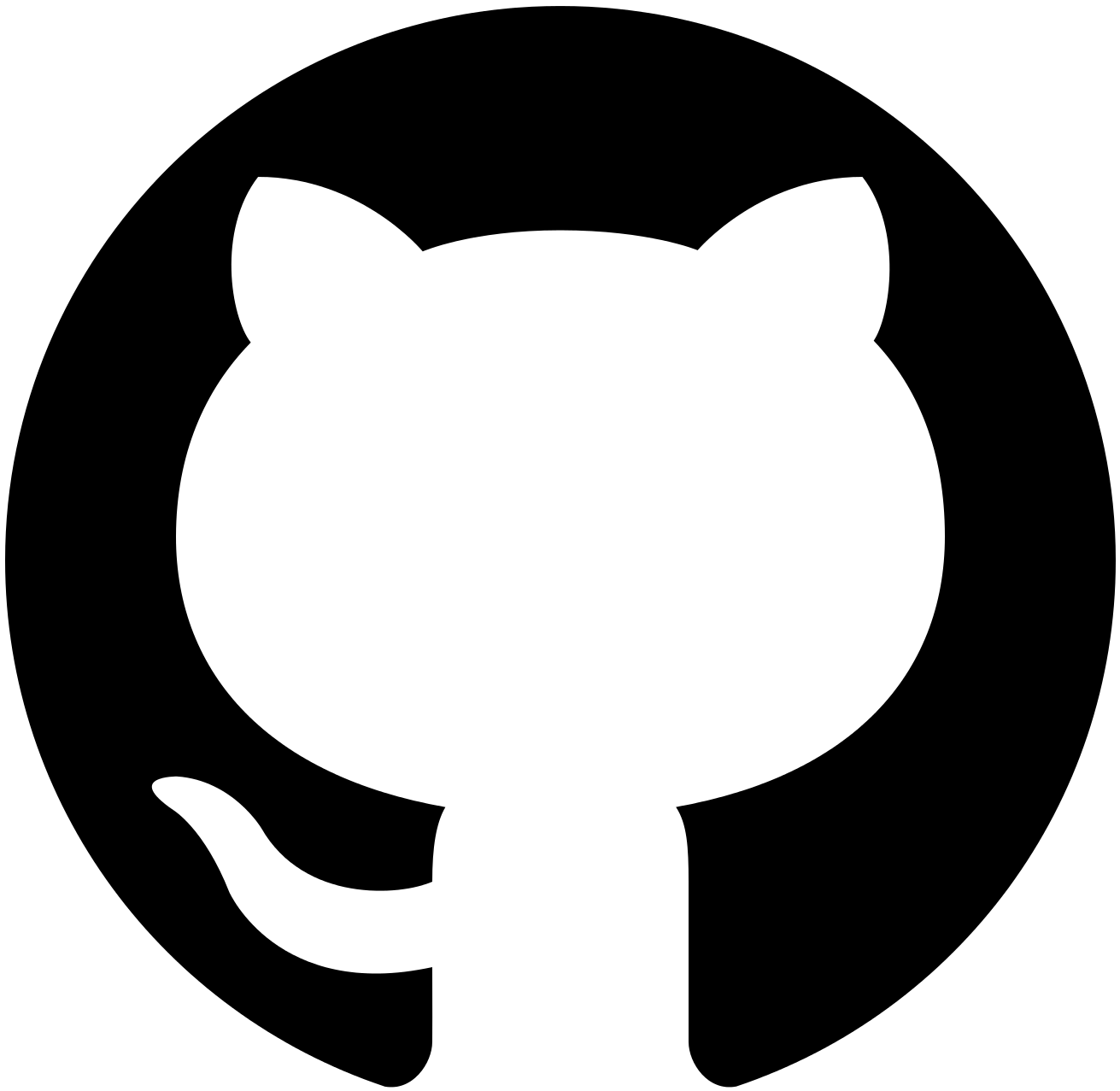}}~~GitHub Repo]{\href{https://github.com/worldbench/WorldLens}{\texttt{https://github.com/worldbench/WorldLens}}
\\[-1.7ex]}

\metadata[
\raisebox{-0.3em}{\includegraphics[width=0.026\linewidth]{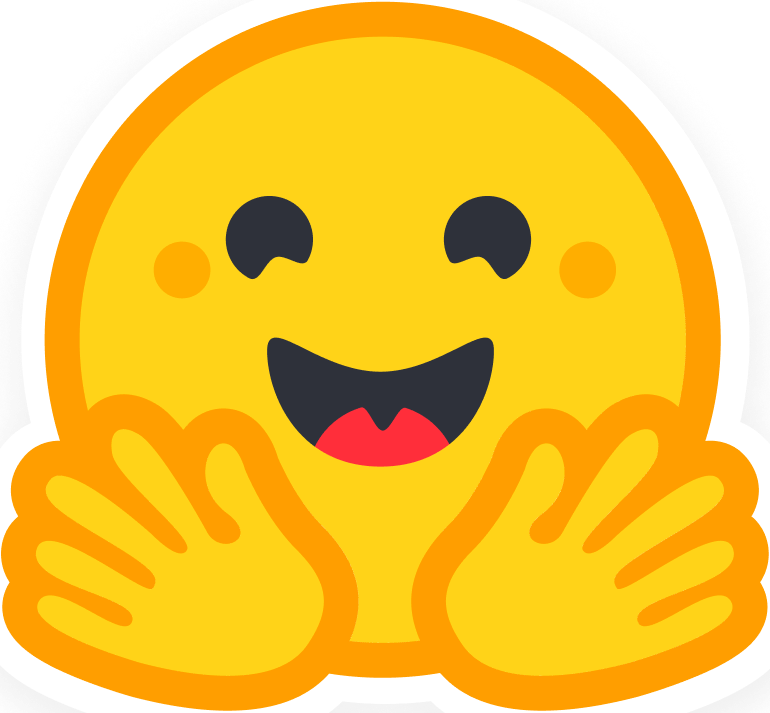}}~~HuggingFace Leaderboard]{\href{https://huggingface.co/spaces/worldbench/WorldLens}{\texttt{https://huggingface.co/spaces/worldbench/WorldLens}}
\\[-1.7ex]}

\metadata[
\raisebox{-0.3em}{\includegraphics[width=0.026\linewidth]{figures/icons/hf.png}}~~HuggingFace Dataset]{\href{https://huggingface.co/datasets/worldbench}{\texttt{https://huggingface.co/datasets/worldbench}}}

\begin{document}

\maketitle

\input{sections/1_intro}
\input{sections/2_related_work}
\input{sections/3_bench}

\input{sections/4_experiments}
\input{sections/5_conclusion}

\section*{Appendix}
\vspace{-0.2cm}
{
\small
\setlength{\parskip}{1pt}
\setlength{\baselineskip}{1.18\baselineskip}
\startcontents[appendices]
\printcontents[appendices]{l}{1}{\setcounter{tocdepth}{2}}
}

\clearpage\clearpage
\input{sections_supp/2_generation}

\clearpage\clearpage
\input{sections_supp/3_reconstruction}

\clearpage\clearpage
\input{sections_supp/4_action_follow}

\clearpage\clearpage
\input{sections_supp/5_downstream}

\clearpage\clearpage
\input{sections_supp/6_human_preference}
\clearpage\clearpage
\input{sections_supp/7_agent}
\clearpage\clearpage
\input{sections_supp/9_impact}
\vspace{0.2cm}
\input{sections_supp/10_ack}

\bibliographystyle{plainnat}
\bibliography{main}

\end{document}

%% file: preamble.tex
\definecolor{w_blue}{RGB}{52,204,204}
\definecolor{w_yellow}{RGB}{255,192,0}

\usepackage{amsfonts}
\usepackage{amsmath}
\usepackage{amssymb}

\usepackage{colortbl}

\usepackage{makecell}
\usepackage{multicol}
\usepackage{multirow}
\usepackage{pifont}

\usepackage{wrapfig}

\definecolor{red}{rgb}{0.8,0,0}  
\definecolor{green}{RGB}{0, 133, 21}  
\definecolor{grey}{rgb}{0.5,0.5,0.5}

\definecolor{w_1}{RGB}{52,204,204}
\definecolor{w_2}{RGB}{70,203,187}
\definecolor{w_3}{RGB}{95,202,161}
\definecolor{w_4}{RGB}{119,200,136}
\definecolor{w_5}{RGB}{155,199,101}
\definecolor{w_6}{RGB}{185,197,70}
\definecolor{w_7}{RGB}{227,194,28}
\definecolor{w_8}{RGB}{243,193,12}
\definecolor{w_9}{RGB}{255,192,0}

\newcommand{\ours}{\textbbf{\textsl{\textcolor{w_1}{W}\textcolor{w_2}{o}\textcolor{w_3}{r}\textcolor{w_4}{l}\textcolor{w_5}{d}\textcolor{w_6}{L}\textcolor{w_7}{e}\textcolor{w_8}{n}\textcolor{w_9}{s}}\nobreak\hspace{0.07em}}}

\newcommand{\textbbf}[1]{\scalebox{1.05}{\textbf{#1}}}

\makeatletter
\def\blfootnote{\xdef\@thefnmark{}\@footnotetext}
\DeclareRobustCommand\onedot{\futurelet\@let@token\@onedot}
\def\@onedot{\ifx\@let@token.\else.\null\fi\xspace}

\makeatother

\usepackage{titletoc}
\usepackage{enumitem}

\tcbset{
  myboxstyle/.style={
    boxrule=1pt,
    boxsep=2pt,
    colback=gray!5,
    fontupper=\scriptsize,
    fonttitle=\color{black},
    title=System prompt for assessing the quality of generated driving videos with VLM,
    coltitle=black,
    colframe=black,
    colbacktitle=blue!10,
    breakable
  }
}

\tcbset{
  myjsonbox/.style={
    enhanced,
    colframe=gray!50,
    colback=white,
    fonttitle=\small\color{white},
    boxrule=0.8pt,
    arc=4pt,
    left=4pt,
    right=4pt,
    top=4pt,
    bottom=4pt,
    boxsep=5pt,
    listing only,
    listing options={
      language=json,
      basicstyle=\scriptsize\ttfamily,
      keywordstyle=\color{blue},
      stringstyle=\color{orange!90!black},
      commentstyle=\color{gray},
      morekeywords={true,false,null},
      literate=%
        *{,}{{\textcolor{red}{,}}}{1}%
         {:}{{\textcolor{red}{:}}}{1}%
         {\{}{{\textcolor{red}{\{}}}{1}%
         {\}}{{\textcolor{red}{\}}}}{1}%
    },
    breakable,
  }
}

\lstdefinelanguage{json}{
    basicstyle=\ttfamily\footnotesize,
    showstringspaces=false,
    breaklines=true,
    breakatwhitespace=true,
    morestring=[b]",
    stringstyle=\color{orange!90!black},
    morecomment=[s]{/*}{*/},
    commentstyle=\color{gray},
    morekeywords={true,false,null},
    keywordstyle=\color{blue},
    literate=%
      *{,}{{\textcolor{red}{,}}}{1}%
       {:}{{\textcolor{red}{:}}}{1}%
       {\{}{{\textcolor{red}{\{}}}{1}%
       {\}}{{\textcolor{red}{\}}}}{1}%
}

%% file: sections/1_intro.tex
\begin{figure}[!ht]
    \centering
    \includegraphics[width=\linewidth]{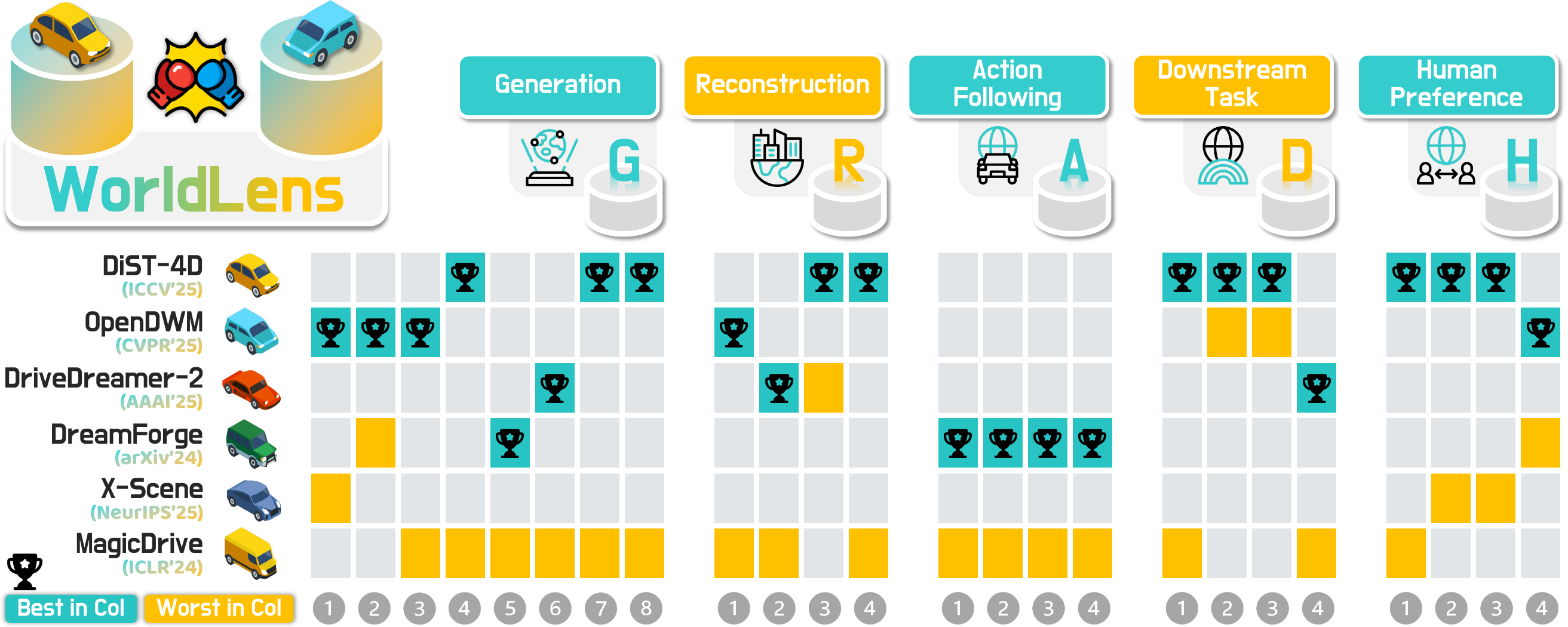}
    \caption{\textbf{Is your driving world model an all-around player?} This work presents \ours, a unified benchmark encompassing evaluations on $^1$\textbf{\textsl{Generation}}, $^2$\textbf{\textsl{Reconstruction}}, $^3$\textbf{\textsl{Action-Following}}, $^4$\textbf{\textsl{Downstream Task}}, and $^5$\textbf{\textsl{Human Preference}}, across a total of $\mathbf{24}$ \textbf{dimensions} spanning visual realism, geometric consistency, functional reliability, and perceptual alignment. We observe \emph{no single model dominates across all axes}, highlighting the need for balanced progress toward physically and behaviorally realistic world modeling.}
    \label{fig:teaser}
\end{figure}

\section{Introduction}
\label{sec:intro}

Generative world models have transformed embodied AI and simulation~\cite{google2024genie2,genie3,ren2025cosmos,russell2025gaia-2,yue2025video}. From text-driven 4D synthesis to controllable driving environments~\cite{opendwm,yan2025drivingsphere,ni2025maskgwm,zhang2025epona,wang2025prophetdwm,wang2025longdwm}, modern systems can produce dash-cam–like sequences with striking visual realism. However, evaluation has not kept pace: the field lacks a standardized way to measure whether generated worlds preserve geometry, respect physics, and support reliable decision-making~\cite{kong2025survey,peper2025survey,wang2025survey}.

Most widely used metrics emphasize frame quality and aesthetics~\cite{ke2021musiq,huang2024vbench,huang2024vbench++,arai2024act-bench}, but reveal little about physical causality, multi-view geometry, or functional reliability under control~\cite{yang2024genad,gao2024vista,bartoccioni2025vavim_vavam,chen2024drivephysica,yan2025rlgf,zhou2024simgen, zheng2025vbench2}. This gap, which is well documented across recent surveys~\cite{kong2025survey,wang2025survey,zhang2025quality,ma2025survey}, has created fragmented progress and incomparable results. While structured maturity scales, \emph{e.g.}, SAE Levels of Driving Automation$^{\text{TM}}$, have clarified autonomy benchmarking~\cite{sae-levels}, an analogous, practice-ready protocol for \textbf{evaluating driving world models} has remained elusive.

To bridge the gap, we build \ours, a full-spectrum benchmark that evaluates how well a world model \emph{builds}, \emph{understands}, and \emph{behaves} within its generated world. As shown in Figure~\ref{fig:teaser}, no existing model excels universally; some achieve strong texture realism but violate physics, while others preserve geometry yet fail behaviorally. 

To reveal these world modeling trade-offs, we decompose evaluation into \textbf{five complementary aspects}:
\begin{itemize}
    \item $^1$\textbf{\textsl{Generation}} — measuring whether a model can synthesize \emph{visually realistic}, \emph{temporally stable}, and \emph{semantically consistent} scenes~\cite{gao2023magicdrive,opendwm,guo2025dist-4d}. Even state-of-the-art models that achieve low perceptual error (\emph{e.g.}, LPIPS, FVD) often suffer from view flickering or motion instability, revealing the limits of current diffusion-based architectures.

    \item $^2$\textbf{\textsl{Reconstruction}} — probing whether generated videos can be reprojected into a \emph{coherent 4D scene} using differentiable rendering \cite{chen2025omnire,kerbl2023-3dgs}. Models that appear sharp in 2D frequently collapse when reconstructed, producing geometric ``floaters'': a gap that exposes how temporal coherence remains weakly coupled in most pipelines.

    \item $^3$\textbf{\textsl{Action-Following}} — testing if a pre-trained action planner \cite{hu2023uniad,jiang2023vad} can \emph{operate safely} inside the generated world. High \emph{open-loop} realism does not guarantee safe \emph{closed-loop} control; almost all existing world models trigger collisions or off-road drifts, underscoring that photometric realism alone cannot yield functional fidelity.

    \item $^4$\textbf{\textsl{Downstream Task}} — evaluating whether the \emph{synthetic data} support downstream perception models trained on real-world datasets~\cite{liu2023bevfusion,tang2024sparseocc,caesar2020nuscenes}. Even visually appealing worlds may degrade detection or segmentation accuracy by $30$–$50\%$, highlighting that alignment to task distributions, not just image quality, is vital for practical usability.

    \item $^5$\textbf{\textsl{Human Preference}} — capturing subjective scores such as \emph{world realism}, \emph{physical plausibility}, and \emph{behavioral safety} through large-scale human annotations. Our study reveals that models with strong geometric consistency are generally rated as more ``real'', confirming that perceptual fidelity is inseparable from structural coherence.
\end{itemize}

To bridge algorithmic metrics with human perception, we curate \textbf{WorldLens-26K}, a large-scale dataset of human-annotated videos covering perceptual, physical, and safety-related dimensions. Each sample contains quantitative scores and textual explanations, capturing how evaluators reason about realism, physical plausibility, and behavioral safety. By pairing human judgments with structured rationales, we aim to transform subjective evaluation into learnable supervision, enabling perception-aligned and interpretable assessment of generative world models.

Leveraging the above, we develop \textbf{WorldLens-Agent}, a feedback-aligned auto-evaluator distilled from human preferences. This agent can predict perceptual and physical scores while generating natural-language explanations consistent with human reasoning. It generalizes well to \emph{unseen} models and enables scalable auto-evaluation of generative worlds without repeated manual labeling.

Together, our benchmark, dataset, and evaluation agent form \textbf{a unified ecosystem} that bridges objective measurement and human interpretation. \emph{We will release the toolkit, dataset, and model to foster standardized, explainable, and human-aligned evaluation} -- guiding future world models not only to \textit{``look''} real, but to \textit{``behave''} reasonably.

%% file: sections/2_related_work.tex
\section{Related Work}
\label{sec:related_work}

\noindent\textbf{Video Generation.}
Recent advances have driven rapid progress in generation \cite{xue2025survey,ma2025survey,li2024survey,omer2024lumiere,wang2025lavie, yue2025video}. Text-to-image  \cite{saharia2022photorealistic,betker2023improving,rombach2022high,wu2023tune-a-video} laid the foundation for high-fidelity synthesis from text, later extended to the temporal domain through text-to-video (T2V) systems \cite{singer2022make,blattmann2023align,pika, ren2025cosmos, fan2025vchitect, si2025RepVideo, imagenvideo, wan2.1, huang2025vchain, kong2024hunyuanvideo, Veo2, kling, sora, yang2024cogvideox}. Building on these foundations, domain-specific methods \cite{sun2019see,li2024drivingdiffusion,wang2024drivedreamer,lin2025drivegen,lu2024wovogen,gao2023magicdrive,hassan2025gem,wu2025umgen,zhou2025hermes} achieved impressive realism using motion-aware conditioning \cite{bian2025dynamiccity,opendwm,huang2025subjectdrive,jia2023adriver-i,wang2024drive-wm,wu2024drivescape,jiang2024dive}. Despite strong perceptual quality, they remain largely \emph{appearance-driven}: they generate visually coherent sequences but lack explicit geometry, physics, or causal control \cite{yan2025drivingsphere,yan2025rlgf,bartoccioni2025vavim_vavam,chen2025geodrive,yang2025dualdiff+}. Without structured world states or dynamics, they cannot model how scenes behave or respond to actions \cite{dauner2024navsim,arai2024act-bench,yang2023bevcontrol}, and metrics focused only on visual fidelity reveal little about whether a model truly understands the world it depicts.

\noindent\textbf{3D \& 4D World Modeling.}
Recent studies move beyond frame-based generation to build \emph{world models} that encode 3D, dynamics, and control-aware representations \cite{kong2025survey}. WonderWorld \cite{yu2025wonderworld}, GAIA-1/2 \cite{hu2023gaia-1,russell2025gaia-2}, Genie-3 \cite{genie3}, and related efforts \cite{yang2024oasis,guo2025mineworld,gao2024vista,yang2024genad} learn physics-grounded latent states representing geometry, occupancy, and motion for conditioned predictions \cite{chen2024drivinggpt,bartoccioni2025vavim_vavam,liang2026lidarcrafter,liu2026lalalidar,zhu2025spiral,yang2025driving,mei2025vision,liang2025pi3det,robodrive_challenge_2024}. DriveArena \cite{yang2024drivearena} and NAVSIM \cite{dauner2024navsim,cao2025navsim-2} integrate perception, planning, and control into generative pipelines that support closed-loop simulation \cite{caesar2021nuplan,dosovitskiy2017carla,yan2024streetgaussian}. Yet, existing paradigms remain limited to perception metrics or qualitative results \cite{kong2025survey}. A unified standard for measuring geometry, 4D consistency, and agent behaviors is missing.

\begin{figure*}[t]
    \centering
    \includegraphics[width=\textwidth]{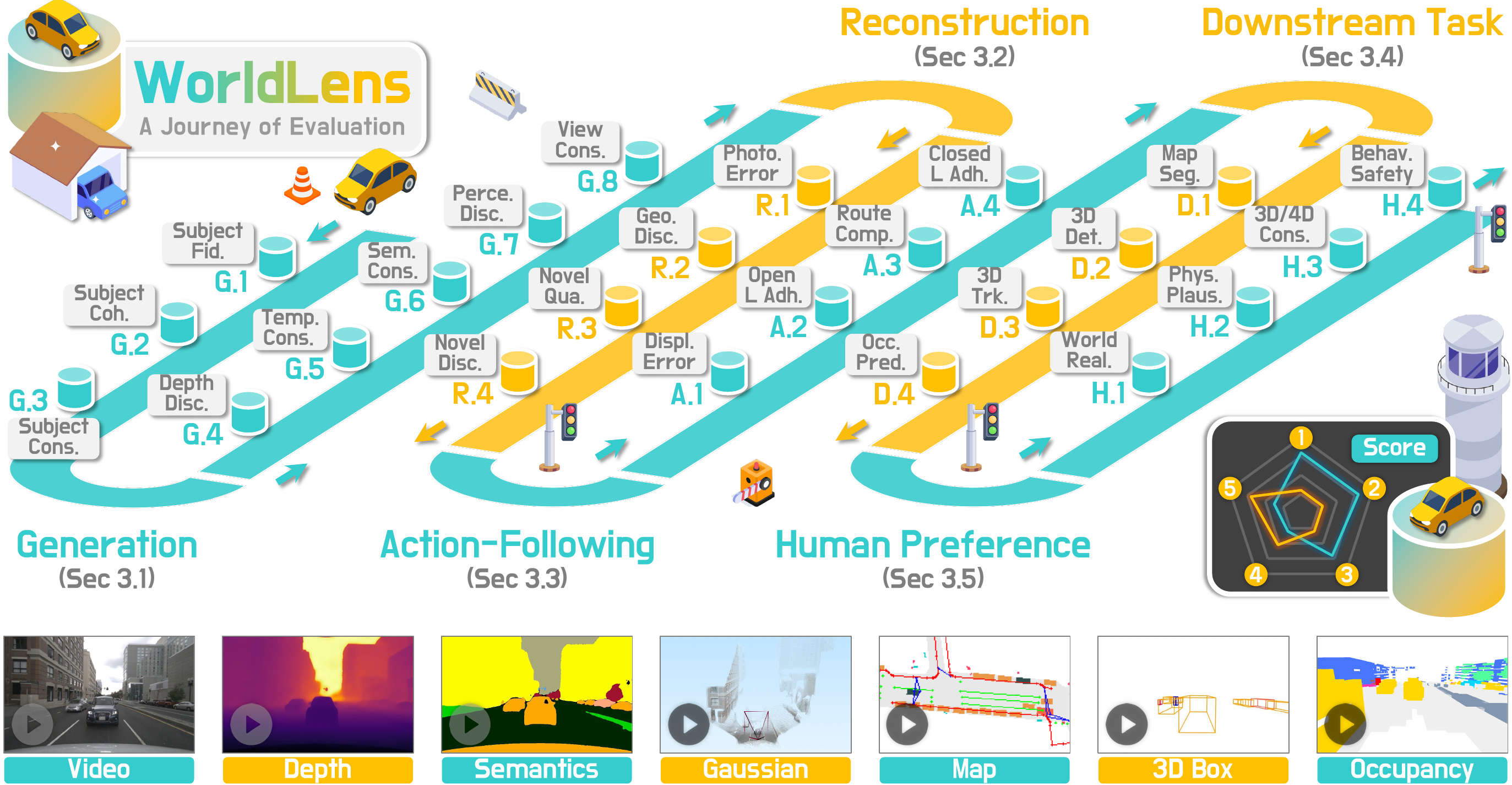}
    \caption{The \ours~evaluation framework unifies five complementary aspects -- $^1$\emph{\textbf{Generation}}, $^2$\emph{\textbf{Reconstruction}}, $^3$\emph{\textbf{Action-Following}}, $^4$\emph{\textbf{Downstream Task}}, and $^5$\emph{\textbf{Human Preference}} --  to assess visual, geometric, functional, and perceptual fidelity of generative world models. Each aspect is decomposed into interpretable dimensions driven by measurable signals such as segmentation, depth, 4D reconstruction, and behavioral simulation, enabling comprehensive and physically grounded evaluations across the full spectrum of world modeling.}
    \label{fig:bench}
\end{figure*}

\noindent\textbf{Evaluation.}
Generative video evaluation has evolved from simple frame-based scores to multi-dimensional benchmarks \cite{liao2024evaluation,li2025k-sort,wang2025aigvqa,liu2024survey, zhang2024evaluationagent}. Early metrics measure distributional similarity and perceptual alignment \cite{heusel2017gans,salimans2016improved,unterthiner2018towards,radford2021learning,ke2021musiq}, while frameworks, \emph{e.g.}, VBench \cite{huang2024vbench, huang2024vbench++}, EvalCrafter \cite{liu2024evalcrafter}, and T2V-CompBench \cite{sun2025t2v}, extend assessment to motion and temporal consistency. More recent efforts, WorldScore \cite{duan2025worldscore} and VideoWorld \cite{ren2025videoworld}, move toward ``world-model'' evaluation using physics-inspired composite scores, yet they remain confined to 2D video settings emphasizing appearance over embodiment \cite{kang2024survey}. In driving, some existing benchmarks \cite{dosovitskiy2017carla,caesar2021nuplan,xie2025vlms} evaluate agents rather than the worlds they inhabit.
\textbf{WorldLens} introduces 4D reconstruction, action-following, and human preference alignment to jointly assess spatial fidelity, behavioral consistency, and physical realism, establishing the \textbf{first} benchmark that measures both the appearance and behavior.

%% file: sections/3_bench.tex
\section{\ours: A Full-Spectrum Benchmark}
\label{sec:bench}

Generative world models must go beyond visual realism to achieve geometric consistency, physical plausibility, and functional reliability. We introduce \textbf{\textsl{WorldLens}}, a unified benchmark that evaluates these capabilities across five complementary aspects -- from \textbf{low-level} appearance fidelity to \textbf{high-level} behavioral realism.
As shown in Figure~\ref{fig:bench}, each aspect is decomposed into fine-grained, interpretable dimensions, forming a comprehensive framework that bridges human perception, physical reasoning, and downstream utility.

\subsection{Generation}
\label{sec:generation}
This aspect decomposes the overall \textbf{generation quality} into eight complementary dimensions that assess appearance fidelity, temporal stability, geometric correctness, and semantic smoothness. Together, these dimensions quantify how ``faithfully'' a model constructs visually and perceptually consistent driving scenes across time and viewpoints.

\noindent{{\raisebox{0.9pt}{\colorbox{w_blue}{\scriptsize{\textbf{\textbf{\textsc{\textcolor{white}{G.1}}}}}}}}}~\textbf{\textsl{Subject Fidelity}} measures the perceptual realism and semantic correctness of individual object instances, \emph{e.g.}, vehicles and pedestrians. For each generated frame, object regions are cropped using bounding boxes and evaluated with class-specific binary classifiers trained on real data~\cite{dosovitskiy2020image}. A high confidence indicates that the generated object visually aligns with its real-world category. This dimension captures localized realism and complements global metrics by focusing on instance-level fidelity.

\noindent{{\raisebox{0.9pt}{\colorbox{w_blue}{\scriptsize{\textbf{\textbf{\textsc{\textcolor{white}{G.2}}}}}}}}}~\textbf{\textsl{Subject Coherence}} assesses the temporal stability of each object’s identity throughout a generated sequence. Using known track IDs, we extract visual embeddings for each instance from a pretrained ReID model~\cite{zuo2024cross, he2021transreid} and compute similarity across frames. A high coherence score indicates that objects maintain consistent shape, color, and texture through motion, reflecting stable appearance and identity preservation over time.

\noindent{{\raisebox{0.9pt}{\colorbox{w_blue}{\scriptsize{\textbf{\textbf{\textsc{\textcolor{white}{G.3}}}}}}}}}~\textbf{\textsl{Subject Consistency}} evaluates fine-grained temporal stability of object-level semantics and geometry. It uses DINO \cite{caron2021dino} features to capture texture and spatial details across frames, ensuring that subjects preserve their structure and semantic meaning without flickering or deformation. High consistency reflects the reliable temporal evolution of objects and smooth changes in appearance under motion.

\noindent{{\raisebox{0.9pt}{\colorbox{w_blue}{\scriptsize{\textbf{\textbf{\textsc{\textcolor{white}{G.4}}}}}}}}}~\textbf{\textsl{Depth Discrepancy}} measures the smoothness of depth variations across time, capturing geometric coherence. We estimate monocular depth for each frame using Depth Anything V2 \cite{yang2024depth-any-v2}, coded with colors, and extract corresponding embeddings through DINO v2 \cite{oquab2024dinov2}. The average feature distance between consecutive frames quantifies the continuity of depth perception. Lower discrepancy indicates geometrically stable and physically plausible scene motion.

\noindent{{\raisebox{0.9pt}{\colorbox{w_blue}{\scriptsize{\textbf{\textbf{\textsc{\textcolor{white}{G.5}}}}}}}}}~\textbf{\textsl{Temporal Consistency}} quantifies global frame-to-frame smoothness in a learned appearance space. Each frame is embedded with the CLIP visual encoder \cite{radford2021learning}, and temporal stability is derived from adjacent-frame similarity, jitter suppression, and motion-rate alignment with real videos. High consistency corresponds to temporally stable dynamics without abrupt or unnatural transitions.

\noindent{{\raisebox{0.9pt}{\colorbox{w_blue}{\scriptsize{\textbf{\textbf{\textsc{\textcolor{white}{G.6}}}}}}}}}~\textbf{\textsl{Semantic Consistency}} evaluates whether the semantic layout of generated scenes evolves smoothly across time. We employ a pretrained  SegFormer \cite{xie2021segformer} to predict frame-wise masks and compute stability across labels, regions, and class distributions. This dimension ensures that generated videos maintain coherent object semantics and scene structures without flickering boundaries or label inconsistencies.

\noindent{{\raisebox{0.9pt}{\colorbox{w_blue}{\scriptsize{\textbf{\textbf{\textsc{\textcolor{white}{G.7}}}}}}}}}~\textbf{\textsl{Perceptual Discrepancy}} quantifies the overall perceptual gap between real and generated videos. We extract spatiotemporal embeddings from a pretrained I3D \cite{carreira2017i3d} and compute the Fréchet Video Distance \cite{unterthiner2018towards} between real and generated distributions. A lower discrepancy indicates closer alignment in appearance and motion statistics, reflecting perceptual realism and temporal coherence.

\noindent{{\raisebox{0.9pt}{\colorbox{w_blue}{\scriptsize{\textbf{\textbf{\textsc{\textcolor{white}{G.8}}}}}}}}}~\textbf{\textsl{Cross-View Consistency}} measures the geometric and photometric alignment between overlapping regions of adjacent camera views. Using LoFTR \cite{sun2021loftr}, we detect feature correspondences between synchronized camera pairs and aggregate their confidence to assess spatial coherence. High consistency indicates better structural alignment and visual continuity across multiple perspectives, ensuring 3D-consistent generation for multi-camera driving systems.

\subsection{Reconstruction}
\label{sec:reconstruction}
This aspect decomposes the overall \textbf{reconstructability} into how well a coherent 4D scene can be recovered from generated videos. Each sequence is lifted into a Gaussian Field and re-rendered under both original and \textbf{novel camera trajectories}, testing spatial interpolation, parallax, and view generalization across representative novel-view paths.

\noindent{{\raisebox{0.9pt}{\colorbox{w_yellow}{\scriptsize{\textbf{\textbf{\textsc{\textcolor{white}{R.1}}}}}}}}}~\textbf{\textsl{Photometric Error}} measures how accurately reconstructed scenes reproduce their input frames. Each generated video is reconstructed into a differentiable 4D representation \cite{chen2025omnire,kerbl2023-3dgs}, and re-rendered at original camera poses. Pixel-level similarities, \emph{i.e.}, LPIPS, PSNR, and SSIM, are computed, reflecting stability of appearance and lighting across time. Lower discrepancy indicates more consistent photometric properties for faithful 4D reconstruction.

\noindent{{\raisebox{0.9pt}{\colorbox{w_yellow}{\scriptsize{\textbf{\textbf{\textsc{\textcolor{white}{R.2}}}}}}}}}~\textbf{\textsl{Geometric Discrepancy}} assesses how well the reconstructed geometry from generated videos aligns with real-world structure. Using identical camera poses, we reconstruct 4D scenes from both generated and ground-truth sequences and compare their rendered depth maps. The Absolute Relative Error (AbsRel) and other related metrics are computed within regions selected by Grounded-SAM 2 \cite{ren2024grounded-sam,ravi2025sam-2}. Lower values indicate more plausible depth and consistent surface geometry with real scenes.

\noindent{{\raisebox{0.9pt}{\colorbox{w_yellow}{\scriptsize{\textbf{\textbf{\textsc{\textcolor{white}{R.3}}}}}}}}}~\textbf{\textsl{Novel-View Quality}} evaluates the perceptual realism of re-rendered frames from unseen camera trajectories. Each reconstructed scene is rendered along novel paths using the same differentiable framework, and visual quality is scored by MUSIQ \cite{ke2021musiq}. A higher score indicates that novel views remain sharp, artifact-free, and visually coherent, demonstrating that the model preserves appearance and illumination consistency beyond training viewpoints.

\noindent{{\raisebox{0.9pt}{\colorbox{w_yellow}{\scriptsize{\textbf{\textbf{\textsc{\textcolor{white}{R.4}}}}}}}}}~\textbf{\textsl{Novel-View Discrepancy}} quantifies the perceptual gap between novel-view renderings from generated and real reconstructions. Both are rendered under identical camera trajectories, and their distance is measured via FVD \cite{unterthiner2018towards} on I3D features \cite{carreira2017i3d}. Lower discrepancy indicates better generalization to unseen viewpoints, maintaining coherent geometry, appearance, and temporal dynamics in 4D space.

\subsection{Action-Following}
\label{sec:action-following}
This aspect evaluates how well the generated worlds support \textbf{plausible driving decisions} when interpreted by pretrained planners, examining whether synthesized scenes provide realistic visual and motion cues that yield real-world-consistent actions. All evaluations are conducted in a generative simulator using custom-designed routes derived from real-world maps against standard benchmarks~\cite{caesar2020nuscenes,caesar2021nuplan}.

\noindent{{\raisebox{0.9pt}{\colorbox{w_blue}{\scriptsize{\textbf{\textbf{\textsc{\textcolor{white}{A.1}}}}}}}}}~\textbf{\textsl{Displacement Error}} measures functional consistency between the trajectories predicted from generated and real videos. Using a pretrained end-to-end planner, UniAD \cite{hu2023uniad} or VAD \cite{jiang2023vad}, both sequences are used to predict future waypoints, and their mean L2 distance is computed. Lower displacement error indicates that the generated scenes preserve motion cues required for accurate trajectory forecasting.

\noindent{{\raisebox{0.9pt}{\colorbox{w_blue}{\scriptsize{\textbf{\textbf{\textsc{\textcolor{white}{A.2}}}}}}}}}~\textbf{\textsl{Open-Loop Adherence}} evaluates how well a pretrained policy performs when operating on generated videos in a generative simulator. In open-loop mode, the policy’s predictions are used purely for evaluation and do not affect the simulated ego-vehicle motion. Following NAVSIM \cite{dauner2024navsim}, we adopt the Predictive Driver Model Score (PDMS), which aggregates safety, progress, and comfort sub-scores computed over a short simulation horizon. Higher PDMS indicates that the policy exhibits realistic, stable, and safe driving behaviors even when guided solely by generated visual input, reflecting reliable short-term functional realism.

\noindent{{\raisebox{0.9pt}{\colorbox{w_blue}{\scriptsize{\textbf{\textbf{\textsc{\textcolor{white}{A.3}}}}}}}}}~\textbf{\textsl{Route Completion}} measures long-horizon navigation stability in closed-loop simulation. It computes the percentage of a predefined route completed before termination due to collision, off-road drift, or timeout. Higher route completion rates signify that generated environments support continuous, physically consistent control over extended trajectories, enabling sustained goal-directed motion.

\noindent{{\raisebox{0.9pt}{\colorbox{w_blue}{\scriptsize{\textbf{\textbf{\textsc{\textcolor{white}{A.4}}}}}}}}}~\textbf{\textsl{Closed-Loop Adherence}} integrates both motion quality and task success into a single metric, the Arena Driving Score (ADS) \cite{yang2024drivearena}. In the closed-loop mode, the planning decisions of the driving agent directly control the ego's actions, thereby influencing the simulated environment. ADS multiplies the PDMS and Route Completion scores, rewarding agents that are both safe and successful. A high ADS implies that the planner achieves realistic, collision-free navigation while completing the route effectively, confirming that generated worlds not only \emph{look} real but also \emph{drive} real within an autonomous-control loop.

\subsection{Downstream Task}
\label{sec:downstream-task}
This aspect evaluates the \textbf{downstream utility} of generated videos by measuring how well 3D perception models pretrained on real data perform when applied to synthetic content. Performance degradation across tasks reflects the realism, fidelity, and transferability of generated scenes.

\noindent{{\raisebox{0.9pt}{\colorbox{w_yellow}{\scriptsize{\textbf{\textbf{\textsc{\textcolor{white}{D.1}}}}}}}}}~\textbf{\textsl{Map Segmentation}} evaluates whether generated data contain sufficient spatial and semantic cues for top-down BEV mapping. A pretrained BEV map segmentation network \cite{liu2023bevfusion,li2025bevformer} predicts semantic maps for each frame, and performance is measured by mean Intersection-over-Union (mIoU). Higher mIoU indicates better structural layout and semantics conducive to accurate BEV reconstruction.

\noindent{{\raisebox{0.9pt}{\colorbox{w_yellow}{\scriptsize{\textbf{\textbf{\textsc{\textcolor{white}{D.2}}}}}}}}}~\textbf{\textsl{3D Object Detection}} tests whether generated data retains geometric cues essential for perceiving traffic participants. A pretrained BEVFusion \cite{liu2023bevfusion} is applied to generated frames, and performance is reported using nuScenes Detection Score (NDS) \cite{caesar2020nuscenes}. Higher values indicate that generated scenes support more accurate 3D object localization.

\noindent{{\raisebox{0.9pt}{\colorbox{w_yellow}{\scriptsize{\textbf{\textbf{\textsc{\textcolor{white}{D.3}}}}}}}}}~\textbf{\textsl{3D Object Tracking}} measures motion consistency and identity information preservation of generated videos across time. A pretrained 3D tracker \cite{ding2024ada} predicts 3D trajectories from each generated video, and performance is quantified by Average Multi-Object Tracking Accuracy (AMOTA) under the nuScenes protocol~\cite{caesar2020nuscenes}. Higher AMOTA reflects more stable temporal dynamics and accurate data association for moving objects from the 3D scene.

\noindent{{\raisebox{0.9pt}{\colorbox{w_yellow}{\scriptsize{\textbf{\textbf{\textsc{\textcolor{white}{D.4}}}}}}}}}~\textbf{\textsl{Occupancy Prediction}} evaluates whether generated scenes enable accurate 3D reconstruction of spatial geometry and semantics. A frozen SparseOcc \cite{tang2024sparseocc} predicts voxel-wise scene representations, and performance is measured using RayIoU, which compares semantic agreement along camera rays rather than volumetric overlap. Higher RayIoU indicates more accurate and depth-consistent occupancy estimation, showing that generated videos preserve 3D structural integrity crucial for downstream scene understanding.

\subsection{Human Preference}
\label{sec:human_preference}
This aspect evaluates alignment with \textbf{human judgment} by assessing how visually authentic, physically coherent, and behaviorally safe the generated videos appear to observers. Each dimension is rated on a `$1$' to `$10$' scale, where higher scores indicate stronger human perceptual fidelity.

\noindent{{\raisebox{0.9pt}{\colorbox{w_blue}{\scriptsize{\textbf{\textbf{\textsc{\textcolor{white}{H.1}}}}}}}}}~\textbf{\textsl{World Realism}} measures the overall authenticity and naturalness of generated videos. We evaluate how closely textures, lighting, and motion resemble those in real-world driving footage. Three sub-dimensions are considered: (1) \emph{Overall Realism}, capturing global scene coherence and natural appearance; (2) \emph{Vehicle Realism}, assessing vehicle geometry, surface reflectance, and motion stability; and (3) \emph{Pedestrian Realism}, focusing on body proportion and walking motion consistency. Higher scores indicate scenes that are visually indistinguishable from real recordings.

\noindent{{\raisebox{0.9pt}{\colorbox{w_blue}{\scriptsize{\textbf{\textbf{\textsc{\textcolor{white}{H.2}}}}}}}}}~\textbf{\textsl{Physical Plausibility}} evaluates whether the scene obeys intuitive physical and causal principles. It focuses on the continuity of motion, correctness of occlusion order, object contact stability, and illumination consistency across time. Scenes exhibiting teleportation, interpenetration, or inconsistent reflections receive lower ratings, while those maintaining smooth transitions and physically coherent dynamics are scored higher.

\noindent{{\raisebox{0.9pt}{\colorbox{w_blue}{\scriptsize{\textbf{\textbf{\textsc{\textcolor{white}{H.3}}}}}}}}}~\textbf{\textsl{3D \& 4D Consistency}} measures spatial-temporal coherence of geometry and appearance across frames. It assesses whether reconstructed 3D structures remain stable over time and whether objects preserve their relative positions, orientations, and trajectories. High consistency reflects that generated videos maintain realistic 3D layout and smooth temporal evolution, forming plausible 4D scenes aligned with real-world dynamics.

\noindent{{\raisebox{0.9pt}{\colorbox{w_blue}{\scriptsize{\textbf{\textbf{\textsc{\textcolor{white}{H.4}}}}}}}}}~\textbf{\textsl{Behavioral Safety}} assesses whether traffic participants behave in predictable and risk-free ways consistent with common driving norms. It focuses on short-term interactions among vehicles, pedestrians, and environmental cues, such as adherence to traffic signals, collision avoidance, and stable lane following. Lower scores indicate abrupt or unsafe behaviors (\emph{e.g.}, sudden collisions or unrealistic crossings), whereas higher scores correspond to smooth, lawful, and controlled motion indicative of safe and credible agent behavior.

\section{Human Annotation \& Evaluation Agent}
\label{sec:human_evalagent}
\subsection{Annotation Process}
\begin{wrapfigure}{r}{0.58\textwidth}
    \begin{minipage}{\linewidth}
        \centering
        \vspace{-1.1cm}
        \includegraphics[width=\linewidth]{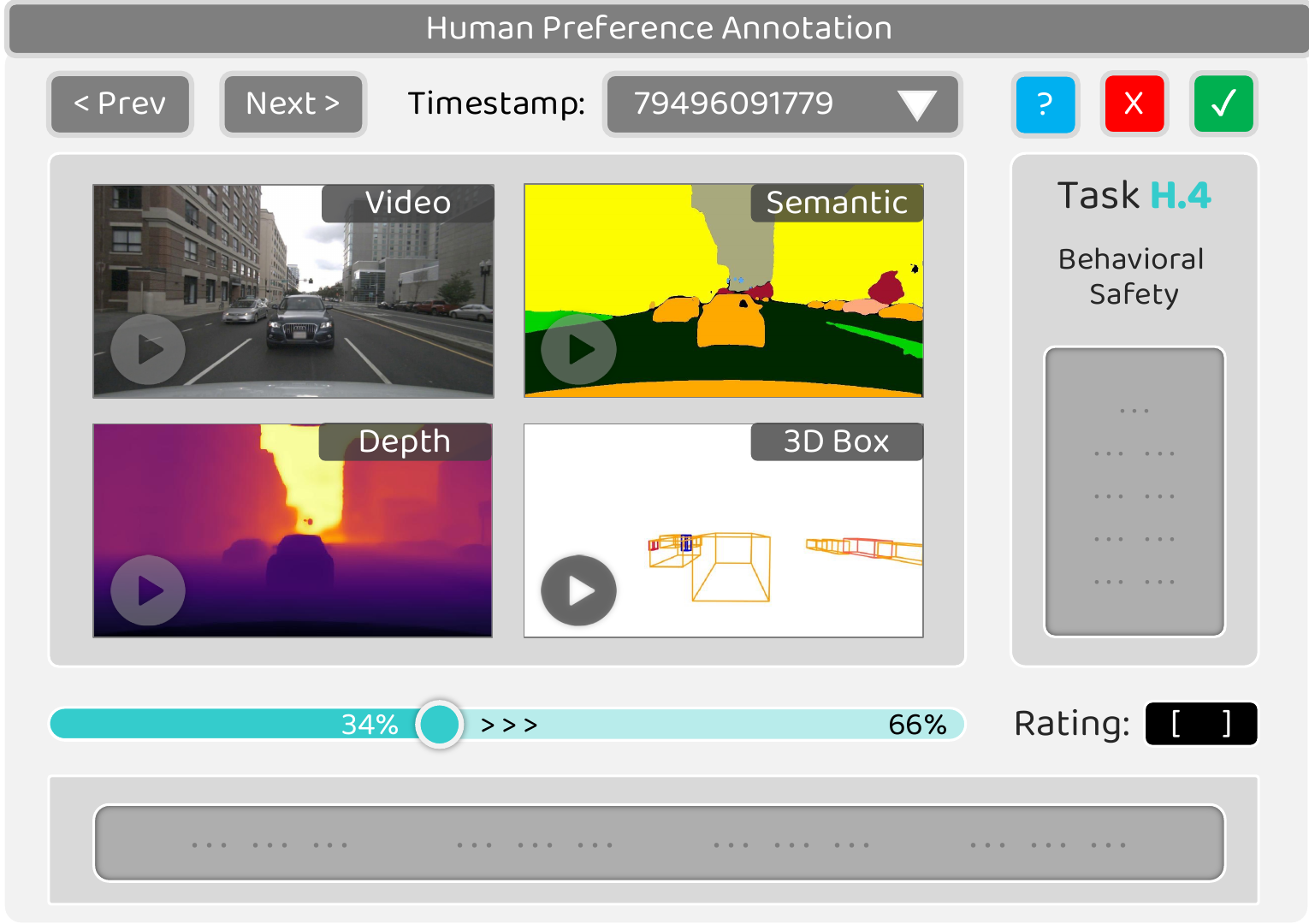}
        \vspace{-0.5cm}
        \caption{Interface for \textbf{Human Preference} annotation process. We present four synchronized views: $^1$generated video, $^2$semantic mask, $^3$depth map, and $^4$3D bounding boxes, enabling comprehensive judgment of realism, physical plausibility, and consistency.}
        \label{fig:interface}
    \end{minipage}
\end{wrapfigure}
To establish reliable human supervision for our benchmark, we designed a structured multi-stage annotation pipeline. Ten annotators were divided into \emph{two independent groups}, each responsible for scoring all videos under the four dimensions defined in Section~\ref{sec:human_preference}. For every video and dimension, the two groups annotated separately; when their ratings diverged, the sample was re-evaluated to ensure consistency. Annotators were presented with \textbf{four synchronized views}: $^1$\textit{generated video}, $^2$\textit{semantic mask}, $^3$\textit{depth map}, and $^4$\textit{3D boxes}, through the interface shown in Figure~\ref{fig:interface}. To promote consistency and domain understanding, all annotators received detailed documentation with examples illustrating each scoring level. On average, each annotation took approximately \textbf{two minutes and eight seconds}, amounting to over $\mathbf{930}$ \textbf{hours}. Further implementation details, documentation, and examples are provided in the \textbf{Appendix}.

\subsection{\textbf{\textsl{WorldLens-26K}}: A Diverse Preference Dataset}
\begin{wrapfigure}{r}{0.57\textwidth}
    \begin{minipage}{\linewidth}
        \centering
        \vspace{-0.2cm}
        \includegraphics[width=\linewidth]{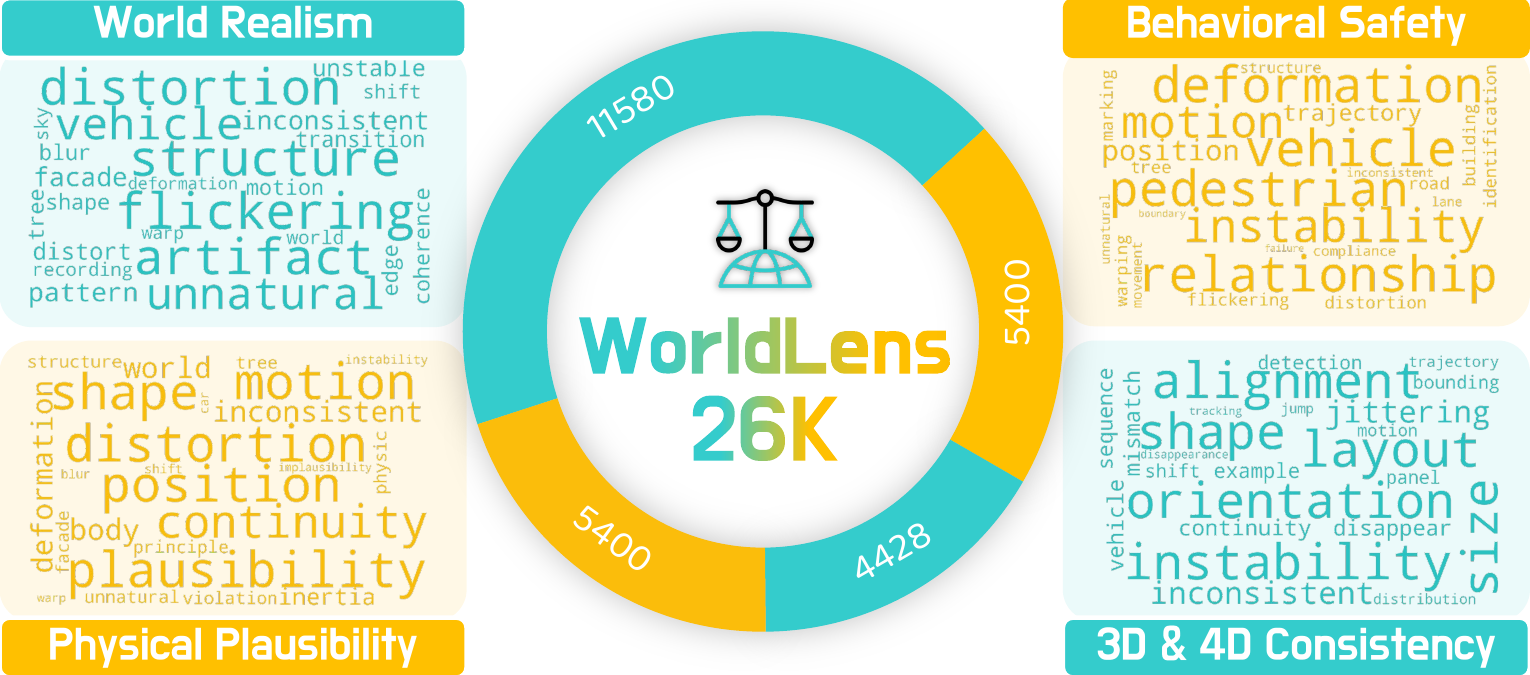}
        \vspace{-0.5cm}
        \caption{The statistics and word clouds of the \textbf{WorldLens-26K} dataset. Frequent keywords align closely with their target criteria (\emph{e.g.}, ``shape'', ``reflection'', ``motion'', ``safety''), confirming that annotators focus more on \textbf{dimension-specific perceptual attributes} and maintain consistent reasoning during the evaluation.}
        \label{fig:dataset}
    \end{minipage}
\end{wrapfigure}

To bridge the gap between human judgment and automated evaluation, we curate a large-scale human-annotated dataset comprising $\mathbf{26{,}808}$ \textbf{scoring records} of generated videos. Each entry includes a \emph{discrete score} and a \emph{concise textual rationale} written by annotators, capturing both quantitative assessment and qualitative explanation. The dataset covers complementary dimensions of perceptual quality (see Section~\ref{sec:human_preference}). This balanced design ensures comprehensive coverage across spatial, temporal, and behavioral aspects of world-model realism. As shown in Figure~\ref{fig:dataset}, the word clouds of textual rationales align closely with their corresponding target dimensions, confirming the validity and interpretability of the collected labels.
We envision \textbf{\textsl{WorldLens-26K}} as a foundational resource for training auto-evaluation agents and constructing human-aligned reward or advantage functions for reinforcement fine-tuning of generative world models.

\clearpage\clearpage
\subsection{\textbf{\textsl{WorldLens-Agent}}: SFT from Human Feedback}
Evaluating generated worlds hinges on human-centered criteria (\emph{physical plausibility}) and subjective preferences (\emph{perceived realism}) that quantitative metrics inherently miss, highlighting the necessity of a human-aligned evaluator. To this end, we introduce \textbf{\textsl{WorldLens-Agent}}, a vision-language critic agent trained on WorldLens-26K. Through LoRA-based supervised fine-tuning, we distill human perceptual and physical judgments into a Qwen3-VL-8B \cite{Qwen2.5-VL}, enabling it to internalize criteria such as realism, plausibility, and behavioral safety. This provides consistent, human-aligned assessments, offering a scalable preference oracle for benchmarking future world models. Kindly refer to Figure~\ref{fig:vis_sft} and the \textbf{Appendix} for automatic scoring and rationale generation examples on \emph{out-of-distribution} videos.

%% file: sections/4_experiments.tex
\section{Experiments}
\label{sec:experiments}

We comprehensively evaluate representative driving world models across all five aspects defined in \ours, covering both quantitative and human-in-the-loop dimensions. Due to space constraints, detailed configurations, metrics, and implementation details are provided in the \textbf{Appendix}.

\input{tables/bench_gen_recon}

\subsection{Per-Aspect Evaluations}

\noindent\textbf{Generation.}
As summarized in Table~\ref{tab:bench_gen_recon}, all existing models remain notably below the `Empirical Max', indicating substantial room for improving the visual and temporal realism of driving world models. Although DiST-4D~\cite{guo2025dist-4d} achieves the lowest \emph{Perceptual Discrepancy}, it underperforms OpenDWM~\cite{opendwm} in \emph{Subject Fidelity} and \emph{View Consistency}, demonstrating that perceptual metrics alone are insufficient for assessing physically coherent scene generation. OpenDWM provides the most balanced overall performance, largely due to large-scale multi-dataset training, while single-dataset models such as MagicDrive~\cite{gao2023magicdrive} and $\mathcal{X}$-Scene~\cite{yang2025x-scene} exhibit limited generalization across all metrics. Notably, conditioned approaches like DiST-4D~\cite{guo2025dist-4d} and DriveDreamer-2~\cite{zhao2024drivedreamer-2} partially overcome this limitation, improving \emph{Depth} and \emph{Cross-View Consistency} by $20$–$30\%$ through the use of ground-truth frames. These results highlight that \emph{the dataset diversity and conditioning strategies are more critical than perceptual fidelity for achieving reliable, temporally consistent world modeling.}

\begin{figure}[t]
    \centering
    \includegraphics[width=\textwidth]{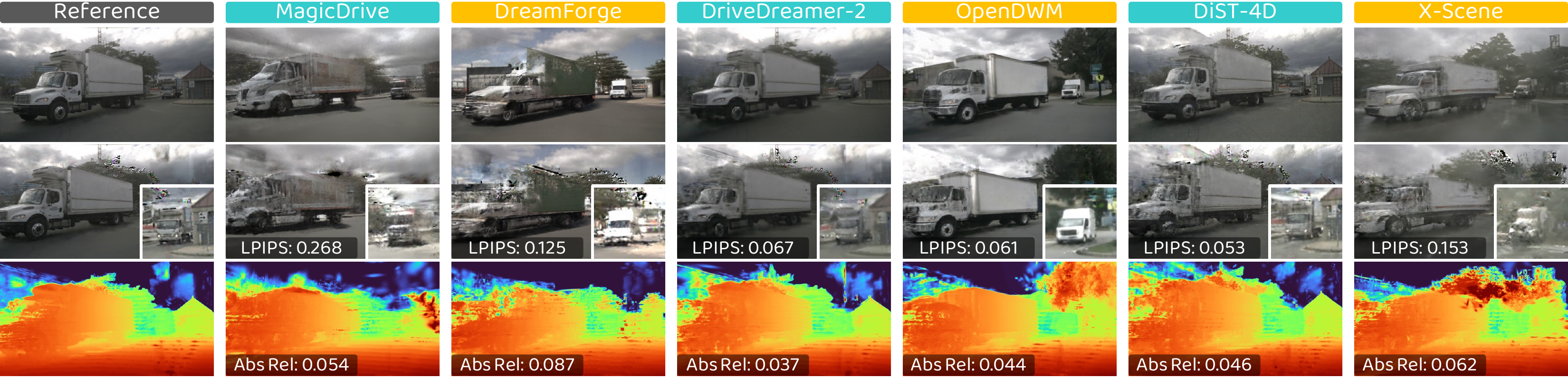}
    \vspace{-0.6cm}
    \caption{
        Qualitative results of \textbf{4D reconstruction} from generated videos. Rows (top to bottom) denote $^1$generated frame, $^2$rendered novel-view frame at a \emph{Lateral Offset}, and $^3$depth map. MagicDrive~\cite{gao2023magicdrive} and DreamForge~\cite{mei2024dreamforge} exhibit dense floaters and geometric distortions, while OpenDWM~\cite{opendwm} and DiST-4D~\cite{guo2025dist-4d} maintain temporally more consistent geometry, aligning with the quantitative results in Table~\ref{tab:bench_gen_recon}.
    }
    \label{fig:vis_recon}
\end{figure}

\noindent\textbf{Reconstruction.}
We assess the spatiotemporal 3D coherence by reconstructing generated videos into 4D Gaussian Fields~\cite{chen2025omnire}, where floaters and geometric instability directly reveal temporal inconsistency. As shown in Table~\ref{tab:bench_gen_recon}, MagicDrive~\cite{gao2023magicdrive} exhibits the weakest reconstruction, with the highest \emph{Photometric Error} and \emph{Geometric Discrepancy}, both over $\mathbf{2\times}$ worse than OpenDWM~\cite{opendwm}. DreamForge~\cite{mei2024dreamforge} shows similar artifacts, indicating limited \emph{3D Consistency}.
In contrast, OpenDWM and DiST-4D~\cite{guo2025dist-4d} achieve markedly better reconstruction, reducing photometric and geometric errors by about $55\%$ and producing more structurally coherent sequences. DiST-4D further attains the best \emph{Novel-View Quality}, likely due to its RGB-D generation design that better preserves depth over time.
As illustrated in Figure~\ref{fig:vis_recon}, MagicDrive and DreamForge produce dense floaters and distortions under lateral views, whereas OpenDWM and DiST-4D maintain clean, stable geometry. Overall, the results highlight that \emph{the temporal stability and geometric consistency are essential for physically realistic and reconstructable world models}.

\input{tables/bench_act}
\noindent\textbf{Action-Following.}
As shown in Table~\ref{tab:bench_act}, we evaluate the functional viability of synthesized environments through closed-loop simulation, where a pretrained planner operates within the ``world'' of each model. Temporal coherence proves critical, as planning agents rely on multi-frame history and ego-state cues; models with weaker temporal stability achieve the lowest \emph{Route Completion} rates. 
A notable finding is the large disparity between open-loop and closed-loop performance. Despite strong open-loop results on \emph{Displacement Error} and \emph{PDMS}, all methods collapse under closed-loop conditions, achieving only marginal \emph{Route Completion} rates. Frequent failures (\emph{e.g.}, collisions, off-road drift) suggest that current synthetic data remain inadequate substitutes for real-world data in high-level control. This highlights a key insight: \emph{enhancing the physical and causal realism of generative world models is indispensable for effective closed-loop deployment}.

\begin{figure}[t]
    \centering
    \includegraphics[width=\textwidth]{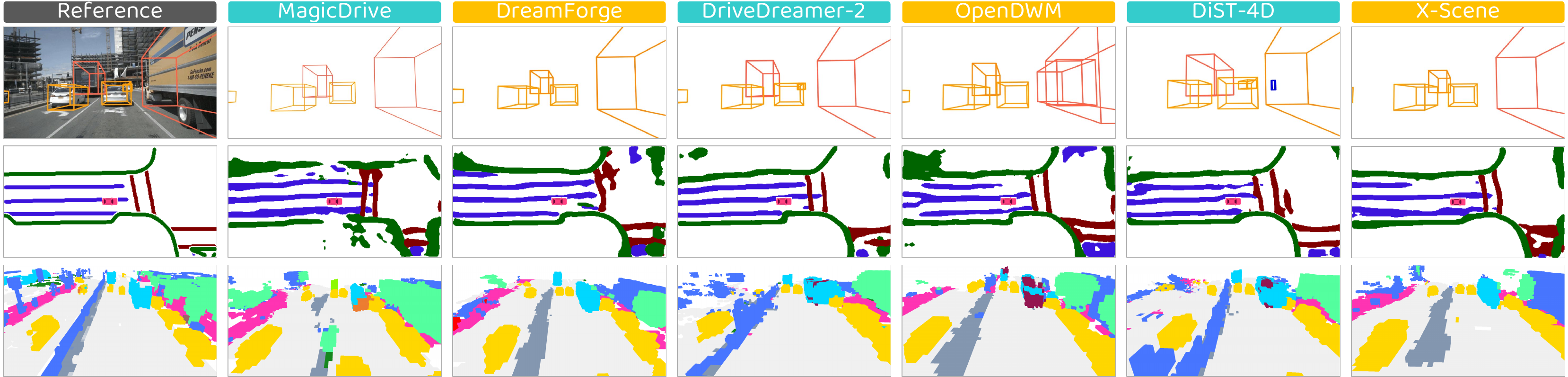}
    \vspace{-0.6cm}
    \caption{
        Qualitative results of \textbf{Downstream Tasks}. Rows (from top to bottom): $^1$3D object detection, $^2$map segmentation, and $^3$semantic occupancy prediction tasks.
    }
    \label{fig:vis_downstream}
\end{figure}

\input{tables/bench_downstream}
\noindent\textbf{Downstream Tasks.}
This aspect directly reflects the practical utility of world models beyond visual realism. As shown in Table~\ref{tab:bench_downstream} and Figure~\ref{fig:teaser}, DiST-4D~\cite{guo2025dist-4d} leads by a large margin across all tasks, (\emph{i.e.}, map segmentation, 3D detection, and tracking), averaging $30$–$40\%$ higher than the next best models. DriveDreamer-2~\cite{zhao2024drivedreamer-2} ranks second, particularly excelling in occupancy prediction, highlighting the advantage of temporal conditioning for consistent video generation. In contrast, MagicDrive~\cite{gao2023magicdrive} performs weakest across all tasks, confirming its limited spatiotemporal coherence. Interestingly, despite strong perceptual quality, OpenDWM~\cite{opendwm} underperforms in detection ($21.9\%$) and tracking ($6.9\%$), suggesting that large-scale multi-domain training may hinder adaptation to specific dataset distributions. Additional qualitative assessments in Figure~\ref{fig:vis_downstream} further verify our observations. Overall, these results indicate that \emph{the temporal conditioning and dataset alignment are critical for task-specific effectiveness for practical usages.}

\subsection{Human Preference Alignments}
\begin{wrapfigure}{r}{0.6\textwidth}
    \begin{minipage}{\linewidth}
    \centering
    \vspace{-0.1cm}
    \begin{subfigure}[t]{0.47\linewidth}
        \centering
        \includegraphics[width=\linewidth]{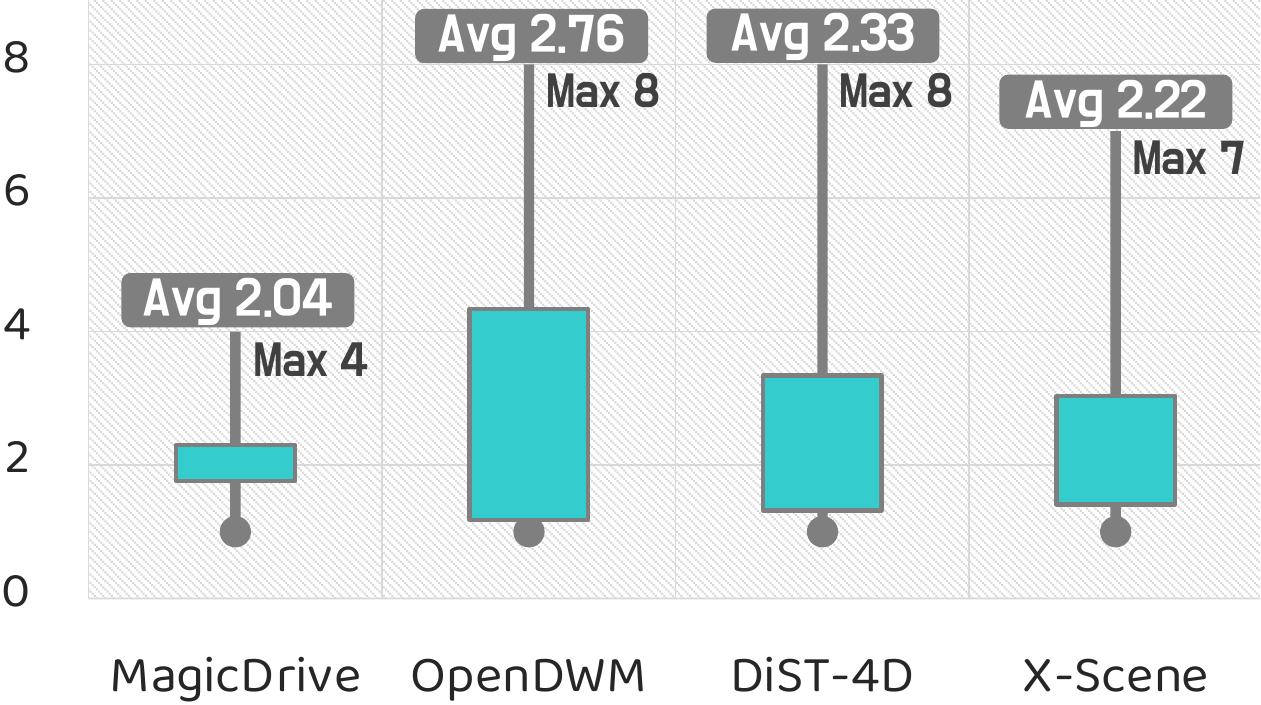}
        \caption{World Realism}
        \label{fig:sub_a}
    \end{subfigure}
    \hfill
    \begin{subfigure}[t]{0.47\linewidth}
        \centering
        \includegraphics[width=\linewidth]{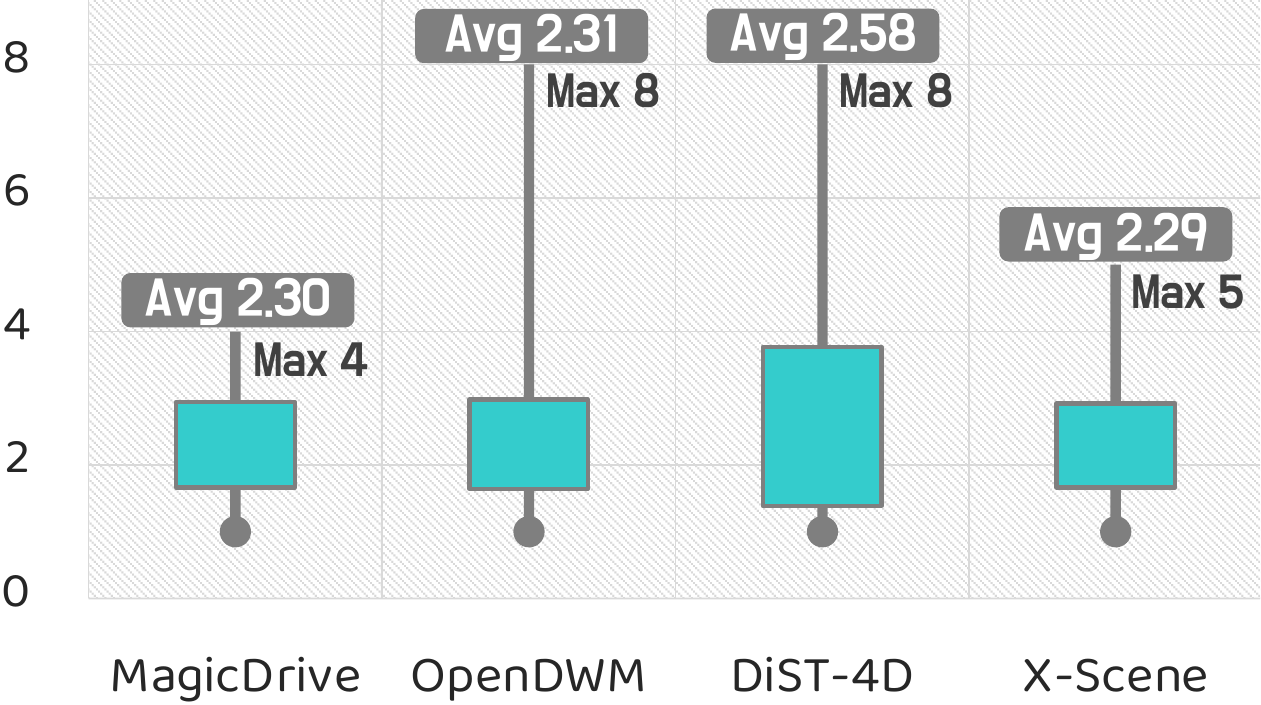}
        \caption{Physical Plausibility}
        \label{fig:sub_b}
    \end{subfigure}
    \vspace{3pt}
    \begin{subfigure}[t]{0.47\linewidth}
        \centering
        \includegraphics[width=\linewidth]{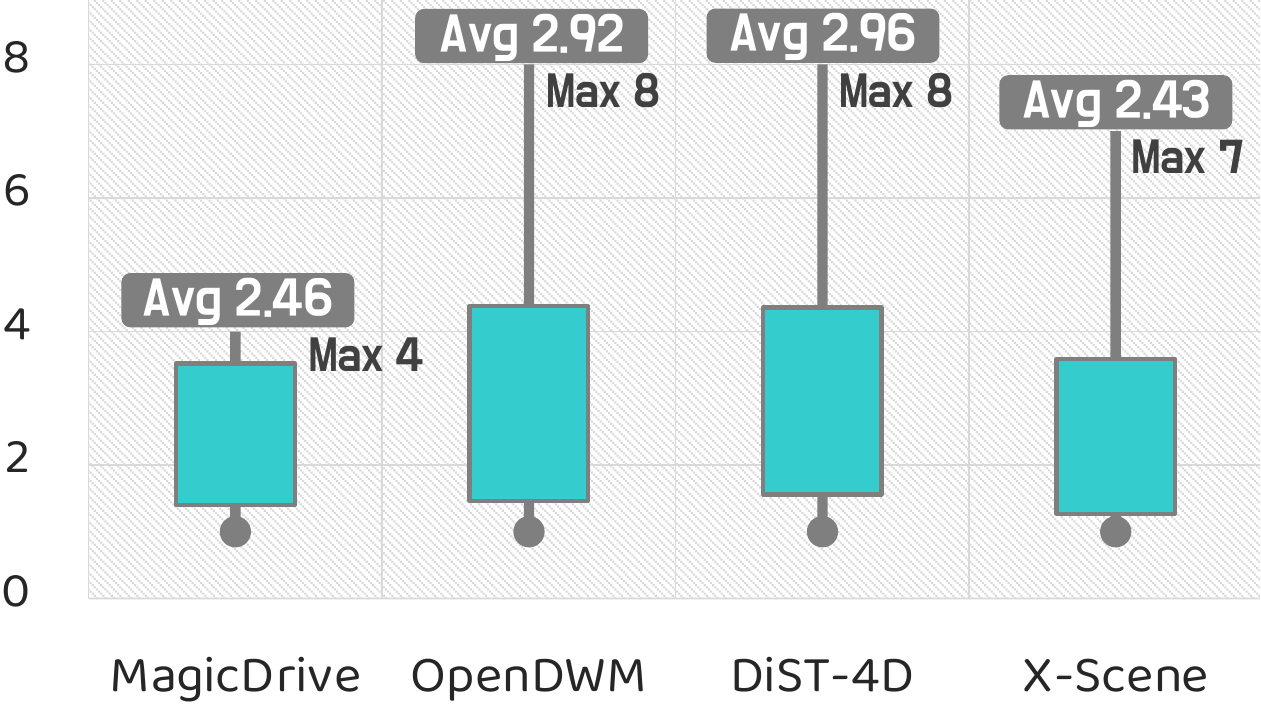}
        \caption{3D \& 4D Consistency}
        \label{fig:sub_c}
    \end{subfigure}
    \hfill
    \begin{subfigure}[t]{0.47\linewidth}
        \centering
        \includegraphics[width=\linewidth]{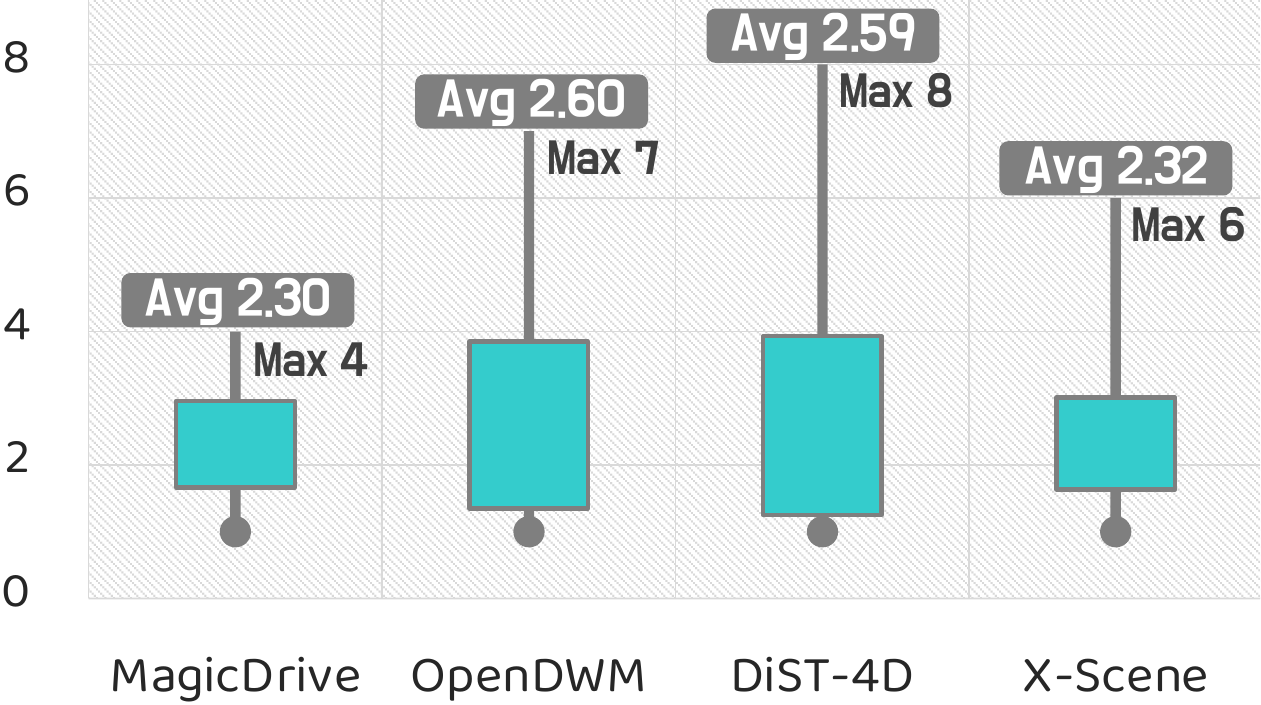}
        \caption{Behavioral Safety}
        \label{fig:sub_d}
    \end{subfigure}
    \vspace{-0.2cm}
    \caption{Summary of alignments to \textbf{Human Preference}, where the max, median, and average scores of each model are compared. For more detailed analyses, kindly refer to the \textbf{Appendix}.}
    \label{fig:human_preference}
    \end{minipage}
\end{wrapfigure}

\noindent\textbf{Subjective Evaluations.}
Since not all aspects of world modeling can be captured by quantitative metrics, we conducted a human evaluation focusing on \emph{World Realism}, \emph{Physical Plausibility}, \emph{3D \& 4D Consistency}, and \emph{Behavioral Safety}.  
As shown in Figure~\ref{fig:human_preference}, overall scores remain modest (on average `$2$'$\sim$`$3$' out of `$10$'), revealing that current world models are far from human-level realism. DiST-4D \cite{guo2025dist-4d} achieves the most balanced scores across all dimensions, leading in physical plausibility (`$2.58$') and behavioral safety (`$2.59$'). OpenDWM \cite{opendwm} attains the highest realism (`$2.76$') but slightly lower physical consistency, while MagicDrive \cite{gao2023magicdrive} ranks lowest overall, reflecting poor coherence. Interestingly, \emph{World Realism} and \emph{Consistency} scores correlate strongly, suggesting that human-perceived realism is tightly coupled with geometric and temporal stability. Overall, these results underscore the \emph{necessity of human-in-the-loop evaluation to complement quantitative benchmarks and provide a holistic assessment of world-model quality.}

\begin{wrapfigure}{r}{0.5\textwidth}
    \begin{minipage}{\linewidth}
        \centering
        \vspace{-0.4cm}
        \includegraphics[width=\linewidth]{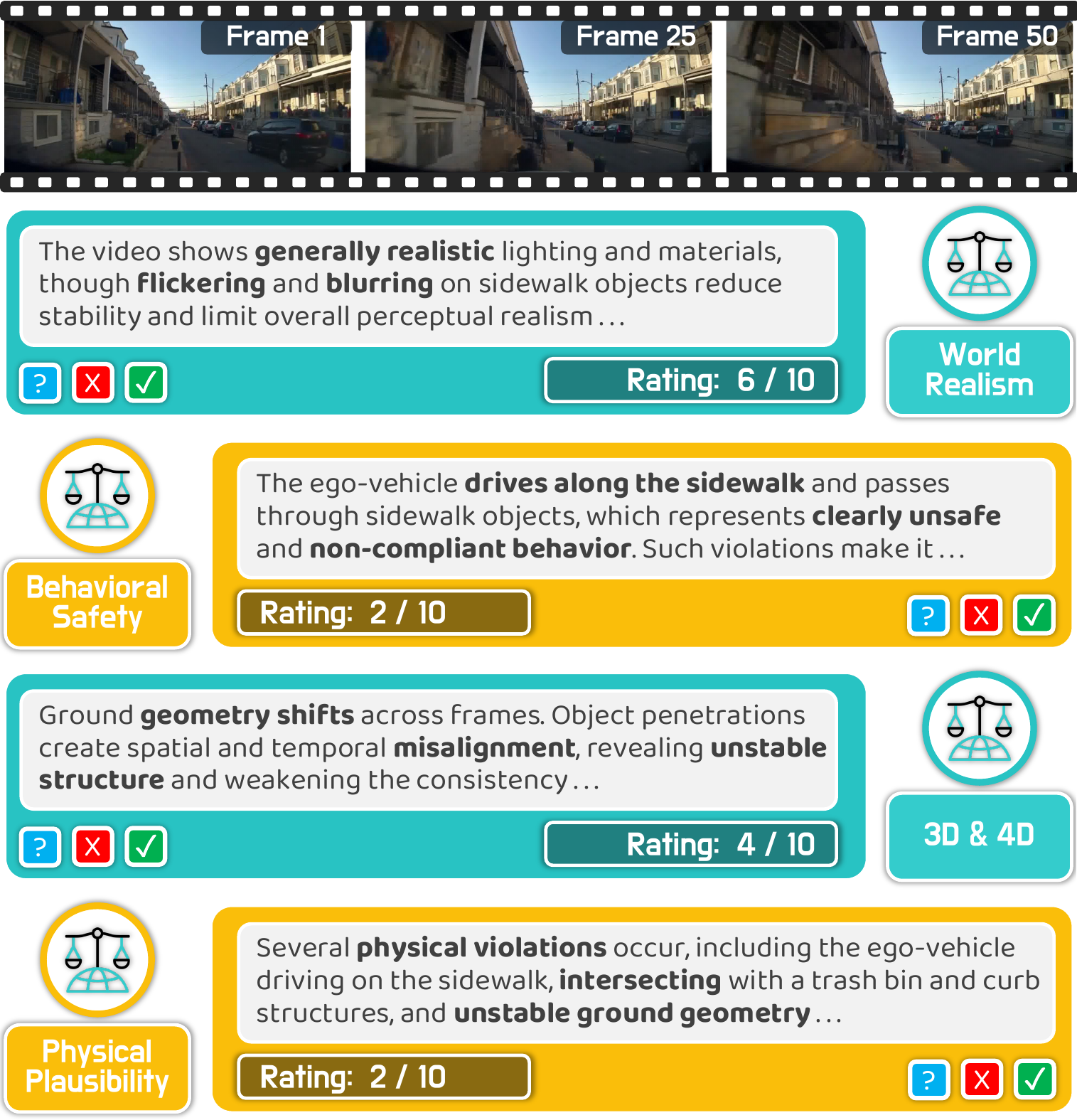}
        \vspace{-0.6cm}
        \caption{Zero-shot evaluations by \textbf{\textsl{WorldLens-Agent}} on unseen videos (from Gen3C \cite{ren2025gen3c}), exhibiting strong alignments with human reasoning. See the \textbf{Appendix} for more examples.}
        \label{fig:vis_sft}
    \end{minipage}
    \vspace{-0.5cm}
\end{wrapfigure}

\noindent\textbf{Human-Agent Alignments.}
To assess the generalizability of our automatic evaluator, \textbf{\textsl{WorldLens-Agent}}, we conduct a zero-shot test on videos generated by Gen3C~\cite{ren2025gen3c}, as shown in Figure~\ref{fig:vis_sft}. The agent’s predicted scores exhibit strong alignment with human annotations across all evaluated dimensions, confirming its ability to capture nuanced subjective preferences. Beyond numerical agreement, the textual rationales generated by the agent closely mirror those written by human annotators, demonstrating both \emph{score-level consistency} and \emph{interpretive coherence}. These results highlight the effectiveness of leveraging human-annotated perception data to train scalable, explainable, and reproducible evaluative agents for future world-model benchmarking.

\subsection{Insights \& Discussions}

\noindent\emph{\textbf{Comprehensive Evaluations are Crucial.}}
No single world model excels in all aspects (Figure~\ref{fig:teaser}): DiST-4D performs best in geometry and novel-view metrics, OpenDWM leads in photometric fidelity, and DriveDreamer-2 achieves the highest depth accuracy. This divergence shows that visual realism, geometric consistency, and downstream usability are \emph{complementary rather than interchangeable, highlighting the necessity of multi-dimensional benchmarking.}

\noindent\emph{\textbf{Perceptual Quality Does Not Imply Usability.}}
Models with strong perceptual scores (\emph{e.g.}, OpenDWM) may underperform on downstream tasks. Despite its visual fidelity, OpenDWM scores \emph{30\% lower} than DiST-4D in 3D detection, indicating that large-scale, multi-domain training can hinder adaptation to task-specific distributions. Hence, \emph{aligning generated data with the target domain is more crucial than perceptual realism for effective downstream use}.

\noindent\emph{\textbf{Geometry Awareness Enables Physical Coherence.}}
The superior reconstruction and novel-view performance of DiST-4D stem from its RGB-D generation and decoupled spatiotemporal diffusion, which jointly model temporal forecasting and spatial synthesis.
This shows that \emph{geometry-aware supervision significantly improves the physical realism and reconstructability of generated scenes.}

\noindent\emph{\textbf{Joint Optimization of Appearance and Geometry.}}
The discrepancy between photometric (LPIPS/PSNR) and geometric metrics (Abs Rel) reveals that current models often treat texture and structure as independent objectives. Geometry-aware supervision stabilizes depth but blurs details, while appearance-driven training sharpens textures yet breaks spatial consistency. A unified formulation that \emph{jointly optimizes appearance and geometry through spatiotemporal regularization yields consistent reconstruction.}

\noindent\emph{\textbf{Guidelines for Future World Model Design.}}
Key principles emerge for developing physically grounded world models: \emph{1) Prioritize geometry as a core objective:} explicit depth prediction or supervision enhances both reconstruction fidelity and downstream perception; \emph{2) Stabilize foreground dynamics:} consistent geometry is essential for reliable motion disentanglement; \emph{3) Ensure autoregressive resilience:} enforcing cross-view and temporal consistency mitigates drift and structural artifacts, while training with self-forcing~\cite{huang2025self, cui2025self} or streaming diffusion~\cite{kodaira2025streamdiffusion} enhances robustness against compounding errors in closed-loop generation, which is crucial for long-horizon stability. Overall, robust world models stem from the \emph{joint optimization of appearance, geometry, and task adaptability}, advancing from \emph{visual realism} toward \emph{physical reliability}.

%% file: tables/bench_gen_recon.tex
\begin{table}[t]
    \centering
    \caption{Benchmarking results of state-of-the-art driving world models for \textbf{Generation} and \textbf{Reconstruction} in WorldLens. }
    \vspace{-0.1cm}
    \label{tab:bench_gen_recon}
    \resizebox{\linewidth}{!}{
    \begin{tabular}{r|r|cccccccc|cccc}
        \toprule
        & & \multicolumn{8}{c|}{\cellcolor{w_blue!24}\textbf{~Aspect: Generation}} & \multicolumn{4}{c}{\cellcolor{w_yellow!24}\textbf{~Aspect: Reconstruction}}
        \\
        \multirow{2.2}{*}{\textbf{Model}} & \multirow{2.2}{*}{\textbf{Venue}} & Subject & Subject & Subject & Depth & Temp. & Sem. & Percept. & View & Photo. & Geo. & Novel & Novel
        \\
        & & ~Fid.$\uparrow$ & ~Cohe.$\uparrow$ & ~Cons.$\uparrow$ & ~Disc.$\downarrow$ & ~Cons.$\uparrow$ & ~Cons.$\uparrow$ & ~Disc.$\downarrow$ & ~Cons.$\uparrow$ & ~Error$\downarrow$ & ~Disc.$\downarrow$ & ~Qual.$\uparrow$ & ~Disc.$\downarrow$
        \\
        & & {{\raisebox{0.9pt}{\colorbox{w_blue}{\scriptsize{\textbf{\textbf{\textsc{\textcolor{white}{~G.1~}}}}}}}}} & {{\raisebox{0.9pt}{\colorbox{w_blue}{\scriptsize{\textbf{\textbf{\textsc{\textcolor{white}{~G.2~}}}}}}}}} & {{\raisebox{0.9pt}{\colorbox{w_blue}{\scriptsize{\textbf{\textbf{\textsc{\textcolor{white}{~G.3~}}}}}}}}} & {{\raisebox{0.9pt}{\colorbox{w_blue}{\scriptsize{\textbf{\textbf{\textsc{\textcolor{white}{~G.4~}}}}}}}}} & {{\raisebox{0.9pt}{\colorbox{w_blue}{\scriptsize{\textbf{\textbf{\textsc{\textcolor{white}{~G.5~}}}}}}}}} & {{\raisebox{0.9pt}{\colorbox{w_blue}{\scriptsize{\textbf{\textbf{\textsc{\textcolor{white}{~G.6~}}}}}}}}} & {{\raisebox{0.9pt}{\colorbox{w_blue}{\scriptsize{\textbf{\textbf{\textsc{\textcolor{white}{~G.7~}}}}}}}}} & {{\raisebox{0.9pt}{\colorbox{w_blue}{\scriptsize{\textbf{\textbf{\textsc{\textcolor{white}{~G.8~}}}}}}}}} & {{\raisebox{0.9pt}{\colorbox{w_yellow}{\scriptsize{\textbf{\textbf{\textsc{\textcolor{white}{~R.1~}}}}}}}}} & {{\raisebox{0.9pt}{\colorbox{w_yellow}{\scriptsize{\textbf{\textbf{\textsc{\textcolor{white}{~R.2~}}}}}}}}} & {{\raisebox{0.9pt}{\colorbox{w_yellow}{\scriptsize{\textbf{\textbf{\textsc{\textcolor{white}{~R.3~}}}}}}}}} & {{\raisebox{0.9pt}{\colorbox{w_yellow}{\scriptsize{\textbf{\textbf{\textsc{\textcolor{white}{~R.4~}}}}}}}}}
        \\
        \midrule
        MagicDrive \cite{gao2023magicdrive} & {\small ICLR'23} & $28.49$ & $75.95$ & $65.22\%$ & $24.19$ & $74.44\%$ & $80.63\%$ & $222.00$ & $185.77$ & $0.140$ & $0.115$ & $39.82\%$ & $427.30$
        \\
        DreamForge \cite{mei2024dreamforge} & {\small arXiv'24} & \underline{$31.99$} & $75.12$ & \underline{$76.40\%$} & $19.27$ & \cellcolor{w_blue!20}$\mathbf{79.82\%}$ & \underline{$84.99\%$} & $189.76$ & $194.99$ & $0.097$ & $0.105$ & $\underline{41.23\%}$ & $347.70$
        \\
        DriveDreamer-2 \cite{zhao2024drivedreamer-2} & {\small AAAI'25} & $27.38$ & $78.97$ & $74.49\%$ & \underline{$17.73$} & $79.51\%$ & \cellcolor{w_blue!20}$\mathbf{85.91\%}$ & $127.07$ & \underline{$302.83$} & $0.093$ & \cellcolor{w_yellow!20}$\mathbf{0.073}$ & $36.10\%$ & $\underline{259.91}$
        \\
        OpenDWM \cite{opendwm} & {\small CVPR'25} & \cellcolor{w_blue!20}$\mathbf{36.30}$ & \cellcolor{w_blue!20}$\mathbf{83.13}$ & \cellcolor{w_blue!20}$\mathbf{78.33\%}$ & $18.17$ & \underline{$79.63\%$} & $84.08\%$ & \underline{$90.42$} & $211.18$ & \cellcolor{w_yellow!20}$\mathbf{0.065}$ & $0.088$ & $39.54\%$ & $287.73$
        \\
        DiST-4D \cite{guo2025dist-4d} & {\small ICCV'25} & $30.32$ & \underline{$79.36$} & $74.69\%$ & \cellcolor{w_blue!20}$\mathbf{17.71}$ & $77.76\%$ & $84.32\%$ & \cellcolor{w_blue!20}$\mathbf{58.08}$ & \cellcolor{w_blue!20}$\mathbf{389.78}$ & $\underline{0.066}$ & $\underline{0.080}$ & \cellcolor{w_yellow!20}$\mathbf{43.09\%}$ & \cellcolor{w_yellow!20}$\mathbf{192.39}$
        \\
        $\mathcal{X}$-Scene \cite{yang2025x-scene} & {\small NeurIPS'25} & $27.17$ & $77.22$ & $74.37\%$ & $20.50$ & $79.41\%$ & $83.80\%$ & $179.74$ & $201.00$ & $0.098$ & $0.096$ & $38.04\%$ & $365.71$
        \\
        \midrule
        \rowcolor{gray!7}\textcolor{gray}{Empirical Max} & \textcolor{gray}{-} & \textcolor{gray}{$60.22$} & \textcolor{gray}{$83.25$} & \textcolor{gray}{$93.66\%$} & \textcolor{gray}{$14.27$} & \textcolor{gray}{$93.24\%$} & \textcolor{gray}{$86.39\%$} & \textcolor{gray}{-} & \textcolor{gray}{$570.75$} & \textcolor{gray}{0.056} & \textcolor{gray}{-} & \textcolor{gray}{$45.69\%$} & \textcolor{gray}{-}
        \\
        \bottomrule
    \end{tabular}}
\end{table}

%% file: tables/bench_act.tex
\begin{wraptable}{r}{0.6\textwidth}
    \centering
    \vspace{-0.5cm}
    \caption{Benchmarking results of state-of-the-art driving world models for \textbf{Action-Following} dimensions in WorldLens.}
    \vspace{-0.1cm}
    \label{tab:bench_act}
    \resizebox{\linewidth}{!}{
    \begin{tabular}{r|r|cccc}
        \toprule
        & & \multicolumn{4}{c}{\cellcolor{w_blue!24}\textbf{~Aspect: Action-Following}}
        \\
        \multirow{2.2}{*}{\textbf{Model}} & \multirow{2.2}{*}{\textbf{Venue}} & Displ. & Open-L & Route & Closed-L
        \\
        & & ~Error$\downarrow$ & ~Adh.$\uparrow$ & ~Compl.$\uparrow$ & ~Adh.$\uparrow$
        \\
        & & {{\raisebox{0.9pt}{\colorbox{w_blue}{\scriptsize{\textbf{\textbf{\textsc{\textcolor{white}{~A.1~}}}}}}}}} & {{\raisebox{0.9pt}{\colorbox{w_blue}{\scriptsize{\textbf{\textbf{\textsc{\textcolor{white}{~A.2~}}}}}}}}} & {{\raisebox{0.9pt}{\colorbox{w_blue}{\scriptsize{\textbf{\textbf{\textsc{\textcolor{white}{~A.3~}}}}}}}}} & {{\raisebox{0.9pt}{\colorbox{w_blue}{\scriptsize{\textbf{\textbf{\textsc{\textcolor{white}{~A.4~}}}}}}}}}
        \\
        \midrule
        MagicDrive \cite{gao2023magicdrive} & {\small ICLR'23} & $0.57$ & $71.23\%$ & $6.89\%$ & $4.82\%$
        \\
        Panacea \cite{wen2024panacea} & {\small CVPR'24} & $0.58$ & - & - & -
        \\
        DreamForge \cite{mei2024dreamforge} & {\small arXiv'24} & $0.55$ & $75.51\%$ & $10.23\%$ & $7.65\%$
        \\
        DrivingSphere \cite{yan2025drivingsphere} & {\small CVPR'25} & \underline{$0.54$} & $76.02\%$ & $11.02\%$ & $8.29\%$
        \\
        MagicDrive-V2 \cite{gao2024magicdrive-v2} & {\small ICCV'25} & \cellcolor{w_blue!20}$\mathbf{0.53}$ & \cellcolor{w_blue!20}$\mathbf{78.91}\%$ & \underline{$12.31\%$} & \underline{$9.50\%$}
        \\
        RLGF \cite{yan2025rlgf} & {\small NeurIPS'25} & \cellcolor{w_blue!20}$\mathbf{0.53}$ & $78.45\%$ & \cellcolor{w_blue!20}$\mathbf{13.51}\%$ & \cellcolor{w_blue!20}$\mathbf{10.59}\%$
        \\
        \midrule
        \rowcolor{gray!7}\textcolor{gray}{Empirical Max} & \textcolor{gray}{-} & \textcolor{gray}{$0.51$} & \textcolor{gray}{-}& \textcolor{gray}{-} & \textcolor{gray}{-}
        \\
        \bottomrule
    \end{tabular}
    }
    \vspace{-0.2cm}
\end{wraptable}

%% file: tables/bench_downstream.tex
\begin{wraptable}{r}{0.6\textwidth}
    \centering
    \vspace{-0.6cm}
    \caption{Summary of benchmarking results of state-of-the-art world models for \textbf{Downstream Task} dimensions in WorldLens.}
    \vspace{-0.2cm}
    \label{tab:bench_downstream}
    \resizebox{\linewidth}{!}{
    \begin{tabular}{r|r|cccc}
        \toprule
        & & \multicolumn{4}{c}{\cellcolor{w_yellow!24}\textbf{~Aspect: Downstream Task}}
        \\
        \multirow{2.2}{*}{\textbf{Model}} & \multirow{2.2}{*}{\textbf{Venue}} & Map & 3D Obj. & 3D Obj. & Occ.
        \\
        & & ~Seg.$\uparrow$ & ~Det.$\uparrow$ & ~Trk.$\uparrow$ & ~Pred.$\uparrow$
        \\
        & & {{\raisebox{0.9pt}{\colorbox{w_yellow}{\scriptsize{\textbf{\textbf{\textsc{\textcolor{white}{~D.1~}}}}}}}}} & {{\raisebox{0.9pt}{\colorbox{w_yellow}{\scriptsize{\textbf{\textbf{\textsc{\textcolor{white}{~D.2~}}}}}}}}} & {{\raisebox{0.9pt}{\colorbox{w_yellow}{\scriptsize{\textbf{\textbf{\textsc{\textcolor{white}{~D.3~}}}}}}}}} & {{\raisebox{0.9pt}{\colorbox{w_yellow}{\scriptsize{\textbf{\textbf{\textsc{\textcolor{white}{~D.4~}}}}}}}}}
        \\
        \midrule
        MagicDrive \cite{gao2023magicdrive} & {\small ICLR'23} & $18.34\%$ & $22.41\%$ & $7.90\%$ & $23.14\%$
        \\
        DreamForge \cite{mei2024dreamforge} & {\small arXiv'24} & $30.31\%$ & $26.71\%$ & $10.30\%$ & $23.71\%$
        \\
        DriveDreamer-2 \cite{zhao2024drivedreamer-2} & {\small AAAI'25} & \underline{$33.62\%$} & \underline{$30.90\%$} & \underline{$13.30\%$} & \cellcolor{w_yellow!20}$\mathbf{26.82\%}$
        \\
        OpenDWM \cite{opendwm} & {\small CVPR'25} & $27.63\%$ & $21.96\%$ & $6.90\%$ & $24.82\%$
        \\
        DiST-4D \cite{guo2025dist-4d} & {\small ICCV'25} & \cellcolor{w_yellow!20}$\mathbf{35.55\%}$ & \cellcolor{w_yellow!20}$\mathbf{33.22}\%$ & \cellcolor{w_yellow!20}$\mathbf{15.30}\%$ & \underline{$26.10\%$}
        \\
        $\mathcal{X}$-Scene \cite{yang2025x-scene} & {\small NeurIPS'25} & $27.24\%$ & $29.89\%$ & $8.80\%$ & $23.68\%$
        \\
        \midrule
        \rowcolor{gray!7}\textcolor{gray}{Empirical Max} & \textcolor{gray}{-} & \textcolor{gray}{$40.64\%$} & \textcolor{gray}{$44.72\%$} & \textcolor{gray}{$36.30\%$} & \textcolor{gray}{$37.05\%$}
        \\
        \bottomrule
    \end{tabular}}
    \vspace{-0.7cm}
\end{wraptable}

%% file: sections/5_conclusion.tex
\section{Conclusion}
\label{sec:conclusion}

We presented \ours, a full-spectrum benchmark that evaluates generative world models across perception, geometry, function, and human alignment perspectives. Through five complementary evaluation aspects and over twenty standardized metrics, it offers a unified protocol for measuring both physical and perceptual realism. Together with \textbf{WorldLens-26K} and \textbf{WorldLens-Agent}, our framework establishes a scalable, interpretable foundation for benchmarking future world models -- guiding progress toward systems that not only \textit{look} real but also \textit{behave} realistically.

%% file: sections_supp/2_generation.tex
\section{~Aspect 1: Generation}

In this section, we detail the metrics used to evaluate the \textbf{quality of generation} of driving world models. This aspect assesses the overall realism, coherence, and physical plausibility of generated driving videos, capturing how well a model reconstructs the spatiotemporal structure of real-world scenes. 

\input{sections_supp/dimensions/gen_subject_fidelity}

\clearpage\clearpage
\input{sections_supp/dimensions/gen_subject_coherence}

\clearpage\clearpage
\input{sections_supp/dimensions/gen_subject_consistency}

\clearpage\clearpage
\input{sections_supp/dimensions/gen_depth_discrepancy}

\clearpage\clearpage
\input{sections_supp/dimensions/gen_temporal_consistency}

\clearpage\clearpage
\input{sections_supp/dimensions/gen_semantic_consistency}

\clearpage\clearpage
\input{sections_supp/dimensions/gen_perceptual_discrepancy}

\clearpage\clearpage
\input{sections_supp/dimensions/gen_cross_view_consistency}

%% file: sections_supp/dimensions/gen_subject_fidelity.tex
\subsection{~Subject Fidelity}

\subsubsection{~Definition}
Subject Fidelity quantifies the perceptual realism of object instances, such as vehicles and pedestrians, that appear in generated driving videos. It focuses on assessing whether each synthesized object visually resembles its real-world counterpart in both appearance and semantic attributes. By isolating individual instances, this metric emphasizes fine-grained visual fidelity that global perceptual measures may overlook, providing an object-centric view of generation quality.

\subsubsection{~Formulation}
\vspace{-0.1cm}
For a generated video $y_j=\{y_j^{(t)}\}_{t=1}^{T}$ with bounding boxes $\{b_{j,k}^{(t)}\}_{k=1}^{K_j^{(t)}}$, we crop object patches $o_{j,k}^{(t)}=\mathrm{Crop}(y_j^{(t)},b_{j,k}^{(t)})$. Let $\mathcal{C}$ denote the evaluated object categories (\emph{e.g.}, vehicle, pedestrian), and $\psi_{\mathrm{CLS}}^{(c)}(\cdot)$ be a pretrained binary classifier for class $c\!\in\!\mathcal{C}$ outputting confidence $p_{j,k}^{(t,c)}\!\in\![0,1]$ that patch $o_{j,k}^{(t)}$ looks real for that class.
Aggregating across all objects, frames, videos, and classes yields the overall Subject Fidelity score:
\begin{equation}
\boxed{\;
\mathcal{S}_{\mathrm{SF}}(\mathcal{Y})
= \tfrac{1}{N_g|\mathcal{C}|}
  \sum\nolimits_{j=1}^{N_g}\!
  \sum\nolimits_{c\in\mathcal{C}}\!
  \tfrac{1}{T}\!
  \sum\nolimits_{t=1}^{T}\!
  \tfrac{1}{K_{j,c}^{(t)}}\!
  \sum\nolimits_{k=1}^{K_{j,c}^{(t)}}\!
  p_{j,k}^{(t,c)}
\;}
\label{eq:subject_fidelity}
\end{equation}
A higher $\mathcal{S}_{\mathrm{SF}}$ score indicates that generated objects are both visually convincing and semantically consistent with their intended categories. Models achieving high fidelity tend to produce realistic textures, shapes, and colors, even under varying viewpoints and lighting conditions. This metric thus complements global measures like FVD or LPIPS by focusing on localized realism at the instance level, offering insights into whether the generated world contains physically believable and semantically meaningful entities.

\subsubsection{~Implementation Details}
We use class-specific confidence scores for evaluation. Pedestrian crops are classified using a pedestrian classifier pretrained on several commonly used pedestrian-datasets~\cite{zheng2015scalable, wei2018person, dalal2005histograms, overett2008new}, while vehicle crops are classified with a ViT-B/16 model (\texttt{`google/vit-base-patch16-224`})~\cite{wu2020visual} pretrained on ImageNet-21k ($14$ million images, $21{,}843$ classes)~\cite{deng2009imagenet}. Category grouping is determined by regex-based label matching against the model’s \texttt{id2label}. Images are resized to $256\times128$ and normalized before inference. For each tracklet, we average the classification confidence of all selected frames, and report the mean confidence as the final score.

\subsubsection{~Examples}
\begin{figure}[t]
    \centering
    \begin{subfigure}[h]{\textwidth}
        \centering
        \includegraphics[width=\linewidth]{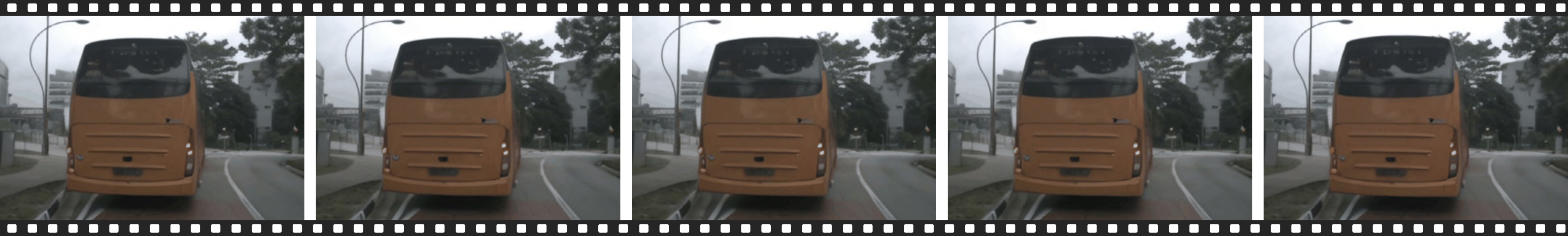}
        \caption{Good example in the \emph{Subject Fidelity} dimension (Score: \textcolor{w_blue}{$94.64\%$})}
        \label{fig:gen_subject_fidelity_1}
    \end{subfigure}
    \\[1ex]
    \begin{subfigure}[h]{\textwidth}
        \centering
        \includegraphics[width=\linewidth]{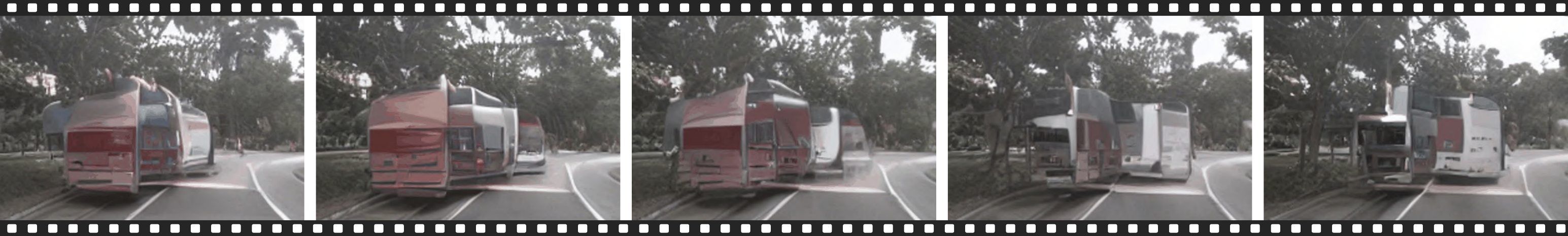}
        \caption{Bad example in the \emph{Subject Fidelity} dimension (Score: \textcolor{red}{$15.42\%$})}
        \label{fig:gen_subject_fidelity_2}
    \end{subfigure}
    \\[3.5ex]
    \begin{subfigure}[h]{\textwidth}
        \centering
        \includegraphics[width=\linewidth]{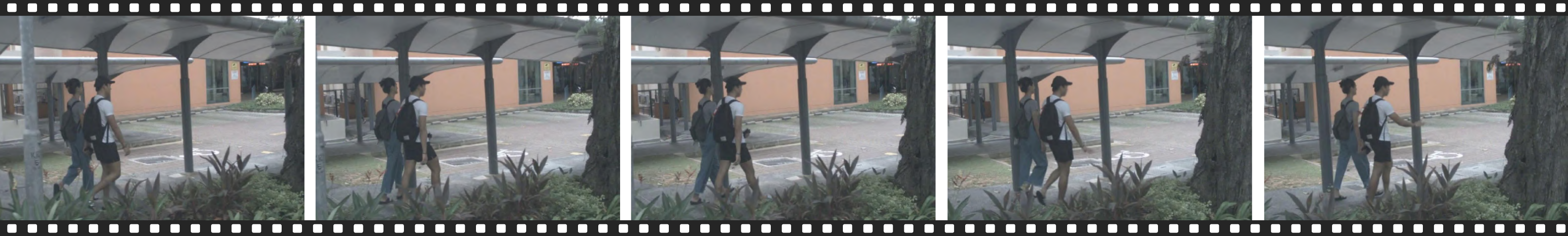}
        \caption{Good example in the \emph{Subject Fidelity} dimension (Score: \textcolor{w_blue}{$96.92\%$})}
        \label{fig:gen_subject_fidelity_3}
    \end{subfigure}
    \\[1ex]
    \begin{subfigure}[h]{\textwidth}
        \centering
        \includegraphics[width=\linewidth]{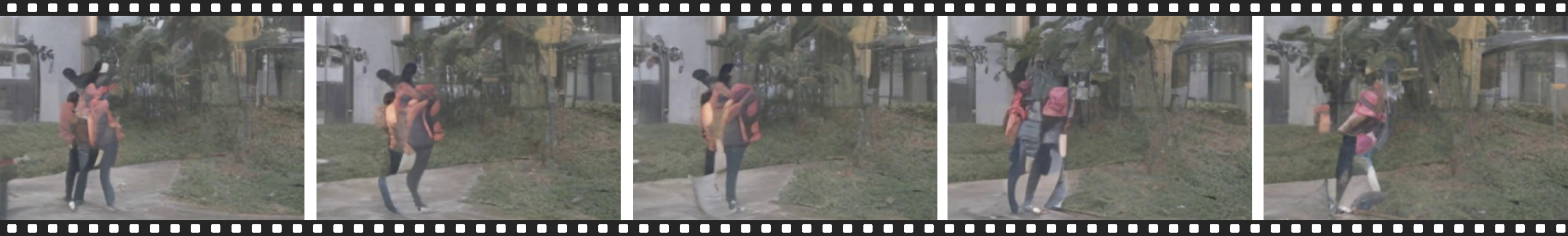}
        \caption{Bad example in the \emph{Subject Fidelity} dimension (Score: \textcolor{red}{$10.14\%$})}
        \label{fig:gen_subject_fidelity_4}
    \end{subfigure}
    \\[3.5ex]
    \begin{subfigure}[h]{\textwidth}
        \centering
        \includegraphics[width=\linewidth]{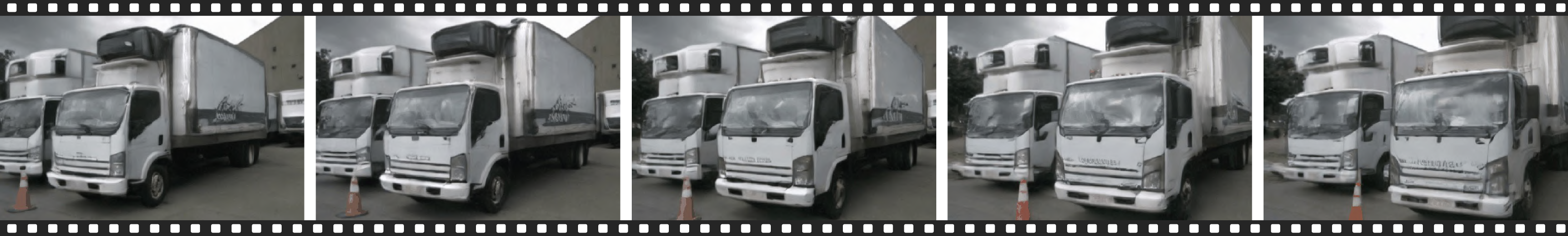}
        \caption{Good example in the \emph{Subject Fidelity} dimension (Score: \textcolor{w_blue}{$91.72\%$})}
        \label{fig:gen_subject_fidelity_5}
    \end{subfigure}
    \\[1ex]
    \begin{subfigure}[h]{\textwidth}
        \centering
        \includegraphics[width=\linewidth]{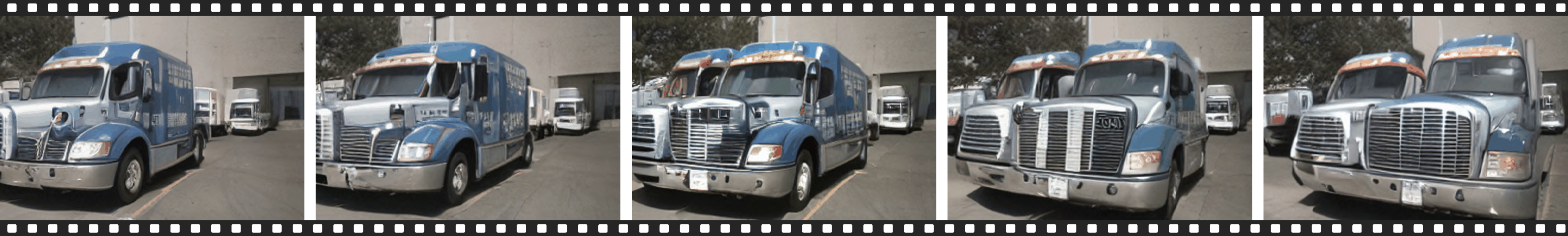}
        \caption{Bad example in the \emph{Subject Fidelity} dimension (Score: \textcolor{red}{$41.75\%$})}
        \label{fig:gen_subject_fidelity_6}
    \end{subfigure}
    \vspace{-0.1cm}
    \caption{Examples of ``good'' and ``bad'' generation qualities in terms of \emph{Subject Fidelity} in WorldLens.}
    \label{fig:gen_subject_fidelity}
\end{figure}

Figure~\ref{fig:gen_subject_fidelity} provides typical examples of videos with good and bad quality in terms of \emph{Subject Fidelity}.

\subsubsection{~Evaluation \& Analysis}
Table~\ref{tab:supp_gen_subject_fidelity} provides the complete results of models in terms of \emph{Subject Fidelity}.

\begin{table*}[h]
    \centering
    \caption{Complete results of state-of-the-art driving world models in terms of \emph{Subject Fidelity} in WorldLens.}
    \vspace{-0.2cm}
    \label{tab:supp_gen_subject_fidelity}
    \resizebox{\linewidth}{!}{
    \begin{tabular}{r|cccccc|c}
        \toprule
        \multirow{2}{*}{$\mathcal{S}_\mathrm{SF}(\cdot)$} & \textbf{MagicDrive} & \textbf{DreamForge}  & \textbf{DriveDreamer-2} & \textbf{OpenDWM} & \textbf{~DiST-4D~} & $\mathcal{X}$\textbf{-Scene} & \textcolor{gray}{\textbf{Empirical}}
        \\
        & \textcolor{gray}{\small[ICLR'24]} & \textcolor{gray}{\small[arXiv'24]} & \textcolor{gray}{\small[AAAI'25]} & \textcolor{gray}{\small[CVPR'25]} & \textcolor{gray}{\small[ICCV'25]} & \textcolor{gray}{\small[NeurIPS'25]} & \textcolor{gray}{\textbf{Max}}
        \\\midrule
        Vehicle ($\uparrow$) & $26.16\%$ & $26.97\%$ & $24.23\%$ & $34.04\%$ & $28.02\%$ & $24.03\%$ & \cellcolor{gray!7}\textcolor{gray}{$56.10\%$}
        \\
        Pedestrian ($\uparrow$) & $49.45\%$ & $77.13\%$ & $55.69\%$ & $56.65\%$ & $50.98\%$ & $55.42\%$ & \cellcolor{gray!7}\textcolor{gray}{$97.27\%$}
        \\
        \textbf{Total ($\uparrow$)} & \cellcolor{w_blue!20}$28.49\%$ &  \cellcolor{w_blue!20}$31.99\%$ & \cellcolor{w_blue!20}$27.38\%$ & \cellcolor{w_blue!20}$36.30\%$ & \cellcolor{w_blue!20}$30.32\%$ & \cellcolor{w_blue!20}$27.17\%$ & \cellcolor{gray!7}\textcolor{gray}{$60.22\%$}
        \\
        \bottomrule
    \end{tabular}}
    \vspace{-0.3cm}
\end{table*}

%% file: sections_supp/dimensions/gen_subject_coherence.tex
\subsection{~Subject Coherence}

\subsubsection{~Definition}
Subject Coherence evaluates the temporal stability of an object’s visual identity across consecutive frames within a generated sequence. It captures whether the same entity -- such as a specific car or pedestrian -- maintains consistent appearance attributes, including color, texture, and shape, over time. This metric assesses not only visual continuity but also the preservation of object identity, which is crucial for generating physically plausible and temporally coherent scenes for autonomous driving applications.

\subsubsection{~Formulation}
For each generated video $y_j=\{y_j^{(t)}\}_{t=1}^{T}$, 
the conditioning provides bounding boxes $\{b_{j,k}^{(t)}\}$ and associated track IDs $r_{j,k}$. 
Object patches are cropped as 
$o_{j,r}^{(t)}=\mathrm{Crop}(y_j^{(t)},b_{j,k}^{(t)})$ for object track $r=r_{j,k}$.
A frozen ReID encoder $\phi_{\mathrm{ReID}}(\cdot)$ extracts $\ell_2$-normalized embeddings:
\[
\mathbf{g}_{j,r}^{(t)}=\phi_{\mathrm{ReID}}\!\left(o_{j,r}^{(t)}\right),
\qquad \|\mathbf{g}_{j,r}^{(t)}\|_2=1.
\]

The dataset-level Subject Coherence is computed as the mean cosine similarity between consecutive embeddings of the same tracked object, aggregated over all tracks, frames, and videos:
\begin{equation}
\boxed{\;
\mathcal{S}_{\mathrm{SC}}(\mathcal{Y})
=\tfrac{1}{N_g}
 \sum\nolimits_{j=1}^{N_g}
 \tfrac{1}{R_j}
 \sum\nolimits_{r=1}^{R_j}
 \tfrac{1}{T_r-1}
 \sum\nolimits_{t=1}^{T_r-1}
 \mathbf{g}_{j,r}^{(t)\top}\mathbf{g}_{j,r}^{(t+1)}
\;}
\label{eq:subject_coherence}
\end{equation}
where $R_j$ is the number of track IDs in video $y_j$ and $T_r$ the number of frames where object $r$ appears. A high $\mathcal{S}_{\mathrm{SC}}$ score reflects consistent and temporally stable object generation, indicating that the model preserves identity-related features despite changes in position, viewpoint, or lighting. In contrast, a low score often signals flickering textures, shape distortions, or identity switches between frames. 

This metric thus serves as a sensitive indicator of temporal realism, distinguishing models that produce temporally coherent scenes from those limited to frame-wise synthesis.

\subsubsection{~Implementation Details}
We compute \textit{Subject Coherence} using embeddings extracted from the Cross-Video ReID model of Zuo et al.~\cite{zuo2024cross}. Frames are filtered using confidence thresholds of $0.25$ for vehicles and $0.50$ for pedestrians before similarity computation. The final score is a combination of both sub-metrics.

\subsubsection{~Examples}
\begin{figure}[t]
    \centering
    \begin{subfigure}[h]{\textwidth}
        \centering
        \includegraphics[width=\linewidth]{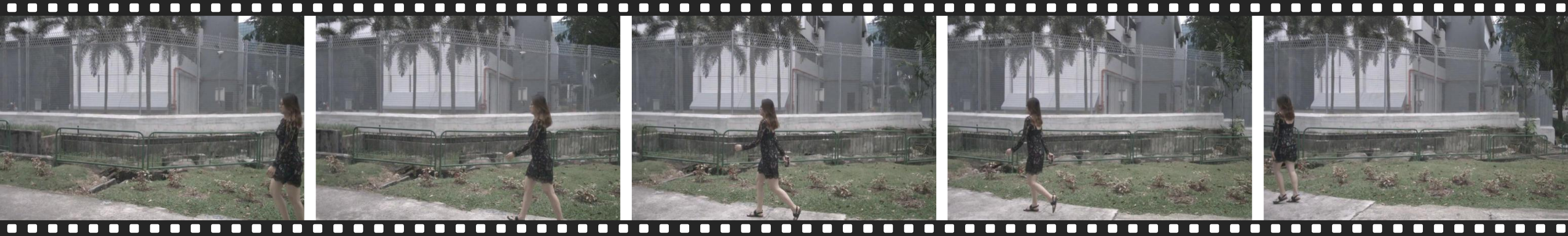}
        \caption{Good example in the \emph{Subject Coherence} dimension (Score: \textcolor{w_blue}{$95.19\%$})}
        \label{fig:gen_subject_coherence_1}
    \end{subfigure}
    \\[1ex]
    \begin{subfigure}[h]{\textwidth}
        \centering
        \includegraphics[width=\linewidth]{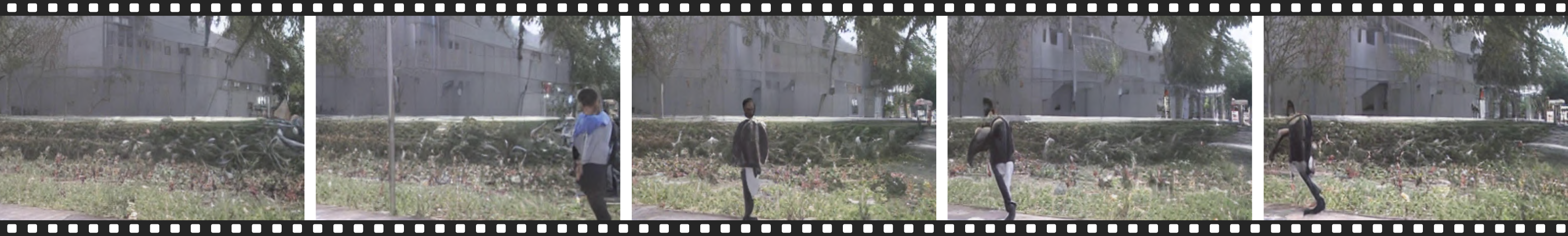}
        \caption{Bad example in the \emph{Subject Coherence} dimension (Score: \textcolor{red}{$54.69\%$})}
        \label{fig:gen_subject_coherence_2}
    \end{subfigure}
    \\[3.5ex]
    \begin{subfigure}[h]{\textwidth}
        \centering
        \includegraphics[width=\linewidth]{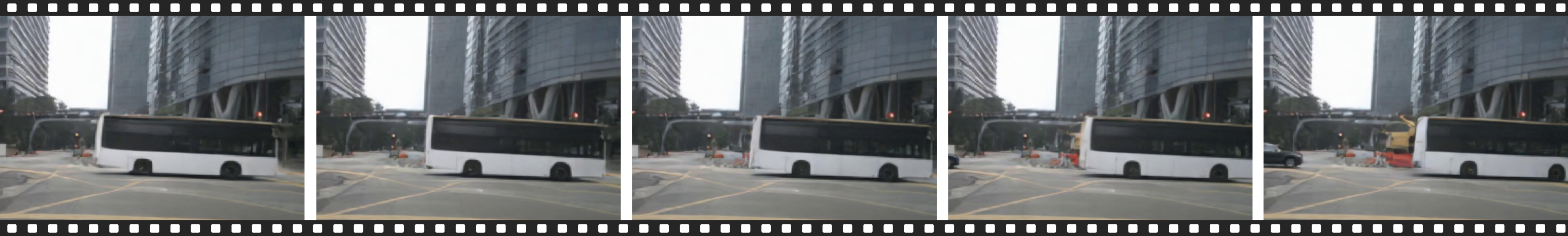}
        \caption{Good example in the \emph{Subject Coherence} dimension (Score: \textcolor{w_blue}{$93.22\%$})}
        \label{fig:gen_subject_coherence_3}
    \end{subfigure}
    \\[1ex]
    \begin{subfigure}[h]{\textwidth}
        \centering
        \includegraphics[width=\linewidth]{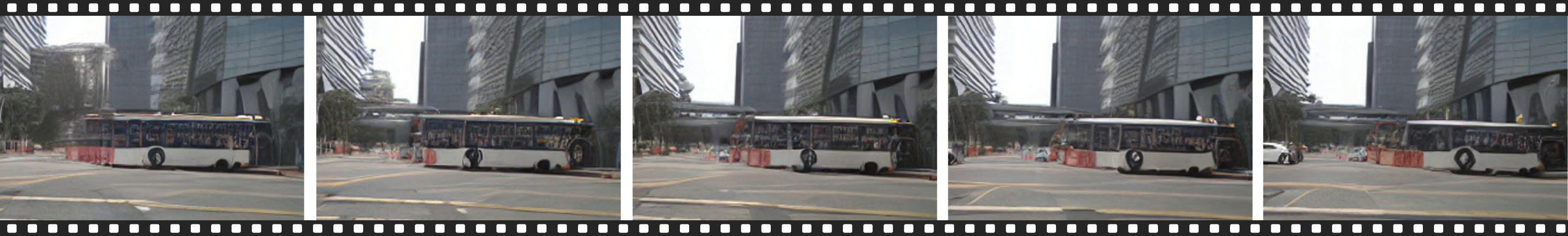}
        \caption{Bad example in the \emph{Subject Coherence} dimension (Score: \textcolor{red}{$65.36\%$})}
        \label{fig:gen_subject_coherence_4}
    \end{subfigure}
    \\[3.5ex]
    \begin{subfigure}[h]{\textwidth}
        \centering
        \includegraphics[width=\linewidth]{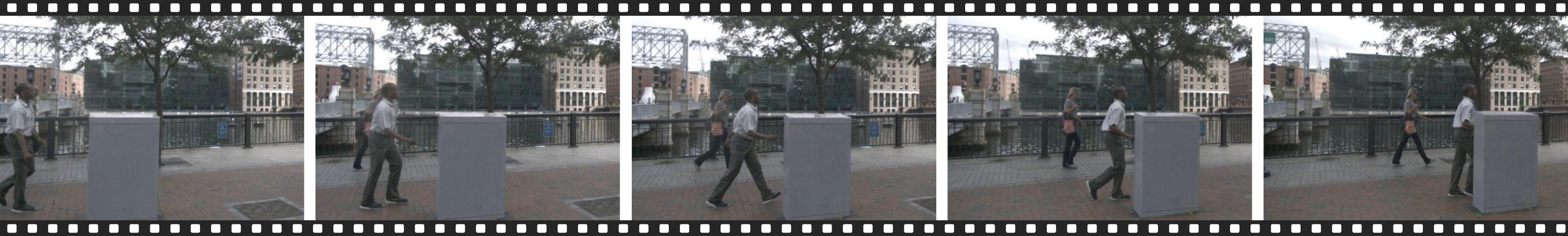}
        \caption{Good example in the \emph{Subject Coherence} dimension (Score: \textcolor{w_blue}{$91.53\%$})}
        \label{fig:gen_subject_coherence_5}
    \end{subfigure}
    \\[1ex]
    \begin{subfigure}[h]{\textwidth}
        \centering
        \includegraphics[width=\linewidth]{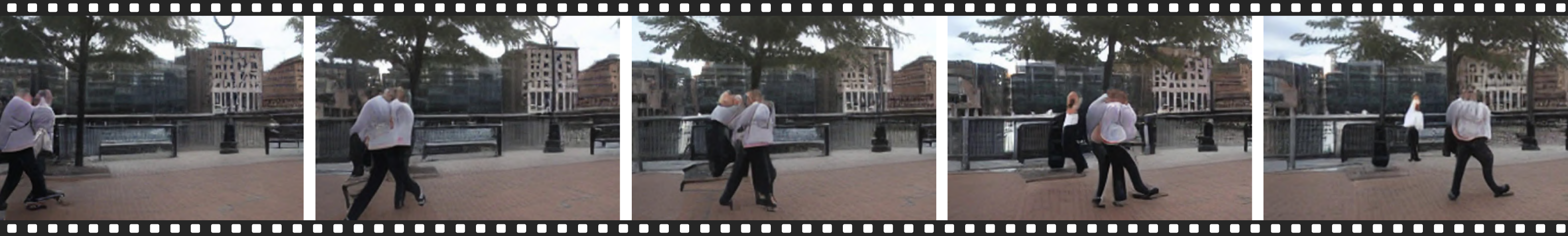}
        \caption{Bad example in the \emph{Subject Coherence} dimension (Score: \textcolor{red}{$66.79\%$})}
        \label{fig:gen_subject_coherence_6}
    \end{subfigure}
    \vspace{-0.1cm} 
    \caption{Examples of ``good'' and ``bad'' generation qualities in terms of \emph{Subject Coherence} in WorldLens.}
    \label{fig:gen_subject_coherence}
\end{figure}

Figure~\ref{fig:gen_subject_coherence} provides typical examples of videos with good and bad quality in terms of \emph{Subject Coherence}.

\subsubsection{~Evaluation \& Analysis}
Table~\ref{tab:supp_gen_subject_coherence} provides the complete results of models in terms of \emph{Subject Coherence}.

\begin{table*}[h]
    \centering
    \vspace{0.3cm}
    \caption{Complete results of state-of-the-art driving world models in terms of \emph{Subject Coherence} in WorldLens.}
    \vspace{-0.2cm}
    \label{tab:supp_gen_subject_coherence}
    \resizebox{\linewidth}{!}{
    \begin{tabular}{r|cccccc|c}
        \toprule
        \multirow{2}{*}{$\mathcal{S}_\mathrm{SC}(\cdot)$} & \textbf{MagicDrive} & \textbf{DreamForge}  & \textbf{DriveDreamer-2} & \textbf{OpenDWM} & \textbf{~DiST-4D~} & $\mathcal{X}$\textbf{-Scene} & \textcolor{gray}{\textbf{Empirical}}
        \\
        & \textcolor{gray}{\small[ICLR'24]} & \textcolor{gray}{\small[arXiv'24]} & \textcolor{gray}{\small[AAAI'25]} & \textcolor{gray}{\small[CVPR'25]} & \textcolor{gray}{\small[ICCV'25]} & \textcolor{gray}{\small[NeurIPS'25]} & \textcolor{gray}{\textbf{Max}}
        \\\midrule
        Vehicle ($\uparrow$) & $72.12\%$ & $72.00\%$ & $77.45\%$ & $82.03\%$ & $78.51\%$ & $74.02\%$ & \textcolor{gray}{$82.86\%$}
        \\
        Pedestrian ($\uparrow$) & $79.78\%$ & $78.23\%$ & $80.48\%$ & $84.22\%$ & $80.20\%$ & $80.42\%$ & \textcolor{gray}{$83.25\%$}
        \\
        \textbf{Total ($\uparrow$)} & \cellcolor{w_blue!20}$75.95\%$ & \cellcolor{w_blue!20}$75.12\%$ & \cellcolor{w_blue!20}$78.97\%$ & \cellcolor{w_blue!20}$83.13\%$ & \cellcolor{w_blue!20}$79.36\%$ & \cellcolor{w_blue!20}$77.22\%$ & \cellcolor{gray!7}\textcolor{gray}{$83.25\%$}
        \\
        \bottomrule
    \end{tabular}}
    \vspace{-0.3cm}
\end{table*}

%% file: sections_supp/dimensions/gen_subject_consistency.tex
\subsection{~Subject Consistency}
\label{subsec:subject_consistency}

\subsubsection{~Definition}
Subject Consistency measures stability of object-level semantics and structural details. It focuses on fine-grained appearance and geometric regularity through DINO features~\cite{caron2021dino}, evaluating whether dynamic subjects maintain consistent texture, shape, and structure over time. High scores indicate that the model preserves the semantic identity and visual integrity of objects throughout motion, avoiding flickering or deformation.

\subsubsection{Formulation}
\vspace{-0.1cm}
For each generated video $y_j=\{y_j^{(t)}\}_{t=1}^{T}$ and its paired ground-truth $x_j=\{x_j^{(t)}\}_{t=1}^{T}$, 
we extract $\ell_2$-normalized DINO embeddings:
$\mathbf{g}_j^{(t)}=\phi_{\mathrm{DINO}}\!(y_j^{(t)})$, $\mathbf{f}_j^{(t)}=\phi_{\mathrm{DINO}}\!(x_j^{(t)})$, and $\|\mathbf{g}_j^{(t)}\|_2=\|\mathbf{f}_j^{(t)}\|_2=1$, where $\phi_{\mathrm{DINO}}(\cdot)$ denotes the frozen DINO feature extractor. To quantify temporal stability, we compute three complementary terms:

\begin{itemize}
    \item \textbf{Adjacent-Frame Smoothness:}
    \vspace{-0.3cm}
    \[
    \mathrm{ACM}(y_j)=\tfrac{1}{T-1}\sum\nolimits_{t=1}^{T-1}\mathbf{g}_j^{(t)\top}\mathbf{g}_j^{(t+1)}~,
    \]
    which measures the average cosine similarity between consecutive frame embeddings.

    \item \textbf{Temporal Jitter Index (TJI):}
    \vspace{-0.3cm}
    \[
    \mathrm{TJI}(y_j)=\tfrac{1}{T-2}\sum\nolimits_{t=2}^{T-1}\frac{\|\mathbf{g}_j^{(t+1)}-2\mathbf{g}_j^{(t)}+\mathbf{g}_j^{(t-1)}\|_2}{\tfrac{1}{2}\bigl(\|\mathbf{g}_j^{(t+1)}-\mathbf{g}_j^{(t)}\|_2+\|\mathbf{g}_j^{(t)}-\mathbf{g}_j^{(t-1)}\|_2\bigr)+\varepsilon}~,
    \]
    which measures normalized second-order fluctuations (lower is smoother).

    \item \textbf{Motion-Rate Similarity (MRS):}
    \vspace{-0.3cm}
    \[
    \mathrm{MRS}(y_j,x_j)=\exp\!\Big(-\beta\,\tfrac{1}{T-1}\sum\nolimits_{t=1}^{T-1}\big|\log\tfrac{\|\mathbf{g}_j^{(t+1)}-\mathbf{g}_j^{(t)}\|_2+\varepsilon}{\|\mathbf{f}_j^{(t+1)}-\mathbf{f}_j^{(t)}\|_2+\varepsilon}\big|\Big)~,
    \]
    which aligns the per-frame feature motion magnitude with that of the ground-truth sequence.
\end{itemize}

The overall Subject Consistency score integrates these terms:
\begin{equation}
\boxed{\;
\mathcal{S}_{\mathrm{SC}}(\mathcal{Y})
=\tfrac{1}{N_g}\sum\nolimits_{j=1}^{N_g}
\frac{\mathrm{ACM}(y_j)}
{1+\mathrm{TJI}(y_j)}
\cdot \mathrm{MRS}(y_j,x_j)^{1/2}
\;}
\label{eq:subject_consistency}
\end{equation}

\subsubsection{~Implementation Details}
We extract frame-wise features using DINO ViT-B/16~\cite{caron2021dino}. These embeddings are used to compute adjacent-frame similarity, temporal jitter, and motion alignment against the corresponding ground-truth videos.

\subsubsection{~Examples}
\begin{figure}[t]
    \centering
    \begin{subfigure}[h]{\textwidth}
        \centering
        \includegraphics[width=\linewidth]{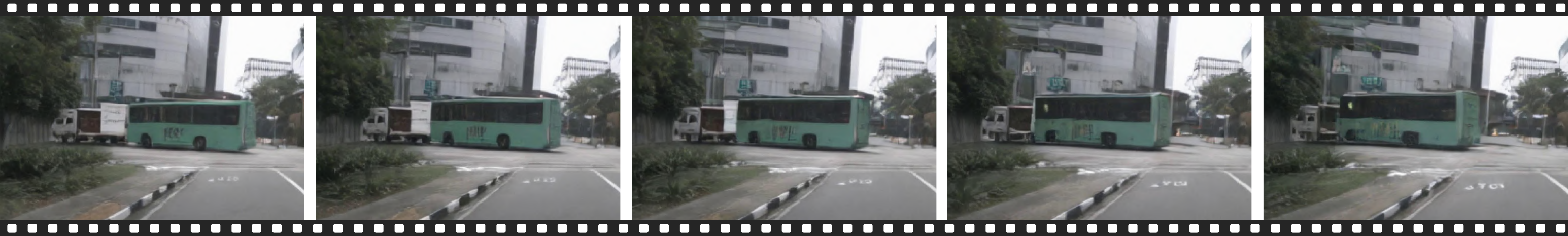}
        \caption{Good example in the \emph{Subject Consistency} dimension (Score: \textcolor{w_blue}{$86.23\%$})}
        \label{fig:gen_subject_consistency_1}
    \end{subfigure}
    \\[1ex]
    \begin{subfigure}[h]{\textwidth}
        \centering
        \includegraphics[width=\linewidth]{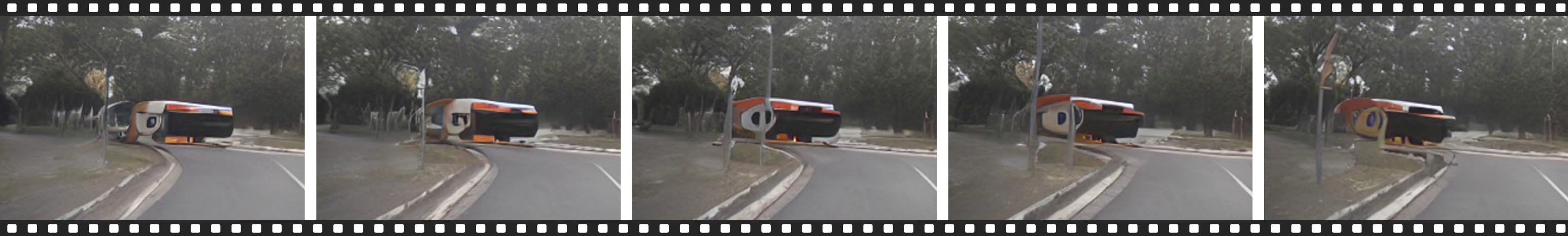}
        \caption{Bad example in the \emph{Subject Consistency} dimension (Score: \textcolor{red}{$42.75\%$})}
        \label{fig:gen_subject_consistency_2}
    \end{subfigure}
    \\[3.5ex]
    \begin{subfigure}[h]{\textwidth}
        \centering
        \includegraphics[width=\linewidth]{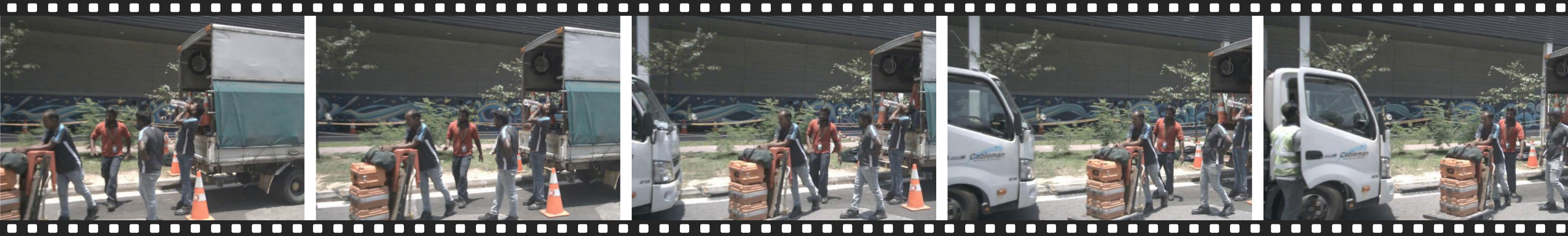}
        \caption{Good example in the \emph{Subject Consistency} dimension (Score: \textcolor{w_blue}{$84.82\%$})}
        \label{fig:gen_subject_consistency_3}
    \end{subfigure}
    \\[1ex]
    \begin{subfigure}[h]{\textwidth}
        \centering
        \includegraphics[width=\linewidth]{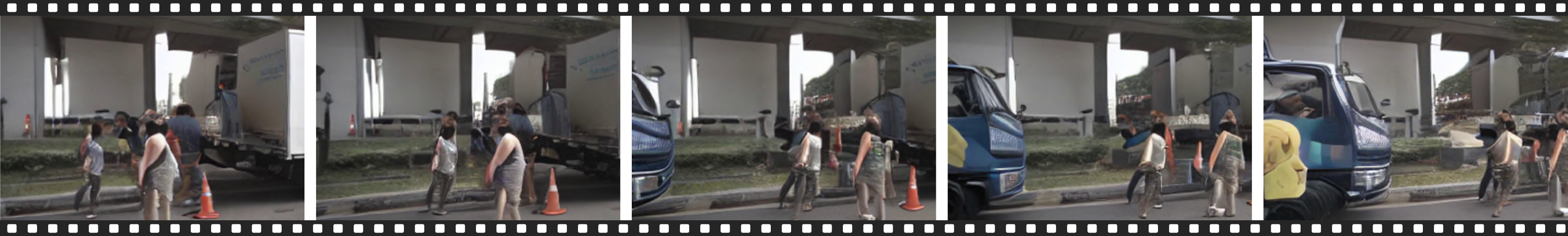}
        \caption{Bad example in the \emph{Subject Consistency} dimension (Score: \textcolor{red}{$43.68\%$})}
        \label{fig:gen_subject_consistency_4}
    \end{subfigure}
    \\[3.5ex]
    \begin{subfigure}[h]{\textwidth}
        \centering
        \includegraphics[width=\linewidth]{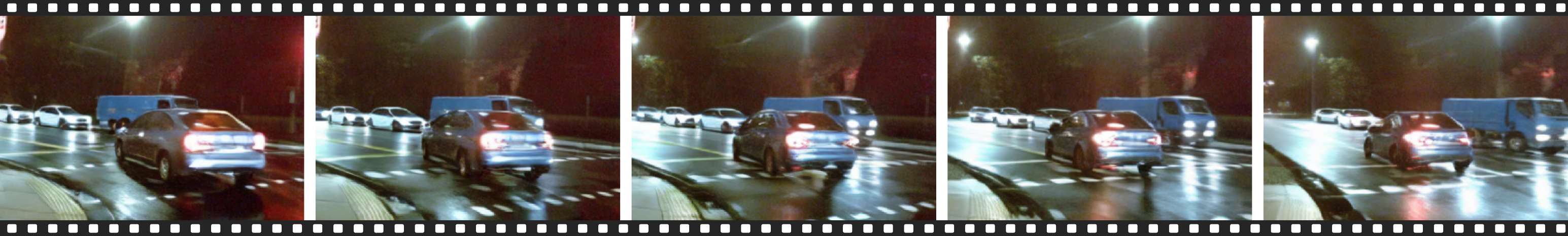}
        \caption{Good example in the \emph{Subject Consistency} dimension (Score: \textcolor{w_blue}{$83.96\%$})}
        \label{fig:gen_subject_consistency_5}
    \end{subfigure}
    \\[1ex]
    \begin{subfigure}[h]{\textwidth}
        \centering
        \includegraphics[width=\linewidth]{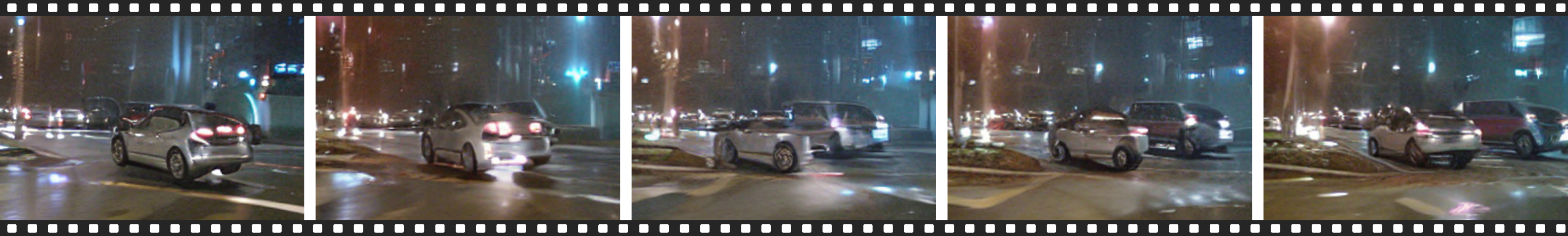}
        \caption{Bad example in the \emph{Subject Consistency} dimension (Score: \textcolor{red}{$42.53\%$})}
        \label{fig:gen_subject_consistency_6}
    \end{subfigure}
    \vspace{-0.1cm}
    \caption{Examples of ``good'' and ``bad'' generation qualities in terms of \emph{Subject Consistency} in WorldLens.}
    \label{fig:gen_subject_consistency}
\end{figure}

Figure~\ref{fig:gen_subject_consistency} provides typical examples of videos with good and bad quality in terms of \emph{Subject Consistency}.

\subsubsection{~Evaluation \& Analysis}
Table~\ref{tab:supp_gen_subject_consistency} provides the complete results of models in terms of \emph{Subject Consistency}.

\begin{table*}[h]
    \centering
    \vspace{0.2cm}
    \caption{Complete results of state-of-the-art driving world models in terms of \emph{Subject Consistency} in WorldLens.}
    \vspace{-0.2cm}
    \label{tab:supp_gen_subject_consistency}
    \resizebox{\linewidth}{!}{
    \begin{tabular}{r|cccccc|c}
        \toprule
        \multirow{2}{*}{$\mathcal{S}_\mathrm{SC}(\cdot)$} & \textbf{MagicDrive} & \textbf{DreamForge}  & \textbf{DriveDreamer-2} & \textbf{OpenDWM} & \textbf{~DiST-4D~} & $\mathcal{X}$\textbf{-Scene} & \textcolor{gray}{\textbf{Empirical}}
        \\
        & \textcolor{gray}{\small[ICLR'24]} & \textcolor{gray}{\small[arXiv'24]} & \textcolor{gray}{\small[AAAI'25]} & \textcolor{gray}{\small[CVPR'25]} & \textcolor{gray}{\small[ICCV'25]} & \textcolor{gray}{\small[NeurIPS'25]} & \textcolor{gray}{\textbf{Max}}
        \\\midrule
        ~ACM ($\uparrow$) & $89.32\%$ & $91.72\%$ & $90.09\%$ & $92.21\%$ & $91.15\%$ & $90.72\%$ & \cellcolor{gray!7}\textcolor{gray}{$93.66\%$}
        \\
        TJI ($\uparrow$) & $44.12\%$ & $43.32\%$ & $45.37\%$ & $44.95\%$ & $45.79\%$ & $43.41\%$ & \cellcolor{gray!7}\textcolor{gray}{$45.94\%$}
        \\
        \textbf{Total ($\uparrow$)} & \cellcolor{w_blue!20}$65.22\%$ & \cellcolor{w_blue!20}$76.40\%$ & \cellcolor{w_blue!20}$74.49\%$ & \cellcolor{w_blue!20}$78.33\%$ & \cellcolor{w_blue!20}$74.69\%$ & \cellcolor{w_blue!20}$74.37\%$ & \cellcolor{gray!7}\textcolor{gray}{$93.66\%$}
        \\
        \bottomrule
    \end{tabular}}
    \vspace{-0.4cm}
\end{table*}

%% file: sections_supp/dimensions/gen_depth_discrepancy.tex
\subsection{~Depth Discrepancy}

\subsubsection{~Definition}
Depth Discrepancy quantifies the temporal stability of depth representations inferred from generated video sequences. In natural driving scenes, the apparent depth of foreground and background objects evolves smoothly with camera motion, whereas inconsistent generation often introduces discontinuous jumps in predicted depth. 
This metric captures such instability by measuring temporal variation in depth embeddings extracted from consecutive frames, providing a geometric complement to perceptual fidelity metrics.

\subsubsection{~Formulation}
For a generated video $y_j=\{y_j^{(t)}\}_{t=1}^{T}$, we estimate per-frame depth maps using a monocular depth estimator $\psi_\mathrm{Depth}(\cdot)$:
\[
d_j^{(t)}=\psi_\mathrm{Depth}\!\left(y_j^{(t)}\right), \qquad 
d_j^{(t)}\in\mathbb{R}^{H\times W}.
\]
Each depth map is RGB-encoded by a fixed colormap $\mathcal{C}$ and processed by a pretrained visual encoder $\phi_\mathrm{DINO}(\cdot)$ to obtain global embeddings:
\[
f_j^{(t)}=\phi_\mathrm{DINO}\!\left(\mathcal{C}(d_j^{(t)})\right),
\qquad 
f_j^{(t)}\in\mathbb{R}^{D}.
\]
Temporal variation in depth representation is then measured by the mean L2 distance between consecutive embeddings:
\[
\mathrm{DD}(y_j)
=\tfrac{1}{T-1}\sum\nolimits_{t=1}^{T-1}\|f_j^{(t)}-f_j^{(t+1)}\|_2~.
\]
Finally, the dataset-level Depth Discrepancy can be calculated as follows:
\begin{equation}
\boxed{\;
\mathcal{S}_{\mathrm{DD}}(\mathcal{Y})
=\tfrac{1}{N_g}\sum\nolimits_{j=1}^{N_g}\mathrm{DD}(y_j)
\;}
\label{eq:depth_discrepancy}
\end{equation}
Lower $\mathcal{S}_{\mathrm{Depth}}$ indicates smoother, more physically consistent depth evolution across time, reflecting stronger temporal geometric stability in the generated videos.

\subsubsection{~Implementation Details}
Depth maps for both generated and ground-truth videos are obtained using Video DepthAnything~\cite{chen2025video}. The predicted depths are directly used to compute the per-frame depth discrepancy.

\subsubsection{~Examples}
\begin{figure}[t]
    \centering
    \begin{subfigure}[h]{\textwidth}
        \centering
        \includegraphics[width=\linewidth]{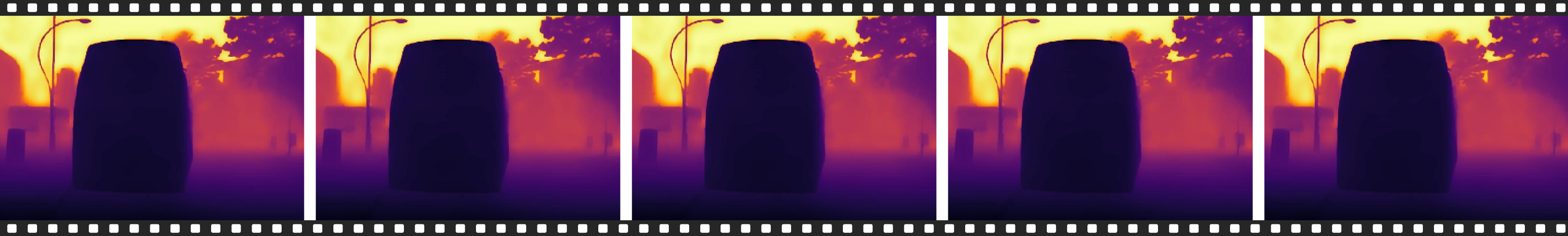}
        \caption{Good example in the \emph{Depth Discrepancy} dimension (Score: \textcolor{w_blue}{$4.43$})}
        \label{fig:gen_depth_discrepancy_1}
    \end{subfigure}
    \\[1ex]
    \begin{subfigure}[h]{\textwidth}
        \centering
        \includegraphics[width=\linewidth]{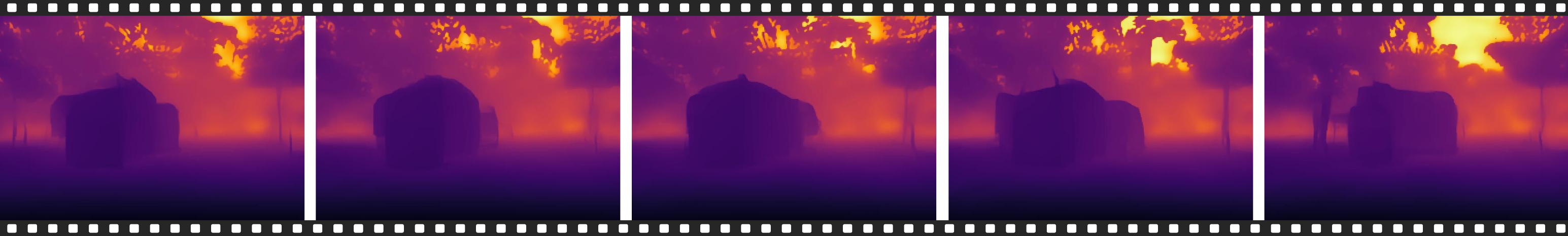}
        \caption{Bad example in the \emph{Depth Discrepancy} dimension (Score: \textcolor{red}{$29.47$})}
        \label{fig:gen_depth_discrepancy_2}
    \end{subfigure}
    \\[3.5ex]
    \begin{subfigure}[h]{\textwidth}
        \centering
        \includegraphics[width=\linewidth]{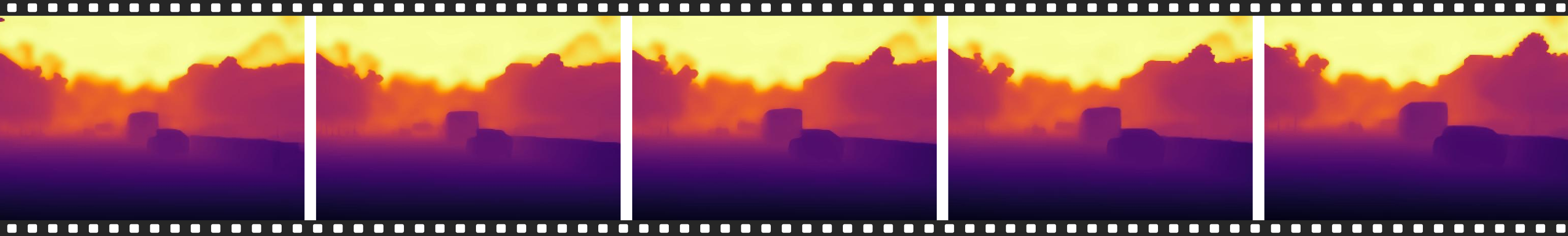}
        \caption{Good example in the \emph{Depth Discrepancy} dimension (Score: \textcolor{w_blue}{$6.23$})}
        \label{fig:gen_depth_discrepancy_3}
    \end{subfigure}
    \\[1ex]
    \begin{subfigure}[h]{\textwidth}
        \centering
        \includegraphics[width=\linewidth]{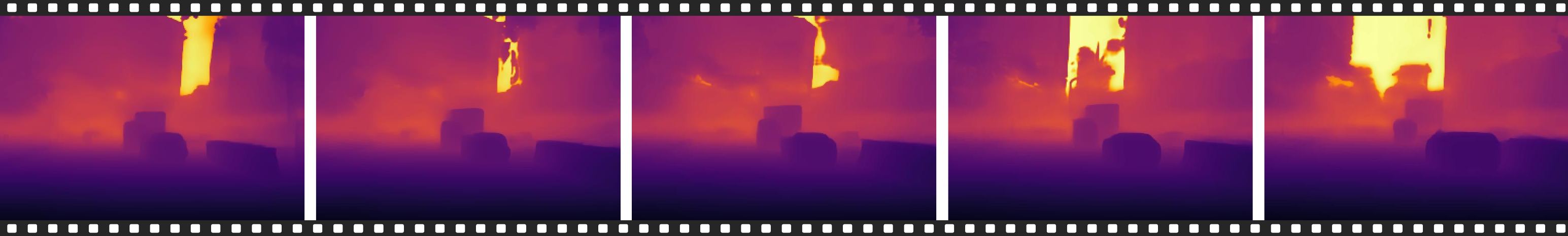}
        \caption{Bad example in the \emph{Depth Discrepancy} dimension (Score: \textcolor{red}{$33.65$})}
        \label{fig:gen_depth_discrepancy_4}
    \end{subfigure}
    \\[3.5ex]
    \begin{subfigure}[h]{\textwidth}
        \centering
        \includegraphics[width=\linewidth]{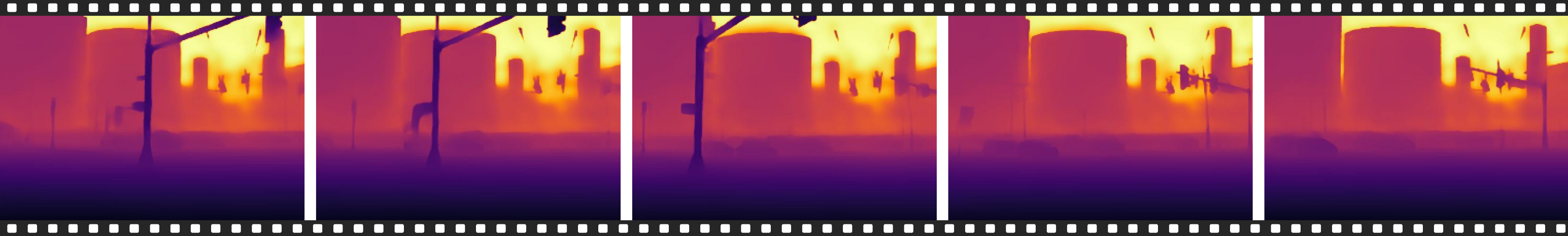}
        \caption{Good example in the \emph{Depth Discrepancy} dimension (Score: \textcolor{w_blue}{$8.67$})}
        \label{fig:gen_depth_discrepancy_5}
    \end{subfigure}
    \\[1ex]
    \begin{subfigure}[h]{\textwidth}
        \centering
        \includegraphics[width=\linewidth]{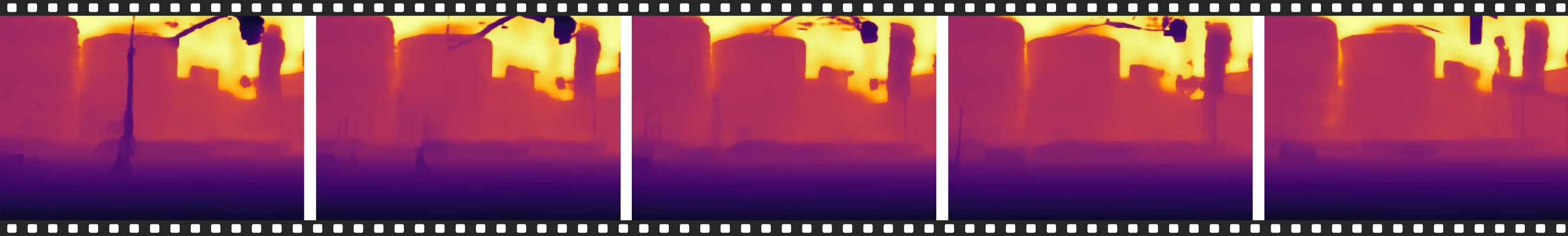}
        \caption{Bad example in the \emph{Depth Discrepancy} dimension (Score: \textcolor{red}{$19.34$})}
        \label{fig:gen_depth_discrepancy_6}
    \end{subfigure}
    \vspace{-0.1cm}
    \caption{Examples of ``good'' and ``bad'' generation qualities in terms of \emph{Depth Discrepancy} in WorldLens.}
    \label{fig:gen_depth_discrepancy}
\end{figure}

Figure~\ref{fig:gen_depth_discrepancy} provides typical examples of videos with good and bad quality in terms of \emph{Depth Discrepancy}.

\subsubsection{~Evaluation \& Analysis}
Table~\ref{tab:supp_gen_depth_discrepancy} provides the complete results of models in terms of \emph{Depth Discrepancy}.

\begin{table*}[h]
    \centering
    \vspace{0.3cm}
    \caption{Complete comparisons of state-of-the-art driving world models in terms of \emph{Depth Discrepancy} in WorldLens.}
    \vspace{-0.2cm}
    \label{tab:supp_gen_depth_discrepancy}
    \resizebox{\linewidth}{!}{
    \begin{tabular}{r|cccccc|c}
        \toprule
        \multirow{2}{*}{$\mathcal{S}_\mathrm{DD}(\cdot)$} & \textbf{MagicDrive} & \textbf{DreamForge}  & \textbf{DriveDreamer-2} & \textbf{OpenDWM} & \textbf{~DiST-4D~} & $\mathcal{X}$\textbf{-Scene} & \textcolor{gray}{\textbf{Empirical}}
        \\
        & \textcolor{gray}{\small[ICLR'24]} & \textcolor{gray}{\small[arXiv'24]} & \textcolor{gray}{\small[AAAI'25]} & \textcolor{gray}{\small[CVPR'25]} & \textcolor{gray}{\small[ICCV'25]} & \textcolor{gray}{\small[NeurIPS'25]} & \textcolor{gray}{\textbf{Max}}
        \\\midrule
        \textbf{Total ($\downarrow$)} & \cellcolor{w_blue!20}$24.19$ & \cellcolor{w_blue!20}$19.27$ & \cellcolor{w_blue!20}$17.73$ & \cellcolor{w_blue!20}$18.17$ & \cellcolor{w_blue!20}$17.71$ & \cellcolor{w_blue!20}$20.50$ & \cellcolor{gray!7}\textcolor{gray}{$14.27$}
        \\
        \bottomrule
    \end{tabular}}
    \vspace{-0.3cm}
\end{table*}

%% file: sections_supp/dimensions/gen_temporal_consistency.tex
\subsection{~Temporal Consistency}
\subsubsection{~Definition}
Temporal Consistency quantifies the frame-to-frame stability of generated videos in a learned appearance space. Using a frozen CLIP encoder, this metric captures whether visual representations evolve smoothly over time without abrupt changes or flickering. It measures: (1) adjacent-frame smoothness, (2) suppression of high-frequency temporal jitter, and (3) alignment of motion magnitudes with real sequences. Together, these components evaluate whether generated videos exhibit physically coherent and temporally realistic dynamics.

\subsubsection{~Formulation}
For each generated video $y_j=\{y_j^{(t)}\}_{t=1}^{T}$ and its paired ground-truth $x_j=\{x_j^{(t)}\}_{t=1}^{T}$, we extract $\ell_2$-normalized CLIP embeddings as follows:
\[
\mathbf{g}_j^{(t)}=\phi_{\mathrm{CLIP}}\!\left(y_j^{(t)}\right),\qquad
\mathbf{f}_j^{(t)}=\phi_{\mathrm{CLIP}}\!\left(x_j^{(t)}\right),\qquad
\|\mathbf{g}_j^{(t)}\|_2=\|\mathbf{f}_j^{(t)}\|_2=1.
\]

Follow the temporal statistics calculations in \emph{Subject Consistency}~(Eq.~\ref{subsec:subject_consistency}), the adjacent-frame smoothness, jitter suppression, and motion-rate alignment are applied in this CLIP space.
Combining these components, the per-video score is defined as:
\[
\mathrm{TC}(y_j)=
\tfrac{\mathrm{ACM}(y_j)}{1+\mathrm{TJI}(y_j)}\,
\mathrm{MRS}(y_j,x_j)^{1/2}.
\]
The dataset-level metric averages per-video scores:
\begin{equation}
\boxed{\;
\mathcal{S}_{\mathrm{TC}}(\mathcal{Y})
=\tfrac{1}{N_g}\sum\nolimits_{j=1}^{N_g}\mathrm{TC}(y_j)
\;}
\label{eq:temporal_consistency}
\end{equation}
with $\varepsilon\!=\!10^{-8}$ and $\beta\!=\!0.5$. 
By construction, $\mathrm{ACM}\!\in\![0,1]$ and $\mathrm{TJI}\!\ge\!0$.

A high $\mathcal{S}_{\mathrm{TC}}$ score indicates that appearance features change gradually across frames, producing smooth motion and physically coherent dynamics. Low scores correspond to flickering, abrupt illumination shifts, or motion discontinuities. This metric captures the degree to which generated sequences maintain continuity in both content and motion, serving as a robust proxy for temporal realism in driving videos.

\subsubsection{~Implementation Details}
\emph{Temporal Consistency} is evaluated using frame-wise features from CLIP ViT-B/32~\cite{radford2021learning} with an input resolution of $224\times224$. The normalized embeddings are used to derive adjacent-frame similarity, a temporal jitter index, and motion alignment between generated and ground-truth videos.

\subsubsection{~Examples}
\begin{figure}[t]
    \centering
    \begin{subfigure}[h]{\textwidth}
        \centering
        \includegraphics[width=\linewidth]{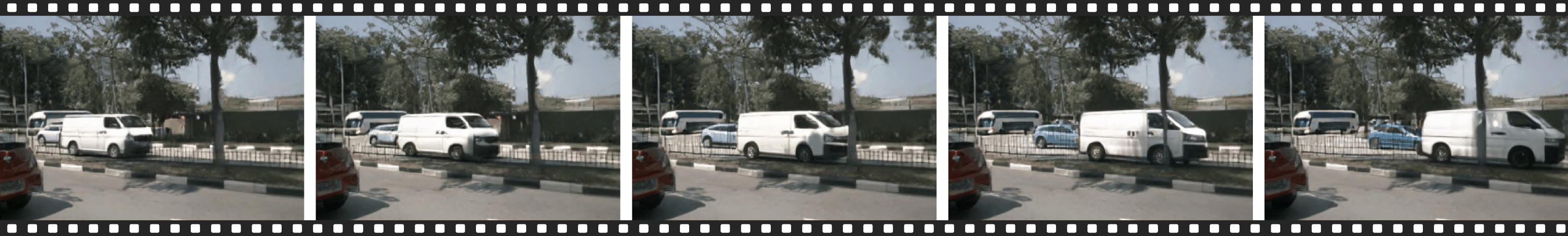}
        \caption{Good example in the \emph{Temporal Consistency} dimension (Score: \textcolor{w_blue}{$87.09\%$})}
        \label{fig:gen_temporal_consistency_1}
    \end{subfigure}
    \\[1ex]
    \begin{subfigure}[h]{\textwidth}
        \centering
        \includegraphics[width=\linewidth]{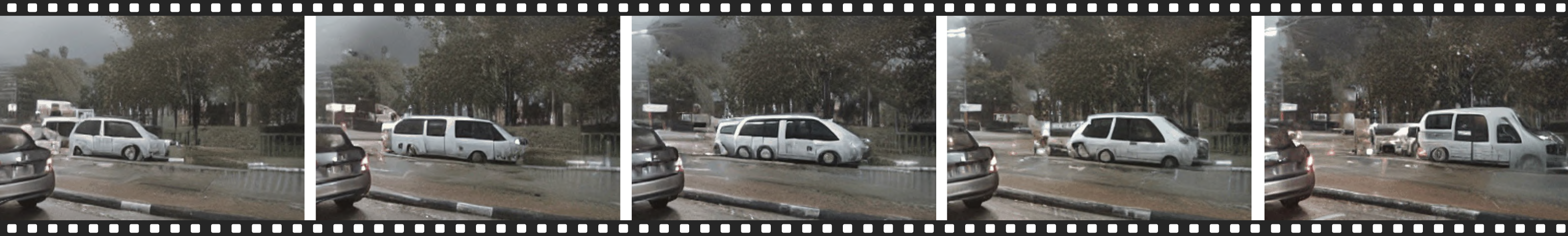}
        \caption{Bad example in the \emph{Temporal Consistency} dimension (Score: \textcolor{red}{$61.31\%$})}
        \label{fig:gen_temporal_consistency_2}
    \end{subfigure}
    \\[3.5ex]
    \begin{subfigure}[h]{\textwidth}
        \centering
        \includegraphics[width=\linewidth]{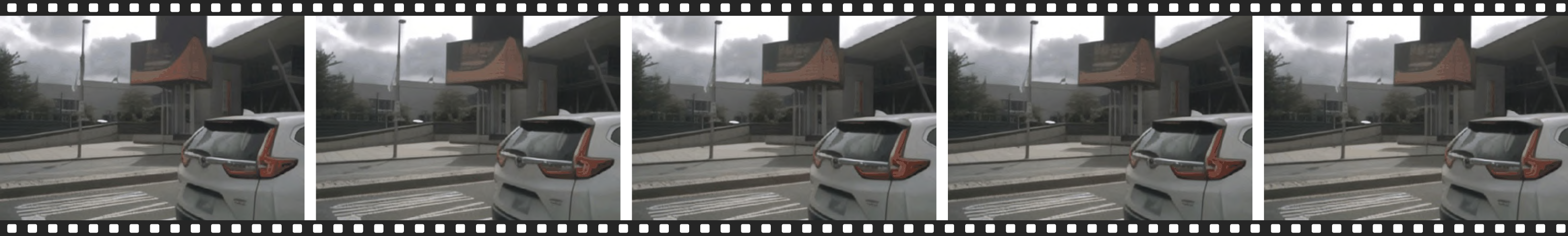}
        \caption{Good example in the \emph{Temporal Consistency} dimension (Score: \textcolor{w_blue}{$88.12\%$})}
        \label{fig:gen_temporal_consistency_3}
    \end{subfigure}
    \\[1ex]
    \begin{subfigure}[h]{\textwidth}
        \centering
        \includegraphics[width=\linewidth]{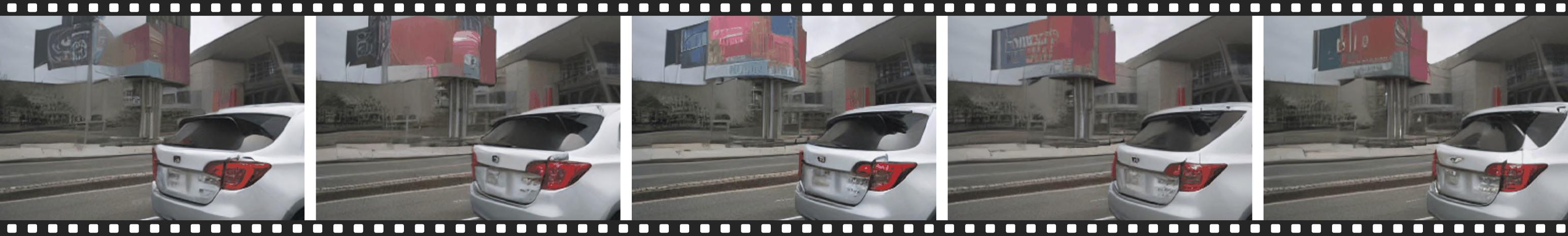}
        \caption{Bad example in the \emph{Temporal Consistency} dimension (Score: \textcolor{red}{$59.37\%$})}
        \label{fig:gen_temporal_consistency_4}
    \end{subfigure}
    \\[3.5ex]
    \begin{subfigure}[h]{\textwidth}
        \centering
        \includegraphics[width=\linewidth]{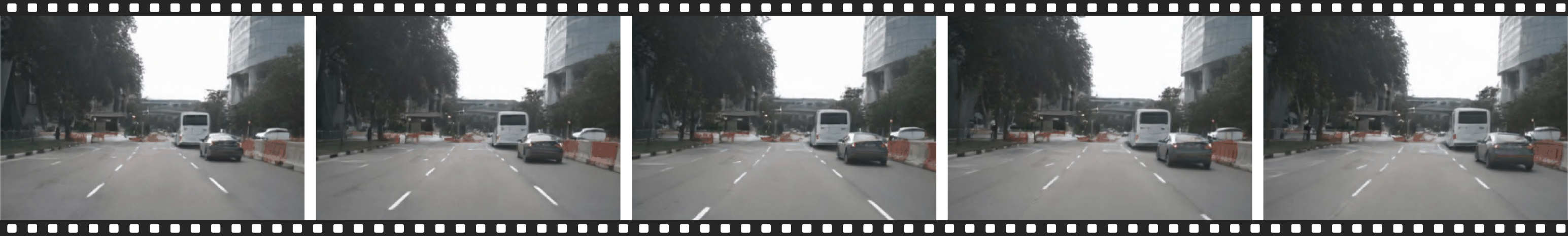}
        \caption{Good example in the \emph{Temporal Consistency} dimension (Score: \textcolor{w_blue}{$85.57\%$})}
        \label{fig:gen_temporal_consistency_5}
    \end{subfigure}
    \\[1ex]
    \begin{subfigure}[h]{\textwidth}
        \centering
        \includegraphics[width=\linewidth]{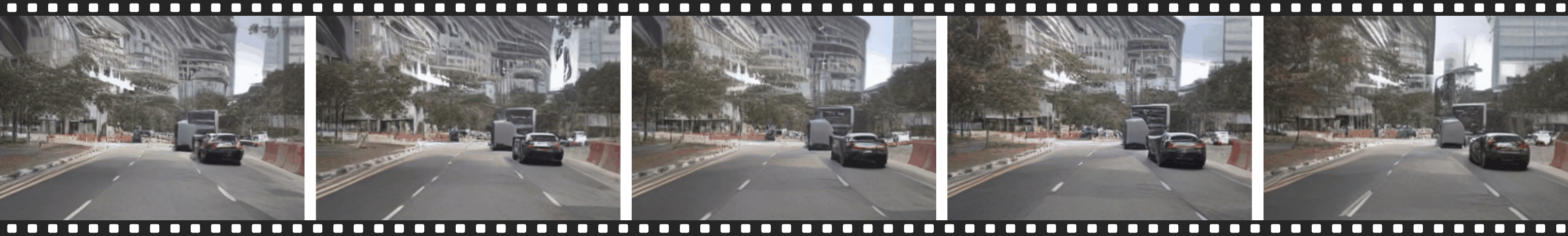}
        \caption{Bad example in the \emph{Temporal Consistency} dimension (Score: \textcolor{red}{$54.45\%$})}
        \label{fig:gen_temporal_consistency_6}
    \end{subfigure}
    \vspace{-0.1cm}
    \caption{Examples of ``good'' and ``bad'' generation qualities in terms of \emph{Temporal Consistency} in WorldLens.}
    \label{fig:gen_temporal_consistency}
\end{figure}

Figure~\ref{fig:gen_temporal_consistency} provides typical examples of videos with good and bad quality in terms of \emph{Temporal Consistency}.

\subsubsection{~Evaluation \& Analysis}
Table~\ref{tab:supp_gen_temporal_consistency} provides the complete results of models in terms of \emph{Temporal Consistency}.

\begin{table*}[h]
    \centering
    \vspace{0.2cm}
    \caption{Complete comparisons of state-of-the-art driving world models in terms of \emph{Temporal Consistency} in WorldLens.}
    \vspace{-0.2cm}
    \label{tab:supp_gen_temporal_consistency}
    \resizebox{\linewidth}{!}{
    \begin{tabular}{r|cccccc|c}
        \toprule
        \multirow{2}{*}{$\mathcal{S}_\mathrm{TC}(\cdot)$} & \textbf{MagicDrive} & \textbf{DreamForge}  & \textbf{DriveDreamer-2} & \textbf{OpenDWM} & \textbf{~DiST-4D~} & $\mathcal{X}$\textbf{-Scene} & \textcolor{gray}{\textbf{Empirical}}
        \\
        & \textcolor{gray}{\small[ICLR'24]} & \textcolor{gray}{\small[arXiv'24]} & \textcolor{gray}{\small[AAAI'25]} & \textcolor{gray}{\small[CVPR'25]} & \textcolor{gray}{\small[ICCV'25]} & \textcolor{gray}{\small[NeurIPS'25]} & \textcolor{gray}{\textbf{Max}}
        \\\midrule
        ACM ($\uparrow$) & $91.43\%$ & $92.69\%$ & $93.65\%$ & $93.55\%$ & $92.27\%$ & $92.26\%$ & \cellcolor{gray!7}\textcolor{gray}{$93.24\%$} 
        \\
        TJI ($\uparrow$) & $43.31\%$ & $42.69\%$ & $44.19\%$ & $43.83\%$ & $44.22\%$ & $42.91\%$ & \cellcolor{gray!7}\textcolor{gray}{$45.87\%$}
        \\
        \textbf{Total ($\uparrow$)} & \cellcolor{w_blue!20}$74.44\%$ & \cellcolor{w_blue!20}$79.82\%$ & \cellcolor{w_blue!20}$79.51\%$ & \cellcolor{w_blue!20}$79.63\%$ & \cellcolor{w_blue!20}$77.76\%$ & \cellcolor{w_blue!20}$79.41\%$ & \cellcolor{gray!7}\textcolor{gray}{$93.24\%$}
        \\
        \bottomrule
    \end{tabular}}
    \vspace{-0.2cm}
\end{table*}

%% file: sections_supp/dimensions/gen_semantic_consistency.tex
\subsection{~Semantic Consistency}

\subsubsection{~Definition}
Semantic Consistency assesses the temporal stability of scene semantics in generated videos, ensuring that the underlying segmentation layout evolves smoothly over time. Using a frozen semantic segmentation model $\psi_{\mathrm{SEG}}(\cdot)$, this metric quantifies how consistently pixel-wise labels, region structures, and global class distributions are preserved between consecutive frames.

\vspace{-0.1cm}
\subsubsection{~Formulation}
For each generated video $y_j=\{y_j^{(t)}\}_{t=1}^{T}$, we obtain frame-wise segmentation masks: $M_j^{(t)}=\psi_{\mathrm{SEG}}\!(y_j^{(t)})\in\{0,\dots,C{-}1\}^{H\times W}$.
Temporal semantic stability is quantified by three complementary components:

\noindent\textbf{Label Flip Rate (LFR)}
measures how rarely \emph{interior} pixels (after class-wise morphological erosion) change their semantic label between consecutive frames. For class $c$, let $\Omega_c^{(t)}$ be the eroded interior region. The flip ratio is the fraction of pixels in $\Omega_c^{(t)}$ whose labels differ in $M_j^{(t+1)}$. The per-video LFR score averages these values across classes and time, then normalizes:
$\mathcal{S}_{\mathrm{LFR}}(y_j)
=1-\tfrac{1}{T-1}\sum\nolimits_{t=1}^{T-1}
\tfrac{\sum\nolimits_{c}\sum\nolimits_{\mathbf{p}\in\Omega_c^{(t)}}
\mathbf{1}[M_j^{(t+1)}(\mathbf{p})\neq c]}
{\sum\nolimits_{c}|\Omega_c^{(t)}|}$.

\noindent\textbf{Segment Association Consistency (SAC)}
measures how consistently connected semantic regions persist over time.  
For each class $c$, connected components in $M_j^{(t)}$ and $M_j^{(t+1)}$ are matched by Hungarian assignment over IoU.  
The score is the pixel-weighted mean IoU of the matched region pairs:
$\mathcal{S}_{\mathrm{SAC}}(y_j)
=\tfrac{1}{T-1}\sum\nolimits_{t=1}^{T-1}
\tfrac{\sum\nolimits_{c}\sum\nolimits_{(R,R')\in\pi_c^{(t)}}|R|\cdot\mathrm{IoU}(R,R')}
{\sum\nolimits_{c}\sum\nolimits_{R\in\mathcal{R}_c^{(t)}}|R|}$, where $\pi_c^{(t)}$ is the optimal region matching.

\noindent\textbf{Class Distribution Stability (CDS)}
compares frame-level class histograms. Let $p^{(t)}$ be the normalized histogram of frame $t$. Global distribution shift is quantified by the Jensen–Shannon divergence:
$\mathcal{S}_{\mathrm{CDS}}(y_j)=1-\tfrac{1}{T-1}
\sum\nolimits_{t=1}^{T-1}\mathrm{JSD}\!\left(p^{(t)}\,\|\,p^{(t+1)}\right)$. 

Each component is normalized to $[0,1]$.  
The final Semantic Consistency score is a weighted combination:
\vspace{-0.2cm}
\begin{equation}
\boxed{\;
\mathcal{S}_{\mathrm{SemC}}(\mathcal{Y})
=\tfrac{1}{N_g}\sum\nolimits_{j=1}^{N_g}
\big[w_1\mathcal{S}_{\mathrm{LFR}}(y_j)
+w_2\mathcal{S}_{\mathrm{SAC}}(y_j)
+w_3\mathcal{S}_{\mathrm{CDS}}(y_j)\big]
\;}
\label{eq:semantic_consistency}
\end{equation}
with $(w_1,w_2,w_3)=(0.5,0.4,0.1)$.
A high $\mathcal{S}_{\mathrm{SemC}}$ score signifies that drivable areas, lane boundaries, and object classes remain stable under temporal changes.

\subsubsection{~Implementation Details}
We obtain frame-wise semantic maps using the panoptic segmentation model from OpenSeeD~\cite{zhang2023simple}. The predicted segments are then converted to label masks via a fixed color palette and used to compute the temporal semantic consistency score.

\vspace{-0.1cm}
\subsubsection{~Examples}
\begin{figure}[t]
    \centering
    \begin{subfigure}[h]{\textwidth}
        \centering
        \includegraphics[width=\linewidth]{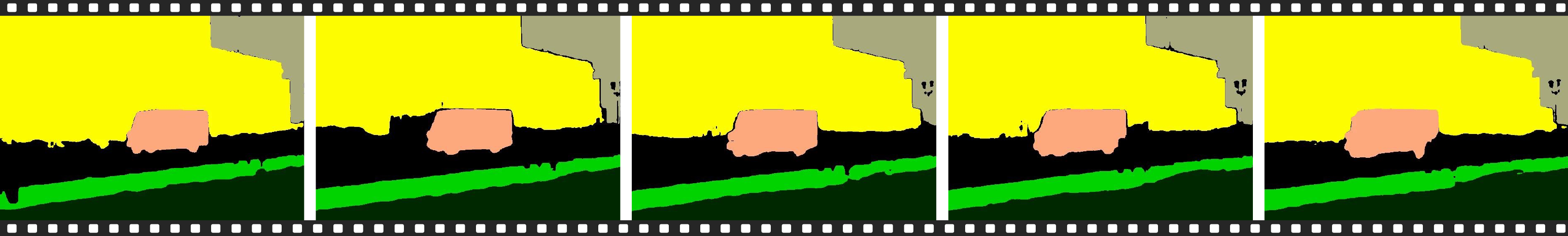}
        \caption{Good example in the \emph{Semantic Consistency} dimension (Score: \textcolor{w_blue}{$94.99\%$})}
        \label{fig:gen_semantic_consistency_1}
    \end{subfigure}
    \\[1ex]
    \begin{subfigure}[h]{\textwidth}
        \centering
        \includegraphics[width=\linewidth]{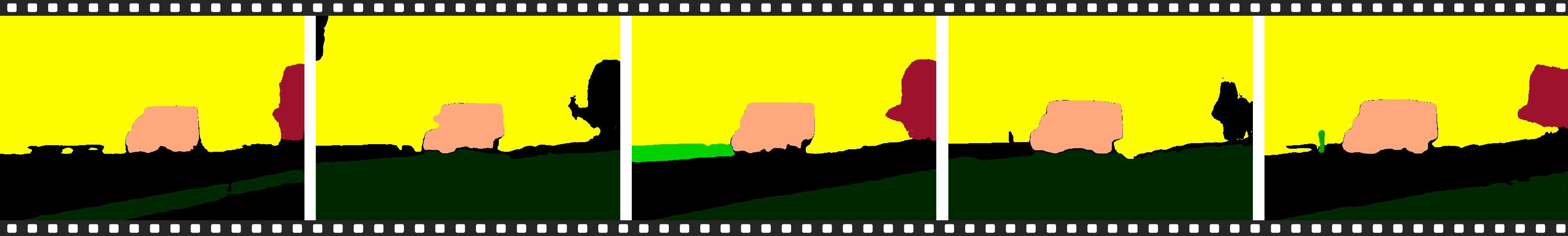}
        \caption{Bad example in the \emph{Semantic Consistency} dimension (Score: \textcolor{red}{$74.70\%$})}
        \label{fig:gen_semantic_consistency_2}
    \end{subfigure}
    \\[3.5ex]
    \begin{subfigure}[h]{\textwidth}
        \centering
        \includegraphics[width=\linewidth]{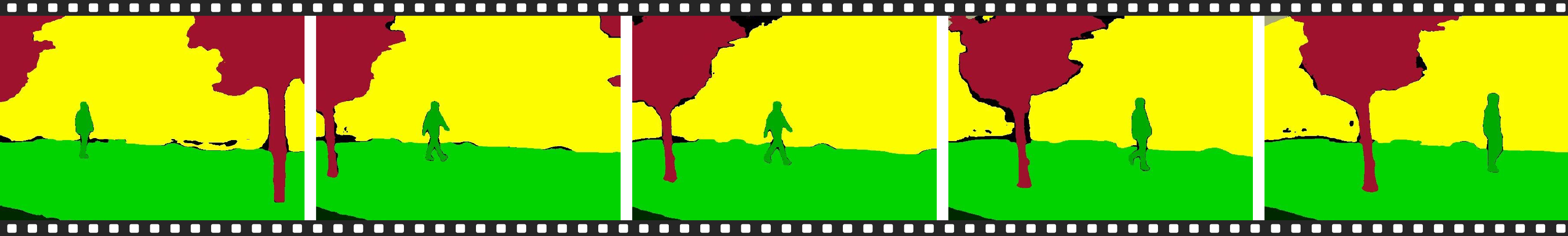}
        \caption{Good example in the \emph{Semantic Consistency} dimension (Score: \textcolor{w_blue}{$95.78\%$})}
        \label{fig:gen_semantic_consistency_3}
    \end{subfigure}
    \\[1ex]
    \begin{subfigure}[h]{\textwidth}
        \centering
        \includegraphics[width=\linewidth]{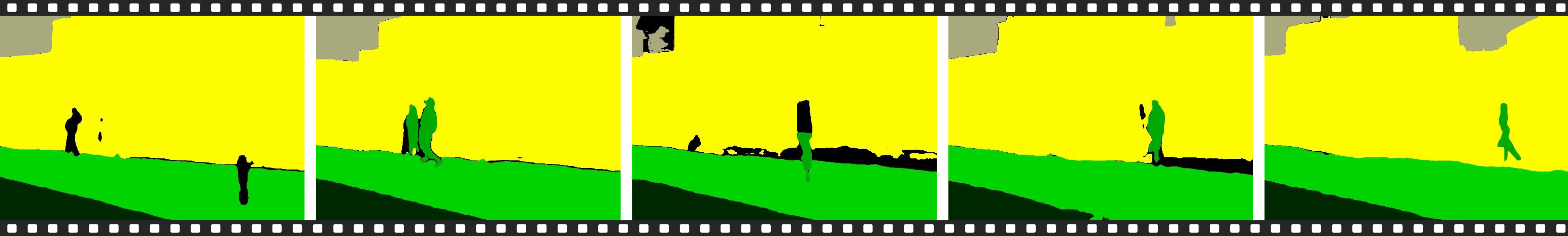}
        \caption{Bad example in the \emph{Semantic Consistency} dimension (Score: \textcolor{red}{$70.14\%$})}
        \label{fig:gen_semantic_consistency_4}
    \end{subfigure}
    \\[3.5ex]
    \begin{subfigure}[h]{\textwidth}
        \centering
        \includegraphics[width=\linewidth]{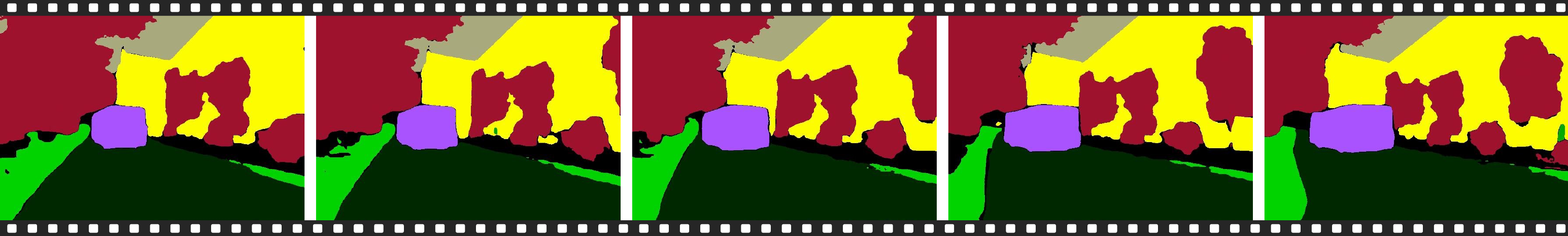}
        \caption{Good example in the \emph{Semantic Consistency} dimension (Score: \textcolor{w_blue}{$93.77\%$})}
        \label{fig:gen_semantic_consistency_5}
    \end{subfigure}
    \\[1ex]
    \begin{subfigure}[h]{\textwidth}
        \centering
        \includegraphics[width=\linewidth]{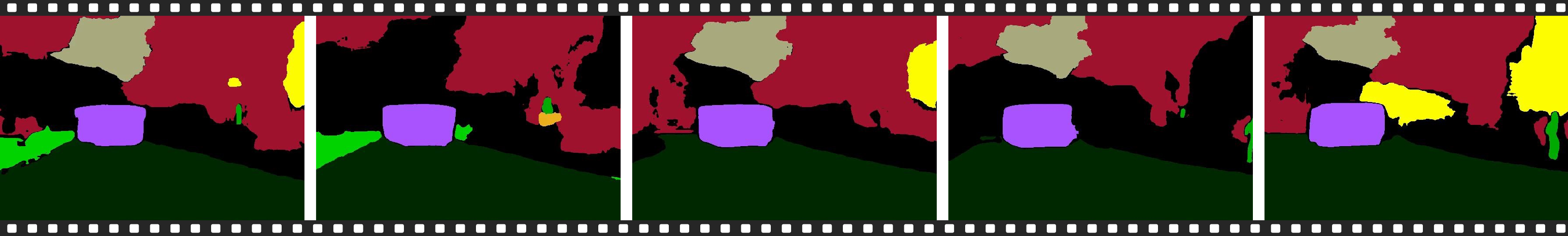}
        \caption{Bad example in the \emph{Semantic Consistency} dimension (Score: \textcolor{red}{$61.82\%$})}
        \label{fig:gen_semantic_consistency_6}
    \end{subfigure}
    \vspace{-0.1cm}
    \caption{Examples of ``good'' and ``bad'' generation qualities in terms of \emph{Semantic Consistency} in WorldLens.}
    \label{fig:gen_semantic_consistency}
\end{figure}

Figure~\ref{fig:gen_semantic_consistency} provides typical examples of videos with good and bad quality in terms of \emph{Semantic Consistency}.

\subsubsection{~Evaluation \& Analysis}
Table~\ref{tab:supp_gen_semantic_consistency} provides the complete results of models in terms of \emph{Semantic Consistency}.

\begin{table*}[h]
    \centering
    \vspace{0.2cm}
    \caption{Complete comparisons of state-of-the-art driving world models in terms of \emph{Semantic Consistency} in WorldLens.}
    \vspace{-0.2cm}
    \label{tab:supp_gen_semantic_consistency}
    \resizebox{\linewidth}{!}{
    \begin{tabular}{r|cccccc|c}
        \toprule
        \multirow{2}{*}{$\mathcal{S}_\mathrm{SemC}(\cdot)$} & \textbf{MagicDrive} & \textbf{DreamForge}  & \textbf{DriveDreamer-2} & \textbf{OpenDWM} & \textbf{~DiST-4D~} & $\mathcal{X}$\textbf{-Scene} & \textcolor{gray}{\textbf{Empirical}}
        \\
        & \textcolor{gray}{\small[ICLR'24]} & \textcolor{gray}{\small[arXiv'24]} & \textcolor{gray}{\small[AAAI'25]} & \textcolor{gray}{\small[CVPR'25]} & \textcolor{gray}{\small[ICCV'25]} & \textcolor{gray}{\small[NeurIPS'25]} & \textcolor{gray}{\textbf{Max}}
        \\\midrule
        Label Flip Rate (LFR, $\uparrow$) & $85.48\%$ & $89.15\%$ & $89.59\%$ & $88.09\%$ & $88.46\%$ & $87.92\%$ & \cellcolor{gray!7}\textcolor{gray}{$90.39\%$} 
        \\
        Segmentation Association (SAC, $\uparrow$) & $75.57\%$ & $80.85\%$ & $82.21\%$ & $79.94\%$ & $80.13\%$ & $79.54\%$ & \cellcolor{gray!7}\textcolor{gray}{$82.48\%$} 
        \\
        Distribution Stability (CDS, $\uparrow$) & $96.40\%$ & $97.31\%$ & $97.05\%$ & $96.86\%$ & $97.00\%$ & $96.95\%$ & \cellcolor{gray!7}\textcolor{gray}{$97.89\%$} 
        \\
        \textbf{Total ($\uparrow$)} & \cellcolor{w_blue!20}$80.63\%$ & \cellcolor{w_blue!20}$84.99\%$ & \cellcolor{w_blue!20}$85.91\%$ & \cellcolor{w_blue!20}$84.08\%$ & \cellcolor{w_blue!20}$84.32\%$ & \cellcolor{w_blue!20}$83.80\%$ & \cellcolor{gray!7}\textcolor{gray}{$86.39\%$}
        \\
        \bottomrule
    \end{tabular}}
    \vspace{-0.2cm}
\end{table*}

%% file: sections_supp/dimensions/gen_perceptual_discrepancy.tex
\subsection{~Perceptual Discrepancy}
\label{subsec:fvd}

\subsubsection{~Definition}
Perceptual Discrepancy evaluates how closely the distribution of generated videos matches that of real ones in a learned video, semantic feature space, typically extracted by a pretrained I3D network~\cite{szegedy2015going} trained on Kinetics~\cite{kay2017kinetics}.

This metric captures both appearance realism and short-range temporal dynamics beyond framewise image-based metrics (\emph{e.g.}, FID), thus reflecting the overall perceptual quality of the synthesized sequences. It is reported as a single scalar, where a lower score indicates higher perceptual similarity to real videos.

\subsubsection{~Formulation}
Let the real and generated video sets be 
$\mathcal{X}=\{x_i\}_{i=1}^{N_r}$ and $\mathcal{Y}=\{y_j\}_{j=1}^{N_g}$. Each video is encoded into a $d$-dimensional feature vector using a fixed video encoder $\phi_{\mathrm{PD}}$:
\[
\mathbf{f}_i = \phi_{\mathrm{PD}}(x_i), 
\qquad 
\mathbf{g}_j = \phi_{\mathrm{PD}}(y_j).
\]
Let $(\boldsymbol{\mu}_x,\boldsymbol{\Sigma}_x)$ and $(\boldsymbol{\mu}_y,\boldsymbol{\Sigma}_y)$ be the empirical means and covariances of the feature sets $\{\mathbf{f}_i\}$ and $\{\mathbf{g}_j\}$, respectively. Following the Fréchet formulation, the Perceptual Fidelity score (equivalent to the Fréchet Video Distance, FVD) is defined as follows:
\begin{equation}
\boxed{\;
\mathcal{S}_\mathrm{PD}(\mathcal{X}, \mathcal{Y})
= \|\boldsymbol{\mu}_x - \boldsymbol{\mu}_y\|_2^2
+ \mathrm{Tr}\!\left(
\boldsymbol{\Sigma}_x + \boldsymbol{\Sigma}_y
- 2(\boldsymbol{\Sigma}_x^{1/2}\boldsymbol{\Sigma}_y\boldsymbol{\Sigma}_x^{1/2})^{1/2}
\right)
\;}
\label{eq:perceptual_fidelity}
\end{equation}
A lower $\mathcal{S}_\mathrm{PD}$ indicates that the generated distribution $\mathcal{Y}$ is perceptually closer to the real distribution $\mathcal{X}$.

\emph{Perceptual Discrepancy} serves as a global perceptual indicator of visual and temporal realism. By comparing distributions in a semantically informed video embedding space, it evaluates not only static appearance but also dynamic motion smoothness and coherence. A low score indicates that the generative model produces sequences with authentic spatial structures, plausible dynamics, and consistent motion statistics, while a high score reveals perceptual drift or domain mismatch.

This metric thus complements fine-grained evaluations by providing an overarching measure of distributional fidelity in world-model generation.

\subsubsection{~Implementation Details}
\emph{Perceptual Discrepancy} is measured using Fréchet Video Distance (FVD). 
We extract video features with a pretrained I3D model~\cite{szegedy2015going} (Kinetics-400~\cite{kay2017kinetics}) 
following the VideoGPT~\cite{yan2021videogpt} protocol, and compute FVD between ground-truth 
and generated feature distributions.

\subsubsection{~Examples}
\begin{figure}[t]
    \centering
    \begin{subfigure}[h]{\textwidth}
        \centering
        \includegraphics[width=\linewidth]{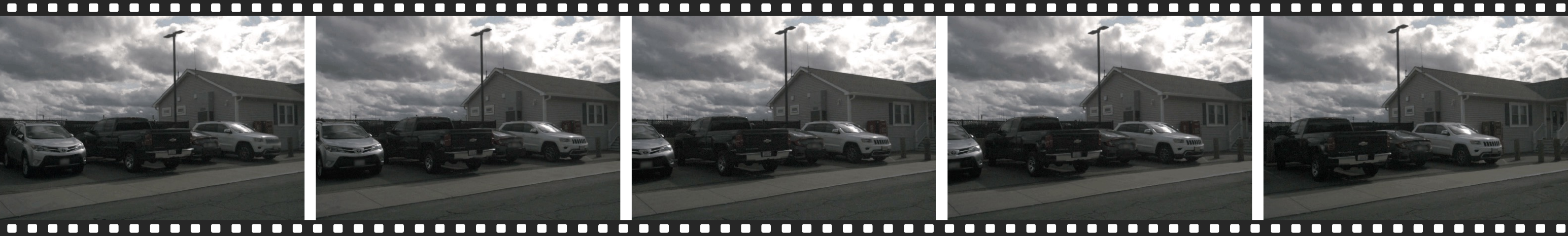}
        \caption{Good example in the \emph{Perceptual Discrepancy} dimension}
        \label{fig:gen_perceptual_discrepancy_1}
    \end{subfigure}
    \\[1ex]
    \begin{subfigure}[h]{\textwidth}
        \centering
        \includegraphics[width=\linewidth]{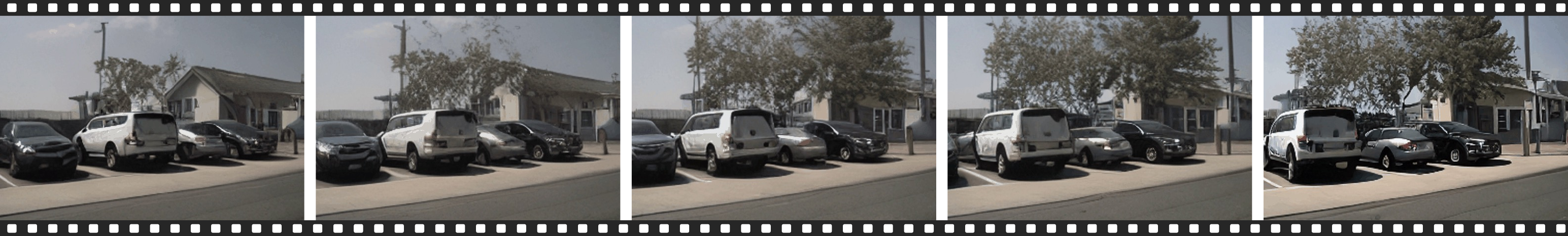}
        \caption{Bad example in the \emph{Perceptual Discrepancy} dimension}
        \label{fig:gen_perceptual_discrepancy_2}
    \end{subfigure}
    \\[3.5ex]
    \begin{subfigure}[h]{\textwidth}
        \centering
        \includegraphics[width=\linewidth]{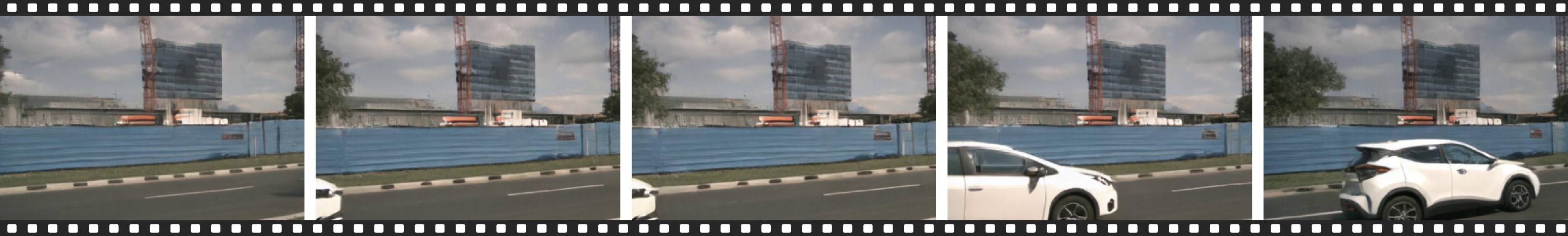}
        \caption{Good example in the \emph{Perceptual Discrepancy} dimension}
        \label{fig:gen_perceptual_discrepancy_3}
    \end{subfigure}
    \\[1ex]
    \begin{subfigure}[h]{\textwidth}
        \centering
        \includegraphics[width=\linewidth]{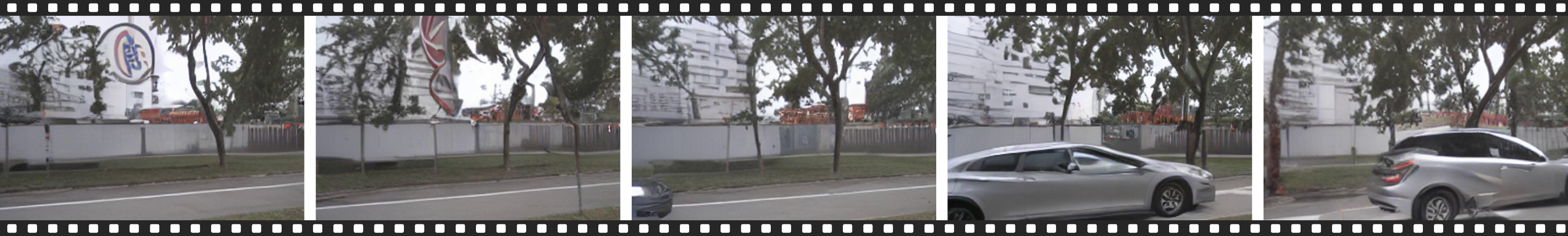}
        \caption{Bad example in the \emph{Perceptual Discrepancy} dimension}
        \label{fig:gen_perceptual_discrepancy_4}
    \end{subfigure}
    \\[3.5ex]
    \begin{subfigure}[h]{\textwidth}
        \centering
        \includegraphics[width=\linewidth]{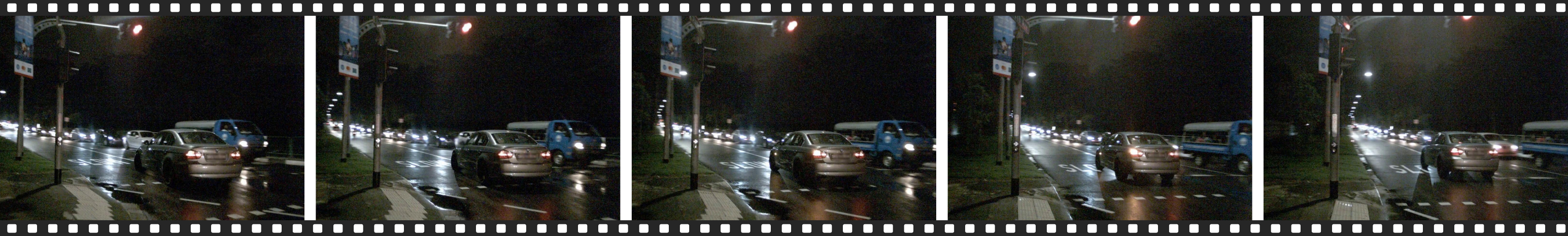}
        \caption{Good example in the \emph{Perceptual Discrepancy} dimension}
        \label{fig:gen_perceptual_discrepancy_5}
    \end{subfigure}
    \\[1ex]
    \begin{subfigure}[h]{\textwidth}
        \centering
        \includegraphics[width=\linewidth]{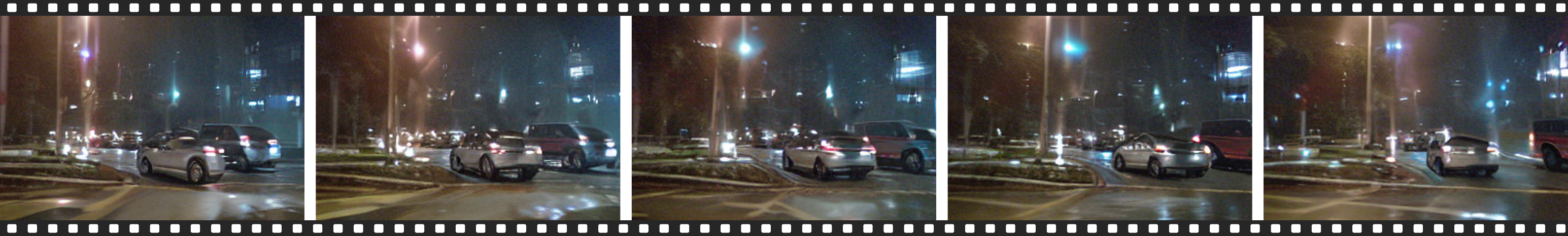}
        \caption{Bad example in the \emph{Perceptual Discrepancy} dimension}
        \label{fig:gen_perceptual_discrepancy_6}
    \end{subfigure}
    \vspace{-0.1cm}
    \caption{Examples of ``good'' and ``bad'' generation qualities in terms of \emph{Perceptual Discrepancy} in WorldLens.}
    \label{fig:gen_perceptual_discrepancy}
\end{figure}

Figure~\ref{fig:gen_perceptual_discrepancy} provides typical examples of videos with good and bad quality in terms of \emph{Perceptual Discrepancy}.

\subsubsection{~Evaluation \& Analysis}
Table~\ref{tab:supp_gen_perceptual_discrepancy} provides the complete results of models in terms of \emph{Perceptual Discrepancy}.

\begin{table*}[h]
    \centering
    \vspace{0.2cm}
    \caption{Complete comparisons of state-of-the-art driving world models in terms of \emph{Perceptual Discrepancy} in WorldLens.}
    \vspace{-0.2cm}
    \label{tab:supp_gen_perceptual_discrepancy}
    \resizebox{\linewidth}{!}{
    \begin{tabular}{r|cccccc|c}
        \toprule
        \multirow{2}{*}{$\mathcal{S}_\mathrm{PD}(\cdot)$} & \textbf{MagicDrive} & \textbf{DreamForge}  & \textbf{DriveDreamer-2} & \textbf{OpenDWM} & \textbf{~DiST-4D~} & $\mathcal{X}$\textbf{-Scene} & \textcolor{gray}{\textbf{Empirical}}
        \\
        & \textcolor{gray}{\small[ICLR'24]} & \textcolor{gray}{\small[arXiv'24]} & \textcolor{gray}{\small[AAAI'25]} & \textcolor{gray}{\small[CVPR'25]} & \textcolor{gray}{\small[ICCV'25]} & \textcolor{gray}{\small[NeurIPS'25]} & \textcolor{gray}{\textbf{Max}}
        \\\midrule
        \textbf{Total ($\downarrow$)} & \cellcolor{w_blue!20}$222.00$ & \cellcolor{w_blue!20}$189.76$ & \cellcolor{w_blue!20}$127.07$ & \cellcolor{w_blue!20}$90.42$ & \cellcolor{w_blue!20}$58.08$ & \cellcolor{w_blue!20}$179.74$ & \cellcolor{gray!7}\textcolor{gray}{$-$}
        \\
        \bottomrule
    \end{tabular}}
    \vspace{-0.2cm}
\end{table*}

%% file: sections_supp/dimensions/gen_cross_view_consistency.tex
\subsection{~Cross-View Consistency}

\subsubsection{~Definition}
Cross-View Consistency evaluates the geometric and photometric coherence across overlapping regions between adjacent camera views in a multi-view driving scene. 
A spatially consistent generation should ensure that content observed from different cameras remains structurally aligned and visually coherent, faithfully representing the same physical world from multiple perspectives. 
This property is critical for autonomous driving, as consistent multi-view generation reflects an accurate understanding of shared 3D geometry and scene semantics.

We quantify this consistency by computing the mean accumulated confidence of feature correspondences between overlapping edge regions of adjacent camera pairs using a pretrained local feature matcher. 
Higher confidence indicates better geometric and appearance alignment across views.

\subsubsection{~Formulation}
For each generated scene $y_j\!\in\!\mathcal{Y}$ with $N_v$ synchronized views and $T$ frames,
a frozen LoFTR matcher $\psi_{\mathrm{LoFTR}}$ produces $M_{ab}^{(t)}$ correspondences with confidence $c_m^{(t)}\!\in\![0,1]$ 
between every adjacent camera pair $(a,b)\!\in\!\mathcal{P}$ at frame $t$.
The overall Cross-View Consistency score averages all confidences across pairs, frames, and videos:
\begin{equation}
\boxed{\;
\mathcal{S}_{\mathrm{CVC}}(\mathcal{Y})
= \frac{1}{N_g|\mathcal{P}|T}\!
\sum\nolimits_{j=1}^{N_g}\!
\sum\nolimits_{(a,b)\in\mathcal{P}}\!
\sum\nolimits_{t=1}^{T}\!
\sum\nolimits_{m=1}^{M_{ab}^{(t)}}\!
c_m^{(t)}
\;}
\label{eq:cross_view_consistency}
\end{equation}
Higher $\mathcal{S}_{\mathrm{CVC}}$ indicates stronger geometric and appearance alignment between adjacent camera views.

A high Cross-View Consistency score signifies that the generated multi-view scene maintains coherent 3D geometry and visual appearance across cameras, implying stable spatial reasoning and accurate scene composition. Conversely, low scores reveal misalignments such as perspective drift, inconsistent object boundaries, or mismatched illumination across views.

This metric thus serves as a key indicator of multi-camera integrity, linking the generative model’s visual realism to its geometric understanding of the physical world.

\subsubsection{~Implementation Details}
The \emph{Cross-View Consistency} score is computed by extracting frame-wise sparse correspondences using the pretrained LoFTR local feature matcher~\cite{sun2021loftr}. Matched keypoints across views are used to assess geometric alignment between generated and ground-truth videos.

\subsubsection{~Examples}
\begin{figure}[t]
    \centering
    \begin{subfigure}[h]{\textwidth}
        \centering
        \includegraphics[width=\linewidth]{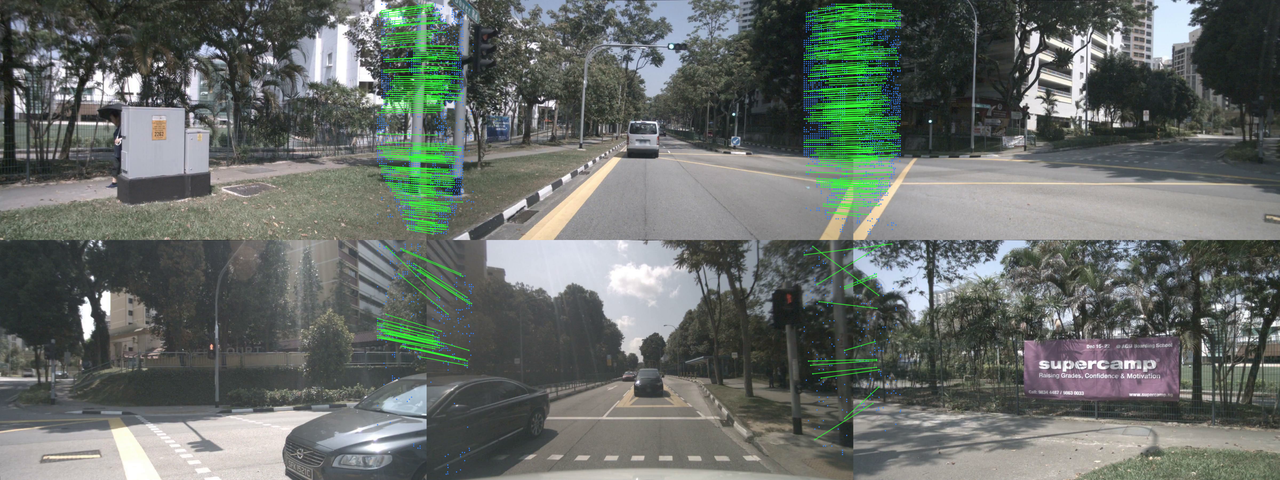}
        \caption{Good example in the \emph{Cross-View Consistency} dimension (Score: \textcolor{w_blue}{$0.74$})}
        \label{fig:gen_cross_view_consistency_1}
    \end{subfigure}
    \\[2.5ex]
    \begin{subfigure}[h]{\textwidth}
        \centering
        \includegraphics[width=\linewidth]{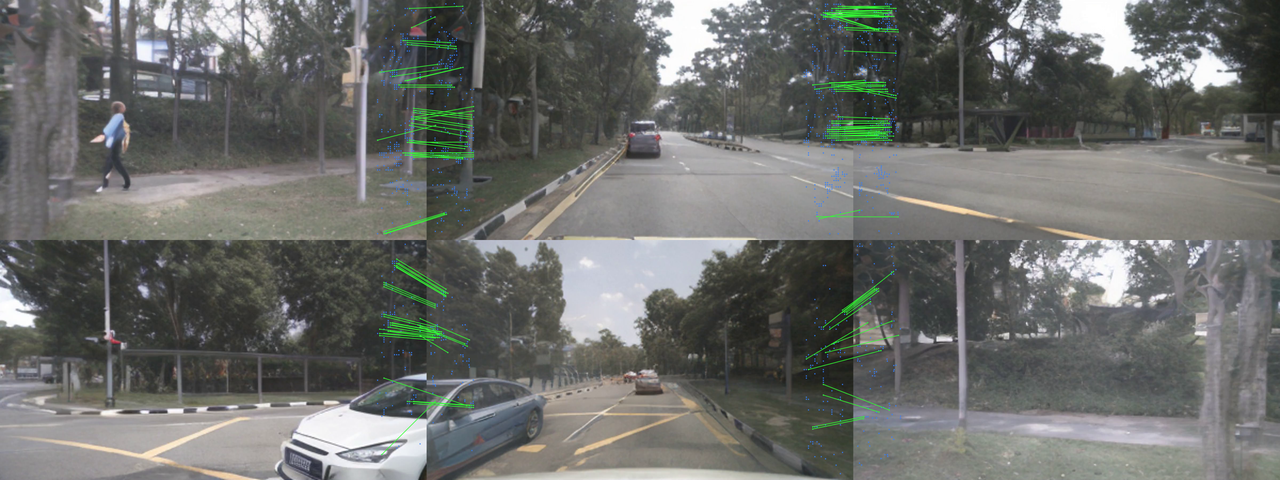}
        \caption{Bad example in the \emph{Cross-View Consistency} dimension (Score: \textcolor{red}{$0.31$})}
        \label{fig:gen_cross_view_consistency_2}
    \end{subfigure}
    \\[2.5ex]
    \begin{subfigure}[h]{\textwidth}
        \centering
        \includegraphics[width=\linewidth]{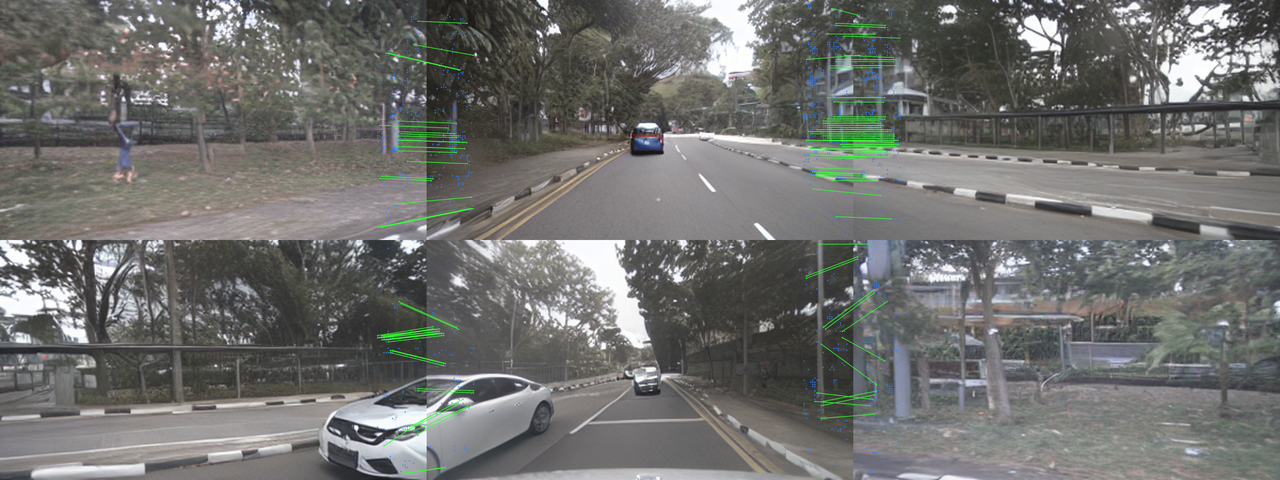}
        \caption{Bad example in the \emph{Cross-View Consistency} dimension (Score: \textcolor{red}{$0.26$})}
        \label{fig:gen_cross_view_consistency_3}
    \end{subfigure}
    \vspace{-0.1cm}
    \caption{Examples of ``good'' and ``bad'' generation qualities in terms of \emph{Cross-View Consistency} in WorldLens.}
    \label{fig:gen_cross_view_consistency}
\end{figure}

Figure~\ref{fig:gen_cross_view_consistency} provides typical examples of videos with good and bad quality in terms of \emph{Cross-View Consistency}.

\subsubsection{~Evaluation \& Analysis}
Table~\ref{tab:supp_gen_cross_view_consistency} provides the complete results of models in terms of \emph{Cross-View Consistency}.

\begin{table*}[h]
    \centering
    \vspace{0.2cm}
    \caption{Complete comparisons of state-of-the-art driving world models in terms of \emph{Cross-View Consistency} in WorldLens.}
    \vspace{-0.2cm}
    \label{tab:supp_gen_cross_view_consistency}
    \resizebox{\linewidth}{!}{
    \begin{tabular}{r|cccccc|c}
        \toprule
        \multirow{2}{*}{$\mathcal{S}_\mathrm{CVC}(\cdot)$} & \textbf{MagicDrive} & \textbf{DreamForge}  & \textbf{DriveDreamer-2} & \textbf{OpenDWM} & \textbf{~DiST-4D~} & $\mathcal{X}$\textbf{-Scene} & \textcolor{gray}{\textbf{Empirical}}
        \\
        & \textcolor{gray}{\small[ICLR'24]} & \textcolor{gray}{\small[arXiv'24]} & \textcolor{gray}{\small[AAAI'25]} & \textcolor{gray}{\small[CVPR'25]} & \textcolor{gray}{\small[ICCV'25]} & \textcolor{gray}{\small[NeurIPS'25]} & \textcolor{gray}{\textbf{Max}}
        \\\midrule
        VC Score ($\uparrow$) & $68.23$ & $72.08$ & $124.51$ & $78.81$ & $184.65$ & $74.97$ & \cellcolor{gray!7}\textcolor{gray}{$319.73$}
        \\
        VC Match ($\uparrow$) & $185.77$ & $194.99$ & $302.83$ & $211.18$ & $389.78$ & $201.00$ & \cellcolor{gray!7}\textcolor{gray}{$570.75$}
        \\
        \textbf{Total ($\uparrow$)} & \cellcolor{w_blue!20}$0.3665$ & \cellcolor{w_blue!20}$0.3686$ & \cellcolor{w_blue!20}$0.4065$ & \cellcolor{w_blue!20}$0.3720$ & \cellcolor{w_blue!20}$0.4574$ & \cellcolor{w_blue!20}$0.3721$ & \cellcolor{gray!7}\textcolor{gray}{$0.5420$}
        \\
        \bottomrule
    \end{tabular}}
    \vspace{-0.2cm}
\end{table*}

%% file: sections_supp/3_reconstruction.tex
\section{~Aspect 2: Reconstruction}

This aspect assesses the \textbf{reconstructability} of generated videos. Given a reconstructed neural 4D representation built from each generated video, we evaluate both its internal fidelity and its rendering performance from novel viewpoints. A high-quality generation should preserve temporally coherent geometry, appearance, and illumination that jointly support faithful 4D reconstruction. We employ differentiable 4D reconstruction to optimize scene geometry and radiance from the generated sequences, then re-render the reconstructed model under both original and unseen camera poses.  


\input{sections_supp/dimensions/recon_photometric_discrepancy}

\clearpage\clearpage
\input{sections_supp/dimensions/recon_geometric_discrepancy}

\clearpage\clearpage
\input{sections_supp/dimensions/recon_novel_view_quality}

\clearpage\clearpage
\input{sections_supp/dimensions/recon_novel_view_discrepancy}

%% file: sections_supp/dimensions/recon_photometric_discrepancy.tex
\subsection{~Photometric Discrepancy}

\subsubsection{~Definition}
Photometric Discrepancy quantifies how accurately the 4D scene reconstructed from a generated video can reproduce its observed frames. Each generated sequence is first converted into a neural radiance field using a differentiable pipeline based on 4D Gaussian Splatting or NeRF-based~\cite{mildenhall2021nerf} video reconstruction. The reconstructed model is then re-rendered from the same camera poses as the input frames, and pixel-wise fidelity is evaluated using standard image quality metrics such as PSNR, SSIM~\cite{zhou2004SSIM}, and LPIPS~\cite{zhang2018LPIPS}. 

\vspace{-0.1cm}
\subsubsection{~Formulation}
Let $\psi_{\mathrm{REC}}(\cdot)$ denote the 4D reconstruction function that produces a radiance field $\hat{\mathcal{R}}_j$ from a generated video $y_j$.  
Rendering this field at the input camera poses yields re-rendered frames: $\hat{y}_j^{(t)}=\mathrm{Render}\!(\hat{\mathcal{R}}_j,\,\mathrm{Pose}(t))$, where $\mathrm{Pose}(t)$ is the camera pose of frame $t$.  
Photometric fidelity is measured by the mean Learned Perceptual Image Patch Similarity (LPIPS) between the reconstructed and original frames:
\begin{equation}
\boxed{\;
\mathcal{S}_{\mathrm{PhoF}}(\mathcal{Y})
=\tfrac{1}{N_gT}\sum\nolimits_{j=1}^{N_g}\sum\nolimits_{t=1}^{T}
\mathrm{LPIPS}\!\left(\hat{y}_j^{(t)},\,y_j^{(t)}\right)
\;}
\label{eq:photometric_fidelity}
\end{equation}
Higher $\mathcal{S}_{\mathrm{PhoF}}$ indicates that the reconstructed radiance fields preserve fine-grained appearance details consistent with the generated frames.

\vspace{-0.1cm}
\subsubsection{~Implementation Details}
We follow the OmniRe~\cite{chen2025omnire} preprocessing pipeline and default configuration on nuScenes~\cite{caesar2020nuscenes}, using the same $6$-camera setup. 
Each generated clip is treated as a short multi-view sequence ($12$\,Hz, $16$ frames per camera at $544{\times}304$ resolution). 
For each clip, we optimize a single 4D Gaussian field for $30\mathrm{k}$ steps, adopting OmniRe’s static- and dynamic-node Gaussian initializations~\cite{chen2025omnire} as well as its batch size, ray-sampling strategy, loss weights, and learning-rate schedule. 
After training, we render all training views and evaluate PSNR, SSIM, and LPIPS averaged over all frames and cameras.

\vspace{-0.1cm}
\subsubsection{~Examples}
\begin{figure}[t]
    \centering
    \begin{subfigure}[h]{\textwidth}
        \centering
        \includegraphics[width=\linewidth]{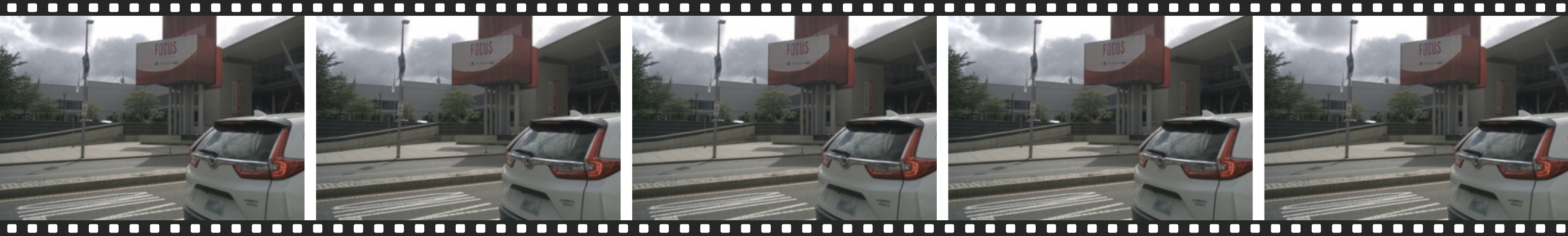}
        \caption{Good example in the \emph{Photometric Discrepancy} dimension (Score: \textcolor{w_blue}{$0.021$})}
        \label{fig:recon_photometric_discrepancy_1}
    \end{subfigure}
    \\[1ex]
    \begin{subfigure}[h]{\textwidth}
        \centering
        \includegraphics[width=\linewidth]{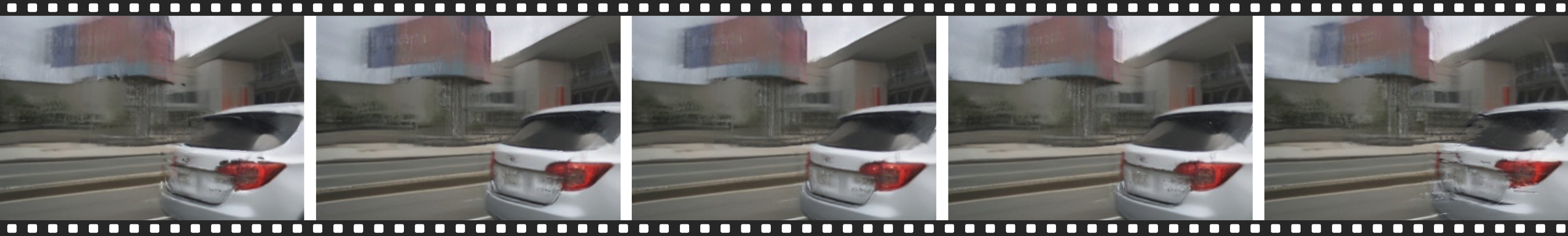}
        \caption{Bad example in the \emph{Photometric Discrepancy} dimension (Score: \textcolor{red}{$0.105$})}
        \label{fig:recon_photometric_discrepancy_2}
    \end{subfigure}
    \\[3.5ex]
    \begin{subfigure}[h]{\textwidth}
        \centering
        \includegraphics[width=\linewidth]{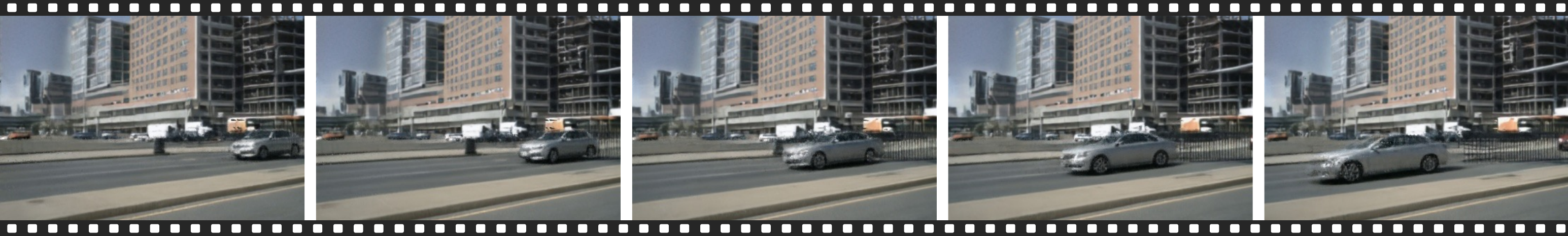}
        \caption{Good example in the \emph{Photometric Discrepancy} dimension (Score: \textcolor{w_blue}{$0.047$})}
        \label{fig:recon_photometric_discrepancy_3}
    \end{subfigure}
    \\[1ex]
    \begin{subfigure}[h]{\textwidth}
        \centering
        \includegraphics[width=\linewidth]{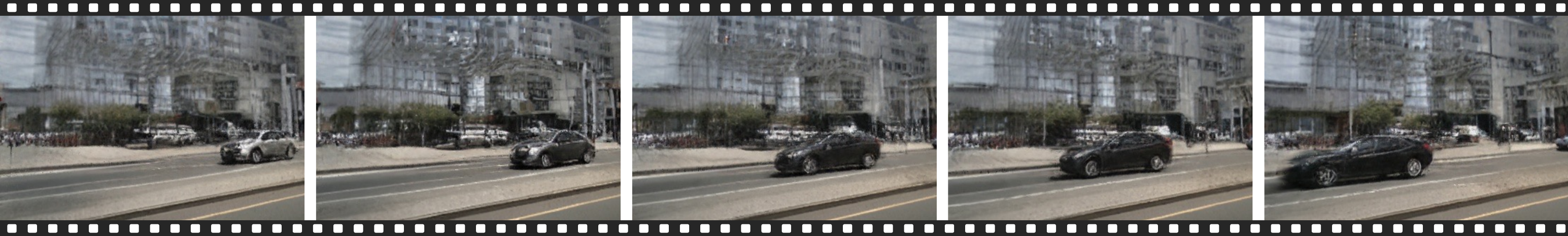}
        \caption{Bad example in the \emph{Photometric Discrepancy} dimension (Score: \textcolor{red}{$0.194$})}
        \label{fig:recon_photometric_discrepancy_4}
    \end{subfigure}
    \\[3.5ex]
    \begin{subfigure}[h]{\textwidth}
        \centering
        \includegraphics[width=\linewidth]{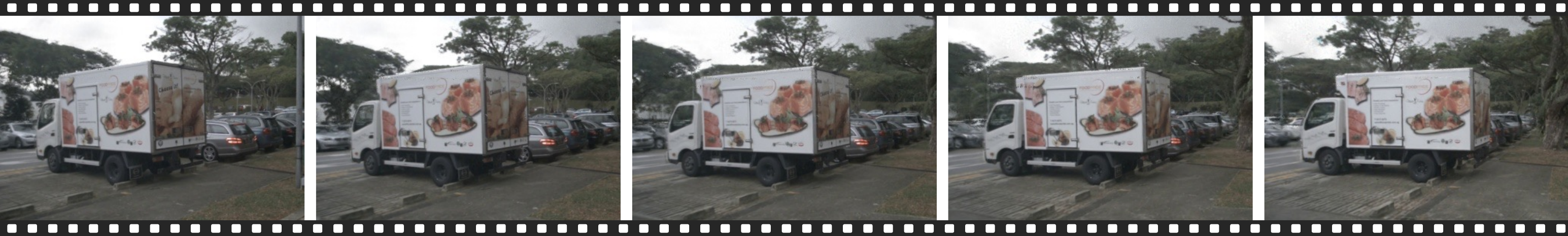}
        \caption{Good example in the \emph{Photometric Discrepancy} dimension (Score: \textcolor{w_blue}{$0.055$})}
        \label{fig:recon_photometric_discrepancy_5}
    \end{subfigure}
    \\[1ex]
    \begin{subfigure}[h]{\textwidth}
        \centering
        \includegraphics[width=\linewidth]{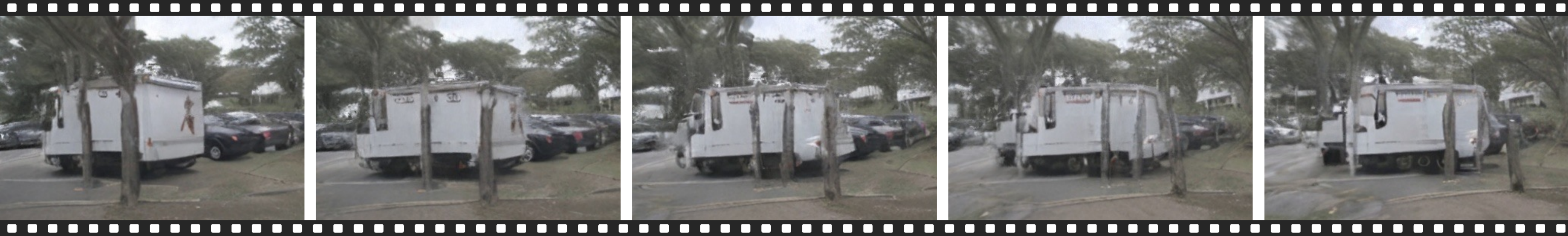}
        \caption{Bad example in the \emph{Photometric Discrepancy} dimension (Score: \textcolor{red}{$0.123$})}
        \label{fig:recon_photometric_discrepancy_6}
    \end{subfigure}
    \vspace{-0.1cm}
    \caption{Examples of ``good'' and ``bad'' reconstruction qualities in terms of \emph{Photometric Discrepancy} in WorldLens.}
    \label{fig:recon_photometric_discrepancy}
\end{figure}

Figure~\ref{fig:recon_photometric_discrepancy} provides typical examples of videos with good and bad quality in terms of \emph{Photometric Discrepancy}.

\vspace{-0.1cm}
\subsubsection{~Evaluation \& Analysis}
Table~\ref{tab:supp_recon_photometric_discrepancy} provides the complete results of models in terms of \emph{Photometric Discrepancy}.

\begin{table*}[h]
    \centering
    \vspace{0.2cm}
    \caption{Complete comparisons of state-of-the-art driving world models in terms of \emph{Photometric Discrepancy} in WorldLens.}
    \vspace{-0.2cm}
    \label{tab:supp_recon_photometric_discrepancy}
    \resizebox{\linewidth}{!}{
    \begin{tabular}{r|cccccc|c}
        \toprule
        \multirow{2}{*}{$\mathcal{S}_\mathrm{PhoF}(\cdot)$} & \textbf{MagicDrive} & \textbf{DreamForge}  & \textbf{DriveDreamer-2} & \textbf{OpenDWM} & \textbf{~DiST-4D~} & $\mathcal{X}$\textbf{-Scene} & \textcolor{gray}{\textbf{Empirical}}
        \\
        & \textcolor{gray}{\small[ICLR'24]} & \textcolor{gray}{\small[arXiv'24]} & \textcolor{gray}{\small[AAAI'25]} & \textcolor{gray}{\small[CVPR'25]} & \textcolor{gray}{\small[ICCV'25]} & \textcolor{gray}{\small[NeurIPS'25]} & \textcolor{gray}{\textbf{Max}}
        \\\midrule
        PSNR~($\uparrow$) & $28.44$ & $29.11$ & $33.15$ & $33.21$ & $32.89$ & $31.25$ & \cellcolor{gray!7}\textcolor{gray}{$34.31$}
        \\
        SSIM~($\uparrow$) & $0.887$ & $0.917$ & $0.946$ & $0.950$ & $0.948$ & $0.926$ & \cellcolor{gray!7}\textcolor{gray}{-}
        \\
        LPIPS~($\downarrow$) & \cellcolor{w_blue!20}$0.140$ & \cellcolor{w_blue!20}$0.097$ & \cellcolor{w_blue!20}$0.093$ & \cellcolor{w_blue!20}$0.065$ & \cellcolor{w_blue!20}$0.066$ & \cellcolor{w_blue!20}$0.098$ & \cellcolor{gray!7}\textcolor{gray}{$0.056$}
        \\
        \bottomrule
    \end{tabular}}
    \vspace{-0.2cm}
\end{table*}

%% file: sections_supp/dimensions/recon_geometric_discrepancy.tex
\subsection{~Geometric Discrepancy}

\subsubsection{~Definition}
Geometric Discrepancy evaluates how faithfully the geometry encoded in a generated video can be recovered after reconstruction. 
For each generated video and its paired ground truth, we reconstruct two 4DGS models using identical camera poses and training parameters, then render per-frame depth maps for both reconstructions. 
Depth consistency is measured using the Absolute Relative Error (Abs Rel) computed on regions defined by Grounded-SAM~2~\cite{ravi2025sam-2} masks that isolate road surfaces and foreground objects.

\vspace{-0.1cm}
\subsubsection{~Formulation}
Let $\psi_{\mathrm{REC}}(\cdot)$ denote the 4D reconstruction function.  
For each generated video $y_j$ and ground truth $x_j$, we obtain two reconstructed fields
$\hat{\mathcal{R}}_j=\psi_{\mathrm{REC}}(y_j)$ 
and 
$\hat{\mathcal{R}}_j^{\mathrm{gt}}=\psi_{\mathrm{REC}}(x_j)$.
At each training pose $\mathrm{Pose}(t)$, the corresponding depth maps are rendered as:
\[
\hat{D}_j^{(t)}=\mathrm{RenderDepth}\!\left(\hat{\mathcal{R}}_j,\mathrm{Pose}(t)\right),
\qquad
D_j^{\mathrm{gt}(t)}=\mathrm{RenderDepth}\!\left(\hat{\mathcal{R}}_j^{\mathrm{gt}},\mathrm{Pose}(t)\right).
\]
Let $M^{(t)}_j$ be the Grounded-SAM~2 mask selecting conditioned pixels.  
The overall Geometric Accuracy score averages the masked AbsRel error over all frames and videos:
\begin{equation}
\boxed{\;
\mathcal{S}_{\mathrm{GeoA}}(\mathcal{Y})
=\tfrac{1}{N_gT}\sum\nolimits_{j=1}^{N_g}\sum\nolimits_{t=1}^{T}
\tfrac{1}{|\mathcal{M}^{(t)}_j|}
\sum\nolimits_{\mathbf{p}\in M^{(t)}_j}
\tfrac{\big|\hat{D}_j^{(t)}(\mathbf{p})-D_j^{\mathrm{gt}(t)}(\mathbf{p})\big|}
{D_j^{\mathrm{gt}(t)}(\mathbf{p})}
\;}
\label{eq:geometric_accuracy}
\end{equation}
Lower $\mathcal{S}_{\mathrm{GeoA}}$ indicates that the reconstructed geometry from generated videos is more consistent with the ground-truth scene structure.

\vspace{-0.1cm}
\subsubsection{~Implementation Details}
This aspect shares the same training setup as the photometric discrepancy. The main difference is in the rendering and metric computation. For each clip, we render per‑pixel depth from the learned 4D Gaussian field for all training views using the default Gaussian rasterizer (GSplat~\cite{ye2025gsplat}) as in OmniRe~\cite{chen2025omnire}, configured in the ``RGB+ED'' mode that outputs both color and Euclidean depth along each camera ray. To obtain fair and semantically meaningful depth metrics, we construct evaluation masks from ground‐truth images using Grounded SAM~2~\cite{ravi2025sam-2}, extracting the union of the road and vehicle regions. Depth errors, \emph{e.g.}, Abs Rel and Root Mean Squared Error (RMSE), are then computed only within these masked pixels by comparing with the depth rendered by the GT‑trained Gaussian field. We also report the threshold accuracy metrics ($\delta_1$, $\delta_2$, $\delta_3$). Per‑clip scores are obtained by averaging over all frames and cameras of the clip.

\vspace{-0.1cm}
\subsubsection{~Example}
\begin{figure}[t]
    \centering
    \begin{subfigure}[h]{\textwidth}
        \centering
        \includegraphics[width=\linewidth]{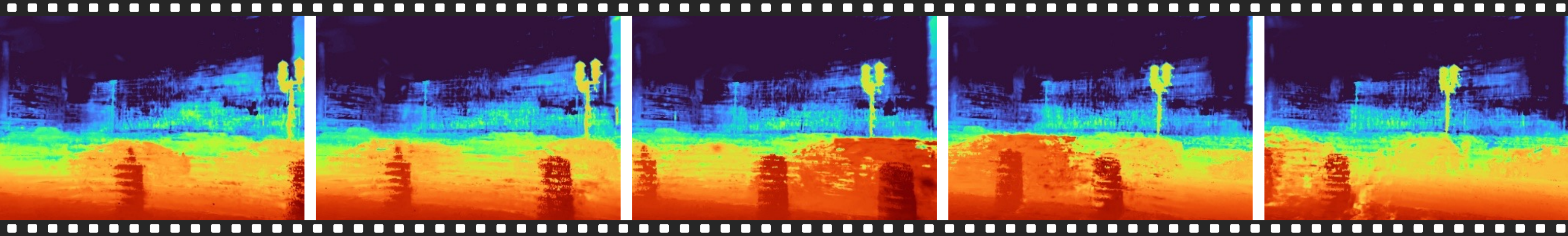}
        \caption{Good example in the \emph{Geometric Discrepancy} dimension (Score: \textcolor{w_blue}{$0.033$})}
        \label{fig:recon_geometric_discrepancy_1}
    \end{subfigure}
    \\[1ex]
    \begin{subfigure}[h]{\textwidth}
        \centering
        \includegraphics[width=\linewidth]{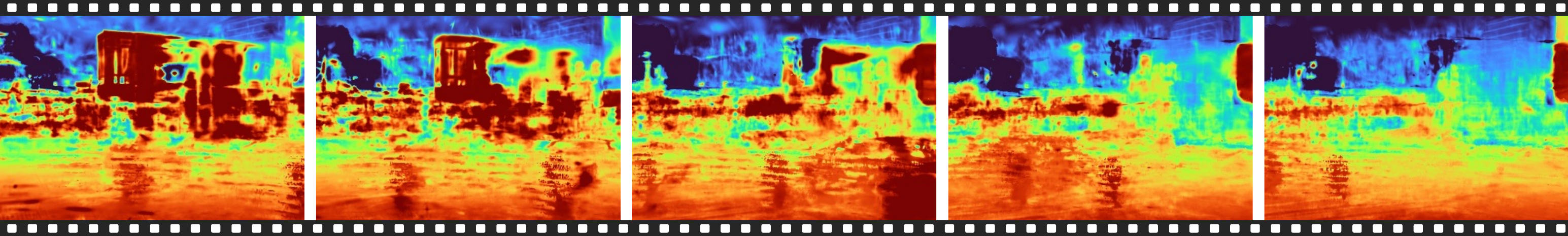}
        \caption{Bad example in the \emph{Geometric Discrepancy} dimension (Score: \textcolor{red}{$0.156$})}
        \label{fig:recon_geometric_discrepancy_2}
    \end{subfigure}
    \\[3.5ex]
    \begin{subfigure}[h]{\textwidth}
        \centering
        \includegraphics[width=\linewidth]{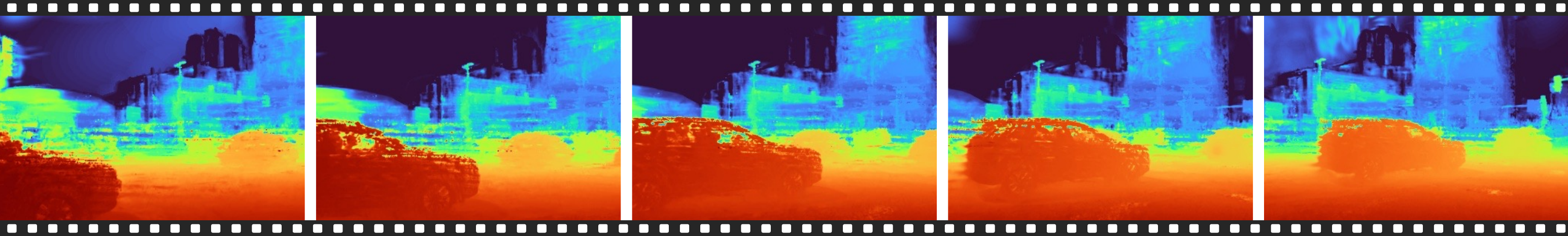}
        \caption{Good example in the \emph{Geometric Discrepancy} dimension (Score: \textcolor{w_blue}{$0.027$})}
        \label{fig:recon_geometric_discrepancy_3}
    \end{subfigure}
    \\[1ex]
    \begin{subfigure}[h]{\textwidth}
        \centering
        \includegraphics[width=\linewidth]{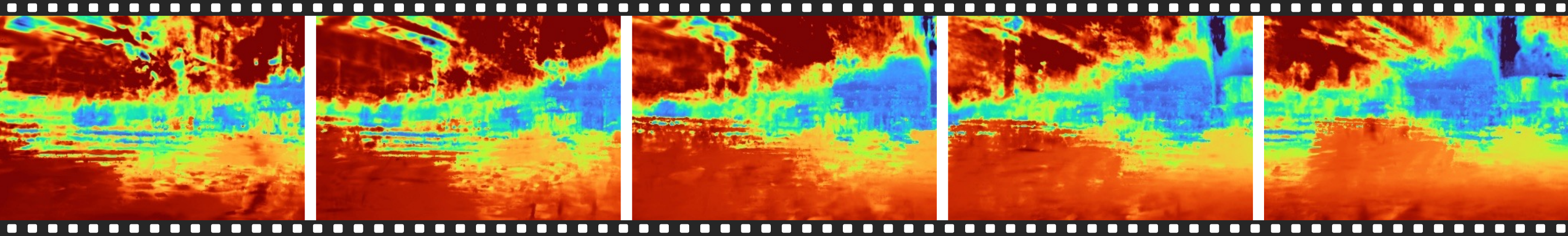}
        \caption{Bad example in the \emph{Geometric Discrepancy} dimension (Score: \textcolor{red}{$0.177$})}
        \label{fig:recon_geometric_discrepancy_4}
    \end{subfigure}
    \\[3.5ex]
    \begin{subfigure}[h]{\textwidth}
        \centering
        \includegraphics[width=\linewidth]{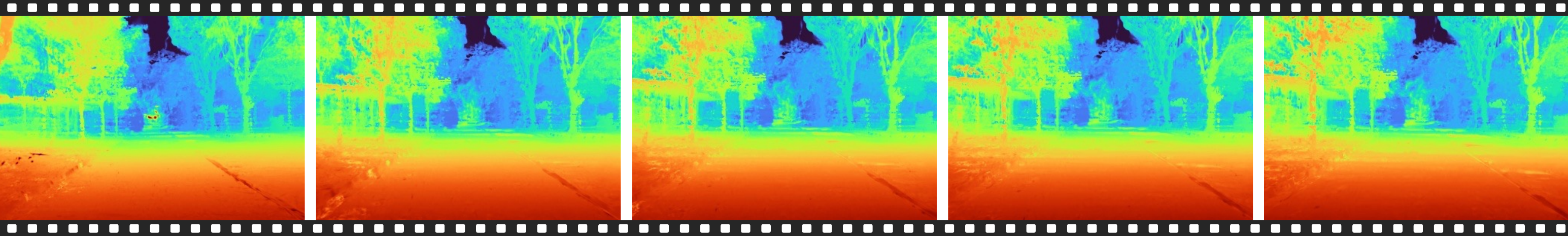}
        \caption{Good example in the \emph{Geometric Discrepancy} dimension (Score: \textcolor{w_blue}{$0.024$})}
        \label{fig:recon_geometric_discrepancy_5}
    \end{subfigure}
    \\[1ex]
    \begin{subfigure}[h]{\textwidth}
        \centering
        \includegraphics[width=\linewidth]{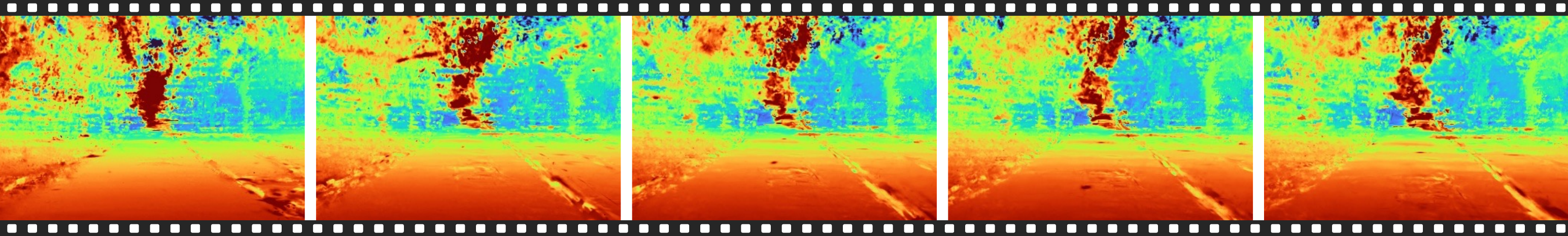}
        \caption{Bad example in the \emph{Geometric Discrepancy} dimension (Score: \textcolor{red}{$0.097$})}
        \label{fig:recon_geometric_discrepancy_6}
    \end{subfigure}
    \vspace{-0.1cm}
    \caption{Examples of ``good'' and ``bad'' reconstruction qualities in terms of \emph{Geometric Discrepancy} in WorldLens.}
    \label{fig:recon_geometric_discrepancy}
\end{figure}

Figure~\ref{fig:recon_geometric_discrepancy} provides typical examples of videos with good and bad quality in terms of \emph{Geometric Discrepancy}.

\vspace{-0.1cm}
\subsubsection{~Evaluation \& Analysis}
Table~\ref{tab:supp_recon_geometric_discrepancy} provides the complete results of models in terms of \emph{Geometric Discrepancy}.

\begin{table*}[h]
    \centering
    \vspace{0.2cm}
    \caption{Complete comparisons of state-of-the-art driving world models in terms of \emph{Geometric Discrepancy} in WorldLens.}
    \vspace{-0.2cm}
    \label{tab:supp_recon_geometric_discrepancy}
    \resizebox{\linewidth}{!}{
    \begin{tabular}{r|cccccc|c}
        \toprule
        \multirow{2}{*}{$\mathcal{S}_\mathrm{GeoA}(\cdot)$} & \textbf{MagicDrive} & \textbf{DreamForge}  & \textbf{DriveDreamer-2} & \textbf{OpenDWM} & \textbf{~DiST-4D~} & $\mathcal{X}$\textbf{-Scene} & \textcolor{gray}{\textbf{Empirical}}
        \\
        & \textcolor{gray}{\small[ICLR'24]} & \textcolor{gray}{\small[arXiv'24]} & \textcolor{gray}{\small[AAAI'25]} & \textcolor{gray}{\small[CVPR'25]} & \textcolor{gray}{\small[ICCV'25]} & \textcolor{gray}{\small[NeurIPS'25]} & \textcolor{gray}{\textbf{Max}}
        \\\midrule
        RMSE~($\downarrow$) & $4.116$ & $4.166$ & $2.869$ & $3.130$ & $2.969$ & $3.594$ & \cellcolor{gray!7}\textcolor{gray}{-}
        \\
        Abs Rel~($\downarrow$) & \cellcolor{w_blue!20}$0.115$ & \cellcolor{w_blue!20}$0.105$ & \cellcolor{w_blue!20}$0.073$ & \cellcolor{w_blue!20}$0.088$ & \cellcolor{w_blue!20}$0.080$ & \cellcolor{w_blue!20}$0.096$ & \cellcolor{gray!7}\textcolor{gray}{-}
        \\
        $\delta_1$~($\uparrow$) & $0.856$ & $0.874$ & $0.923$ & $0.914$ & $0.910$ & $0.889$ & \cellcolor{gray!7}\textcolor{gray}{-}
        \\
        $\delta_2$~($\uparrow$) & $0.925$ & $0.940$ & $0.968$ & $0.961$ & $0.962$ & $0.946$ & \cellcolor{gray!7}\textcolor{gray}{-}
        \\
        $\delta_3$~($\uparrow$) & $0.953$ & $0.966$ & $0.983$ & $0.978$ & $0.981$ & $0.969$ & \cellcolor{gray!7}\textcolor{gray}{-}
        \\
        \bottomrule
    \end{tabular}}
\end{table*}

%% file: sections_supp/dimensions/recon_novel_view_quality.tex
\subsection{~Novel-View Quality}
\label{subsec:recon_nvq}

\subsubsection{~Definition}\label{subsubsec:recon_nvq_definition}
Novel View Quality (NVQ) assesses the perceptual quality of rendered frames from unseen camera trajectories, complementing Novel View Fidelity by focusing on frame-level realism rather than distributional similarity. For each novel-view trajectory, we render novel-view videos from reconstructed radiance fields and evaluate the perceptual quality of each frame using the pretrained MUSIQ model~\cite{ke2021musiq}.
The novel-view trajectories are:
\begin{itemize}
    \item ``\texttt{front\_center\_interp}'', which smoothly interpolates along the original front-center (ID~0) camera path by selecting four key poses at indices $0$, $\lfloor N/4\rfloor$, $\lfloor N/2\rfloor$, and $\lfloor3N/4\rfloor$, and generating intermediate 4{\texttimes}4 poses through linear translation and spherical linear interpolation (Slerp) of orientations. 
    \vspace{-0.1cm}
    \item ``\texttt{s\_curve}'', which constructs an S-shaped trajectory by traversing five key poses from front-left (ID~1), front-center (ID~0), and front-right (ID~2) cameras, at indices $(0)$, $\lfloor N/4\rfloor$, $\lfloor N/2\rfloor$, $\lfloor3N/4\rfloor$, and $(N{-}1)$, yielding a smooth left–center–right–center motion.
    \vspace{-0.1cm}
    \item ``\texttt{lateral\_offset}'', which generates a parallel-view sequence by shifting each front-camera pose (ID~0) laterally by a fixed offset along its local $+x$ axis while preserving orientation, followed by temporal resampling through linear and Slerp interpolation. All trajectories are resampled to a fixed target length.
\end{itemize}

\subsubsection{~Formulation}
Given the re-rendered novel-view videos 
$y_j^{\ast}=\{\mathrm{Render}(\hat{\mathcal{R}}_j,\,\mathrm{Pose}^{\ast}(t))\}_{t=1}^{T}$ 
under any of the novel-view settings, we compute frame-level perceptual quality scores via the pretrained image-quality assessor $\phi_{\mathrm{MUSIQ}}(\cdot)$.

Each frame receives a quality score $q_j^{(t)}=\phi_{\mathrm{MUSIQ}}(y_j^{\ast(t)})$, and the overall dataset-level Novel View Quality is obtained by averaging across all frames and videos:
\begin{equation}
\boxed{\;
\mathcal{S}_{\mathrm{NVQ}}(\mathcal{Y})
=\tfrac{1}{N_gT}\sum\nolimits_{j=1}^{N_g}\sum\nolimits_{t=1}^{T} q_j^{(t)}
\;}
\label{eq:novel_view_quality}
\end{equation}
Higher $\mathcal{S}_{\mathrm{NVQ}}$ indicates better perceptual quality of novel-view renderings, reflecting sharper appearance, fewer artifacts, and more realistic content across unseen trajectories.

\subsubsection{~Implementation Details}
We render videos from four novel viewpoints using the Gaussian Fields trained by each world model, following the definition provided in Section \ref{subsubsec:recon_nvq_definition}, where the lateral offset is set to 1~m. Novel-view image quality is assessed using the pretrained MUSIQ model \cite{ke2021musiq}. Each rendered novel-view video is processed frame-by-frame (resized to a maximum spatial dimension of $512$ pixels), and the MUSIQ scores are averaged across all frames and videos within each view condition.

\vspace{-0.1cm}
\subsubsection{~Examples}
\begin{figure}[t]
    \centering
    \begin{subfigure}[h]{\textwidth}
        \centering
        \includegraphics[width=\linewidth]{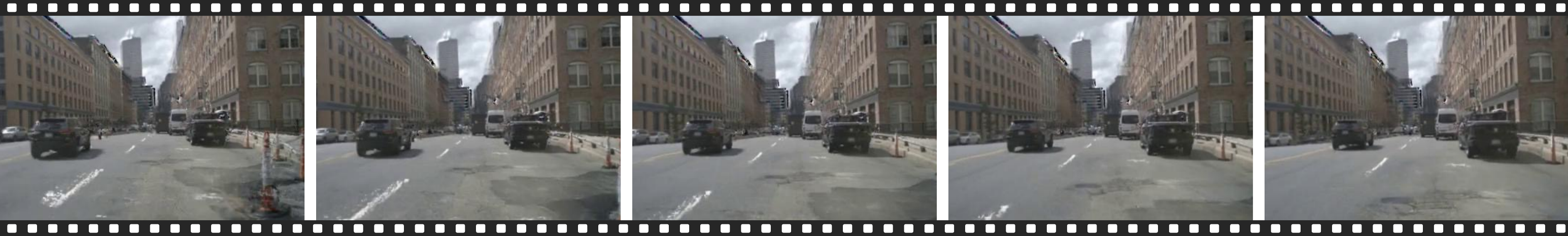}
        \caption{Good example in the \emph{Novel-View Quality} dimension (Score: \textcolor{w_blue}{$54.90\%$})}
        \label{fig:recon_novel_view_quality_1}
    \end{subfigure}
    \\[1ex]
    \begin{subfigure}[h]{\textwidth}
        \centering
        \includegraphics[width=\linewidth]{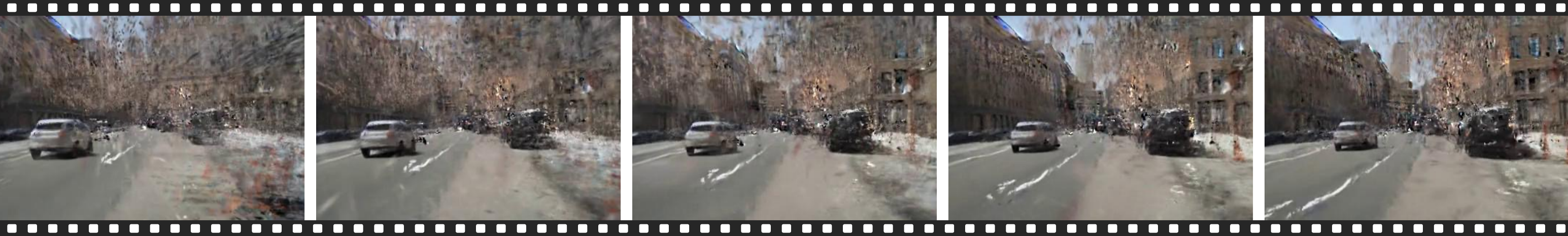}
        \caption{Bad example in the \emph{Novel-View Quality} dimension (Score: \textcolor{red}{$22.77\%$})}
        \label{fig:recon_novel_view_quality_2}
    \end{subfigure}
    \\[3.5ex]
    \begin{subfigure}[h]{\textwidth}
        \centering
        \includegraphics[width=\linewidth]{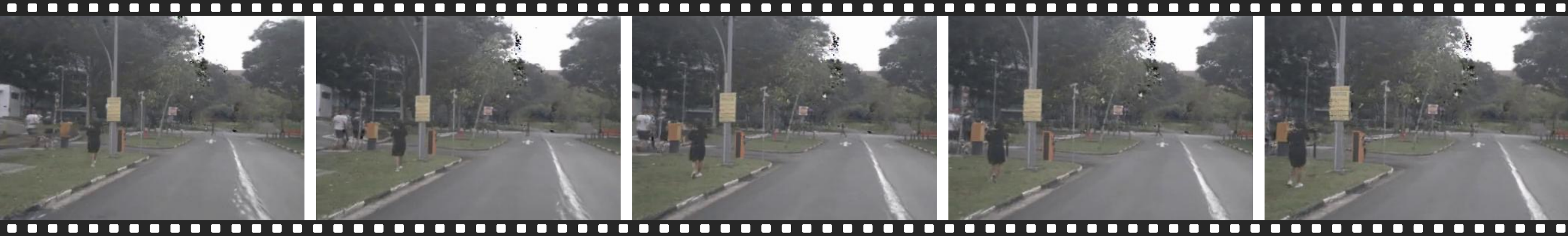}
        \caption{Good example in the \emph{Novel-View Quality} dimension (Score: \textcolor{w_blue}{$48.68\%$})}
        \label{fig:recon_novel_view_quality_3}
    \end{subfigure}
    \\[1ex]
    \begin{subfigure}[h]{\textwidth}
        \centering
        \includegraphics[width=\linewidth]{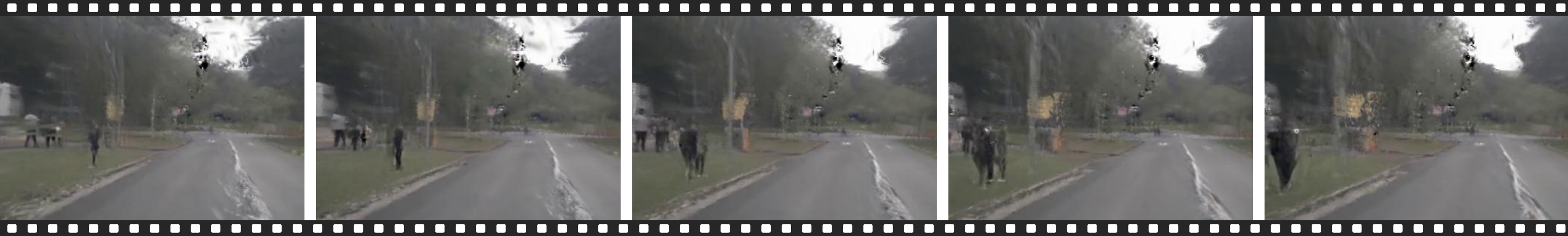}
        \caption{Bad example in the \emph{Novel-View Quality} dimension (Score: \textcolor{red}{$25.92\%$})}
        \label{fig:recon_novel_view_quality_4}
    \end{subfigure}
    \\[3.5ex]
    \begin{subfigure}[h]{\textwidth}
        \centering
        \includegraphics[width=\linewidth]{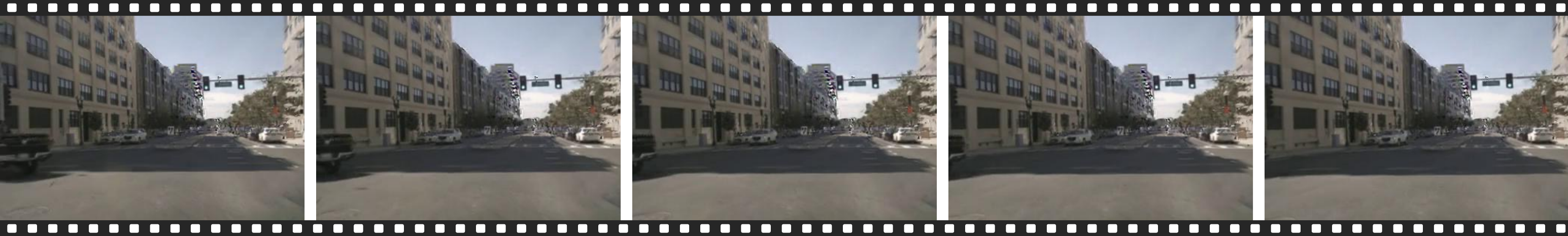}
        \caption{Good example in the \emph{Novel-View Quality} dimension (Score: \textcolor{w_blue}{$53.89\%$})}
        \label{fig:recon_novel_view_quality_5}
    \end{subfigure}
    \\[1ex]
    \begin{subfigure}[h]{\textwidth}
        \centering
        \includegraphics[width=\linewidth]{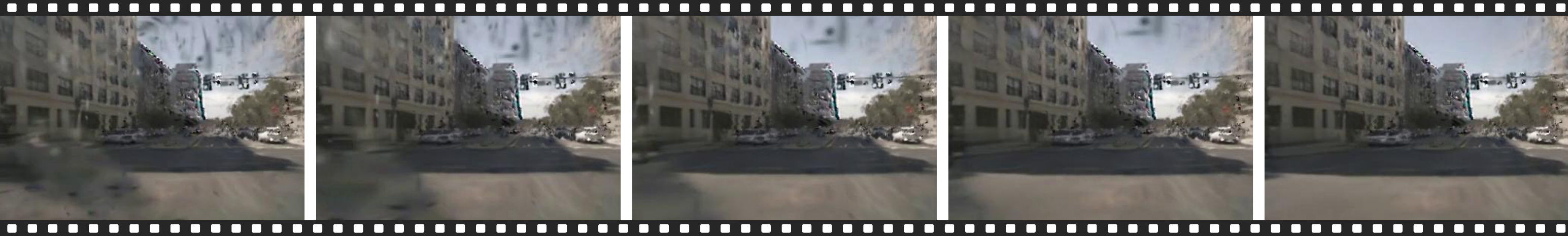}
        \caption{Bad example in the \emph{Novel-View Quality} dimension (Score: \textcolor{red}{$28.12\%$})}
        \label{fig:recon_novel_view_quality_6}
    \end{subfigure}
    \vspace{-0.1cm}
    \caption{Examples of ``good'' and ``bad'' reconstruction qualities in terms of \emph{Novel-View Quality} in WorldLens.}
    \label{fig:recon_novel_view_quality}
\end{figure}

Figure~\ref{fig:recon_novel_view_quality} provides typical examples of videos with good and bad quality in terms of \emph{Novel-View Quality}.

\vspace{-0.1cm}
\subsubsection{~Evaluation \& Analysis}
Table~\ref{tab:supp_recon_novel_view_quality} provides the complete results of models in terms of \emph{Novel-View Quality}.

\begin{table*}[h]
    \centering
    \vspace{0.2cm}
    \caption{Complete comparisons of state-of-the-art driving world models in terms of \emph{Novel-View Quality} in WorldLens.}
    \vspace{-0.2cm}
    \label{tab:supp_recon_novel_view_quality}
    \resizebox{\linewidth}{!}{
    \begin{tabular}{r|cccccc|c}
        \toprule
        \multirow{2}{*}{$\mathcal{S}_\mathrm{NVQ}(\cdot)$} & \textbf{MagicDrive} & \textbf{DreamForge}  & \textbf{DriveDreamer-2} & \textbf{OpenDWM} & \textbf{~DiST-4D~} & $\mathcal{X}$\textbf{-Scene} & \textcolor{gray}{\textbf{Empirical}}
        \\
        & \textcolor{gray}{\small[ICLR'24]} & \textcolor{gray}{\small[arXiv'24]} & \textcolor{gray}{\small[AAAI'25]} & \textcolor{gray}{\small[CVPR'25]} & \textcolor{gray}{\small[ICCV'25]} & \textcolor{gray}{\small[NeurIPS'25]} & \textcolor{gray}{\textbf{Max}}
        \\\midrule
        Center Interpolation~($\uparrow$) & $39.31\%$ & $42.66\%$ & $37.40\%$ & $40.71\%$ & $44.67\%$ & $38.11\%$ & \cellcolor{gray!7}\textcolor{gray}{-}
        \\
        S-Curve~($\uparrow$) & $39.67\%$ & $41.57\%$ & $35.15\%$ & $38.99\%$ & $42.64\%$ & $38.07\%$ &\cellcolor{gray!7}\textcolor{gray}{-}
        \\
        Left Lateral Offset~($\uparrow$) & $40.28\%$ & $40.56\%$ & $36.03\%$ & $39.20\%$ & $42.64\%$ & $38.25\%$ &\cellcolor{gray!7}\textcolor{gray}{-}
        \\
        Right Lateral Offset~($\uparrow$) & $40.02\%$ & $40.14\%$ & $35.82\%$ & $39.24\%$ & $42.42\%$ & $37.74\%$ & \cellcolor{gray!7}\textcolor{gray}{-}
        \\
        \textbf{Average~($\uparrow$)} & \cellcolor{w_blue!20}$39.82\%$ & \cellcolor{w_blue!20}$41.23\%$ & \cellcolor{w_blue!20}$36.10\%$ & \cellcolor{w_blue!20}$39.54\%$ & \cellcolor{w_blue!20}$43.09\%$ & \cellcolor{w_blue!20}$38.04\%$ & \cellcolor{gray!7}\textcolor{gray}{-}
        \\
        \bottomrule
    \end{tabular}}
    \vspace{-0.3cm}
\end{table*}

%% file: sections_supp/dimensions/recon_novel_view_discrepancy.tex
\subsection{~Novel-View Discrepancy}

\subsubsection{~Definition}
Novel-View Discrepancy measures the perceptual realism of newly rendered videos under unseen camera trajectories reconstructed from generated scenes.

Given the reconstructed neural radiance field of each generated video, we render novel-view sequences at held-out camera poses and compare them against ground-truth novel-view renderings of the corresponding real scenes.

\subsubsection{~Formulation}
Let $\hat{\mathcal{R}}_j$ and $\hat{\mathcal{R}}_j^{\mathrm{gt}}$ denote the reconstructed radiance fields from the generated and ground-truth videos, respectively.  
Rendering each field along a novel trajectory $\{\mathrm{Pose}^{\ast}(t)\}_{t=1}^{T}$ yields two new video sequences:
\begin{align}
y_j^{\ast}&=\{\mathrm{Render}(\hat{\mathcal{R}}_j,\,\mathrm{Pose}^{\ast}(t))\}_{t=1}^{T}~,
\\
x_j^{\ast}&=\{\mathrm{Render}(\hat{\mathcal{R}}_j^{\mathrm{gt}},\,\mathrm{Pose}^{\ast}(t))\}_{t=1}^{T}~.
\end{align}
We compute the Fréchet Video Distance (FVD) between the distributions of generated and ground-truth novel-view videos using Eq.~\eqref{eq:perceptual_fidelity}, where the feature extractor $\phi_{\mathrm{PF}}$ (I3D on Kinetics) remains the same.

The dataset-level \emph{Novel View Fidelity} is thus defined as:
\begin{equation}
\boxed{\;
\mathcal{S}_{\mathrm{NVD}}(\mathcal{Y})
=\mathcal{S}_{\mathrm{PF}}\big(\{x_j^{\ast}\},\{y_j^{\ast}\}\big)
\;}
\label{eq:novel_view_fidelity}
\end{equation}
Lower $\mathcal{S}_{\mathrm{NVD}}$ indicates higher perceptual fidelity of the reconstructed scenes when viewed from unseen camera trajectories.

\subsubsection{~Implementation Details}
The selection of novel viewpoints and the rendering configurations are kept consistent with those in Section~\ref{subsec:recon_nvq}.
For \emph{Novel-View Discrepancy}, the calculation process follows the same setting of Section~\ref{subsec:fvd}.

\subsubsection{~Example}
\begin{figure}[t]
    \centering
    \begin{subfigure}[h]{\textwidth}
        \centering
        \includegraphics[width=\linewidth]{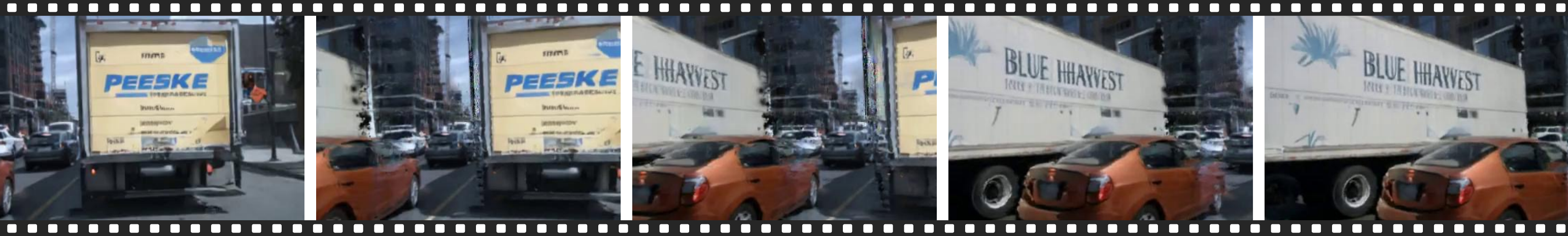}
        \caption{Good example in the \emph{Novel-View Discrepancy} dimension}
        \label{fig:recon_novel_view_discrepancy_1}
    \end{subfigure}
    \\[1ex]
    \begin{subfigure}[h]{\textwidth}
        \centering
        \includegraphics[width=\linewidth]{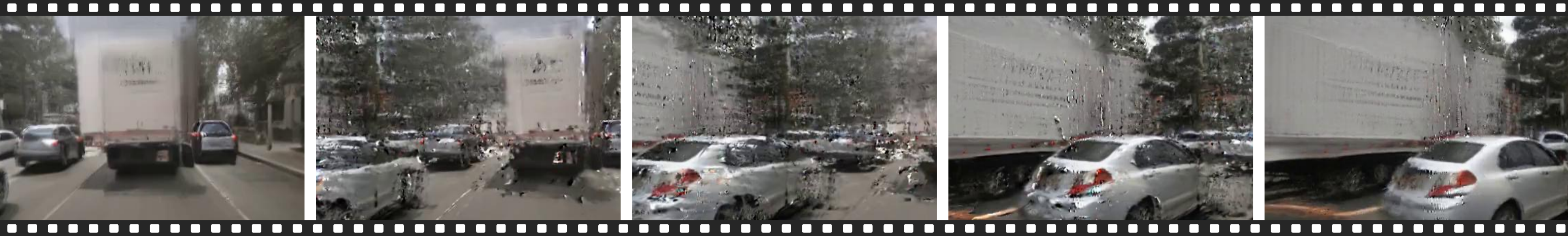}
        \caption{Bad example in the \emph{Novel-View Discrepancy} dimension}
        \label{fig:recon_novel_view_discrepancy_2}
    \end{subfigure}
    \\[3.5ex]
    \begin{subfigure}[h]{\textwidth}
        \centering
        \includegraphics[width=\linewidth]{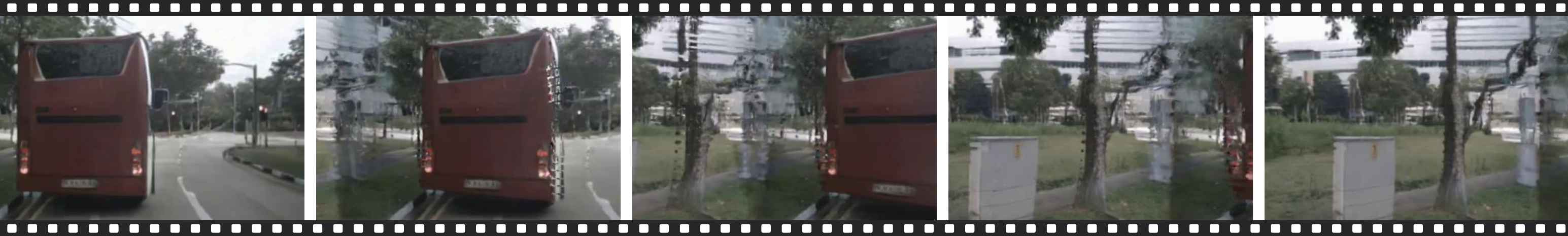}
        \caption{Good example in the \emph{Novel-View Discrepancy} dimension}
        \label{fig:recon_novel_view_discrepancy_3}
    \end{subfigure}
    \\[1ex]
    \begin{subfigure}[h]{\textwidth}
        \centering
        \includegraphics[width=\linewidth]{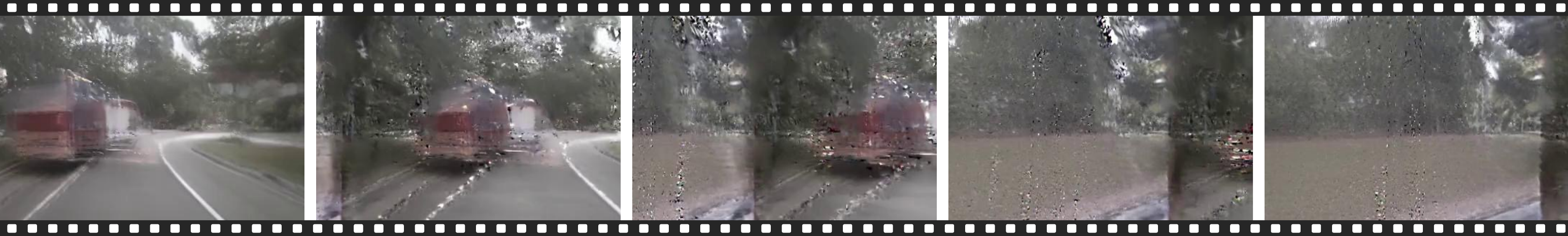}
        \caption{Bad example in the \emph{Novel-View Discrepancy} dimension}
        \label{fig:recon_novel_view_discrepancy_4}
    \end{subfigure}
    \\[3.5ex]
    \begin{subfigure}[h]{\textwidth}
        \centering
        \includegraphics[width=\linewidth]{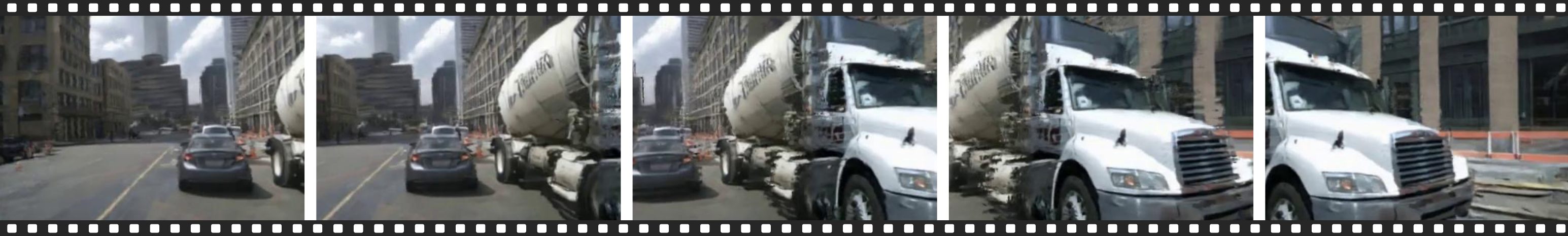}
        \caption{Good example in the \emph{Novel-View Discrepancy} dimension}
        \label{fig:recon_novel_view_discrepancy_5}
    \end{subfigure}
    \\[1ex]
    \begin{subfigure}[h]{\textwidth}
        \centering
        \includegraphics[width=\linewidth]{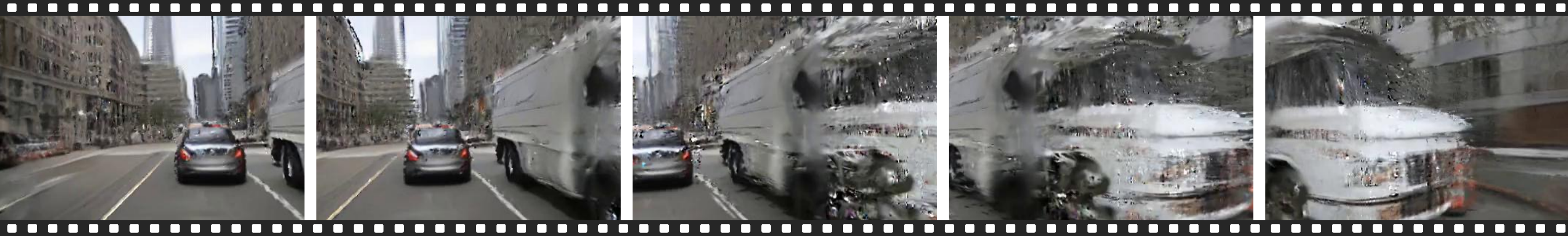}
        \caption{Bad example in the \emph{Novel-View Discrepancy} dimension}
        \label{fig:recon_novel_view_discrepancy_6}
    \end{subfigure}
    \vspace{-0.1cm}
    \caption{Examples of ``good'' and ``bad'' reconstruction qualities in terms of \emph{Novel-View Discrepancy} in WorldLens.}
    \label{fig:recon_novel_view_discrepancy}
\end{figure}

Figure~\ref{fig:recon_novel_view_discrepancy} provides typical examples of videos with good and bad quality in terms of \emph{Novel-View Discrepancy}.

\subsubsection{~Evaluation \& Analysis}
Table~\ref{tab:supp_recon_novel_view_discrepancy} provides the complete results of models in terms of \emph{Novel-View Discrepancy}.

\begin{table*}[h]
    \centering
    \vspace{0.2cm}
    \caption{Complete comparisons of state-of-the-art driving world models in terms of \emph{Novel-View Discrepancy} in WorldLens.}
    \vspace{-0.2cm}
    \label{tab:supp_recon_novel_view_discrepancy}
    \resizebox{\linewidth}{!}{
    \begin{tabular}{r|cccccc|c}
        \toprule
        \multirow{2}{*}{$\mathcal{S}_\mathrm{NVD}(\cdot)$} & \textbf{MagicDrive} & \textbf{DreamForge}  & \textbf{DriveDreamer-2} & \textbf{OpenDWM} & \textbf{~DiST-4D~} & $\mathcal{X}$\textbf{-Scene} & \textcolor{gray}{\textbf{Empirical}}
        \\
        & \textcolor{gray}{\small[ICLR'24]} & \textcolor{gray}{\small[arXiv'24]} & \textcolor{gray}{\small[AAAI'25]} & \textcolor{gray}{\small[CVPR'25]} & \textcolor{gray}{\small[ICCV'25]} & \textcolor{gray}{\small[NeurIPS'25]} & \textcolor{gray}{\textbf{Max}}
        \\\midrule
        Center Interpolation~($\downarrow$) & $448.62$ & $403.47$ & $259.96$ & $339.85$ & $190.17$ & $376.67$ & \cellcolor{gray!7}\textcolor{gray}{-}
        \\
        S-Curve~($\downarrow$) & $281.91$ & $171.93$ & $132.68$ & $159.84$ & $96.90$ & $219.89$ &\cellcolor{gray!7}\textcolor{gray}{-}
        \\
        Left Lateral Offset~($\downarrow$) & $492.97$ & $400.68$ & $326.95$ & $318.26$ & $237.08$ & $435.50$ &\cellcolor{gray!7}\textcolor{gray}{-}
        \\
        Right Lateral Offset~($\downarrow$) & $485.70$ & $414.72$ & $320.05$ & $332.97$ & $245.42$ & $430.78$ & \cellcolor{gray!7}\textcolor{gray}{-}
        \\
        \textbf{Average~($\downarrow$)} & \cellcolor{w_blue!20}$427.30$ & \cellcolor{w_blue!20}$347.70$ &
        \cellcolor{w_blue!20}$259.91$ &
        \cellcolor{w_blue!20}$287.73$ &
        \cellcolor{w_blue!20}$192.39$ &
        \cellcolor{w_blue!20}$365.71$ &
        \cellcolor{gray!7}\textcolor{gray}{-}
        \\
        \bottomrule
    \end{tabular}}
    \vspace{-0.2cm}
\end{table*}

%% file: sections_supp/4_action_follow.tex
\section{~Aspect 3: Action-Following}
In this section, we evaluate the \textbf{Action-Following} capability of driving world models, which reflects how well the generated videos preserve the functional cues necessary for downstream decision-making and control. Here, we assess the \emph{functional alignment} between generated content and real-world driving behavior. Specifically, we examine how the visual information synthesized influences an end-to-end planning agent in both \textbf{open-loop} and \textbf{closed-loop} simulation settings. A model with strong action-following ability should not only generate visually convincing scenes but also guide a pretrained planner to produce trajectories and control actions that are consistent with those derived from real-world videos.

\input{sections_supp/dimensions/act_displacement_error}

\clearpage\clearpage
\input{sections_supp/dimensions/act_open_loop_adherence}

\clearpage\clearpage
\input{sections_supp/dimensions/act_route_completion}

\clearpage\clearpage
\input{sections_supp/dimensions/act_closed_loop_adherence}

%% file: sections_supp/dimensions/act_displacement_error.tex
\subsection{~Displacement Error}

\subsubsection{~Definition}
Displacement Error (L2) evaluates the functional consistency of generated videos on the downstream task of motion planning. Instead of measuring perceptual realism or pixel-level accuracy, this metric assesses whether a generated video can serve as a reliable input for an end-to-end planner. It measures how closely the predicted trajectory inferred from a generated video aligns with the trajectory predicted from the corresponding ground-truth video. A lower displacement error indicates that the generated sequence preserves the semantic and motion cues necessary for robust trajectory forecasting, demonstrating that it is not only visually plausible but also functionally faithful to real-world driving dynamics.

\subsubsection{~Formulation}
We employ a pretrained end-to-end planning network $\psi_{\mathrm{Plan}}(\cdot)$ to predict trajectories from both generated and ground-truth videos.  
Given paired sequences $y_j$ and $x_j$, the model produces corresponding planned trajectories
\[
\hat{\tau}_j^{\mathrm{gen}}=\psi_{\mathrm{Plan}}(y_j), 
\qquad 
\hat{\tau}_j^{\mathrm{gt}}=\psi_{\mathrm{Plan}}(x_j),
\]
where each trajectory $\hat{\tau}\!\in\!\mathbb{R}^{T_p\times 2}$ contains $T_p$ future waypoints in 2D ground-plane coordinates. The Displacement Error is computed as the mean L2 distance between corresponding waypoints:
\begin{equation}
\boxed{\;
\mathcal{S}_{\mathrm{DE}}(\mathcal{Y})
=\frac{1}{N_gT_p}\sum_{j=1}^{N_g}
\sum_{t=1}^{T_p}
\big\|
\hat{\tau}_j^{\mathrm{gen}}(t)-\hat{\tau}_j^{\mathrm{gt}}(t)
\big\|_2
\;}
\label{eq:displacement_error}
\end{equation}
Lower $\mathcal{S}_{\mathrm{DE}}$ indicates that the generated videos induce planning behaviors that are more consistent with those derived from real-world observations, reflecting higher functional fidelity.

\subsubsection{~Implementation Details}
We conduct the \emph{Displacement Error} evaluation on the official nuScenes validation set, which consists of 150 diverse driving scenes. For each test case, the driving world model generates a video sequence conditioned on the initial context. These synthesized videos are then used as input for UniAD~\cite{hu2023uniad}, a state-of-the-art end-to-end planning network, to infer future ego-motion trajectories. Following the standard protocol, we extract the planned trajectory for a horizon of $1$ second (covering the immediate future waypoints). The \emph{Displacement Error} is calculated as the L2 distance between the trajectory predicted from the generated video and the trajectory predicted from the ground-truth video. This metric strictly isolates the impact of visual generation quality on downstream perception and planning accuracy in an open-loop setting.

\subsubsection{~Examples}
\begin{figure}[t]
    \centering
    \begin{subfigure}[h]{\textwidth}
        \centering
        \includegraphics[width=\linewidth]{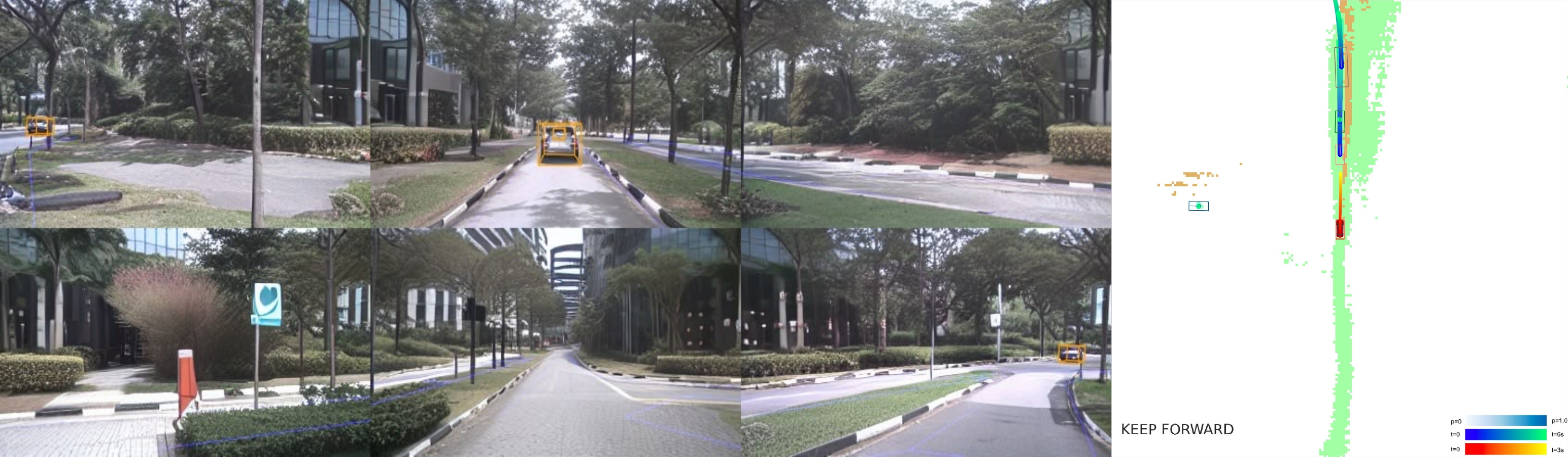}
        \caption{Good example in the \emph{Displacement Error} dimension (Score: \textcolor{w_blue}{$0.43$})}
        \label{fig:act_displacement_error_1}
    \end{subfigure}
    \\[5ex]
    \begin{subfigure}[h]{\textwidth}
        \centering
        \includegraphics[width=\linewidth]{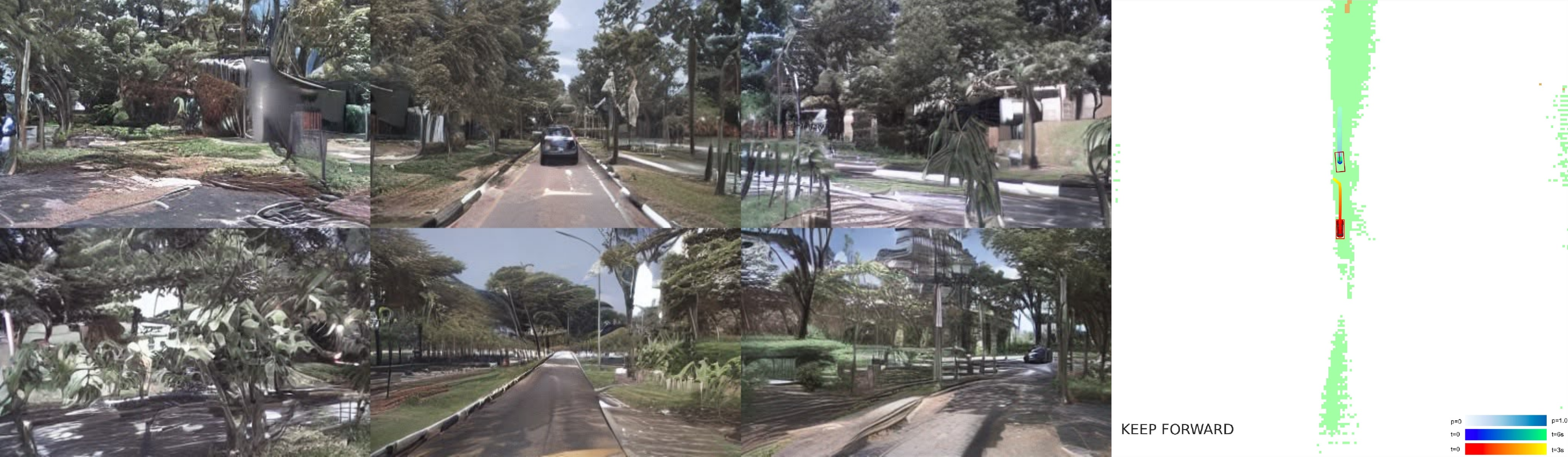}
        \caption{Bad example in the \emph{Displacement Error} dimension (Score: \textcolor{red}{0.63})}
        \label{fig:act_displacement_error_2}
    \end{subfigure}
    \\[5ex]
    \begin{subfigure}[h]{\textwidth}
        \centering
        \includegraphics[width=\linewidth]{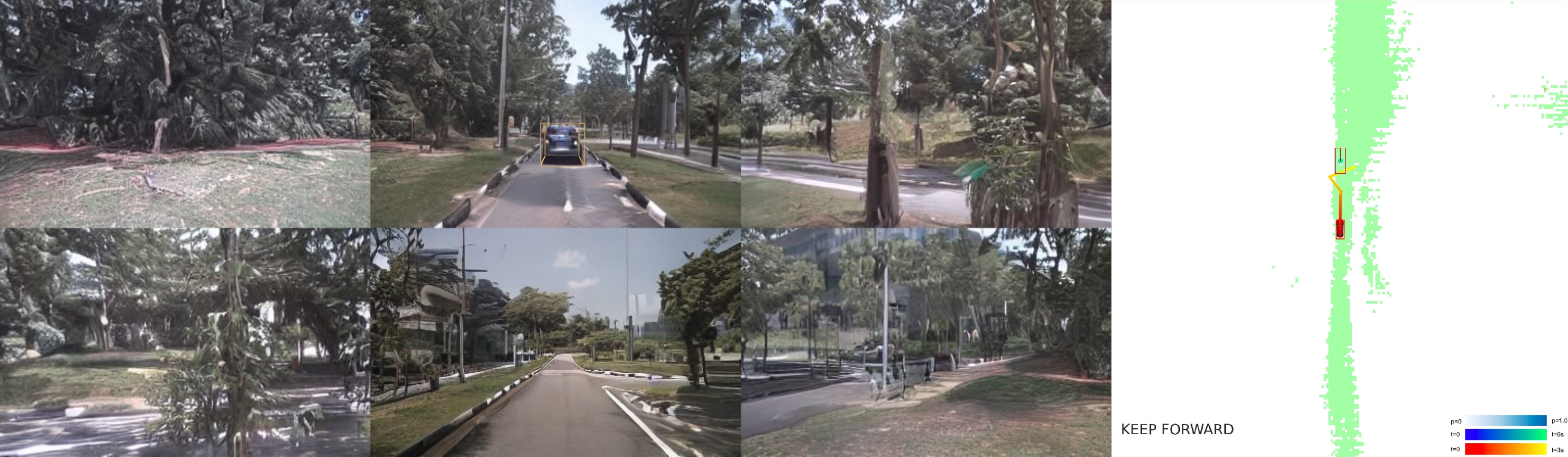}
        \caption{Bad example in the \emph{Displacement Error} dimension (Score: \textcolor{red}{0.71})}
        \label{fig:act_displacement_error_3}
    \end{subfigure}
    \vspace{-0.05cm}
    \caption{Examples of ``good'' and ``bad'' action-following performances in terms of \emph{Displacement Error} in WorldLens.}
    \label{fig:act_displacement_error}
\end{figure}

Figure~\ref{fig:act_displacement_error} provides typical examples of videos with good and bad quality in terms of \emph{Displacement Error}.

\subsubsection{~Evaluation \& Analysis}
Table~\ref{tab:supp_act_displacement_error} provides the complete results of models in terms of \emph{Displacement Error}.

\begin{table*}[h]
    \centering
    \caption{Complete comparisons of state-of-the-art driving world models in terms of \emph{Displacement Error} in WorldLens.}
    \vspace{-0.2cm}
    \label{tab:supp_act_displacement_error}
    \resizebox{\linewidth}{!}{
    \begin{tabular}{r|cccccc|c}
        \toprule
        \multirow{2}{*}{$\mathcal{S}_\mathrm{DE}(\cdot)$} & \textbf{MagicDrive} & \textbf{DreamForge} & \textbf{Panacea} & \textbf{DrivingSphere} & \textbf{MagicDrive-V2} & \textbf{RLGF} & \textcolor{gray}{\textbf{Empirical}}
        \\
        & \textcolor{gray}{\small[ICLR'24]} & \textcolor{gray}{\small[arXiv'24]} & \textcolor{gray}{\small [CVPR'24]} & \textcolor{gray}{\small[CVPR'25]} & \textcolor{gray}{\small[ICCV'25]}  & \textcolor{gray}{\small[NeurIPS'25]} & \textcolor{gray}{\textbf{Max}}
        \\\midrule
        \textbf{Total~($\downarrow$)} & \cellcolor{w_blue!20}$0.57$ & \cellcolor{w_blue!20}$0.57$ & \cellcolor{w_blue!20}$0.58$ & \cellcolor{w_blue!20}$0.55$ & \cellcolor{w_blue!20}$0.54$ & \cellcolor{w_blue!20}$0.53$ & \cellcolor{gray!7}\textcolor{gray}{$0.51$}
        \\
        \bottomrule
    \end{tabular}}
    \vspace{-0.2cm}
\end{table*}

%% file: sections_supp/dimensions/act_open_loop_adherence.tex
\subsection{~Open-Loop Adherence}

\subsubsection{~Definition}
Open-Loop Adherence evaluates the functional reliability of generated videos by measuring how well an end-to-end driving policy can perform when operating on the generated input in a non-reactive simulation environment. Following NAVSIM~\cite{dauner2024navsim}, we use the \emph{Predictive Driver Model Score (PDMS)} to quantify adherence between the policy behavior induced by generated videos and that observed under real data.

\subsubsection{~Formulation}
Given a pretrained planner $\psi_{\mathrm{Plan}}(\cdot)$ and its predicted trajectory $\hat{\tau}_j$ from a generated video $y_j$, we simulate the resulting ego motion over a fixed horizon (\emph{e.g.}, $4\,\mathrm{s}$) in a non-reactive setting where other agents follow their recorded trajectories. At each timestep, sub-scores are computed for: \emph{no collision} (NC), \emph{drivable-area compliance} (DAC), \emph{ego progress} (EP), \emph{time-to-collision} (TTC), and \emph{comfort} (C). Penalties (NC, DAC) suppress inadmissible behaviors, while the remaining terms are averaged with fixed weights. The PDMS is defined as:
\[
\mathrm{PDMS}
=\big(\!\!\!\prod_{m\in\{\mathrm{NC},\mathrm{DAC}\}}\!\!\!\mathrm{score}_m\big)
\cdot
\tfrac{\sum\nolimits_{w\in\{\mathrm{EP},\mathrm{TTC},\mathrm{C}\}}
\mathrm{weight}_w\,\mathrm{score}_w}
{\sum\nolimits_{w\in\{\mathrm{EP},\mathrm{TTC},\mathrm{C}\}}\mathrm{weight}_w}~,
\label{eq:pdms}
\]
with default weights $\mathrm{weight}_{\mathrm{EP}}{=}\mathrm{weight}_{\mathrm{TTC}}{=}5$ and $\mathrm{weight}_{\mathrm{C}}{=}2$ as in~\cite{dauner2024navsim}.  
We report the dataset-level score as the mean PDMS across all evaluated videos:
\begin{equation}
\boxed{\;
\mathcal{S}_{\mathrm{PDMS}}(\mathcal{Y})
=\tfrac{1}{N_g}\sum\nolimits_{j=1}^{N_g}\mathrm{PDMS}(y_j)
\;}
\end{equation}
Higher $\mathcal{S}_{\mathrm{PDMS}}$ indicates stronger alignment between generated and real scenes in terms of functional behavior.

\subsubsection{~Implementation Details} 
We support two map environments, \emph{singapore-onenorth} and \emph{boston-seaport}, aligned with the DriveArena platform~\cite{yang2024drivearena}. 
A total of five simulation sequences are defined for validation, enabling the evaluation of driving agents in both open-loop and closed-loop modes.
In our implementation, the traffic flow engine~\cite{wen2023limsim} operates at a frequency of $10$ Hz, while the control signals are set to $2$ Hz. Every $0.5$ simulation seconds, the 2D traffic flow engine updates its state and renders multi-view layouts as conditions for the video generation model. Video generation models use the last $3$ frames as reference images to generate $448\times 800$ images, which are subsequently resized to $224\times 400$ to serve as input for the driving agent.

\subsubsection{~Examples}
\begin{figure}[t]
    \centering
    \begin{subfigure}[h]{\textwidth}
        \centering
        \includegraphics[width=\linewidth]{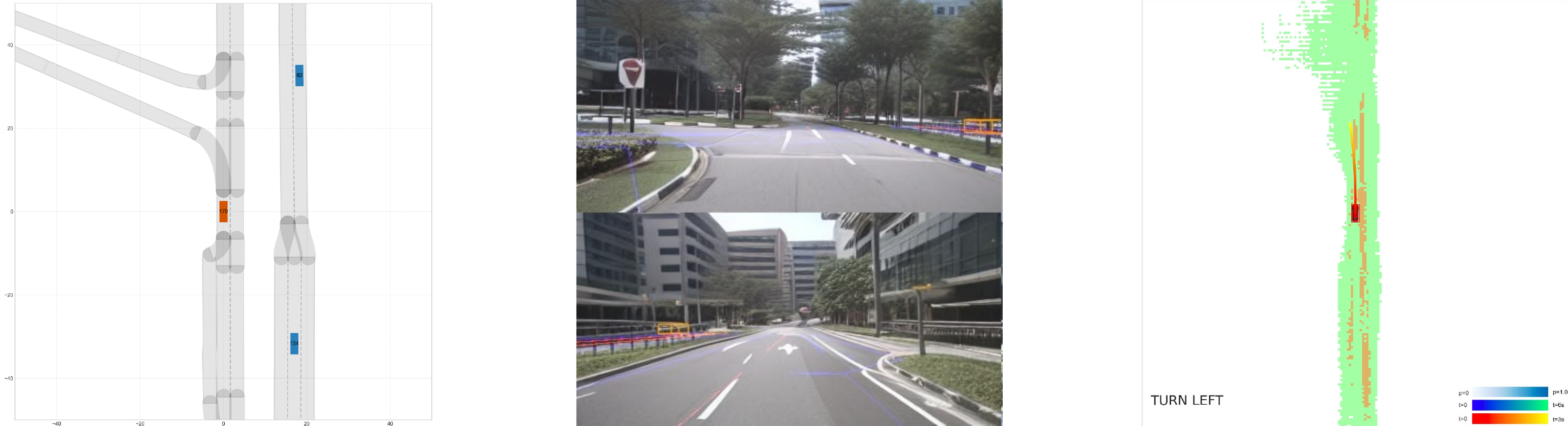}
        \caption{Good example in the \emph{Open-Loop Adherence} dimension (Score: \textcolor{w_blue}{$0.812$})}
        \label{fig:act_open_loop_1}
    \end{subfigure}
    \\[2ex]
    \begin{subfigure}[h]{\textwidth}
        \centering
        \includegraphics[width=\linewidth]{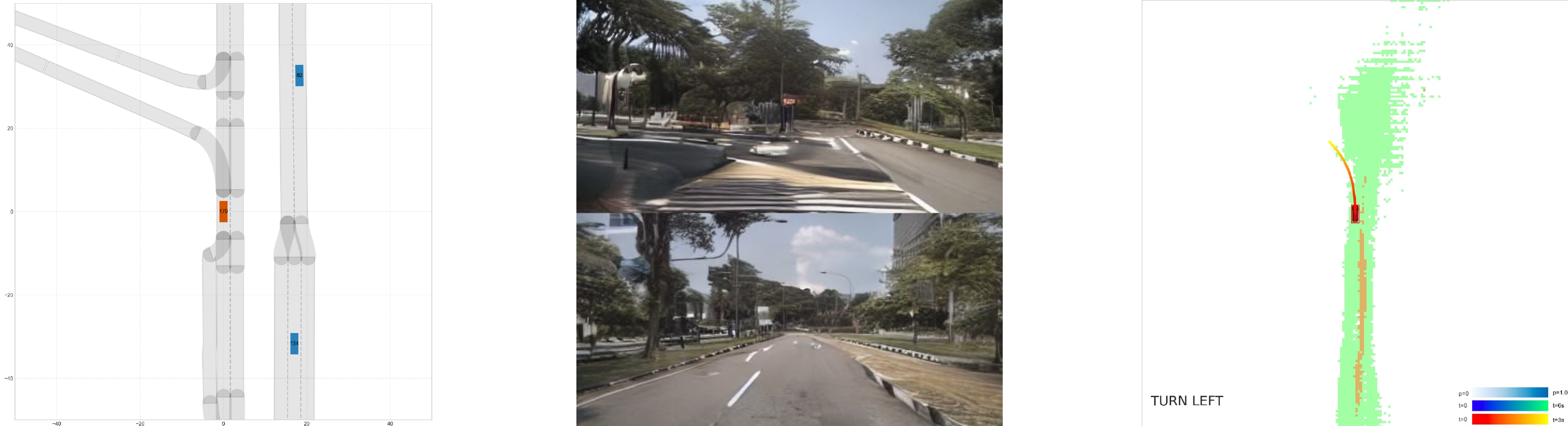}
        \caption{Bad example in the \emph{Open-Loop Adherence} dimension (Score: \textcolor{red}{0.708})}
        \label{fig:act_open_loop_2}
    \end{subfigure}
    \\[2ex]
    \begin{subfigure}[h]{\textwidth}
        \centering
        \includegraphics[width=\linewidth]{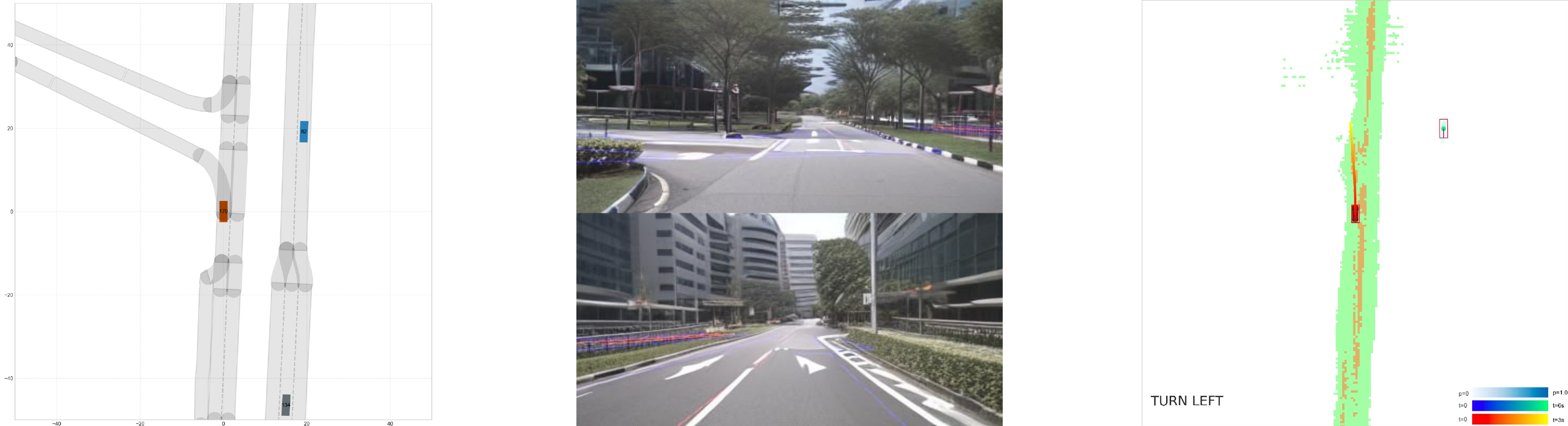}
        \caption{Good example in the \emph{Open-Loop Adherence} dimension (Score: \textcolor{w_blue}{$0.745$})}
        \label{fig:act_open_loop_3}
    \end{subfigure}
    \\[2ex]
    \begin{subfigure}[h]{\textwidth}
        \centering
        \includegraphics[width=\linewidth]{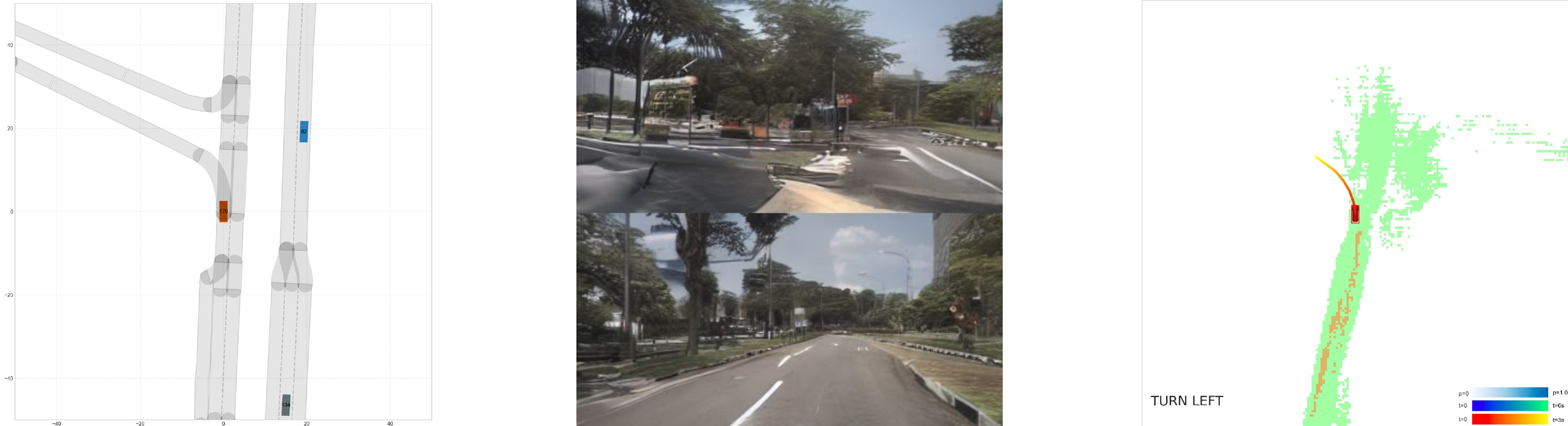}
        \caption{Bad example in the \emph{Open-Loop Adherence} dimension (Score: \textcolor{red}{$0.621$})}
        \label{fig:act_open_loop_4}
    \end{subfigure}
    \\
    \vspace{-0.1cm}
    \caption{Examples of ``good'' and ``bad'' action-following performances in terms of \emph{Open-Loop Adherence} in WorldLens.}
    \label{fig:act_open_loop}
\end{figure}

Figure~\ref{fig:act_open_loop} provides typical examples of videos with good and bad quality in terms of \emph{Open-Loop Adherence}.

\subsubsection{~Evaluation \& Analysis}
Table~\ref{tab:supp_act_open_loop} provides the complete results of models in terms of \emph{Open-Loop Adherence}.

\begin{table*}[h]
    \centering
    \vspace{0.2cm}
    \caption{Complete comparisons of state-of-the-art driving world models in terms of \emph{Open-Loop Adherence} in WorldLens.}
    \vspace{-0.2cm}
    \label{tab:supp_act_open_loop}
    \resizebox{\linewidth}{!}{
    \begin{tabular}{r|ccccc|c}
        \toprule
        \multirow{2}{*}{$\mathcal{S}_\mathrm{PDMS}(\cdot)$} & \textbf{MagicDrive} & \textbf{DreamForge} &  \textbf{DrivingSphere} & \textbf{MagicDrive-V2} & \textbf{RLGF} & \textcolor{gray}{\textbf{Empirical}}
        \\
        & \textcolor{gray}{\small[ICLR'24]} & \textcolor{gray}{\small[arXiv'24]}  & \textcolor{gray}{\small[CVPR'25]} & \textcolor{gray}{\small[ICCV'25]}  & \textcolor{gray}{\small[NeurIPS'25]} & \textcolor{gray}{\textbf{Max}}
        \\\midrule
        No Collision (NC, $\uparrow$) & $0.885$    & $0.915$ & $0.932$ & $0.968$ & $0.975$ & \cellcolor{gray!7}\textcolor{gray}{-}
        \\
        Compliance (DAC, $\uparrow$) & $0.955$    & $0.970$ &$ 0.978$ & $0.985$ & $0.988$ & \cellcolor{gray!7}\textcolor{gray}{-}
        \\
        Ego Progress (EP, $\uparrow$) &$0.825$    &$0.832$  &$0.835$  &$0.842$  &$0.838$  & \cellcolor{gray!7}\textcolor{gray}{-}
        \\
        Time-to-Collision (TTC, $\uparrow$) &$0.840$    &$0.855$  &$0.860$  &$0.865$  &$ 0.850$ & \cellcolor{gray!7}\textcolor{gray}{-}
        \\
        Comfort (C, $\uparrow$) & $0.815$  & $0.825$   & $0.830$ & $0.835$  & $0.828$ & \cellcolor{gray!7}\textcolor{gray}{-}
        \\
        \textbf{Total~($\uparrow$)} & \cellcolor{w_blue!20}$0.712$ & \cellcolor{w_blue!20}$0.755$  & \cellcolor{w_blue!20}$0.760$ & \cellcolor{w_blue!20}$0.789$ & \cellcolor{w_blue!20}$0.784$ & \cellcolor{gray!7}\textcolor{gray}{-}
        \\
        \bottomrule
    \end{tabular}
    }
    \vspace{-0.3cm}
\end{table*}

%% file: sections_supp/dimensions/act_route_completion.tex
\subsection{~Route Completion}
\label{sec:supp_act_route_completion}

\subsubsection{Definition}
Route Completion (RC) measures the ability of an autonomous driving agent to complete a predefined navigation route in closed-loop simulation. It quantifies the percentage of the total planned route distance successfully traveled by the ego agent before simulation termination (\emph{e.g.}, collision, off-road, or timeout). Higher RC values indicate better long-horizon stability and control consistency, reflecting how well the generated video enables the policy to sustain safe driving behavior throughout the route.

\subsubsection{Formulation}
Let $D_{\mathrm{total}}$ denote the total length of the planned route, and $D_{\mathrm{completed}}$ the distance actually traveled by the ego agent before termination. Following~\cite{yang2024drivearena,dauner2024navsim}, \emph{Route Completion} is defined as the ratio between the completed and total distances:
\[
\mathrm{RC}=\frac{D_{\mathrm{completed}}}{D_{\mathrm{total}}}~.
\label{eq:route_completion}
\]
We report the dataset-level metric as the mean RC across all evaluated closed-loop rollouts:
\begin{equation}
\boxed{\;
\mathcal{S}_{\mathrm{RC}}(\mathcal{Y})
=\tfrac{1}{N_g}\sum\nolimits_{j=1}^{N_g}\mathrm{RC}(y_j)
\;}
\end{equation}
Higher $\mathcal{S}_{\mathrm{RC}}$ indicates that the generated scenes enable the planner to complete longer portions of the route, implying greater action stability and environmental consistency.

\subsubsection{~Implementation Details}
Different from the open-loop evaluation (Displacement Error), both Route Completion and Closed-Loop Adherence are evaluated in a fully reactive closed-loop mode. In this setting, the ego-vehicle’s trajectory is not determined by pre-recorded logs but is driven by the agent's decisions.

Specifically, the planning agent processes the video generated by the world model, outputs a control signal, and this signal updates the ego-vehicle’s state within the simulator.

The world model then generates the next frame based on this new state, creating a continuous feedback loop. A simulation episode continues until one of the following termination criteria is met:

\begin{enumerate}
    \item \emph{Completion:} The agent successfully reaches the destination and finishes the predefined route.

    \item \emph{Failure:} The simulation is terminated early due to safety-critical infractions, specifically collision with other objects or driving off-road (exiting the drivable area).
\end{enumerate}

This setup evaluates the ability of the generative driving world model to support long-horizon consistency and error-free decision-making.

\subsubsection{~Examples}
\begin{figure}[t]
    \centering
    \begin{subfigure}[h]{\textwidth}
        \centering
        \includegraphics[width=\linewidth]{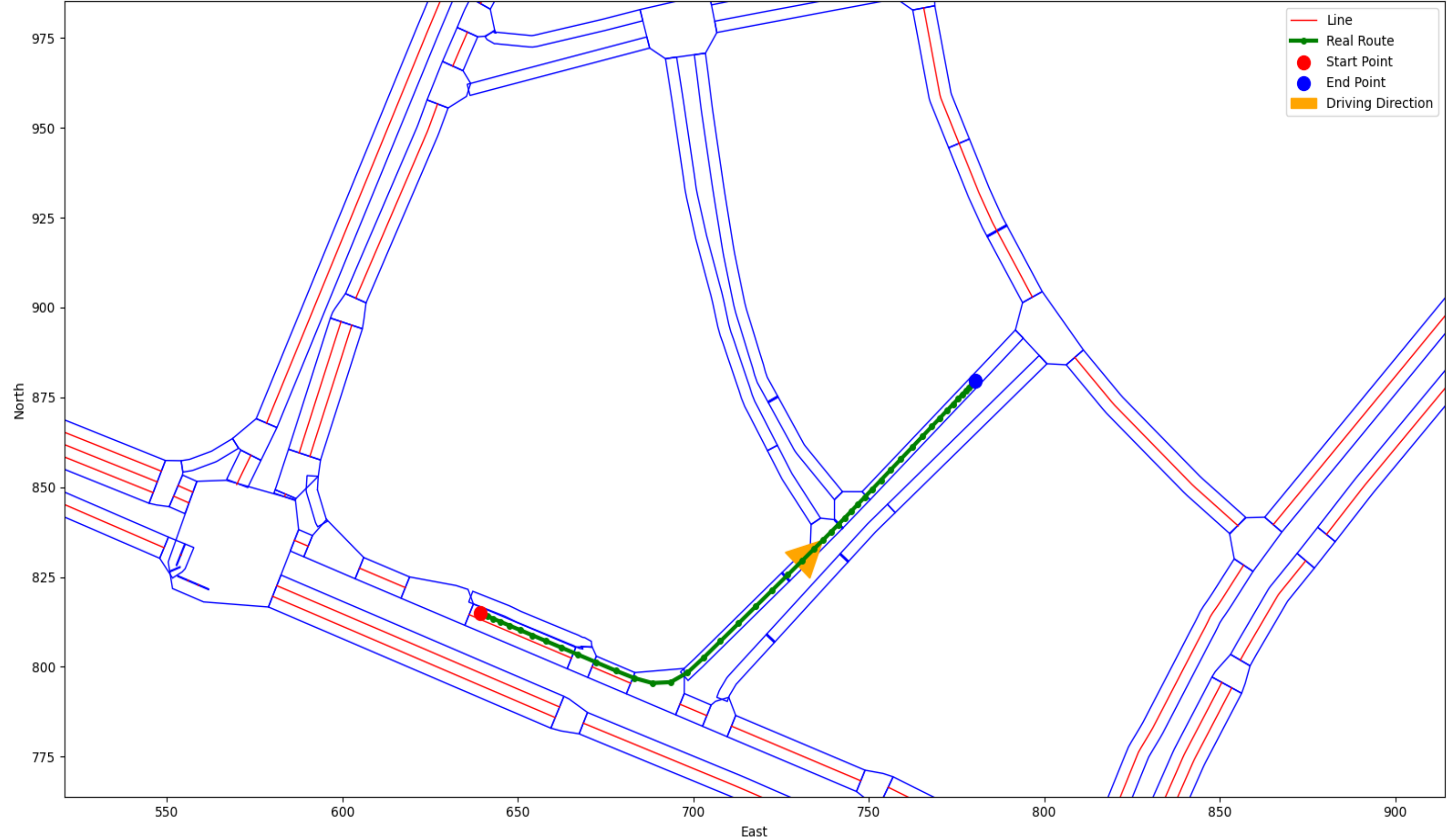}
        \caption{Good example in the \emph{Route Completion} dimension (Score: \textcolor{w_blue}{$18.7\%$})}
        \label{fig:act_route_completion_1}
    \end{subfigure}
    \\[3.5ex]
    \begin{subfigure}[h]{\textwidth}
        \centering
        \includegraphics[width=\linewidth]{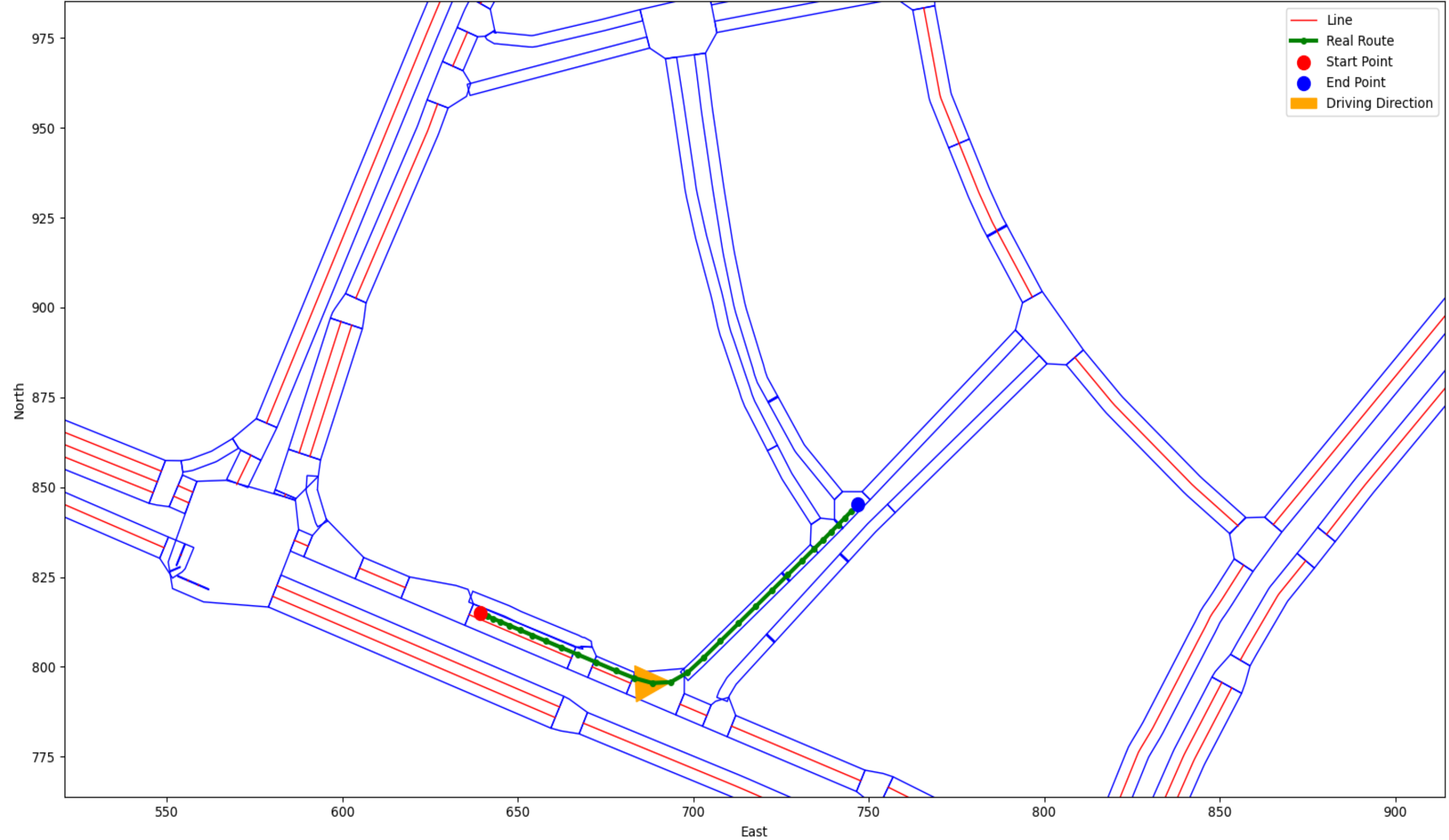}
        \caption{Bad example in the \emph{Route Completion} dimension (Score: \textcolor{red}{11.2\%})}
        \label{fig:act_route_completion_2}
    \end{subfigure}
    \\
    \vspace{-0.1cm}
    \caption{Examples of ``good'' and ``bad'' action-following performances in terms of \emph{Route Completion} in WorldLens.}
    \label{fig:act_route_completion}
\end{figure}

Figure~\ref{fig:act_route_completion} provides typical examples of videos with good and bad quality in terms of \emph{Route Completion}.

\subsubsection{~Evaluation \& Analysis}
Table~\ref{tab:supp_act_route_completion} provides the complete results of models in terms of \emph{Route Completion}.

\begin{table*}[h]
    \centering
    \vspace{0.2cm}
    \caption{Complete comparisons of state-of-the-art driving world models in terms of \emph{Route Completion} in WorldLens}
    \vspace{-0.2cm}
    \label{tab:supp_act_route_completion}
    \resizebox{\linewidth}{!}{
    \begin{tabular}{r|cccccc|c}
        \toprule
        \multirow{2}{*}{$\mathcal{S}_\mathrm{RC}(\cdot)$} & \textbf{MagicDrive} & \textbf{DreamForge} & \textbf{Panacea} & \textbf{DrivingSphere} & \textbf{MagicDrive-V2} & \textbf{RLGF} & \textcolor{gray}{\textbf{Empirical}}
        \\
        & \textcolor{gray}{\small[ICLR'24]} & \textcolor{gray}{\small[arXiv'24]} & \textcolor{gray}{\small [CVPR'24]} & \textcolor{gray}{\small[CVPR'25]} & \textcolor{gray}{\small[ICCV'25]}  & \textcolor{gray}{\small[NeurIPS'25]} & \textcolor{gray}{\textbf{Max}}
        \\\midrule
        \textbf{Total~($\uparrow$)} & \cellcolor{w_blue!20}$6.89\%$ & \cellcolor{w_blue!20}$10.23\%$ & \cellcolor{w_blue!20}- & \cellcolor{w_blue!20}$11.02\%$ & \cellcolor{w_blue!20}$12.31\%$ & \cellcolor{w_blue!20}$13.51\%$ & \cellcolor{gray!7}\textcolor{gray}{-}
        \\
        \bottomrule
    \end{tabular}}
    \vspace{-0.3cm}
\end{table*}

%% file: sections_supp/dimensions/act_closed_loop_adherence.tex
\subsection{~Closed-Loop Adherence}
\label{sec:closed}

\subsubsection{~Definition}
Closed-Loop Adherence measures the overall driving performance of an autonomous agent in a closed-loop simulation. It is represented by the \emph{Arena Driving Score (ADS)}~\cite{yang2024drivearena}, which jointly accounts for both driving quality and task completion.

While the PDMS score reflects the safety, comfort, and stability of the predicted trajectory, the Route Completion (RC) measures how much of the planned route is successfully finished without failure. The multiplicative formulation ensures that an agent must be both competent (high PDMS) and consistent (high RC) to achieve a strong overall score. Agents that drive perfectly but crash early, or complete the route with poor motion quality, will both be penalized accordingly.

\subsubsection{~Formulation}
Given the PDMS and RC metrics defined in \eqref{eq:pdms} and~\eqref{eq:route_completion}, the \emph{Arena Driving Score (ADS)} is computed as follows:
\[
\mathrm{ADS} = \mathrm{RC} \times \mathrm{PDMS}~,
\label{eq:ads}
\]
where $\mathrm{RC}\in[0,1]$ denotes route completion.
For a dataset of generated videos $\mathcal{Y}$, the final closed-loop adherence is reported as the mean ADS across all evaluated driving episodes:
\begin{equation}
\boxed{\;
\mathcal{S}_{\mathrm{ADS}}(\mathcal{Y})
=\tfrac{1}{N_g}\sum\nolimits_{j=1}^{N_g}\mathrm{ADS}(y_j)
\;}
\end{equation}
Higher $\mathcal{S}_{\mathrm{ADS}}$ indicates that the generated videos yield planners capable of both safe and complete driving behavior in closed-loop simulation.

\subsubsection{~Implementation Details}
\textit{Closed-Loop Adherence} shares the same experiment environment with \textit{Route Completion}. The implementation details can be found in Section~\ref{sec:supp_act_route_completion}.

\subsubsection{~Examples}
\begin{figure}[t]
    \centering
    \begin{subfigure}[h]{\textwidth}
        \centering
        \includegraphics[width=\linewidth]{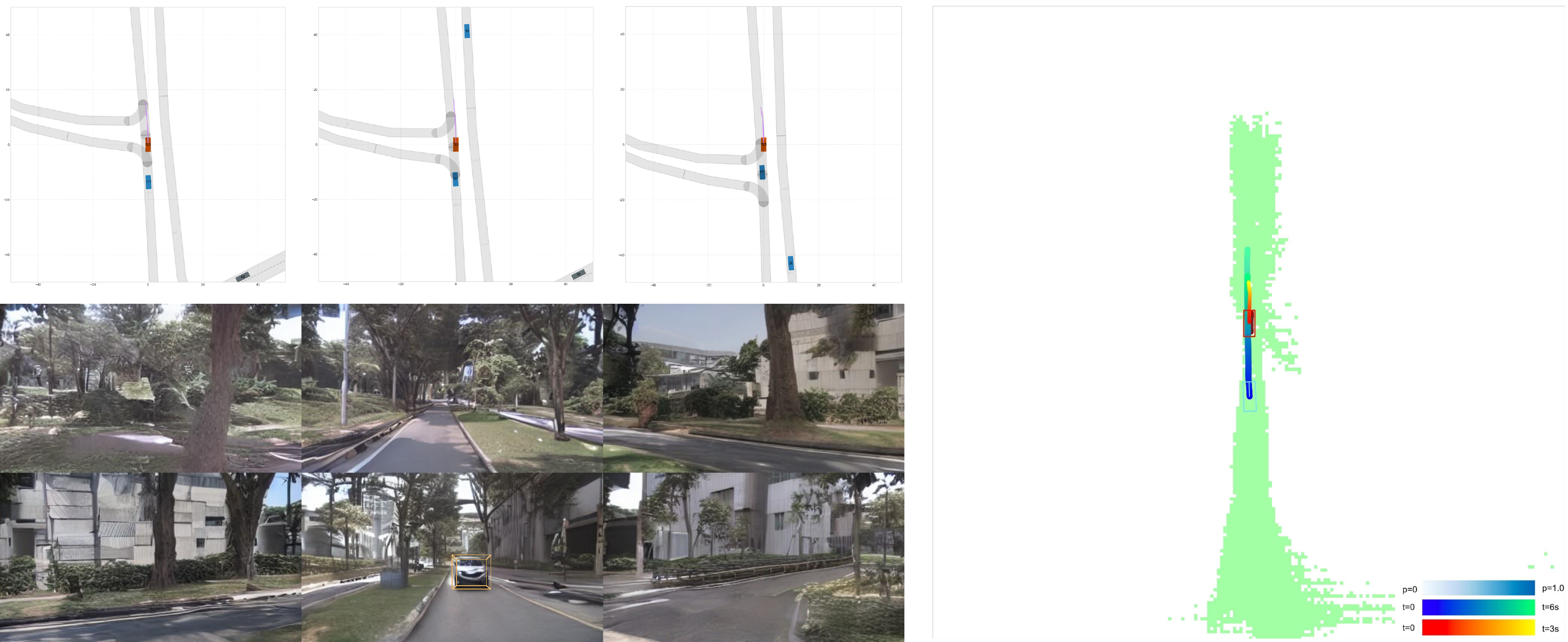}
        \caption{Good example in the \emph{Closed-Loop Adherence} dimension (Score: \textcolor{w_blue}{$0.103$})}
        \label{fig:act_closed_loop_1}
    \end{subfigure}
    \\[3.5ex]
    \begin{subfigure}[h]{\textwidth}
        \centering
        \includegraphics[width=\linewidth]{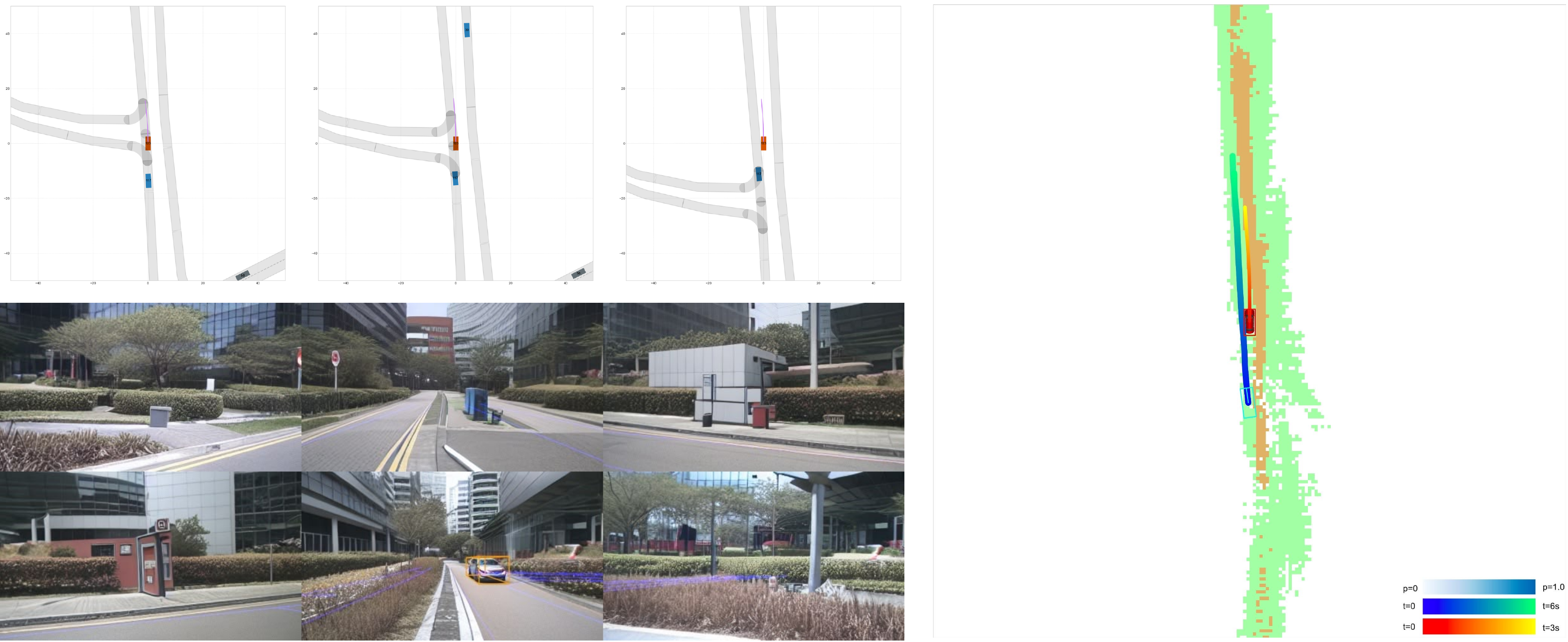}
        \caption{Bad example in the \emph{Closed-Loop Adherence} dimension (Score: \textcolor{red}{0.062})}
        \label{fig:act_closed_loop_2}
    \end{subfigure}
    \\
    \vspace{-0.05cm}
    \caption{Examples of ``good'' and ``bad'' action-following performances in terms of \emph{Closed-Loop Adherence} in WorldLens.}
    \label{fig:act_closed_loop}
\end{figure}

Figure~\ref{fig:act_closed_loop} provides typical examples of videos with good and bad quality in terms of \emph{Closed-Loop Adherence}.

\subsubsection{~Evaluation \& Analysis}
Table~\ref{tab:supp_act_closed_loop} provides the complete results of models in terms of \emph{Closed-Loop Adherence}.

\begin{table*}[h]
    \centering
    \vspace{0.3cm}
    \caption{Complete comparisons of state-of-the-art driving world models in terms of \emph{Closed-Loop Adherence} in WorldLens.}
    \vspace{-0.2cm}
    \label{tab:supp_act_closed_loop}
    \resizebox{\linewidth}{!}{
    \begin{tabular}{r|ccccc|c}
        \toprule
        \multirow{2}{*}{$\mathcal{S}_\mathrm{ADS}(\cdot)$} & \textbf{MagicDrive} & \textbf{DreamForge}  & \textbf{DrivingSphere} & \textbf{MagicDrive-V2} & \textbf{RLGF} & \textcolor{gray}{\textbf{Empirical}}
        \\
        & \textcolor{gray}{\small[ICLR'24]} & \textcolor{gray}{\small[arXiv'24]}  & \textcolor{gray}{\small[CVPR'25]} & \textcolor{gray}{\small[ICCV'25]}  & \textcolor{gray}{\small[NeurIPS'25]} & \textcolor{gray}{\textbf{Max}}
        \\\midrule
        No Collision (NC, $\uparrow$) & $0.815$ & $0.855$  & $0.858$ & $0.885$ & $0.912$ & \cellcolor{gray!7}\textcolor{gray}{-}
        \\
        Compliance (DAC, $\uparrow$) & $0.910$ &$ 0.930$   & $0.935$ &$ 0.948$ & $0.965$ & \cellcolor{gray!7}\textcolor{gray}{-}
        \\
        Ego Progress (EP, $\uparrow$) & $0.712$ & $0.740$   &$ 0.745$ &$ 0.770$ & $0.985$ & \cellcolor{gray!7}\textcolor{gray}{-}
        \\
        Time-to-Collision (TTC, $\uparrow$) & $0.745$   &$ 0.765$  &$ 0.772$ & $0.795$ &$ 0.905$ & \cellcolor{gray!7}\textcolor{gray}{-}
        \\
        Comfort (C, $\uparrow$)&$ 0.720$ &$ 0.745$   & $0.750$ & $0.765$  &$0.850$  & \cellcolor{gray!7}\textcolor{gray}{-}
        \\
        Route Completion (RC, $\uparrow$) & $0.068$ & $0.102$ & $0.110$ & $0.123$  & $0.135$ & \cellcolor{gray!7}\textcolor{gray}{-}
        \\
        \textbf{Total~($\uparrow$)} & \cellcolor{w_blue!20}$0.048$ & \cellcolor{w_blue!20}$0.077$ & \cellcolor{w_blue!20}$0.083$ & \cellcolor{w_blue!20}$0.095$ & \cellcolor{w_blue!20}$0.106$ & \cellcolor{gray!7}\textcolor{gray}{-}
        \\
        \bottomrule
    \end{tabular}}
    \vspace{-0.3cm}
\end{table*}

%% file: sections_supp/5_downstream.tex
\section{~Aspect 4: Downstream Task}

In this section, we evaluate the \textbf{downstream task utility} of generated videos by assessing how well pretrained perception models perform when applied to synthetic data. Rather than measuring visual realism or temporal stability directly, this aspect examines whether a generative world model can produce data that is \textit{useful} for real-world perception tasks. Specifically, we test \textbf{four representative downstream tasks} that span spatial understanding, object reasoning, and 3D scene interpretation. For each task, a perception model is pretrained on the corresponding ground-truth dataset and then evaluated on videos generated by the world model. Performance degradation relative to the ground truth reflects the distribution gap introduced by generation. 

\input{sections_supp/dimensions/downstream_map_segmentation}

\clearpage\clearpage
\input{sections_supp/dimensions/downstream_3d_det}

\clearpage\clearpage
\input{sections_supp/dimensions/downstream_3d_track}

\clearpage\clearpage
\input{sections_supp/dimensions/downstream_occ}

%% file: sections_supp/dimensions/downstream_map_segmentation.tex
\subsection{~Map Segmentation}
\label{subsec:map_seg}

\subsubsection{~Definition}
BEV (Bird’s-Eye-View) Map Segmentation evaluates whether individual generated frames contain sufficient spatial and semantic cues for top-down mapping. A pretrained perception network $\psi_{\mathrm{BEV}}(\cdot)$ takes each generated frame $y_j^{(t)}$ as input and predicts a BEV semantic map, which is compared with the corresponding ground-truth annotation using mean Intersection-over-Union (mIoU).  Higher scores indicate that the generated frames preserve structural layout and scene semantics conducive to reliable map inference.

\subsubsection{~Formulation}
For each generated frame $y_j^{(t)}$, the pretrained model predicts a BEV map:
$\hat{B}_j^{(t)}=\psi_{\mathrm{BEV}}(y_j^{(t)})$ and
$\hat{B}_j^{(t)}\in\{0,\dots,C_{\mathrm{BEV}}{-}1\}^{H_b\times W_b}$, and $B_j^{\mathrm{gt}(t)}$ denotes the corresponding ground-truth BEV annotation. The per-frame mean IoU is computed as:
\[
S_{\mathrm{BEV}}^{(t)}(y_j)=
\tfrac{1}{C_{\mathrm{BEV}}}\sum\nolimits_{c=1}^{C_{\mathrm{BEV}}}
\tfrac{|\hat{B}_j^{(t,c)}\cap B_j^{\mathrm{gt}(t,c)}|}
     {|\hat{B}_j^{(t,c)}\cup B_j^{\mathrm{gt}(t,c)}|}~.
\]
The dataset-level Map Segmentation score averages over all frames and videos, that is:
\begin{equation}
\boxed{\;
\mathcal{S}_{\mathrm{Seg}}(\mathcal{Y})
=\tfrac{1}{N_gT}\sum\nolimits_{j=1}^{N_g}\sum\nolimits_{t=1}^{T} S_{\mathrm{BEV}}^{(t)}(y_j)
\;}
\label{eq:map_segmentation}
\end{equation}
where $C_{\mathrm{BEV}}$ is the number of BEV categories and $(H_b,W_b)$ the BEV map resolution.

\subsubsection{~Implementation Details}
We employ the pretrained BEVFusion 
multi-task model of Liu \emph{et al.}~\cite{liu2023bevfusion}, using its camera-only configuration with a ResNet-101~\cite{he2016deep} backbone and BEVFormer encoder. The model predicts BEV semantic maps on a $150\times150$ grid covering a $[-30,30]\!\times\![-15,15]\text{ m}$ 
region, which are used for mIoU evaluation.

\subsubsection{~Examples}
\begin{figure}[t]
    \centering
    \begin{subfigure}[h]{\textwidth}
        \centering
        \includegraphics[width=\linewidth]{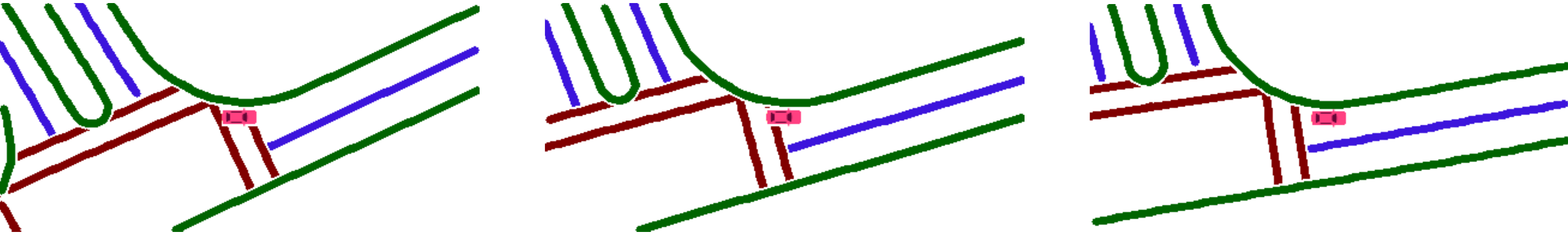}
        \caption{Good example in the \emph{Map Segmentation} dimension (Score: \textcolor{w_blue}{$100.00\%$})}
        \label{fig:downstream_map_seg_1}
    \end{subfigure}
    \\[1ex]
    \begin{subfigure}[h]{\textwidth}
        \centering
        \includegraphics[width=\linewidth]{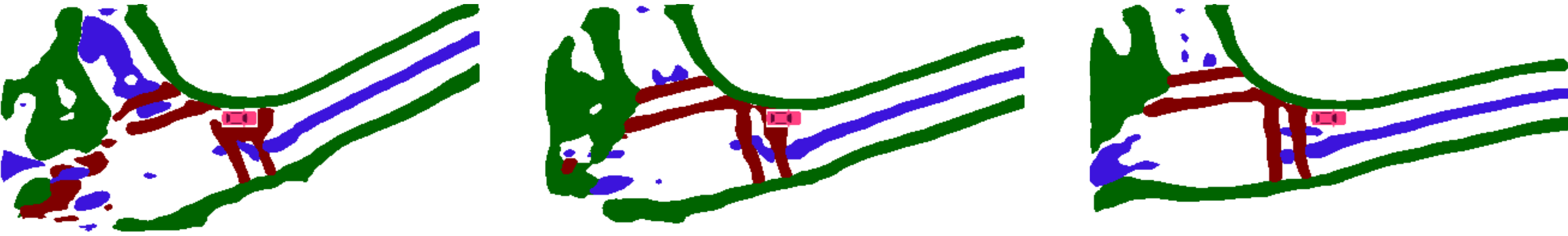}
        \caption{Bad example in the \emph{Map Segmentation} dimension (Score: \textcolor{red}{$9.46\%$})}
        \label{fig:downstream_map_seg_2}
    \end{subfigure}
    \\[3.5ex]
    \begin{subfigure}[h]{\textwidth}
        \centering
        \includegraphics[width=\linewidth]{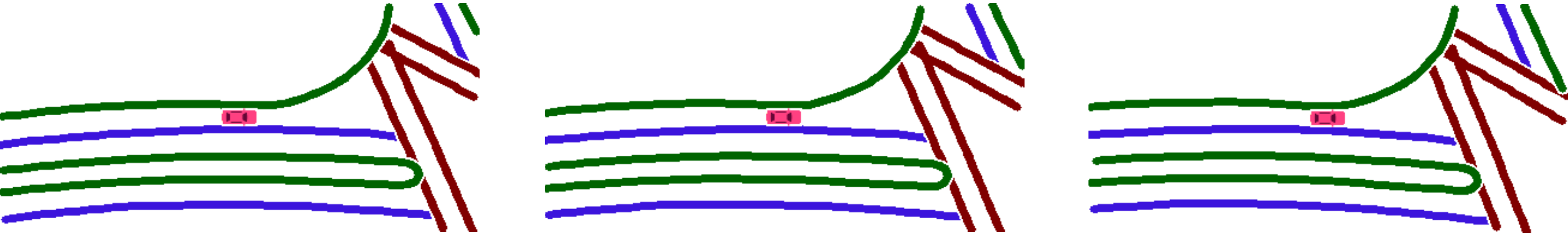}
        \caption{Good example in the \emph{Map Segmentation} dimension (Score: \textcolor{w_blue}{$100.00\%$})}
        \label{fig:downstream_map_seg_3}
    \end{subfigure}
    \\[1ex]
    \begin{subfigure}[h]{\textwidth}
        \centering
        \includegraphics[width=\linewidth]{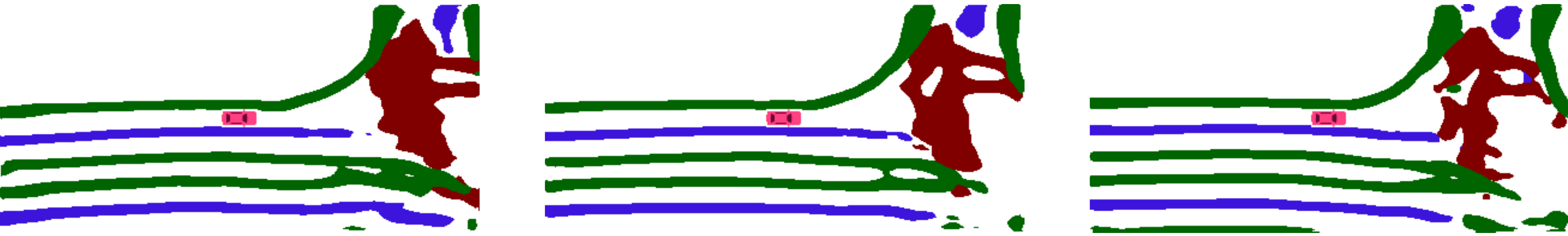}
        \caption{Bad example in the \emph{Map Segmentation} dimension (Score: \textcolor{red}{$11.27\%$})}
        \label{fig:downstream_map_seg_4}
    \end{subfigure}
    \\[3.5ex]
    \begin{subfigure}[h]{\textwidth}
        \centering
        \includegraphics[width=\linewidth]{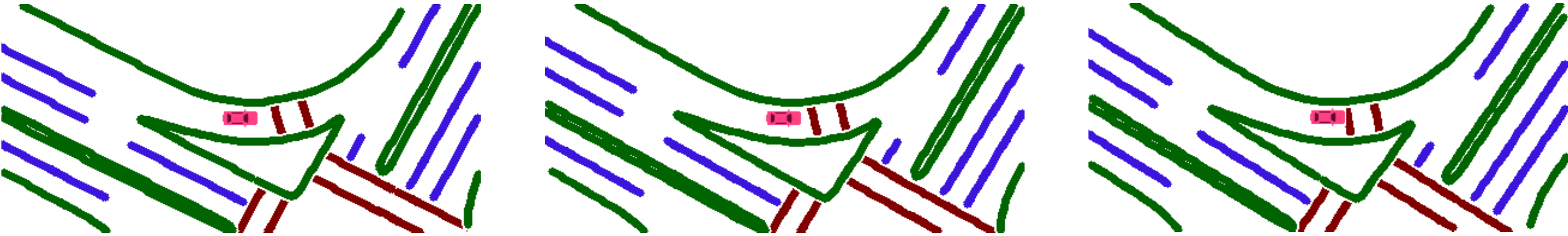}
        \caption{Good example in the \emph{Map Segmentation} dimension (Score: \textcolor{w_blue}{$100.00\%$})}
        \label{fig:downstream_map_seg_5}
    \end{subfigure}
    \\[1ex]
    \begin{subfigure}[h]{\textwidth}
        \centering
        \includegraphics[width=\linewidth]{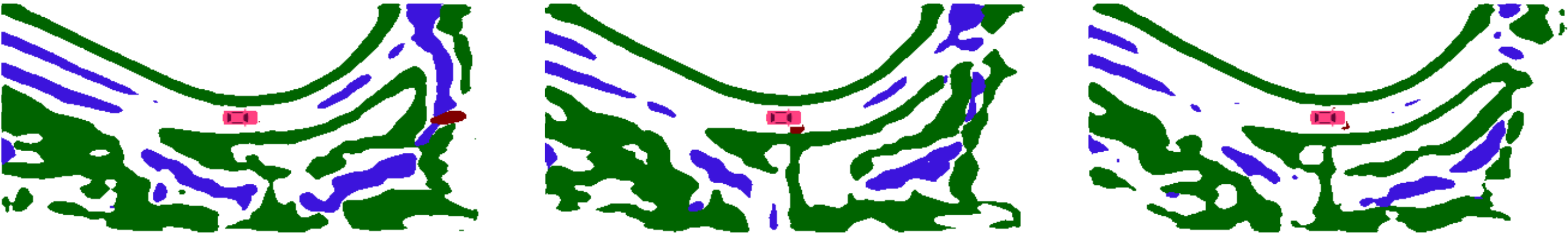}
        \caption{Bad example in the \emph{Map Segmentation} dimension (Score: \textcolor{red}{$7.88\%$})}
        \label{fig:downstream_map_seg_6}
    \end{subfigure}
    \vspace{-0.05cm}
    \caption{Examples of ``good'' and ``bad'' downstream task performances in terms of \emph{Map Segmentation} in WorldLens.}
    \label{fig:downstream_map_seg}
\end{figure}

Figure~\ref{fig:downstream_map_seg} provides typical examples of videos with good and bad quality in terms of \emph{Map Segmentation}.

\subsubsection{~Evaluation \& Analysis}
Table~\ref{tab:supp_downstream_map_seg} provides the complete results of models in terms of \emph{Map Segmentation}.

\begin{table*}[h]
    \centering
    \vspace{0.2cm}
    \caption{Complete comparisons of state-of-the-art driving world models in terms of \emph{Map Segmentation} in WorldLens.}
    \vspace{-0.2cm}
    \label{tab:supp_downstream_map_seg}
    \resizebox{\linewidth}{!}{
    \begin{tabular}{r|cccccc|c}
        \toprule
        \multirow{2}{*}{$\mathcal{S}_\mathrm{Seg}(\cdot)$} & \textbf{MagicDrive} & \textbf{DreamForge}  & \textbf{DriveDreamer-2} & \textbf{OpenDWM} & \textbf{~DiST-4D~} & $\mathcal{X}$\textbf{-Scene} & \textcolor{gray}{\textbf{Empirical}}
        \\
        & \textcolor{gray}{\small[ICLR'24]} & \textcolor{gray}{\small[arXiv'24]} & \textcolor{gray}{\small[AAAI'25]} & \textcolor{gray}{\small[CVPR'25]} & \textcolor{gray}{\small[ICCV'25]} & \textcolor{gray}{\small[NeurIPS'25]} & \textcolor{gray}{\textbf{Max}}
        \\\midrule
        Divider~($\uparrow$) & $23.39\%$ & $34.44\%$ & $39.69\%$ & $31.88\%$ & $41.26\%$ & $31.84\%$ & \cellcolor{gray!7}\textcolor{gray}{$46.08\%$}
        \\
        Ped. Crossing~($\uparrow$) & $9.77\%$ & $21.18\%$ & $24.12\%$ & $20.27\%$ & $26.17\%$ & $17.67\%$ & \cellcolor{gray!7}\textcolor{gray}{$30.38\%$}
        \\
        Boundary~($\uparrow$) & $21.87\%$ & $35.31\%$ & $37.03\%$ & $30.74\%$ & $39.23\%$ & $32.22\%$ & \cellcolor{gray!7}\textcolor{gray}{$45.45\%$}
        \\
        \textbf{Average~($\uparrow$)} & \cellcolor{w_blue!20}$18.34\%$ & \cellcolor{w_blue!20}$30.31\%$ & \cellcolor{w_blue!20}$33.62\%$ & \cellcolor{w_blue!20}$27.63\%$ & \cellcolor{w_blue!20}$35.55\%$ & \cellcolor{w_blue!20}$27.24\%$ & \cellcolor{gray!7}\textcolor{gray}{$40.64\%$}
        \\
        \bottomrule
    \end{tabular}}
    \vspace{-0.3cm}
\end{table*}

%% file: sections_supp/dimensions/downstream_3d_det.tex
\subsection{~3D Object Detection}

\subsubsection{~Definition}
3D Object Detection evaluates whether generated frames preserve the geometric and motion cues necessary for accurate perception of traffic participants.

A pretrained detector $\psi_{\mathrm{DET}}(\cdot)$, trained on ground-truth data, is applied to each generated frame $y_j^{(t)}$ to predict 3D bounding boxes with category, position, scale, and velocity information. Following the nuScenes detection protocol~\cite{caesar2020nuscenes}, detections are compared against ground-truth boxes to compute mean Average Precision (mAP) and the consolidated nuScenes Detection Score (NDS).

Higher mAP and NDS indicate that the generated data retains faithful 3D spatial structure and dynamic cues consistent with real-world scenes.

\subsubsection{~Formulation}
For each frame $y_j^{(t)}$, the pretrained detector predicts a set of 3D bounding boxes:
\[
\hat{\mathcal{B}}_j^{(t)}=\psi_{\mathrm{DET}}\!\left(y_j^{(t)}\right),
\qquad 
\mathcal{B}_j^{\mathrm{gt}(t)} \text{~ denotes the corresponding ground-truth set.}
\]
Per-frame detection metrics (mAP and NDS) are computed following~\cite{li2025bevformer,liu2023bevfusion} using standard matching and error terms. 
The dataset-level 3D detection score averages these values across all generated frames:
\begin{equation}
\boxed{\;
\mathcal{S}_{\mathrm{Det}}(\mathcal{Y})
=\tfrac{1}{N_gT}\sum\nolimits_{j=1}^{N_g}\sum\nolimits_{t=1}^{T}\mathrm{NDS}\!\left(y_j^{(t)}\right)
\;}
\label{eq:object_detection}
\end{equation}
Higher $\mathcal{S}_{\mathrm{Det}}$ (and mAP) indicates that the generated frames support more accurate 3D reasoning and reliable downstream perception for autonomous driving.

\subsubsection{~Implementation Details}
The 3D detection evaluation uses the same pretrained BEVFusion model as in Section~\ref{subsec:map_seg}, with its detection head producing 3D bounding boxes on the nuScenes BEV range $[-51.2,51.2]\text{ m}^2$ and a $150\times150$ grid. Predicted boxes are evaluated against ground-truth annotations using standard nuScenes 3D detection metrics.

\subsubsection{~Examples}
\begin{figure}[t]
    \centering
    \begin{subfigure}[h]{\textwidth}
        \centering
        \includegraphics[width=\linewidth]{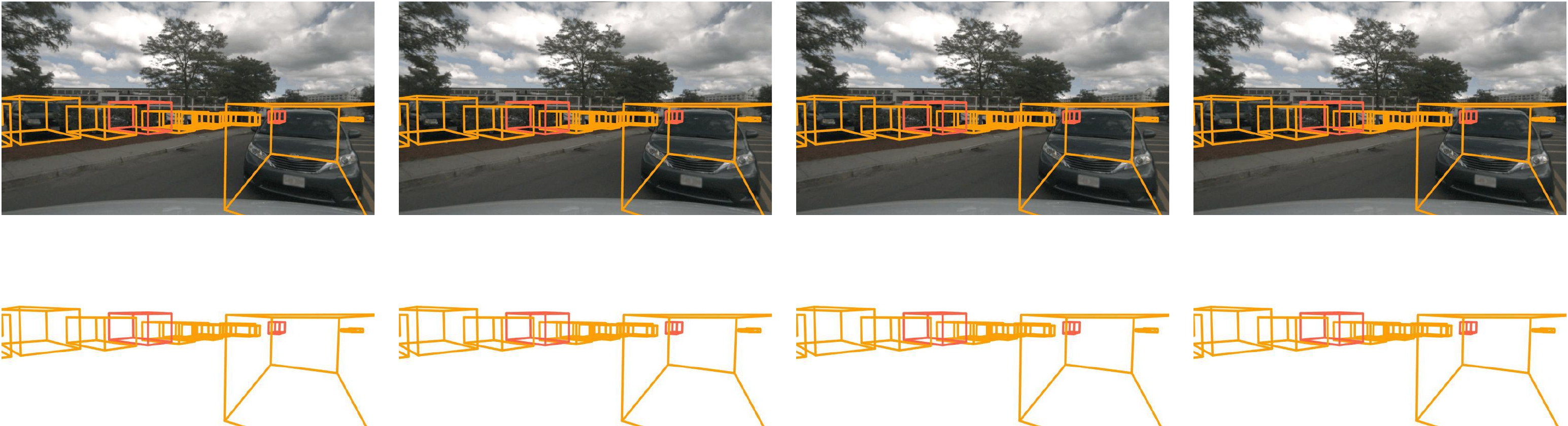}
        \caption{Good example in the \emph{3D Object Detection} dimension}
        \label{fig:downstream_3d_det_1}
    \end{subfigure}
    \\[2ex]
    \begin{subfigure}[h]{\textwidth}
        \centering
        \includegraphics[width=\linewidth]{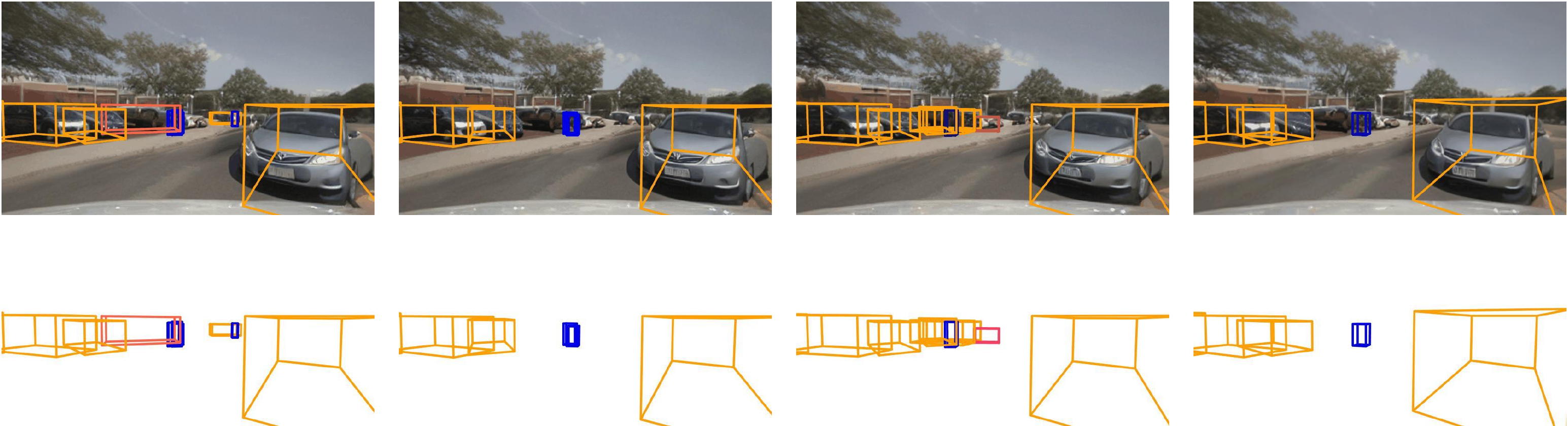}
        \caption{Bad example in the \emph{3D Object Detection} dimension}
        \label{fig:downstream_3d_det_2}
    \end{subfigure}
    \\[2ex]
    \begin{subfigure}[h]{\textwidth}
        \centering
        \includegraphics[width=\linewidth]{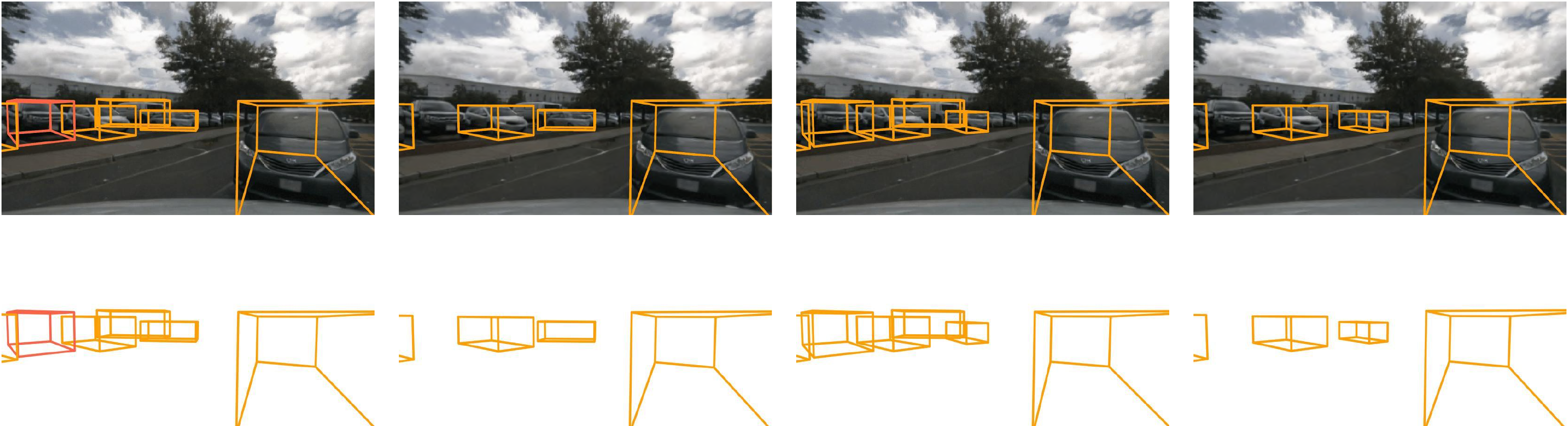}
        \caption{Bad example in the \emph{3D Object Detection} dimension}
        \label{fig:downstream_3d_det_3}
    \end{subfigure}
    \\[2ex]
    \begin{subfigure}[h]{\textwidth}
        \centering
        \includegraphics[width=\linewidth]{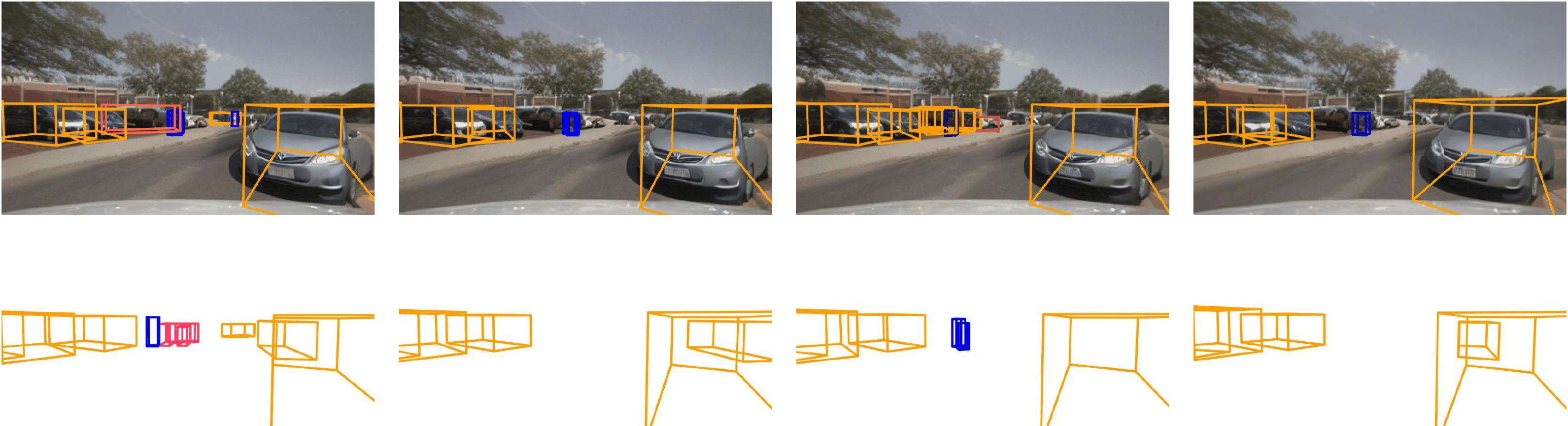}
        \caption{Bad example in the \emph{3D Object Detection} dimension}
        \label{fig:downstream_3d_det_4}
    \end{subfigure}
    \\
    \vspace{-0.05cm}
    \caption{Examples of ``good'' and ``bad'' downstream task performances in terms of \emph{3D Object Detection} in WorldLens.}
    \label{fig:downstream_3d_det}
\end{figure}

Figure~\ref{fig:downstream_3d_det} provides typical examples of videos with good and bad quality in terms of \emph{3D Object Detection}.

\subsubsection{~Evaluation \& Analysis}
Table~\ref{tab:supp_downstream_3d_det} provides the complete results of models in terms of \emph{3D Object Detection}.

\begin{table*}[h]
    \centering
    \vspace{0.2cm}
    \caption{Complete comparisons of state-of-the-art driving world models in terms of \emph{3D Object Detection} in WorldLens.}
    \vspace{-0.2cm}
    \label{tab:supp_downstream_3d_det}
    \resizebox{\linewidth}{!}{
    \begin{tabular}{r|cccccc|c}
        \toprule
        \multirow{2}{*}{$\mathcal{S}_\mathrm{Det}(\cdot)$} & \textbf{MagicDrive} & \textbf{DreamForge}  & \textbf{DriveDreamer-2} & \textbf{OpenDWM} & \textbf{~DiST-4D~} & $\mathcal{X}$\textbf{-Scene} & \textcolor{gray}{\textbf{Empirical}}
        \\
        & \textcolor{gray}{\small[ICLR'24]} & \textcolor{gray}{\small[arXiv'24]} & \textcolor{gray}{\small[AAAI'25]} & \textcolor{gray}{\small[CVPR'25]} & \textcolor{gray}{\small[ICCV'25]} & \textcolor{gray}{\small[NeurIPS'25]} & \textcolor{gray}{\textbf{Max}}
        \\\midrule
        mAP~($\uparrow$) & $0.1178$ & $0.1636$ & $0.1961$ & $0.944$ & $0.2242$ & $0.1562$ & \cellcolor{gray!7}\textcolor{gray}{$0.3657$}
        \\
        mATE~($\downarrow$) & $0.9435$ & $0.9469$ & $0.8443$ & $1.0354$ & $0.9256$ & $0.8870$ & \cellcolor{gray!7}\textcolor{gray}{$0.7356$}
        \\
        mASE~($\downarrow$) & $0.3400$ & $0.3207$ & $0.3273$ & $0.3479$ & $0.3214$ & $0.3218$ & \cellcolor{gray!7}\textcolor{gray}{$0.2919$}
        \\
        mAOE~($\downarrow$) & $0.7834$ & $0.8237$ & $0.5930$ & $0.7734$ & $0.5252$ & $0.6509$ & \cellcolor{gray!7}\textcolor{gray}{$0.4400$}
        \\
        mAVE~($\downarrow$) & $1.0133$ & $0.8039$ & $0.8904$ & $0.8629$ & $0.7897$ & $0.7061$ & \cellcolor{gray!7}\textcolor{gray}{$0.6821$}
        \\
        mAAE~($\downarrow$) & $0.2814$ & $0.2520$ & $0.2349$ & $0.2917$ & $0.2374$ & $0.2265$ & \cellcolor{gray!7}\textcolor{gray}{$0.2072$}
        \\
        \textbf{NDS~($\uparrow$)} & \cellcolor{w_blue!20}$0.2241$ & \cellcolor{w_blue!20}$0.2671$ & \cellcolor{w_blue!20}$0.3090$ & \cellcolor{w_blue!20}$0.2196$ & \cellcolor{w_blue!20}$0.3322$ & \cellcolor{w_blue!20}$0.2989$ & \cellcolor{gray!7}\textcolor{gray}{$0.4472$}
        \\
        \bottomrule
    \end{tabular}}
    \vspace{-0.3cm}
\end{table*}

%% file: sections_supp/dimensions/downstream_3d_track.tex
\subsection{~3D Object Tracking}

\subsubsection{~Definition}
3D Object Tracking evaluates whether generated videos preserve consistent object motion and identity information that supports temporal data association. A pretrained tracker $\psi_{\mathrm{TRK}}(\cdot)$, trained on ground-truth sequences, is applied to each generated video to estimate 3D trajectories of dynamic objects.

Following the nuScenes tracking protocol~\cite{caesar2020nuscenes}, tracking performance is measured using the Average Multi-Object Tracking Accuracy (AMOTA), which integrates precision, recall, and association quality across recall thresholds. 
Higher AMOTA values indicate that the generated videos exhibit realistic temporal dynamics, enabling stable object tracking over time.

\subsubsection{~Formulation}
For each generated video $y_j=\{y_j^{(t)}\}_{t=1}^{T}$, the tracker predicts a set of object trajectories:
\[
\hat{\mathcal{T}}_j=\psi_{\mathrm{TRK}}(y_j)
=\big\{\hat{\tau}_n=\{\hat{\mathbf{b}}_n^{(t)}\}_{t\in\mathcal{I}_n}\big\}_{n=1}^{N_{\mathrm{trk}}},
\]
and $\mathcal{T}_j^{\mathrm{gt}}$ denotes the corresponding ground-truth trajectories.  
Tracking accuracy is evaluated using the official nuScenes metrics~\cite{caesar2020nuscenes}, including MOTA and AMOTA, where higher scores indicate more reliable data association and motion continuity.

The dataset-level 3D tracking metric averages per-video AMOTA over all generated sequences:
\begin{equation}
\boxed{\;
\mathcal{S}_{\mathrm{Trk}}(\mathcal{Y})
=\tfrac{1}{N_g}\sum\nolimits_{j=1}^{N_g}\mathrm{AMOTA}(y_j)
\;}
\label{eq:object_tracking}
\end{equation}
Higher $\mathcal{S}_{\mathrm{Trk}}$ indicates that generated videos maintain realistic and temporally coherent object motion, supporting accurate long-term tracking.

\subsubsection{~Implementation Details}
We evaluate the 3D object tracking performance using the pretrained camera-only ADA-Track~\cite{ding2024ada}, following its official nuScenes configuration. The tracker is run directly on the generated multi-view videos.

\subsubsection{~Examples}
\begin{figure}[t]
    \centering
    \begin{subfigure}[h]{\textwidth}
        \centering
        \includegraphics[width=\linewidth]{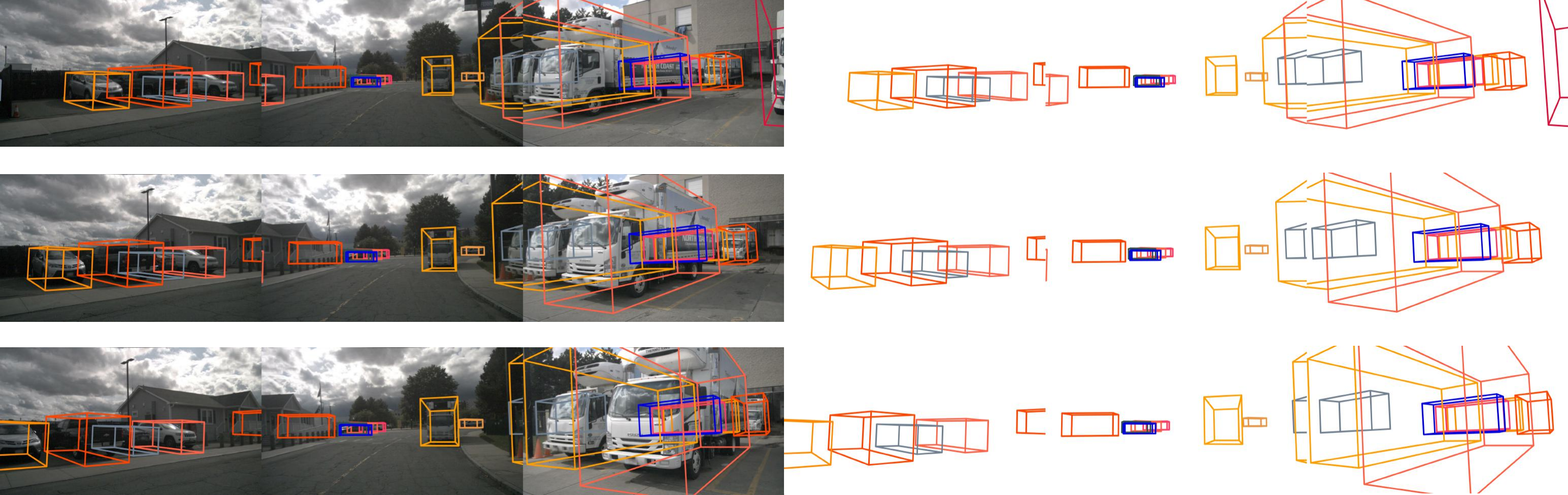}
        \caption{Good example in the \emph{3D Object Tracking} dimension}
        \label{fig:downstream_3d_track_1}
    \end{subfigure}
    \\[3.5ex]
    \begin{subfigure}[h]{\textwidth}
        \centering
        \includegraphics[width=\linewidth]{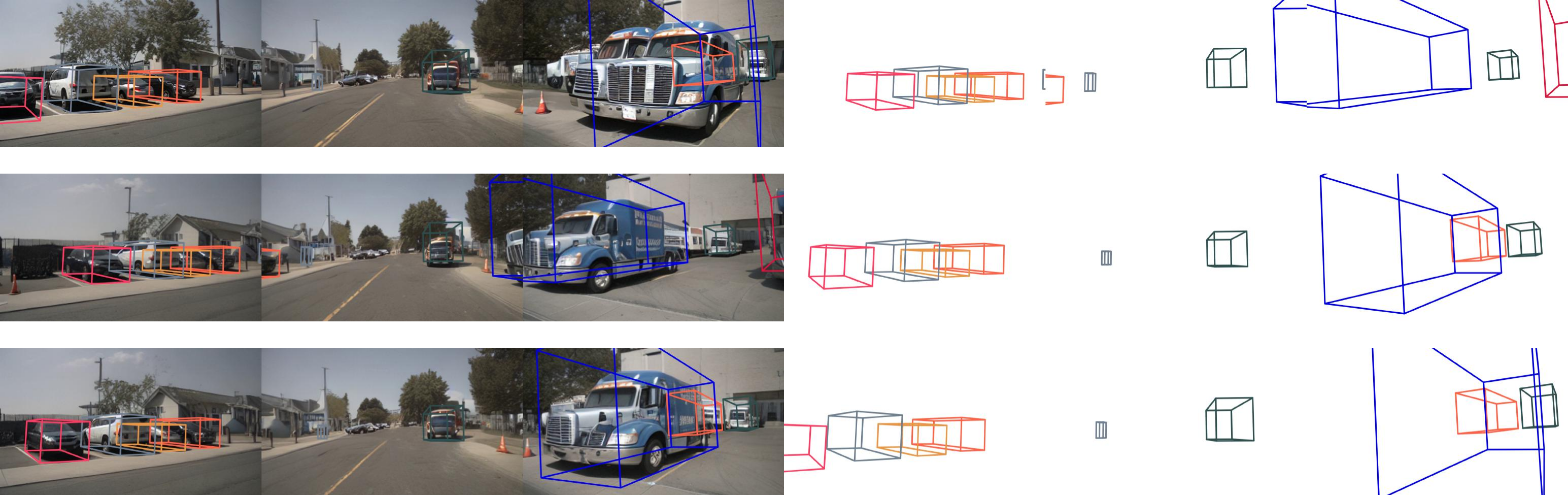}
        \caption{Bad example in the \emph{3D Object Tracking} dimension}
        \label{fig:downstream_3d_track_2}
    \end{subfigure}
    \\[3.5ex]
    \begin{subfigure}[h]{\textwidth}
        \centering
        \includegraphics[width=\linewidth]{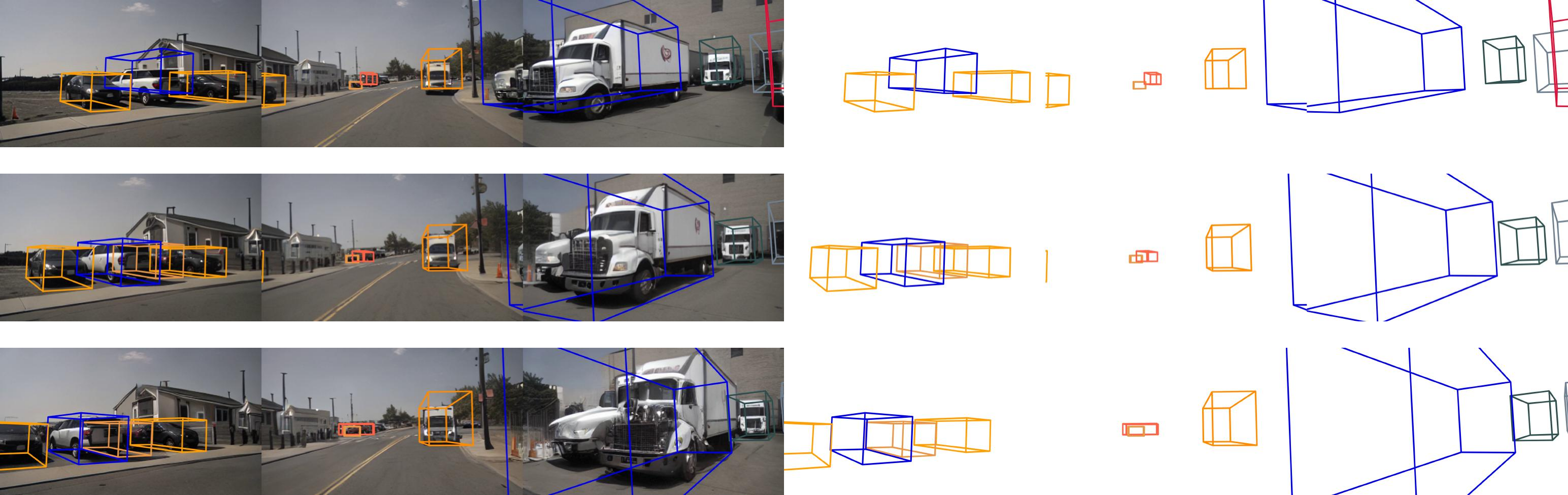}
        \caption{Bad example in the \emph{3D Object Tracking} dimension}
        \label{fig:downstream_3d_track_3}
    \end{subfigure}.
    \vspace{-0.05cm}
    \caption{Examples of ``good'' and ``bad'' downstream task performances in terms of \emph{3D Object Tracking} in WorldLens.}
    \label{fig:downstream_3d_track}
\end{figure}

Figure~\ref{fig:downstream_3d_track} provides typical examples of videos with good and bad quality in terms of \emph{3D Object Tracking}.

\subsubsection{~Evaluation \& Analysis}
Table~\ref{tab:supp_downstream_3d_track} provides the complete results of models in terms of \emph{3D Object Tracking}.

\begin{table*}[h]
    \centering
    \vspace{0.2cm}
    \caption{Complete comparisons of state-of-the-art driving world models in terms of \emph{3D Object Tracking} in WorldLens.}
    \vspace{-0.2cm}
    \label{tab:supp_downstream_3d_track}
    \resizebox{\linewidth}{!}{
    \begin{tabular}{r|cccccc|c}
        \toprule
        \multirow{2}{*}{$\mathcal{S}_\mathrm{Trk}(\cdot)$} & \textbf{MagicDrive} & \textbf{DreamForge}  & \textbf{DriveDreamer-2} & \textbf{OpenDWM} & \textbf{~DiST-4D~} & $\mathcal{X}$\textbf{-Scene} & \textcolor{gray}{\textbf{Empirical}}
        \\
        & \textcolor{gray}{\small[ICLR'24]} & \textcolor{gray}{\small[arXiv'24]} & \textcolor{gray}{\small[AAAI'25]} & \textcolor{gray}{\small[CVPR'25]} & \textcolor{gray}{\small[ICCV'25]} & \textcolor{gray}{\small[NeurIPS'25]} & \textcolor{gray}{\textbf{Max}}
        \\\midrule
        AMOTP~($\downarrow$) & $1.843$ & $1.812$ & $1.778$ & $1.848$ & $1.705$ & $1.829$ & \cellcolor{gray!7}\textcolor{gray}{$1.405$}
        \\
        Recall~($\uparrow$) & $15.50\%$ & $16.10\%$ & $19.50\%$ & $12.40\%$ & $29.20\%$ & $15.56\%$ & \cellcolor{gray!7}\textcolor{gray}{$45.30\%$}
        \\
        MOTA~($\uparrow$) & $8.70\%$ & $9.70\%$ & $12.70\%$ & $7.40\%$ & $15.10\%$ & $10.60\%$ & \cellcolor{gray!7}\textcolor{gray}{$34.30\%$}
        \\
        FN~($\downarrow$) & $5093$ & $4622$ & $4361$ & $4820$ & $3725$ & $4799$ & \cellcolor{gray!7}\textcolor{gray}{$2678$}
        \\
        TP~($\uparrow$) & $1615$ & $2084$ & $2343$ & $1887$ & $2977$ & $1909$ & \cellcolor{gray!7}\textcolor{gray}{$4027$}
        \\
        \textbf{AMOTA~($\uparrow$)} & \cellcolor{w_blue!20}$7.90\%$ & \cellcolor{w_blue!20}$10.30\%$& \cellcolor{w_blue!20}$13.30\%$ & \cellcolor{w_blue!20}$6.90\%$ & \cellcolor{w_blue!20}$15.30\%$ & \cellcolor{w_blue!20}$8.80\%$ & \cellcolor{gray!7}\textcolor{gray}{$36.30\%$}
        \\
        \bottomrule
    \end{tabular}}
    \vspace{-0.3cm}
\end{table*}

%% file: sections_supp/dimensions/downstream_occ.tex
\subsection{~Occupancy Prediction}

\subsubsection{~Definition}
Occupancy Prediction evaluates whether generated videos enable accurate 3D reconstruction of scene geometry and semantics. 
We adopt the RayIoU metric~\cite{tang2024sparseocc}, which measures semantic and geometric agreement \emph{along camera rays} rather than voxel overlap. 
For each ray, RayIoU compares the \emph{frontmost} occupied voxel in the predicted and ground-truth volumes, requiring both class correctness and depth proximity within a tolerance $\delta$.
This ray-wise formulation avoids the depth-ambiguity of voxel mIoU (which may reward thick surfaces) and naturally supports multi-pose scene completion evaluation via ray casting.

\subsubsection{~Formulation}
A frozen occupancy estimator $\psi_{\mathrm{Occ}}(\cdot)$ predicts a probabilistic 3D volume for each generated video:
\[
\hat{\mathbf{O}}_j=\psi_{\mathrm{Occ}}(y_j)~,
\qquad 
\hat{\mathbf{O}}_j\in[0,1]^{X\times Y\times Z}~.
\]
Let $\mathcal{R}$ denote the set of sampled query rays (with distance-balanced resampling).  
For each ray $r\in\mathcal{R}$, denote the frontmost occupied voxel in prediction and ground truth by $(\hat d_r,\hat c_r)$ and $(d_r^{\mathrm{gt}},c_r^{\mathrm{gt}})$.  
A prediction is correct if $\hat c_r=c_r^{\mathrm{gt}}$ and $|\,\hat d_r-d_r^{\mathrm{gt}}\,|\le\delta$.  
The RayIoU at tolerance $\delta$ is defined as:
\[
\mathrm{RayIoU}@\delta
=\tfrac{1}{C}\sum\nolimits_{c=1}^{C}
\tfrac{\mathrm{TP}_c(\delta)}
{\mathrm{TP}_c(\delta)+\mathrm{FP}_c(\delta)+\mathrm{FN}_c(\delta)}~,
\]
and the mean RayIoU (mRayIoU) aggregates multiple tolerances:
\[
\mathrm{mRayIoU}
=\tfrac{1}{3}\sum\nolimits_{\delta\in\{1,2,4\}}\mathrm{RayIoU}@\delta~.
\]
The dataset-level semantic occupancy score averages mRayIoU across all generated videos:
\begin{equation}
\boxed{\;
\mathcal{S}_{\mathrm{Occ}}(\mathcal{Y})
=\tfrac{1}{N_g}\sum\nolimits_{j=1}^{N_g}\mathrm{mRayIoU}(y_j)
\;}
\label{eq:occupancy_prediction}
\end{equation}
Higher $\mathcal{S}_{\mathrm{Occ}}$ indicates that generated scenes enable more accurate, depth-consistent, and semantically faithful occupancy reconstruction.

\subsubsection{~Implementation Details}
We perform occupancy prediction using the pretrained SparseOcc model~\cite{tang2024sparseocc}. The model is applied to the generated multi-view frames following the official nuScenes camera-only configuration, producing voxel-wise semantic occupancy grids within a $[-40,40]\!\times[-40,40]\!\times[-1,5.4]\text{ m}$ 3D volume for evaluation.

\subsubsection{~Examples}
\begin{figure}[t]
    \centering
    \begin{subfigure}[h]{\textwidth}
        \centering
        \includegraphics[width=\linewidth]{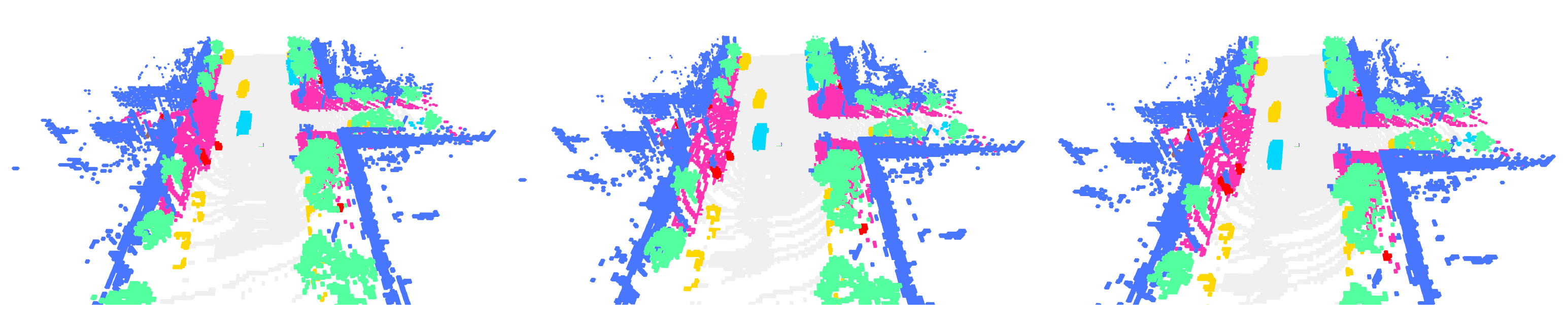}
        \caption{Good example in the \emph{Occupancy Prediction} dimension (Score: \textcolor{w_blue}{$100.00\%$})}
        \label{fig:downstream_occ_predict_1}
    \end{subfigure}
    \\[2ex]
    \begin{subfigure}[h]{\textwidth}
        \centering
        \includegraphics[width=\linewidth]{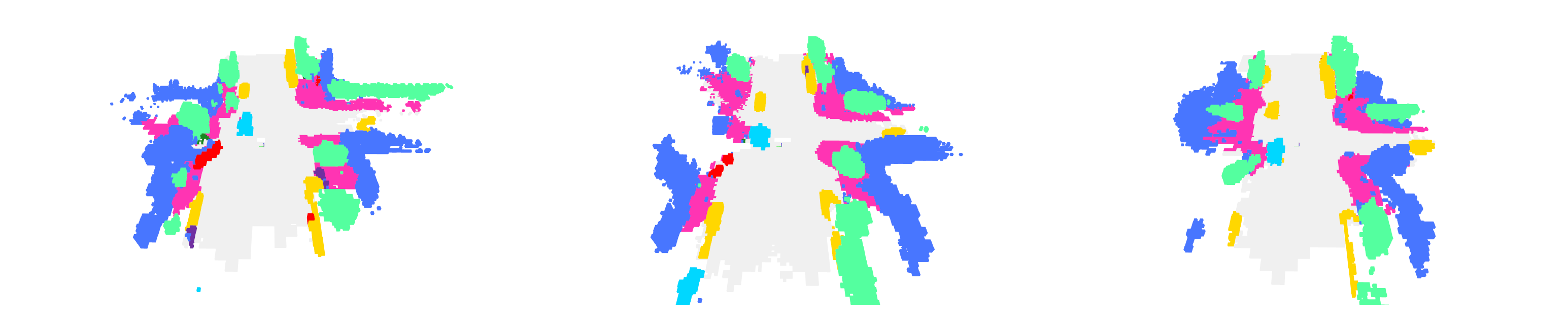}
        \caption{Bad example in the \emph{Occupancy Prediction} dimension (Score: \textcolor{red}{$9.42\%$})}
        \label{fig:downstream_occ_predict_2}
    \end{subfigure}
    \\[3.5ex]
    \begin{subfigure}[h]{\textwidth}
        \centering
        \includegraphics[width=\linewidth]{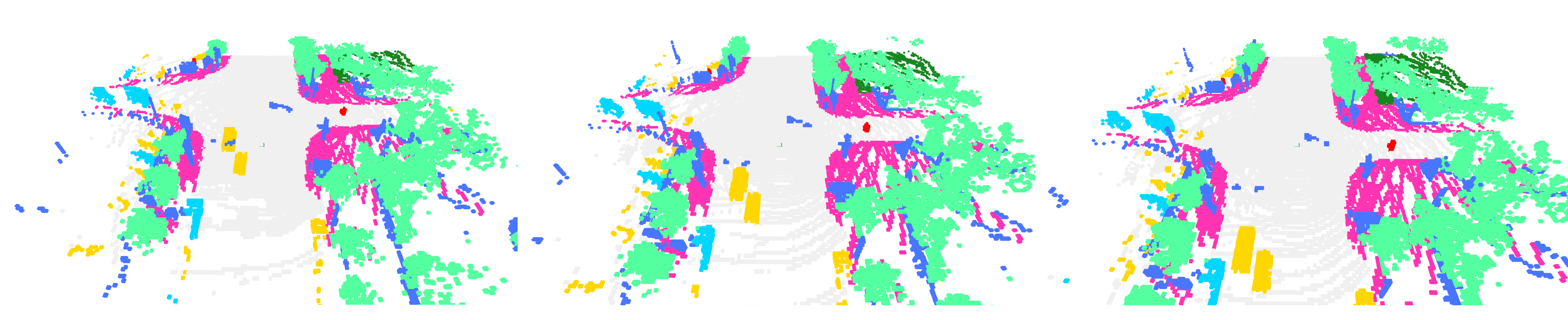}
        \caption{Good example in the \emph{Occupancy Prediction} dimension (Score: \textcolor{w_blue}{$100\%$})}
        \label{fig:downstream_occ_predict_3}
    \end{subfigure}
    \\[2ex]
    \begin{subfigure}[h]{\textwidth}
        \centering
        \includegraphics[width=\linewidth]{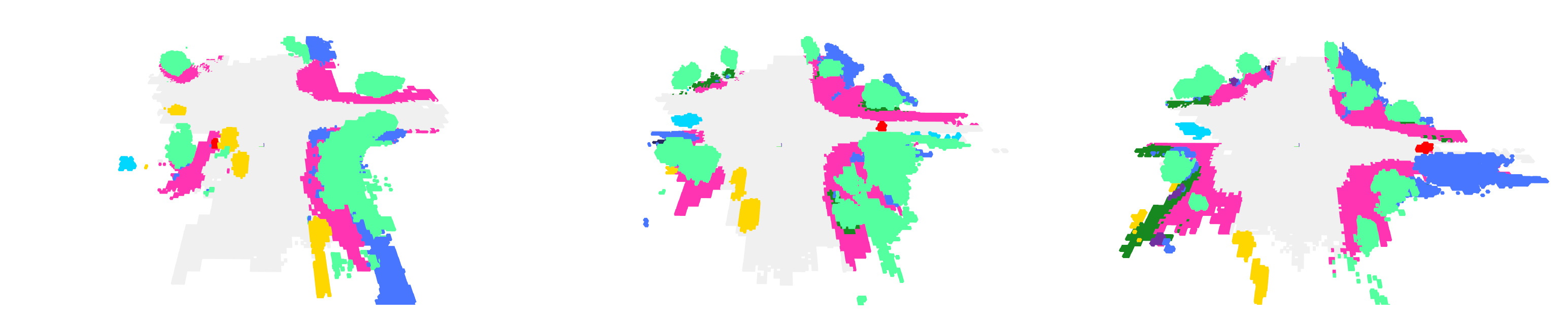}
        \caption{Bad example in the \emph{Occupancy Prediction} dimension (Score: \textcolor{red}{$11.27\%$})}
        \label{fig:downstream_occ_predict_4}
    \end{subfigure}
    \vspace{-0.05cm}
    \caption{Examples of ``good'' and ``bad'' downstream task performances in terms of \emph{Occupancy Prediction} in WorldLens.}
    \label{fig:downstream_occ_predict}
\end{figure}

Figure~\ref{fig:downstream_occ_predict} provides typical examples of videos with good and bad quality in terms of \emph{Occupancy Prediction}.

\subsubsection{~Evaluation \& Analysis}
Table~\ref{tab:supp_downstream_occ_predict} provides the complete results of models in terms of \emph{Occupancy Prediction}.

\begin{table*}[h]
    \centering
    \vspace{0.2cm}
    \caption{Complete comparisons of state-of-the-art driving world models in terms of \emph{Occupancy Prediction} in WorldLens.}
    \vspace{-0.2cm}
    \label{tab:supp_downstream_occ_predict}
    \resizebox{\linewidth}{!}{
    \begin{tabular}{r|cccccc|c}
        \toprule
        \multirow{2}{*}{$\mathcal{S}_\mathrm{Occ}(\cdot)$} & \textbf{MagicDrive} & \textbf{DreamForge}  & \textbf{DriveDreamer-2} & \textbf{OpenDWM} & \textbf{~DiST-4D~} & $\mathcal{X}$\textbf{-Scene} & \textcolor{gray}{\textbf{Empirical}}
        \\
        & \textcolor{gray}{\small[ICLR'24]} & \textcolor{gray}{\small[arXiv'24]} & \textcolor{gray}{\small[AAAI'25]} & \textcolor{gray}{\small[CVPR'25]} & \textcolor{gray}{\small[ICCV'25]} & \textcolor{gray}{\small[NeurIPS'25]} & \textcolor{gray}{\textbf{Max}}
        \\\midrule
        RayIoU@1m~($\uparrow$) & $17.24\%$ & $18.24\%$ & $20.32\%$ & $17.78\%$ & $18.81\%$ & $18.00\%$ & \cellcolor{gray!7}\textcolor{gray}{$29.04\%$}
        \\
        RayIoU@2m~($\uparrow$) & $23.53\%$ & $24.10\%$ & $27.36\%$ & $25.35\%$ & $26.50\%$ & $23.93\%$ & \cellcolor{gray!7}\textcolor{gray}{$38.17\%$}
        \\
        RayIoU@4m~($\uparrow$) & $28.65\%$ & $28.79\%$ & $32.77\%$ & $31.32\%$ & $33.00\%$ & $29.12\%$ & \cellcolor{gray!7}\textcolor{gray}{$43.93\%$}
        \\
        \textbf{Average~($\uparrow$)} & \cellcolor{w_blue!20}$23.14\%$ & \cellcolor{w_blue!20}$23.71\%$ & \cellcolor{w_blue!20}$26.82\%$ & \cellcolor{w_blue!20}$24.82\%$ & \cellcolor{w_blue!20}$26.10\%$ & \cellcolor{w_blue!20}$23.68\%$ & \cellcolor{gray!7}\textcolor{gray}{$37.05\%$}
        \\
        \bottomrule
    \end{tabular}}
    \vspace{-0.3cm}
\end{table*}

%% file: sections_supp/6_human_preference.tex
\section{~Aspect 5: Human Preference}

This section presents human-centered evaluations. While quantitative measures capture specific aspects of fidelity, consistency, and geometric accuracy, they cannot fully reflect \textbf{how humans perceive realism, stability, and overall scene quality}. To bridge this gap, we introduce a human preference study that scores generated videos across multiple dimensions, providing a holistic and perceptually grounded assessment of model performance.

\input{sections_supp/dimensions/human_overall_realism}

\clearpage\clearpage
\input{sections_supp/dimensions/human_vehicle_realism}

\clearpage\clearpage
\input{sections_supp/dimensions/human_pedestrian_realism}

\clearpage\clearpage
\input{sections_supp/dimensions/human_physical_plausibility}

\clearpage\clearpage
\input{sections_supp/dimensions/human_3d_4d}

\clearpage\clearpage
\input{sections_supp/dimensions/human_behavioral_safety}

%% file: sections_supp/dimensions/human_overall_realism.tex
\subsection{~World Realism - Overall Realism}

Overall Realism measures the global visual believability. Annotators judge whether the generated video ``looks like a real-world driving recording''. They are instructed to judge each clip based on the following criteria:
\begin{itemize}
    \item Structural and perspective coherence of the environment.
    \vspace{-0.1cm}
    \item Visual stability without severe flicker, tearing, or geometric warping.
    \vspace{-0.1cm}
    \item Realistic lighting, shadows, and surface textures.
    \vspace{-0.1cm}
    \item Consistent composition of static (roads, buildings, sky) and dynamic (vehicles, pedestrians) elements.
\end{itemize}

Higher \emph{Overall Realism} indicates that generated scenes are globally coherent, visually stable, and perceptually indistinguishable from real-world videos.

\subsubsection{~Protocol}
Each generated video is rated on a $1$–$10$ scale of perceived realism:

\begin{table}[h]
    \centering
    \label{tab:human_overall_realism}
    \resizebox{\linewidth}{!}{
    \begin{tabular}{|c|c|p{350pt}<{\raggedright}|}
        \hline
        \textbf{Score} & \textbf{Level} & \textbf{Description}
        \\
        \hline\hline
        \multirow{2}{*}{$1$} & \multirow{2}{*}{Extremely Unrealistic} & Severe structural and temporal defects; global flicker or collapse of scene; 
        \\
        & & lighting/texture incoherent; scene immediately identifiable as synthetic.
        \\
        \hline
        \multirow{2}{*}{$3$} & \multirow{2}{*}{Unrealistic} & Local artifacts such as inconsistent textures, ghosting, or unstable motion, but 
        \\
        & & the overall layout remains interpretable.
        \\
        \hline
        \multirow{2}{*}{$5$} & \multirow{2}{*}{Moderately Realistic} & Global appearance mostly coherent with minor motion discontinuities or soft 
        \\
        & & blur;  realism is partially convincing.
        \\
        \hline
        \multirow{2}{*}{$7$} & \multirow{2}{*}{Realistic} & Scene composition, motion continuity, and lighting are natural; just some small 
        \\
        & & imperfections but do not affect perceived realism.
        \\
        \hline
        \multirow{2}{*}{$9$} & \multirow{2}{*}{Highly Realistic} & Scene fully photorealistic in both space and time; almost indistinguishable from 
        \\
        & & real-world footage by human eyes.
        \\
        \hline
        \multirow{2}{*}{$10$} & \multirow{2}{*}{Ground Truth} & \multirow{2}{*}{-}
        \\
        & & 
        \\
        \hline
    \end{tabular}}
\end{table}

\vspace{-0.3cm}
\subsubsection{~Examples}
\begin{figure}[t]
    \centering
    \begin{subfigure}[h]{\textwidth}
        \centering
        \includegraphics[width=\linewidth]{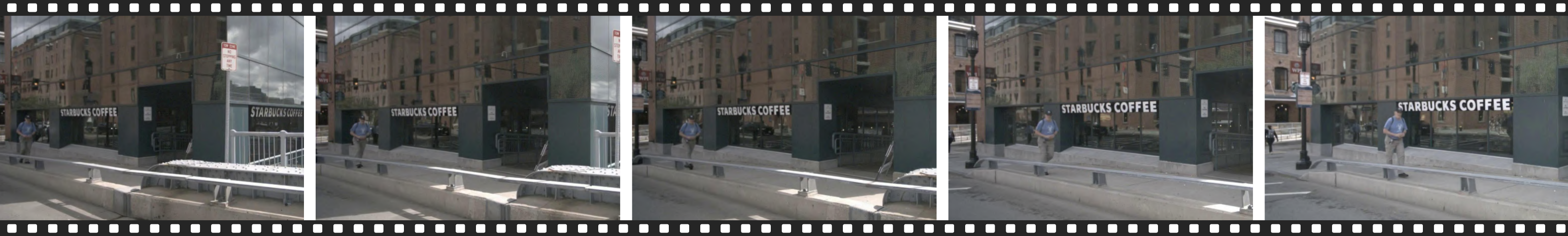}
        \caption{Good example in the \emph{Overall Realism} dimension (Score: \textcolor{w_blue}{$10$})}
        \label{fig:human_overall_realism_1}
    \end{subfigure}
    \\[1ex]
    \begin{subfigure}[h]{\textwidth}
        \centering
        \includegraphics[width=\linewidth]{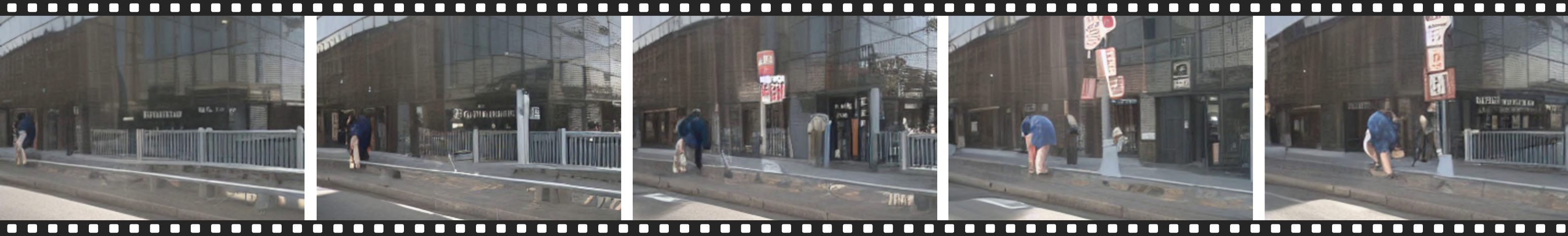}
        \caption{Bad example in the \emph{Overall Realism} dimension (Score: \textcolor{red}{$1$})}
        \label{fig:human_overall_realism_2}
    \end{subfigure}
    \\[3.5ex]
    \begin{subfigure}[h]{\textwidth}
        \centering
        \includegraphics[width=\linewidth]{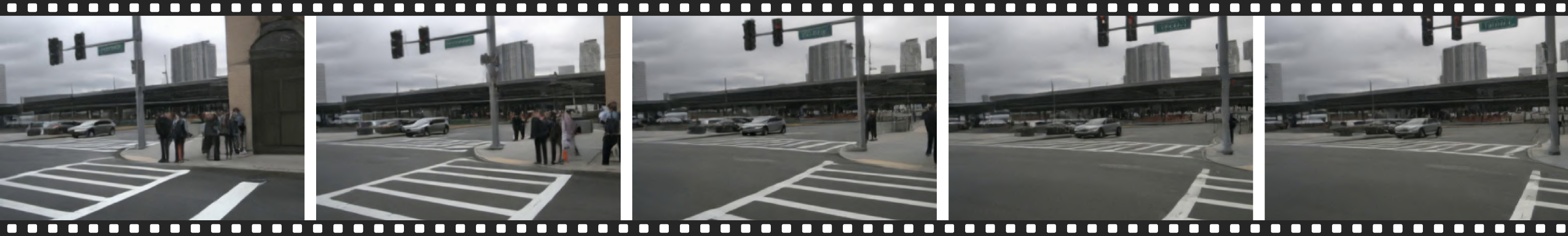}
        \caption{Good example in the \emph{Overall Realism} dimension (Score: \textcolor{w_blue}{$6$})}
        \label{fig:human_overall_realism_3}
    \end{subfigure}
    \\[1ex]
    \begin{subfigure}[h]{\textwidth}
        \centering
        \includegraphics[width=\linewidth]{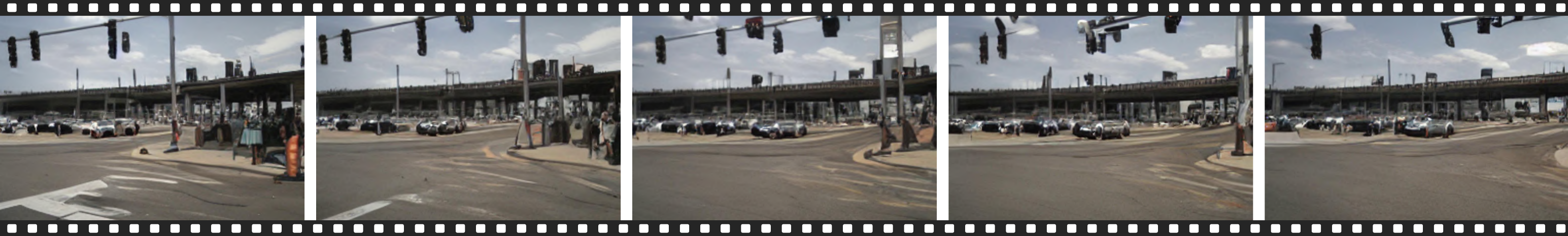}
        \caption{Bad example in the \emph{Overall Realism} dimension (Score: \textcolor{red}{$1$})}
        \label{fig:human_overall_realism_4}
    \end{subfigure}
    \\[3.5ex]
    \begin{subfigure}[h]{\textwidth}
        \centering
        \includegraphics[width=\linewidth]{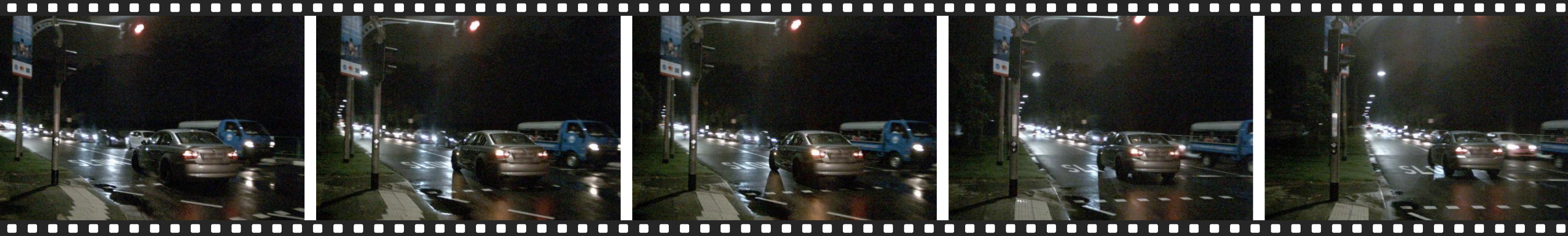}
        \caption{Good example in the \emph{Overall Realism} dimension (Score: \textcolor{w_blue}{$8$})}
        \label{fig:human_overall_realism_5}
    \end{subfigure}
    \\[1ex]
    \begin{subfigure}[h]{\textwidth}
        \centering
        \includegraphics[width=\linewidth]{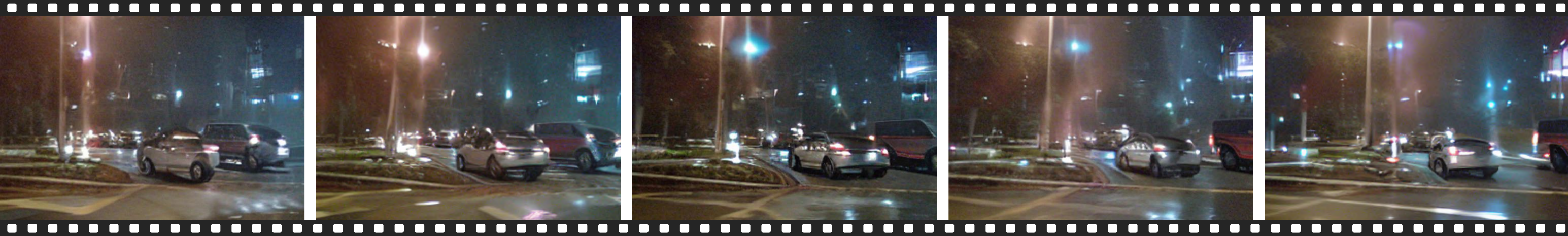}
        \caption{Bad example in the \emph{Overall Realism} dimension (Score: \textcolor{red}{$1$})}
        \label{fig:human_overall_realism_6}
    \end{subfigure}
    \vspace{-0.1cm}
    \caption{Examples of ``good'' and ``bad'' human preference alignments in terms of \emph{Overall Realism} in WorldLens.}
    \label{fig:human_overall_realism}
\end{figure}

Figure~\ref{fig:human_overall_realism} provides typical examples of videos with good and bad quality in terms of \emph{Overall Realism}.

\vspace{-0.1cm}
\subsubsection{~Evaluation \& Analysis}
Table~\ref{tab:supp_human_overall_realism} provides the complete results of models in terms of \emph{Overall Realism}.

\begin{table*}[h]
    \centering
    \vspace{0.2cm}
    \caption{Complete comparisons of state-of-the-art driving world models in terms of \emph{Overall Realism} in WorldLens.}
    \vspace{-0.2cm}
    \label{tab:supp_human_overall_realism}
    \resizebox{\linewidth}{!}{
    \begin{tabular}{r|cccccc|c}
        \toprule
        Overall & \textbf{MagicDrive} & \textbf{DreamForge}  & \textbf{DriveDreamer-2} & \textbf{OpenDWM} & \textbf{~DiST-4D~} & $\mathcal{X}$\textbf{-Scene} & \textcolor{gray}{\textbf{Empirical}} \\
        Realism & \textcolor{gray}{\small[ICLR'24]} & \textcolor{gray}{\small[arXiv'24]} & \textcolor{gray}{\small[AAAI'25]} & \textcolor{gray}{\small[CVPR'25]} & \textcolor{gray}{\small[ICCV'25]} & \textcolor{gray}{\small[NeurIPS'25]} & \textcolor{gray}{\textbf{Max}} 
        \\
        \midrule
        \textsl{min} 
            & $2.0$ & $2.0$ & $2.0$ & $2.0$ & $2.0$ & $2.0$ 
            & \cellcolor{gray!7}\textcolor{gray}{-} \\
        \textsl{max} 
            & $4.0$ & $6.0$ & $6.0$ & $6.0$ & $6.0$ & $4.0$
            & \cellcolor{gray!7}\textcolor{gray}{-} \\
        \textbf{\textsl{mean}}
            & \cellcolor{w_blue!20}$2.062$ & \cellcolor{w_blue!20}$2.204$ & \cellcolor{w_blue!20}$2.256$ & \cellcolor{w_blue!20}$2.209$ & \cellcolor{w_blue!20}$2.320$ & \cellcolor{w_blue!20}$2.080$
            & \cellcolor{gray!7}\textcolor{gray}{$10$} \\
        \textsl{std} 
            & $0.347$ & $0.620$ & $0.865$ & $0.801$ & $0.912$ & $0.392$
            & \cellcolor{gray!7}\textcolor{gray}{-} \\
        \textsl{median} 
            & $2.0$ & $2.0$ & $2.0$ & $2.0$ & $2.0$ & $2.0$
            & \cellcolor{gray!7}\textcolor{gray}{-} \\
        \textsl{q25} 
            & $2.0$ & $2.0$ & $2.0$ & $2.0$ & $2.0$ & $2.0$
            & \cellcolor{gray!7}\textcolor{gray}{-} \\
        \textsl{q75} 
            & $2.0$ & $2.0$ & $2.0$ & $2.0$ & $2.0$ & $2.0$
            & \cellcolor{gray!7}\textcolor{gray}{-} \\
        \bottomrule
    \end{tabular}}
    \vspace{-0.4cm}
\end{table*}

%% file: sections_supp/dimensions/human_vehicle_realism.tex
\subsection{~World Realism - Vehicle Realism}
Vehicle Realism isolates the perceptual authenticity of vehicles within the scene, focusing solely on their visual appearance. Annotators evaluate whether vehicles ``look like real cars". Annotators are instructed to judge each clip according to the following criteria:

\begin{itemize}
    \item Correct body shape, door/roof/wheel-arch proportions, and stable contours without deformation.
    \vspace{-0.1cm}
    \item Realistic metallic paint, plastic, glass, tires, and recognizable small components (logos, grilles, lamps).
    \vspace{-0.1cm}
    \item Natural highlights, shadows, and reflections under various weather and illumination conditions.
    \vspace{-0.1cm}
    \item Color, texture, and boundary stability across adjacent frames.
\end{itemize}

High \emph{Vehicle Realism} reflects consistent car geometry, convincing materials, physically plausible reflections, and temporally stable rendering. Low scores correspond to warped, ``rubber-like" cars with flickering colors, melted textures, or incoherent lighting.

\subsubsection{~Protocol}
Each generated video is rated on a $1$–$10$ scale of perceived realism:

\begin{table}[h]
    \centering
    \label{tab:human_vehicle_realism}
    \resizebox{\linewidth}{!}{
    \begin{tabular}{|c|c|p{350pt}<{\raggedright}|}
        \hline
        \textbf{Score} & \textbf{Level} & \textbf{Description}
        \\
        \hline\hline
        \multirow{2}{*}{$1$} & \multirow{2}{*}{Extremely Unrealistic} & Vehicle geometry or texture severely distorted; missing parts, collapsed meshes,
        \\
        & & or flickering silhouettes; clearly fake appearance.
        \\
        \hline
        \multirow{2}{*}{$3$} & \multirow{2}{*}{Unrealistic} & Vehicles roughly shaped but show color inconsistency, unstable reflections, or
        \\
        & & unnatural motion patterns.
        \\
        \hline
        \multirow{2}{*}{$5$} & \multirow{2}{*}{Moderately Realistic} & Vehicles recognizable with mostly correct proportions and materials; small surface
        \\
        & & or temporal noise visible.
        \\
        \hline
        \multirow{2}{*}{$7$} & \multirow{2}{*}{Realistic} & Vehicle shape, motion, and illumination coherent and stable; only have some 
        \\
        & & minor local imperfections.
        \\
        \hline
        \multirow{2}{*}{$9$} & \multirow{2}{*}{Highly Realistic} & Fully natural vehicles with correct proportions, lighting response, and dynamic 
        \\
        & & reflections; seamlessly integrated in the scene.
        \\
        \hline
        \multirow{2}{*}{$10$} & \multirow{2}{*}{Ground Truth} & \multirow{2}{*}{-}
        \\
        & & 
        \\
        \hline
    \end{tabular}}
\end{table}

\vspace{-0.3cm}
\subsubsection{~Examples}
\begin{figure}[t]
    \centering
    \begin{subfigure}[h]{\textwidth}
        \centering
        \includegraphics[width=\linewidth]{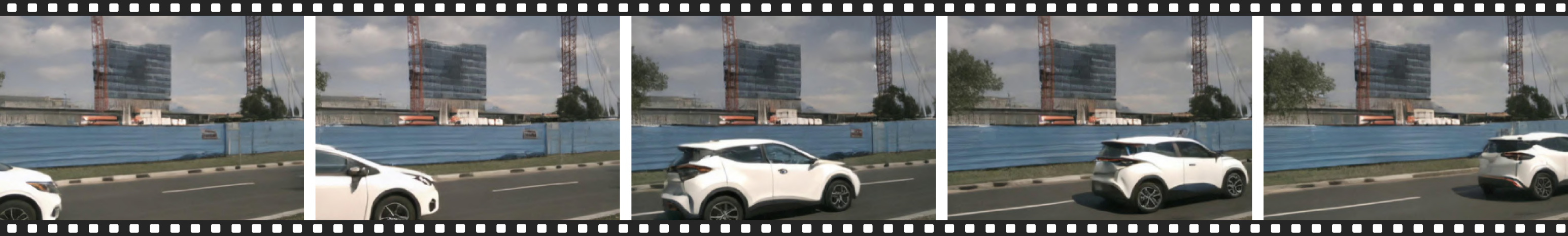}
        \caption{Good example in the \emph{Vehicle Realism} dimension (Score: \textcolor{w_blue}{$8$})}
        \label{fig:human_vehicle_realism_1}
    \end{subfigure}
    \\[1ex]
    \begin{subfigure}[h]{\textwidth}
        \centering
        \includegraphics[width=\linewidth]{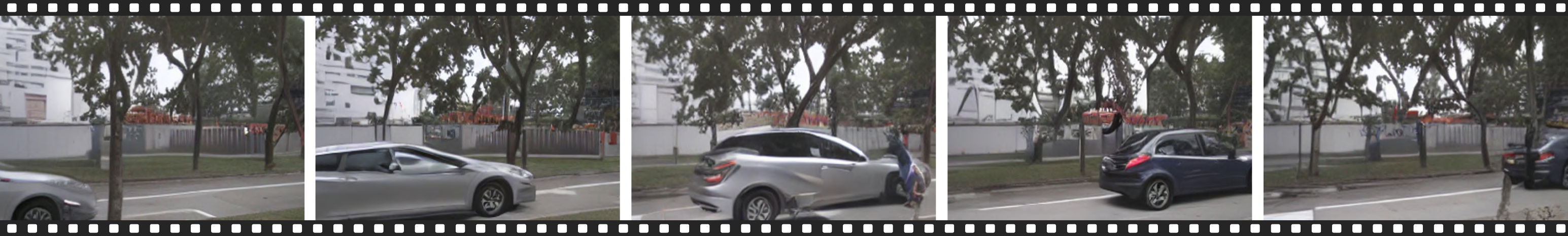}
        \caption{Bad example in the \emph{Vehicle Realism} dimension (Score: \textcolor{red}{$1$})}
        \label{fig:human_vehicle_realism_2}
    \end{subfigure}
    \\[3.5ex]
    \begin{subfigure}[h]{\textwidth}
        \centering
        \includegraphics[width=\linewidth]{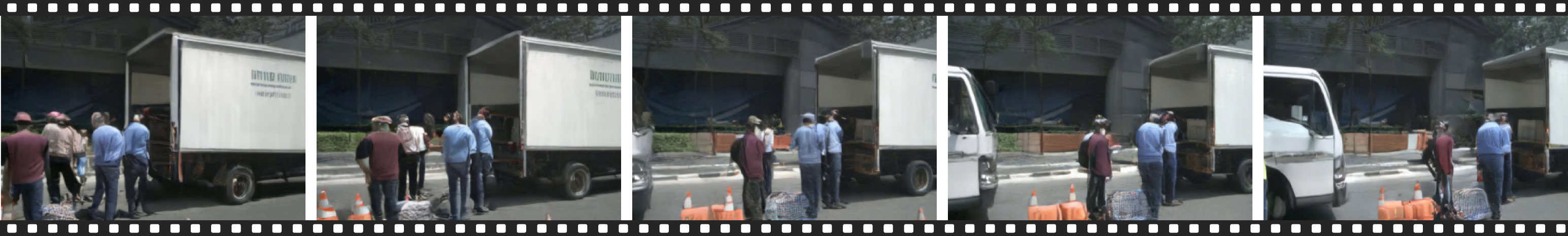}
        \caption{Good example in the \emph{Vehicle Realism} dimension (Score: \textcolor{w_blue}{$6$})}
        \label{fig:human_vehicle_realism_3}
    \end{subfigure}
    \\[1ex]
    \begin{subfigure}[h]{\textwidth}
        \centering
        \includegraphics[width=\linewidth]{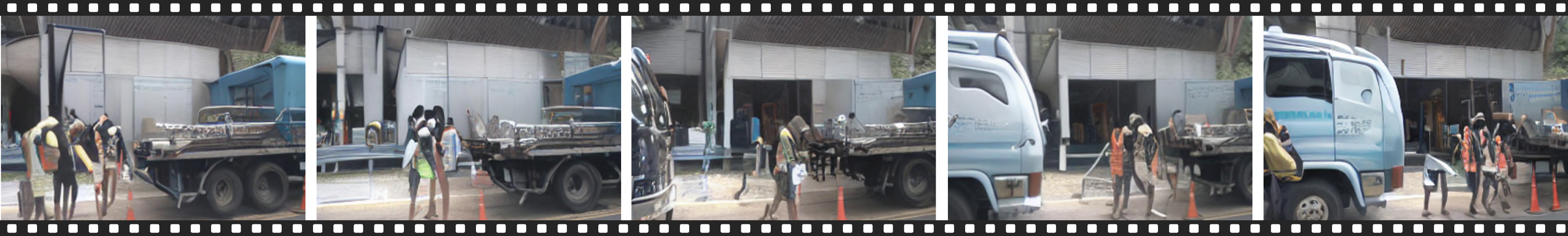}
        \caption{Bad example in the \emph{Vehicle Realism} dimension (Score: \textcolor{red}{$1$})}
        \label{fig:human_vehicle_realism_4}
    \end{subfigure}
    \\[3.5ex]
    \begin{subfigure}[h]{\textwidth}
        \centering
        \includegraphics[width=\linewidth]{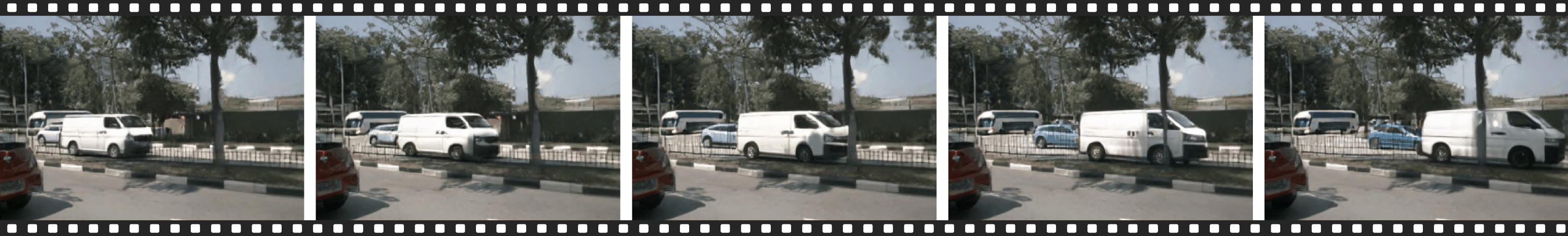}
        \caption{Good example in the \emph{Vehicle Realism} dimension (Score: \textcolor{w_blue}{$8$})}
        \label{fig:human_vehicle_realism_5}
    \end{subfigure}
    \\[1ex]
    \begin{subfigure}[h]{\textwidth}
        \centering
        \includegraphics[width=\linewidth]{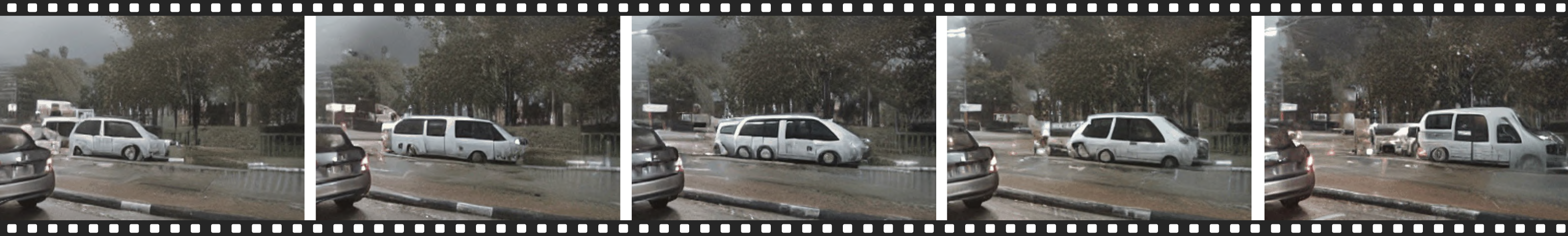}
        \caption{Bad example in the \emph{Vehicle Realism} dimension (Score: \textcolor{red}{$1$})}
        \label{fig:human_vehicle_realism_6}
    \end{subfigure}
    \vspace{-0.1cm}
    \caption{Examples of ``good'' and ``bad'' human preference alignments in terms of \emph{Vehicle Realism} in WorldLens.}
    \label{fig:human_vehicle_realism}
\end{figure}

Figure~\ref{fig:human_vehicle_realism} provides typical examples of videos with good and bad quality in terms of \emph{Vehicle Realism}.

\subsubsection{~Evaluation \& Analysis}
Table~\ref{tab:supp_human_vehicle_realism} provides the complete results of models in terms of \emph{Vehicle Realism}.

\begin{table*}[h]
    \centering
    \vspace{0.2cm}
    \caption{Complete comparisons of state-of-the-art driving world models in terms of \emph{Vehicle Realism} in WorldLens.}
    \vspace{-0.2cm}
    \label{tab:supp_human_vehicle_realism}
    \resizebox{\linewidth}{!}{
    \begin{tabular}{r|cccccc|c}
        \toprule
        Vehicle & \textbf{MagicDrive} & \textbf{DreamForge}  & \textbf{DriveDreamer-2} & \textbf{OpenDWM} & \textbf{~DiST-4D~} & $\mathcal{X}$\textbf{-Scene} & \textcolor{gray}{\textbf{Empirical}} \\
        Realism & \textcolor{gray}{\small[ICLR'24]} & \textcolor{gray}{\small[arXiv'24]} & \textcolor{gray}{\small[AAAI'25]} & \textcolor{gray}{\small[CVPR'25]} & \textcolor{gray}{\small[ICCV'25]} & \textcolor{gray}{\small[NeurIPS'25]} & \textcolor{gray}{\textbf{Max}} 
        \\
        \midrule
        \textsl{min} 
            & $2.0$ & $2.0$ & $2.0$ & $2.0$ & $2.0$ & $2.0$
            & \cellcolor{gray!7}\textcolor{gray}{-} \\
        \textsl{max} 
            & $4.0$ & $6.0$ & $8.0$ & $8.0$ & $8.0$ & $8.0$
            & \cellcolor{gray!7}\textcolor{gray}{-} \\
        \textbf{\textsl{mean}}
            & \cellcolor{w_blue!20}$2.036$ 
            & \cellcolor{w_blue!20}$2.043$ 
            & \cellcolor{w_blue!20}$2.720$ 
            & \cellcolor{w_blue!20}$2.757$ 
            & \cellcolor{w_blue!20}$2.328$ 
            & \cellcolor{w_blue!20}$2.216$
            & \cellcolor{gray!7}\textcolor{gray}{$10$} \\
        \textsl{std} 
            & $0.268$ & $0.328$ & $1.700$ & $1.584$ & $1.011$ & $0.808$
            & \cellcolor{gray!7}\textcolor{gray}{-} \\
        \textsl{median} 
            & $2.0$ & $2.0$ & $2.0$ & $2.0$ & $2.0$ & $2.0$
            & \cellcolor{gray!7}\textcolor{gray}{-} \\
        \textsl{q25} 
            & $2.0$ & $2.0$ & $2.0$ & $2.0$ & $2.0$ & $2.0$
            & \cellcolor{gray!7}\textcolor{gray}{-} \\
        \textsl{q75} 
            & $2.0$ & $2.0$ & $2.0$ & $2.0$ & $2.0$ & $2.0$
            & \cellcolor{gray!7}\textcolor{gray}{-} \\
        \bottomrule
    \end{tabular}}
    \vspace{-0.4cm}
\end{table*}

%% file: sections_supp/dimensions/human_pedestrian_realism.tex
\subsection{~World Realism - Pedestrian Realism}

Pedestrian Realism measures whether humans in generated videos look and move like real people. It focuses on anatomical plausibility, natural appearance, and temporal stability of pedestrians. Annotators are instructed to judge each clip according to the following criteria:
\begin{itemize}
    \item Realistic head-torso-limb ratios, joint positions, and poses without twisted or intersecting limbs.
    \vspace{-0.1cm}
    \item Plausible garment structure, texture clarity, and consistency of accessories. 
    \vspace{-0.1cm}
    \item Smooth and natural shading without wax-like or distorted faces.
    \vspace{-0.1cm}
    \item Continuous appearance without flickering, sliding, or sudden disappearance.
    \vspace{-0.1cm}
    \item Whether pedestrians resemble real filmed humans rather than avatars or composites.
\end{itemize}

Higher \emph{Pedestrian Realism} indicates pedestrians with stable body structures, coherent motion, realistic textures, and natural temporal behavior.

\subsubsection{~Protocol}
Each generated video is rated on a $1$–$10$ scale of perceived realism:

\begin{table}[h]
    \centering
    \label{tab:human_pedestrian_realism}
    \resizebox{\linewidth}{!}{
    \begin{tabular}{|c|c|p{350pt}<{\raggedright}|}
        \hline
        \textbf{Score} & \textbf{Level} & \textbf{Description}
        \\
        \hline\hline
        \multirow{2}{*}{$1$} & \multirow{2}{*}{Extremely Unrealistic} & Human figures deformed or incomplete; limbs twisted or merged; motion violates body 
        \\
        & & mechanics; instantly recognizable as artificial.
        \\
        \hline
        \multirow{2}{*}{$3$} & \multirow{2}{*}{Unrealistic} & Human silhouettes intact but with visible motion or shape glitches, coarse skin/cloth 
        \\
        & & texture, or flicker; gait unnatural.
        \\
        \hline
        \multirow{2}{*}{$5$} & \multirow{2}{*}{Moderately Realistic} &  Pedestrians generally human-like with slight stiffness or occasional temporal instability, 
        \\
        & & shape distortions, or texture issues.
        \\
        \hline
        \multirow{2}{*}{$7$} & \multirow{2}{*}{Realistic} & Natural body proportions, coherent motion, and stable clothing appearance; plausible 
        \\
        & & human dynamics.
        \\
        \hline
        \multirow{2}{*}{$9$} & \multirow{2}{*}{Highly Realistic} & Anatomically and kinematically accurate humans with smooth gait, fine-grained 
        \\
        & & details, and temporally consistent appearance.
        \\
        \hline
        \multirow{2}{*}{$10$} & \multirow{2}{*}{Ground Truth} & \multirow{2}{*}{-}
        \\
        & & 
        \\
        \hline
    \end{tabular}}
\end{table}

\vspace{-0.3cm}
\subsubsection{~Examples}
\begin{figure}[t]
    \centering
    \begin{subfigure}[h]{\textwidth}
        \centering
        \includegraphics[width=\linewidth]{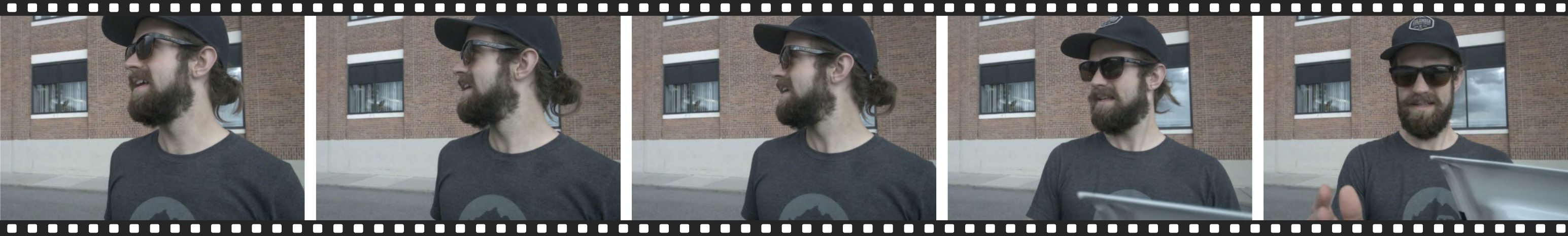}
        \caption{Good example in the \emph{Pedestrian Realism} dimension (Score: \textcolor{w_blue}{$10$})}
        \label{fig:human_pedestrian_realism_1}
    \end{subfigure}
    \\[1ex]
    \begin{subfigure}[h]{\textwidth}
        \centering
        \includegraphics[width=\linewidth]{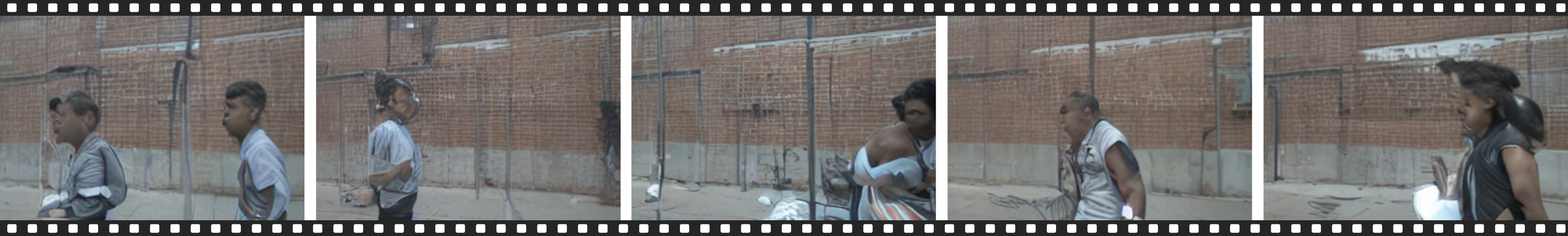}
        \caption{Bad example in the \emph{Pedestrian Realism} dimension (Score: \textcolor{red}{$1$})}
        \label{fig:human_pedestrian_realism_2}
    \end{subfigure}
    \\[3.5ex]
    \begin{subfigure}[h]{\textwidth}
        \centering
        \includegraphics[width=\linewidth]{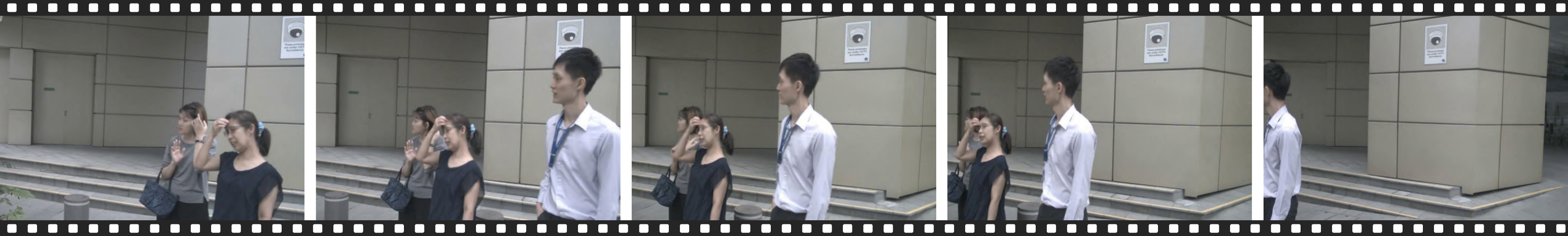}
        \caption{Good example in the \emph{Pedestrian Realism} dimension (Score: \textcolor{w_blue}{$10$})}
        \label{fig:human_pedestrian_realism_3}
    \end{subfigure}
    \\[1ex]
    \begin{subfigure}[h]{\textwidth}
        \centering
        \includegraphics[width=\linewidth]{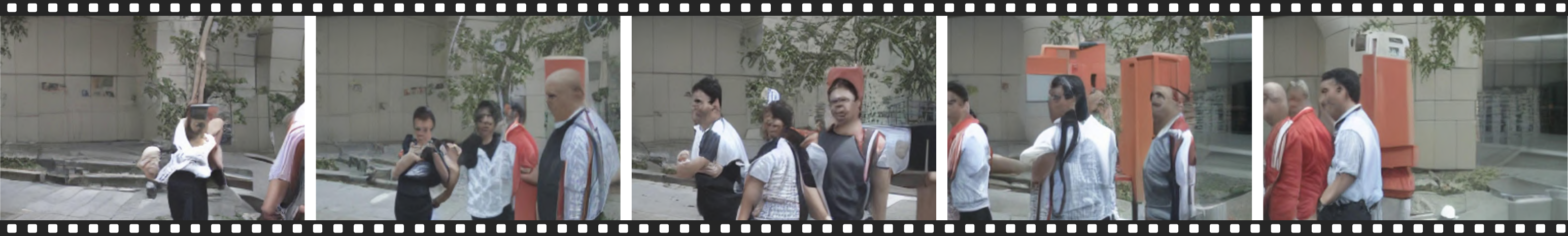}
        \caption{Bad example in the \emph{Pedestrian Realism} dimension (Score: \textcolor{red}{$1$})}
        \label{fig:human_pedestrian_realism_4}
    \end{subfigure}
    \\[3.5ex]
    \begin{subfigure}[h]{\textwidth}
        \centering
        \includegraphics[width=\linewidth]{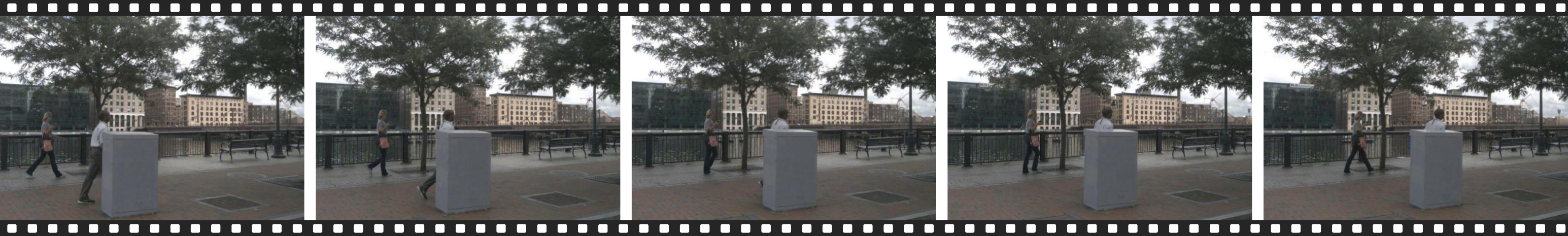}
        \caption{Good example in the \emph{Pedestrian Realism} dimension (Score: \textcolor{w_blue}{$10$})}
        \label{fig:human_pedestrian_realism_5}
    \end{subfigure}
    \\[1ex]
    \begin{subfigure}[h]{\textwidth}
        \centering
        \includegraphics[width=\linewidth]{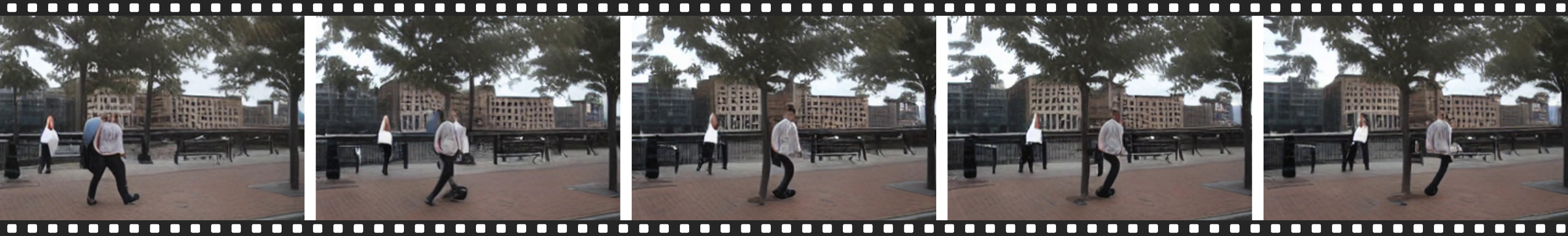}
        \caption{Bad example in the \emph{Pedestrian Realism} dimension (Score: \textcolor{red}{$1$})}
        \label{fig:human_pedestrian_realism_6}
    \end{subfigure}
    \vspace{-0.1cm}
    \caption{Examples of ``good'' and ``bad'' human preference alignments in terms of \emph{Pedestrian Realism} in WorldLens.}
    \label{fig:human_pedestrian_realism}
\end{figure}

Figure~\ref{fig:human_pedestrian_realism} provides typical examples of videos with good and bad quality in terms of \emph{Pedestrian Realism}.

\subsubsection{~Evaluation \& Analysis}
Table~\ref{tab:supp_human_pedestrian_realism} provides the complete results of models in terms of \emph{Pedestrian Realism}.

\begin{table*}[h]
    \centering
    \vspace{0.2cm}
    \caption{Complete comparisons of state-of-the-art driving world models in terms of \emph{Pedestrian Realism} in WorldLens.}
    \vspace{-0.2cm}
    \label{tab:supp_human_pedestrian_realism}
    \resizebox{\linewidth}{!}{
    \begin{tabular}{r|cccccc|c}
        \toprule
        Pedestrian & \textbf{MagicDrive} & \textbf{DreamForge}  & \textbf{DriveDreamer-2} & \textbf{OpenDWM} & \textbf{~DiST-4D~} & $\mathcal{X}$\textbf{-Scene} & \textcolor{gray}{\textbf{Empirical}} \\
        Realism & \textcolor{gray}{\small[ICLR'24]} & \textcolor{gray}{\small[arXiv'24]} & \textcolor{gray}{\small[AAAI'25]} & \textcolor{gray}{\small[CVPR'25]} & \textcolor{gray}{\small[ICCV'25]} & \textcolor{gray}{\small[NeurIPS'25]} & \textcolor{gray}{\textbf{Max}} 
        \\
        \midrule
        \textsl{min} 
            & $2.0$ & $2.0$ & $2.0$ & $2.0$ & $2.0$ & $2.0$
            & \cellcolor{gray!7}\textcolor{gray}{-} \\
        \textsl{max} 
            & $4.0$ & $6.0$ & $6.0$ & $4.0$ & $6.0$ & $4.0$
            & \cellcolor{gray!7}\textcolor{gray}{-} \\
        \textbf{\textsl{mean}}
            & \cellcolor{w_blue!20}$2.288$
            & \cellcolor{w_blue!20}$2.352$
            & \cellcolor{w_blue!20}$2.341$
            & \cellcolor{w_blue!20}$2.325$
            & \cellcolor{w_blue!20}$2.406$
            & \cellcolor{w_blue!20}$2.293$
            & \cellcolor{gray!7}\textcolor{gray}{$10$} \\
        \textsl{std} 
            & $0.618$ & $0.727$ & $0.703$ & $0.671$ & $0.832$ & $0.629$
            & \cellcolor{gray!7}\textcolor{gray}{-} \\
        \textsl{median} 
            & $2.0$ & $2.0$ & $2.0$ & $2.0$ & $2.0$ & $2.0$
            & \cellcolor{gray!7}\textcolor{gray}{-} \\
        \textsl{q25} 
            & $2.0$ & $2.0$ & $2.0$ & $2.0$ & $2.0$ & $2.0$
            & \cellcolor{gray!7}\textcolor{gray}{-} \\
        \textsl{q75} 
            & $2.0$ & $2.0$ & $2.0$ & $2.0$ & $2.0$ & $2.0$
            & \cellcolor{gray!7}\textcolor{gray}{-} \\
        \bottomrule
    \end{tabular}}
    \vspace{-0.4cm}
\end{table*}

%% file: sections_supp/dimensions/human_physical_plausibility.tex
\subsection{~Physical Plausibility}

Physical Plausibility evaluates whether the motions, interactions, and visual evolution of a generated driving scene are consistent with basic physical laws and causal structure in the real world. This dimension explicitly targets \emph{physics and dynamics}: whether objects move, collide, occlude, and respond in ways that respect continuity, inertia, contact, and depth ordering. Annotators are instructed to judge each clip according to the following criteria:

\begin{itemize}
    \item Positions, velocities, colors, and textures should evolve smoothly over time, without teleportation, duplication, spontaneous appearance or disappearance, or violent jumps in shape or brightness.
    \vspace{-0.1cm}
    \item Vehicles, pedestrians, and static elements (barriers, poles, buildings) should not interpenetrate. Feet should visually remain on the ground when walking, and objects should not float or sink into surfaces.
    \vspace{-0.1cm}
    \item Foreground and background elements should obey consistent occlusion relationships. Distant objects should not suddenly occlude closer ones, and elements like traffic lights, fences, and signboards should not phase through other geometry.
    \vspace{-0.1cm}
    \item Highlights, reflections, glare, and shadows should change smoothly with camera motion and object movement, without unexplained flashes, patches of incoherent reflection, or abrupt brightness jumps.
\end{itemize}

Higher \emph{Physical Plausibility} indicates that generated worlds exhibit more physically consistent dynamics.

\subsubsection{~Protocol}
Each generated video is rated on a $1$–$10$ scale of perceived realism:

\begin{table}[h]
    \centering
    \label{tab:human_physical_plausibility}
    \resizebox{\linewidth}{!}{
    \begin{tabular}{|c|c|p{350pt}<{\raggedright}|}
        \hline
        \textbf{Score} & \textbf{Level} & \textbf{Description}
        \\
        \hline\hline
        \multirow{2}{*}{$1$} & \multirow{2}{*}{Extremely Implausible} & Frequent physics violations: teleportation, penetration, inconsistent occlusion, 
        \\
        & & abrupt geometry or lighting jumps.
        \\
        \hline
        \multirow{2}{*}{$3$} & \multirow{2}{*}{Implausible} & Localized but noticeable non-physical events (\emph{e.g.}, a single penetration or 
        \\
        & & transient reflection anomaly); scene remains somewhat coherent.
        \\
        \hline
        \multirow{2}{*}{$5$} & \multirow{2}{*}{Moderately Plausible} & Motion and contact mostly realistic with occasional small violations (\emph{e.g.}, light  
        \\
        & & flicker, minor occlusion inversion); perceptually acceptable but imperfect.
        \\
        \hline
        \multirow{2}{*}{$7$} & \multirow{2}{*}{Plausible} & Motion, occlusion, and lighting largely follow physical laws; minor irregularities 
        \\
        & & remain in non-critical regions.
        \\
        \hline
        \multirow{2}{*}{$9$} & \multirow{2}{*}{Highly Plausible} & Entire clip adheres to continuity, contact, reflection, and causality constraints; 
        \\
        & & fully consistent with real-world physics.
        \\
        \hline
        \multirow{2}{*}{$10$} & \multirow{2}{*}{Ground Truth} & \multirow{2}{*}{-}
        \\
        & & 
        \\
        \hline
    \end{tabular}}
\end{table}

\vspace{-0.3cm}
\subsubsection{~Examples}
\begin{figure}[t]
    \centering
    \begin{subfigure}[h]{\textwidth}
        \centering
        \includegraphics[width=\linewidth]{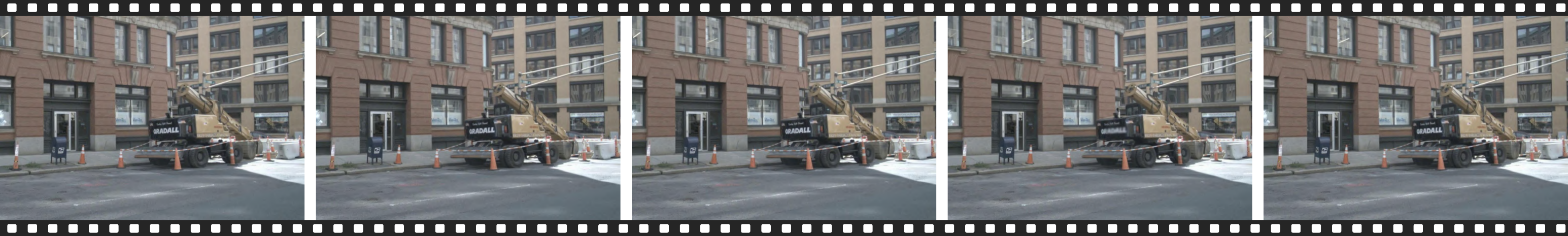}
        \caption{Good example in the \emph{Physical Plausibility} dimension (Score: \textcolor{w_blue}{$10$})}
        \label{fig:human_physical_plausibility_1}
    \end{subfigure}
    \\[1ex]
    \begin{subfigure}[h]{\textwidth}
        \centering
        \includegraphics[width=\linewidth]{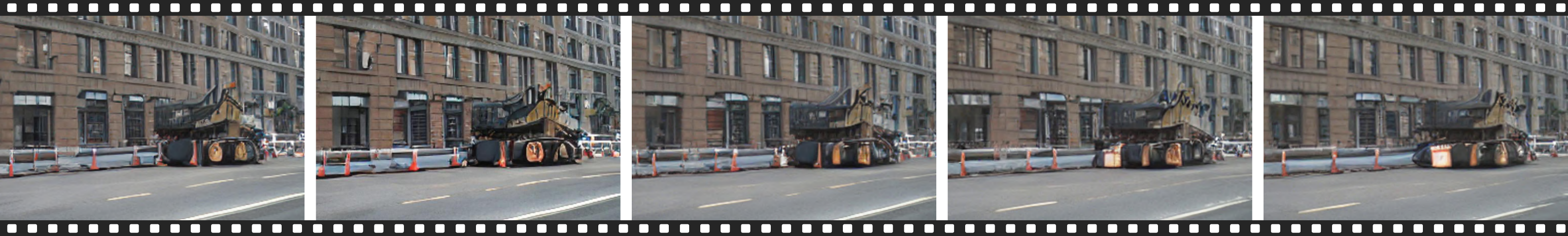}
        \caption{Bad example in the \emph{Physical Plausibility} dimension (Score: \textcolor{red}{$1$})}
        \label{fig:human_physical_plausibility_2}
    \end{subfigure}
    \\[3.5ex]
    \begin{subfigure}[h]{\textwidth}
        \centering
        \includegraphics[width=\linewidth]{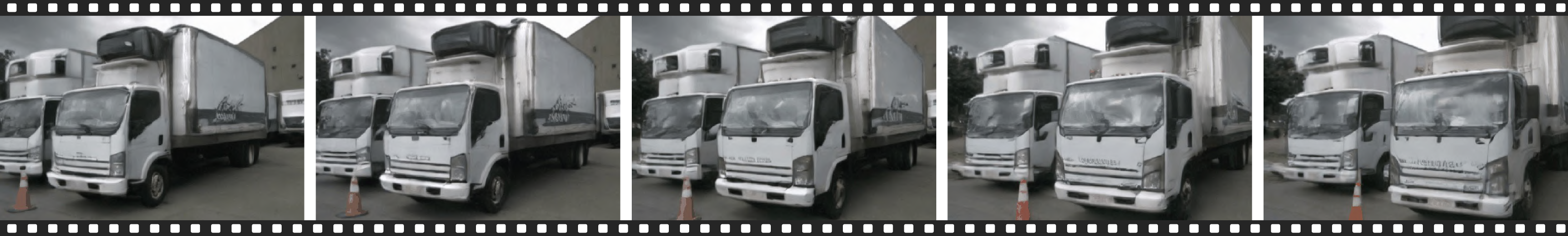}
        \caption{Good example in the \emph{Physical Plausibility} dimension (Score: \textcolor{w_blue}{$8$})}
        \label{fig:human_physical_plausibility_3}
    \end{subfigure}
    \\[1ex]
    \begin{subfigure}[h]{\textwidth}
        \centering
        \includegraphics[width=\linewidth]{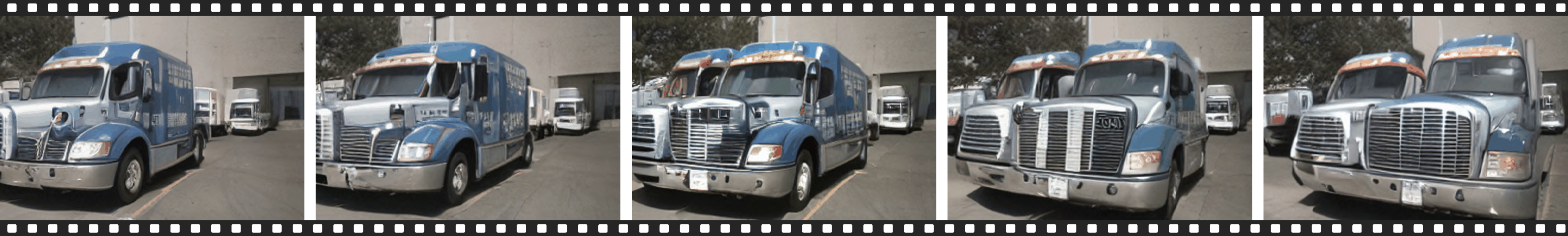}
        \caption{Bad example in the \emph{Physical Plausibility} dimension (Score: \textcolor{red}{$1$})}
        \label{fig:human_physical_plausibility_4}
    \end{subfigure}
    \\[3.5ex]
    \begin{subfigure}[h]{\textwidth}
        \centering
        \includegraphics[width=\linewidth]{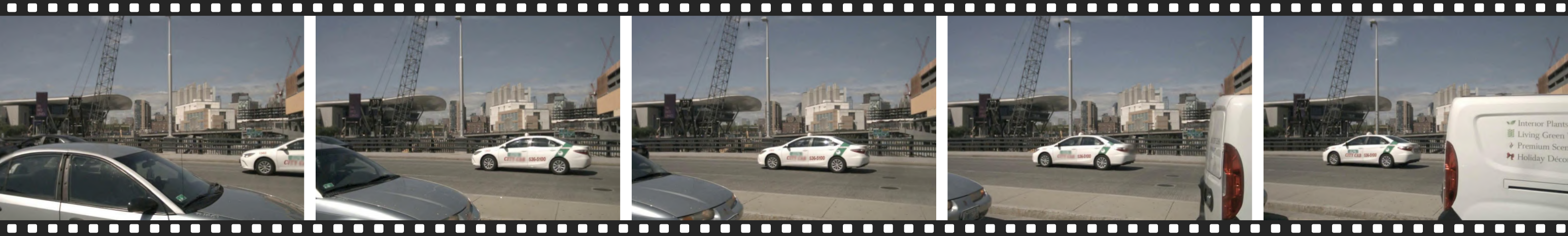}
        \caption{Good example in the \emph{Physical Plausibility} dimension (Score: \textcolor{w_blue}{$10$})}
        \label{fig:human_physical_plausibility_5}
    \end{subfigure}
    \\[1ex]
    \begin{subfigure}[h]{\textwidth}
        \centering
        \includegraphics[width=\linewidth]{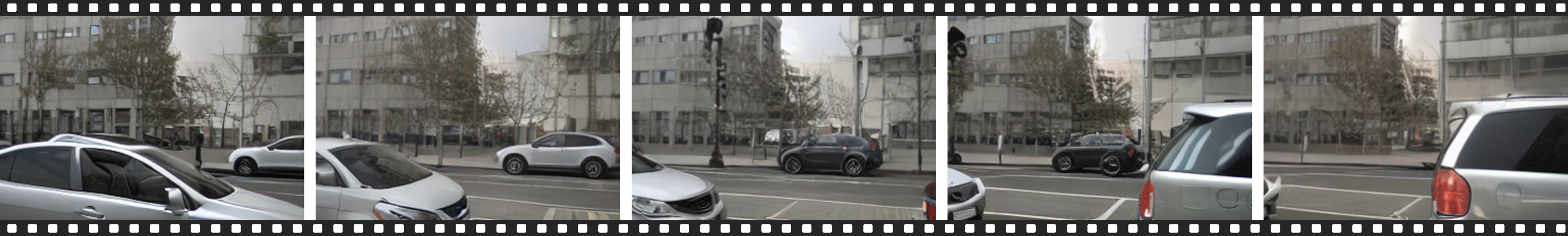}
        \caption{Bad example in the \emph{Physical Plausibility} dimension (Score: \textcolor{red}{$1$})}
        \label{fig:human_physical_plausibility_6}
    \end{subfigure}
    \vspace{-0.1cm}
    \caption{Examples of ``good'' and ``bad'' human preference alignments in terms of \emph{Physical Plausibility} in WorldLens.}
    \label{fig:human_physical_plausibility}
\end{figure}

Figure~\ref{fig:human_physical_plausibility} provides typical examples of videos with good and bad quality in terms of \emph{Physical Plausibility}.

\subsubsection{~Evaluation \& Analysis}
Table~\ref{tab:supp_human_physical_plausibility} provides the complete results of models in terms of \emph{Physical Plausibility}.

\begin{table*}[h]
    \centering
    \vspace{0.2cm}
    \caption{Complete comparisons of state-of-the-art driving world models in terms of \emph{Physical Plausibility} in WorldLens.}
    \vspace{-0.2cm}
    \label{tab:supp_human_physical_plausibility}
    \resizebox{\linewidth}{!}{
    \begin{tabular}{r|cccccc|c}
        \toprule
        Physical & \textbf{MagicDrive} & \textbf{DreamForge}  & \textbf{DriveDreamer-2} & \textbf{OpenDWM} & \textbf{~DiST-4D~} & $\mathcal{X}$\textbf{-Scene} & \textcolor{gray}{\textbf{Empirical}} \\
        Plausibility & \textcolor{gray}{\small[ICLR'24]} & \textcolor{gray}{\small[arXiv'24]} & \textcolor{gray}{\small[AAAI'25]} & \textcolor{gray}{\small[CVPR'25]} & \textcolor{gray}{\small[ICCV'25]} & \textcolor{gray}{\small[NeurIPS'25]} & \textcolor{gray}{\textbf{Max}} 
        \\
        \midrule
        \textsl{min} 
            & $2.0$ & $2.0$ & $2.0$ & $2.0$ & $2.0$ & $2.0$
            & \cellcolor{gray!7}\textcolor{gray}{-} \\
        \textsl{max} 
            & $4.0$ & $4.0$ & $8.0$ & $8.0$ & $10.0$ & $4.0$
            & \cellcolor{gray!7}\textcolor{gray}{-} \\
        \textbf{\textsl{mean}}
            & \cellcolor{w_blue!20}$2.300$
            & \cellcolor{w_blue!20}$2.300$
            & \cellcolor{w_blue!20}$2.380$
            & \cellcolor{w_blue!20}$2.312$
            & \cellcolor{w_blue!20}$2.583$
            & \cellcolor{w_blue!20}$2.292$
            & \cellcolor{gray!7}\textcolor{gray}{$10$} \\
        \textsl{std} 
            & $0.640$ & $0.640$ & $0.783$ & $0.674$ & $1.187$ & $0.626$
            & \cellcolor{gray!7}\textcolor{gray}{-} \\
        \textsl{median} 
            & $2.0$ & $2.0$ & $2.0$ & $2.0$ & $2.0$ & $2.0$
            & \cellcolor{gray!7}\textcolor{gray}{-} \\
        \textsl{q25} 
            & $2.0$ & $2.0$ & $2.0$ & $2.0$ & $2.0$ & $2.0$
            & \cellcolor{gray!7}\textcolor{gray}{-} \\
        \textsl{q75} 
            & $2.0$ & $2.0$ & $2.0$ & $2.0$ & $3.0$ & $2.0$
            & \cellcolor{gray!7}\textcolor{gray}{-} \\
        \bottomrule
    \end{tabular}}
    \vspace{-0.4cm}
\end{table*}

%% file: sections_supp/dimensions/human_3d_4d.tex
\subsection{~3D \& 4D Consistency}

Physical Plausibility measures how well the 3D structure and temporal evolution of objects in a generated video align with those in the corresponding real (ground-truth) sequence.  
Rather than judging raw pixels, this dimension focuses on the stability and accuracy of 3D bounding boxes over time, as estimated by a pretrained tracking or detection model applied to both generated and real videos. Annotators are instructed to judge each clip according to the following criteria:

\begin{itemize}
    \item For each object, the 3D box size, orientation, and position should evolve smoothly over time, without jitter, sudden jumps, unnatural scaling, or misalignment with the underlying object.
    \vspace{-0.1cm}
    \item Tracks should persist as long as the object is visible, without frequent flickering, disappearing-and-reappearing, or drifting away from the target.
    \vspace{-0.1cm}
    \item The number and spatial arrangement of boxes in the generated view should broadly match those in the ground-truth view, especially for prominent nearby objects.
        \vspace{-0.1cm}
    \item In dynamic scenes, the motion direction and speed of boxes in the generated view should be close to those in the ground-truth view; in static scenes, boxes should remain essentially still.
\end{itemize}
Higher scores indicate that 3D box trajectories in generated videos closely track those in real scenes.

\subsubsection{~Protocol}
Each generated video is rated on a $1$–$10$ scale of perceived realism:

\begin{table}[h]
    \centering
    \label{tab:human_3d_4d}
    \resizebox{\linewidth}{!}{
    \begin{tabular}{|c|c|p{350pt}<{\raggedright}|}
        \hline
        \textbf{Score} & \textbf{Level} & \textbf{Description}
        \\
        \hline\hline
        \multirow{2}{*}{$1$} & \multirow{2}{*}{Extremely Inconsistent} & 3D boxes unstable or mismatched; severe jitter, missing frames, or large 
        \\
        & & misalignment from the ground truth.
        \\
        \hline
        \multirow{2}{*}{$3$} & \multirow{2}{*}{Inconsistent} & Rough trajectory trend visible but with clear instability or mismatched counts; 
        \\
        & & large misalignment from the ground truth.
        \\
        \hline
        \multirow{2}{*}{$5$} & \multirow{2}{*}{Moderately Consistent} & 3D boxes mostly aligned with ground truth; however, there are still minor jitter 
        \\
        & & or missing boxes that can be detected by human eyes.
        \\
        \hline
        \multirow{2}{*}{$7$} & \multirow{2}{*}{Consistent} & Smooth, stable trajectories and coherent spatial alignment; only slight temporal 
        \\
        & & noise; the number of detected 3D boxes mostly aligns with the ground truth.
        \\
        \hline
        \multirow{2}{*}{$9$} & \multirow{2}{*}{Highly Consistent} & Generated 3D boxes match ground-truth positions and motions almost perfectly 
        \\
        & & across time; the inconsistency can hardly be detected by human eyes.
        \\
        \hline
        \multirow{2}{*}{$10$} & \multirow{2}{*}{Ground Truth} & \multirow{2}{*}{-}
        \\
        & & 
        \\
        \hline
    \end{tabular}}
\end{table}

\vspace{-0.3cm}
\subsubsection{~Examples}
\begin{figure}[t]
    \centering
    \begin{subfigure}[h]{\textwidth}
        \centering
        \includegraphics[width=\linewidth]{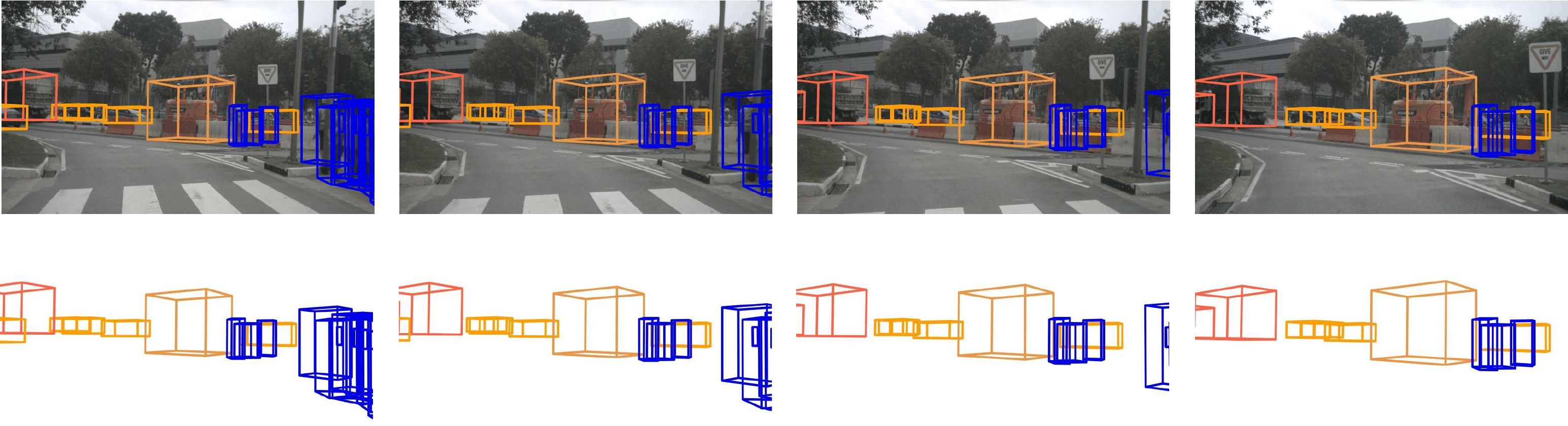}
        \caption{Good example in the \emph{3D \& 4D Consistency} dimension (Score: \textcolor{w_blue}{$10$})}
        \label{fig:human_3d_4d_1}
    \end{subfigure}
    \\[2ex]
    \begin{subfigure}[h]{\textwidth}
        \centering
        \includegraphics[width=\linewidth]{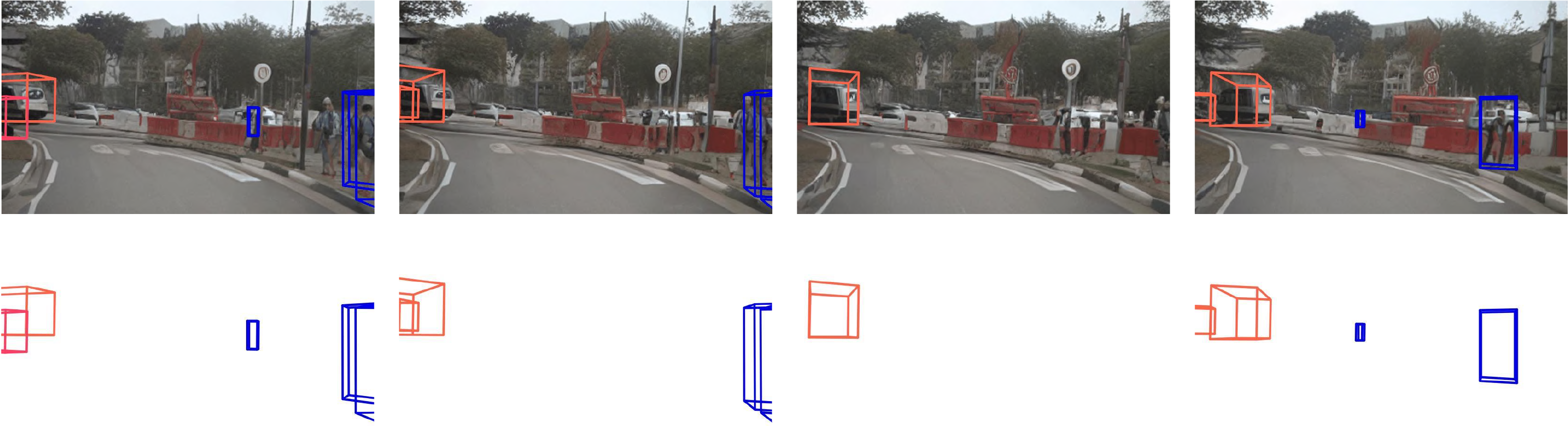}
        \caption{Bad example in the \emph{3D \& 4D Consistency} dimension (Score: \textcolor{red}{$1$})}
        \label{fig:human_3d_4d_2}
    \end{subfigure}
    \\[2ex]
    \begin{subfigure}[h]{\textwidth}
        \centering
        \includegraphics[width=\linewidth]{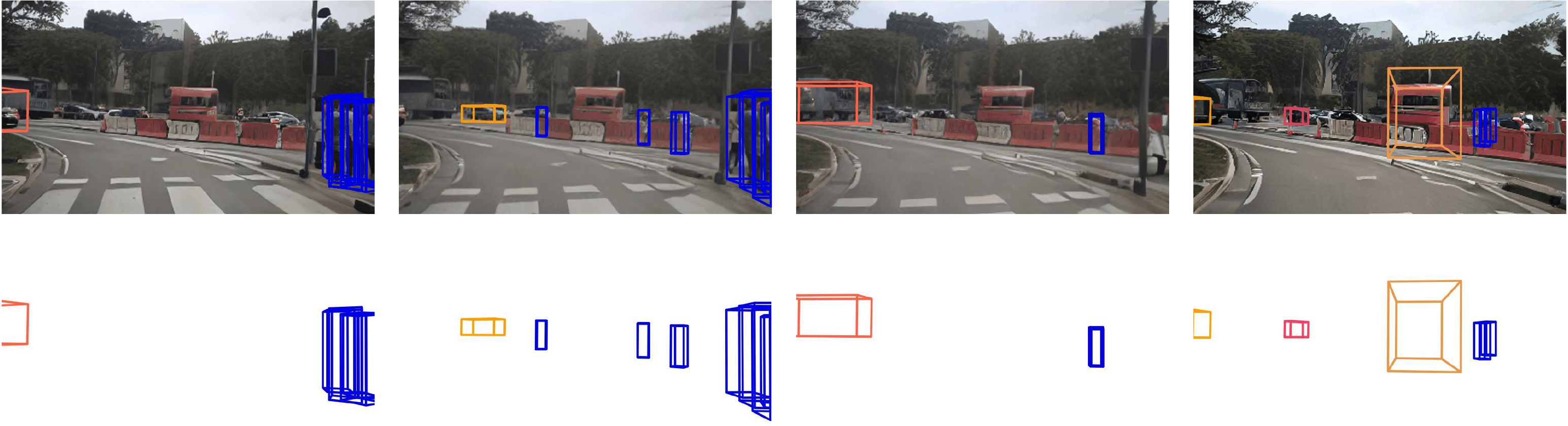}
        \caption{Bad example in the \emph{3D \& 4D Consistency} dimension (Score: \textcolor{red}{$1$})}
        \label{fig:human_3d_4d_3}
    \end{subfigure}
    \\[2ex]
    \begin{subfigure}[h]{\textwidth}
        \centering
        \includegraphics[width=\linewidth]{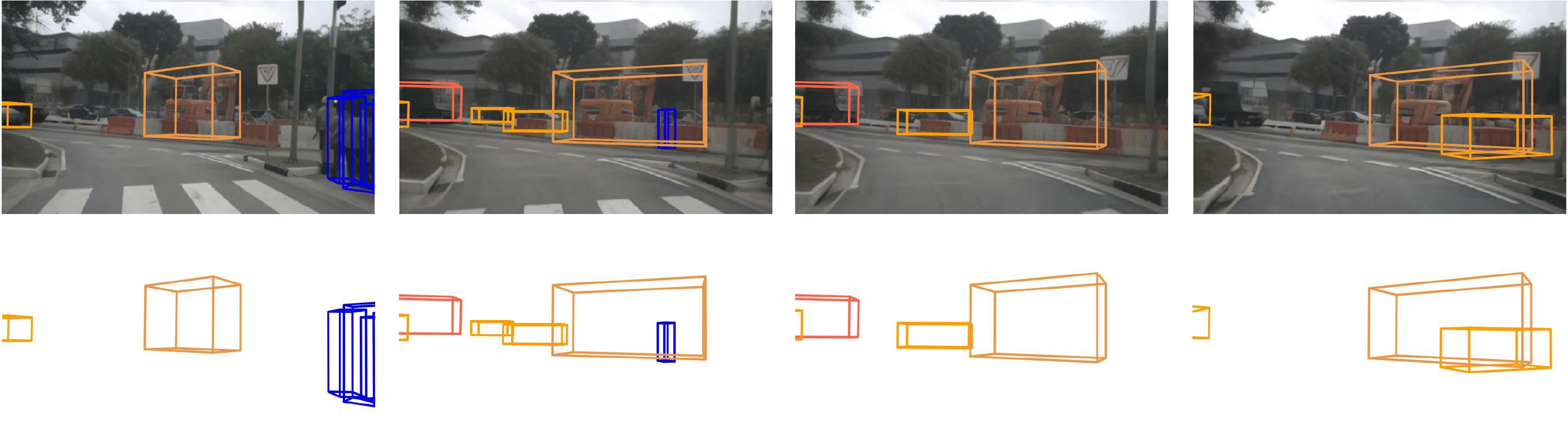}
        \caption{Bad example in the \emph{3D \& 4D Consistency} dimension (Score: \textcolor{red}{$1$})}
        \label{fig:human_3d_4d_4}
    \end{subfigure}
    \\
    \vspace{-0.05cm}
    \caption{Examples of ``good'' and ``bad'' human preference alignments in terms of \emph{3D \& 4D Consistency} in WorldLens.}
    \label{fig:human_3d_4d}
\end{figure}

Figure~\ref{fig:human_3d_4d} provides typical examples of videos with good and bad quality in terms of \emph{3D \& 4D Consistency}.

\subsubsection{~Evaluation \& Analysis}
Table~\ref{tab:supp_human_3d_4d} provides the complete results of models in terms of \emph{3D \& 4D Consistency}.

\begin{table*}[h]
    \centering
    \vspace{0.2cm}
    \caption{Complete comparisons of state-of-the-art driving world models in terms of \emph{3D \& 4D Consistency} in WorldLens.}
    \vspace{-0.2cm}
    \label{tab:supp_human_3d_4d}
    \resizebox{\linewidth}{!}{
    \begin{tabular}{r|cccccc|c}
        \toprule
        3D \& 4D & \textbf{MagicDrive} & \textbf{DreamForge}  & \textbf{DriveDreamer-2} & \textbf{OpenDWM} & \textbf{~DiST-4D~} & $\mathcal{X}$\textbf{-Scene} & \textcolor{gray}{\textbf{Empirical}}
        \\
        Consistency & \textcolor{gray}{\small[ICLR'24]} & \textcolor{gray}{\small[arXiv'24]} & \textcolor{gray}{\small[AAAI'25]} & \textcolor{gray}{\small[CVPR'25]} & \textcolor{gray}{\small[ICCV'25]} & \textcolor{gray}{\small[NeurIPS'25]} & \textcolor{gray}{\textbf{Max}}
        \\\midrule
        \textsl{min} 
            & $2.0$ & $2.0$ & $2.0$ & $2.0$ & $2.0$ & $2.0$
            & \cellcolor{gray!7}\textcolor{gray}{-} \\
        \textsl{max} 
            & $10.0$ & $10.0$ & $10.0$ & $10.0$ & $10.0$ & $10.0$
            & \cellcolor{gray!7}\textcolor{gray}{-} \\
        \textbf{\textsl{mean}}
            & \cellcolor{w_blue!20}$2.455$
            & \cellcolor{w_blue!20}$2.743$
            & \cellcolor{w_blue!20}$2.751$
            & \cellcolor{w_blue!20}$2.920$
            & \cellcolor{w_blue!20}$2.961$
            & \cellcolor{w_blue!20}$2.431$
            & \cellcolor{gray!7}\textcolor{gray}{$10$} \\
        \textsl{std} 
            & $1.061$ & $1.378$ & $1.530$ & $1.467$ & $1.405$ & $1.161$
            & \cellcolor{gray!7}\textcolor{gray}{-} \\
        \textsl{median} 
            & $2.0$ & $2.0$ & $2.0$ & $2.0$ & $2.0$ & $2.0$
            & \cellcolor{gray!7}\textcolor{gray}{-} \\
        \textsl{q25} 
            & $2.0$ & $2.0$ & $2.0$ & $2.0$ & $2.0$ & $2.0$
            & \cellcolor{gray!7}\textcolor{gray}{-} \\
        \textsl{q75} 
            & $2.0$ & $4.0$ & $3.0$ & $4.0$ & $4.0$ & $2.0$
            & \cellcolor{gray!7}\textcolor{gray}{-} \\
        \bottomrule
    \end{tabular}}
    \vspace{-0.4cm}
\end{table*}

%% file: sections_supp/dimensions/human_behavioral_safety.tex
\subsection{~Behavioral Safety}

Behavioral Safety measures how safe and predictable the visible behavior of traffic participants appears in generated driving videos, as judged by human observers. Rather than evaluating visual realism alone, this dimension focuses on whether vehicles, pedestrians, cyclists, and other agents interact with each other and with key scene elements in a way that is consistent with basic traffic rules and low-risk driving. Annotators are instructed to judge each clip according to the following criteria:
\begin{itemize}
    \item Obvious impossible behaviors, \emph{e.g.}, sudden teleportation, splitting or merging of agents, agents appearing or disappearing without cause, or severe shape deformation that destroys basic spatial relations.
    \vspace{-0.1cm}
    \item Whether agent behavior clearly contradicts prominent traffic signals, signs, or lane markings (for example, ignoring a red light, driving against traffic, or violating stop or yield indications).
    \vspace{-0.1cm}
    \item Whether vehicles and road users maintain reasonable gaps, avoid implausible near-collisions or illegal crossings, and follow trajectories that are smooth and predictable rather than erratic or conflict-prone.
    \vspace{-0.1cm}
    \item Whether scene distortions, flickering, or object deformations directly impair the ability to read safety-critical cues, such as lane boundaries, signal states, and relative positions of agents.
\end{itemize}

Higher \emph{Behavioral Safety} indicates that generated videos tend to display traffic behavior that raters judge as safe and consistent with basic road rules.

\subsubsection{Protocol}
Each generated video is rated on a $1$–$10$ scale of perceived realism:

\begin{table}[h]
    \centering
    \label{tab:human_behavioral_safety}
    \resizebox{\linewidth}{!}{
    \begin{tabular}{|c|c|p{350pt}<{\raggedright}|}
        \hline
        \textbf{Score} & \textbf{Level} & \textbf{Description}
        \\
        \hline\hline
        \multirow{2}{*}{$1$} & \multirow{2}{*}{Extremely Unsafe} & Catastrophic anomalies or impossible events (teleportation, splitting, collisions, 
        \\
        & & severe signal violations) that make the scene unsafe at a glance.
        \\
        \hline
        \multirow{2}{*}{$3$} & \multirow{2}{*}{Mostly Unsafe} & Partial or localized safety violations, \emph{e.g.}, brief teleportation, collisions, or 
        \\
        & & unreadable signal states, that clearly degrade perceived safety.
        \\
        \hline
        \multirow{2}{*}{$5$} & \multirow{2}{*}{Moderately Safe} & Generally safe behavior with mild instability or motion artifacts; no major 
        \\
        & & conflicts,  though realism is limited.
        \\
        \hline
        \multirow{2}{*}{$7$} & \multirow{2}{*}{Safe} & Predictable and stable behavior; vehicles and pedestrians maintain proper 
        \\
        & & spacing and respect traffic logic.
        \\
        \hline
        \multirow{2}{*}{$9$} & \multirow{2}{*}{Highly Safe} & Fully natural and rule-abiding behavior; smooth trajectories, compliant with 
        \\
        & & signals, and entirely risk-free appearance.
        \\
        \hline
        \multirow{2}{*}{$10$} & \multirow{2}{*}{Ground Truth} & \multirow{2}{*}{-}
        \\
        & & 
        \\
        \hline
    \end{tabular}}
\end{table}

\vspace{-0.3cm}
\subsubsection{~Examples}
\begin{figure}[t]
    \centering
    \begin{subfigure}[h]{\textwidth}
        \centering
        \includegraphics[width=\linewidth]{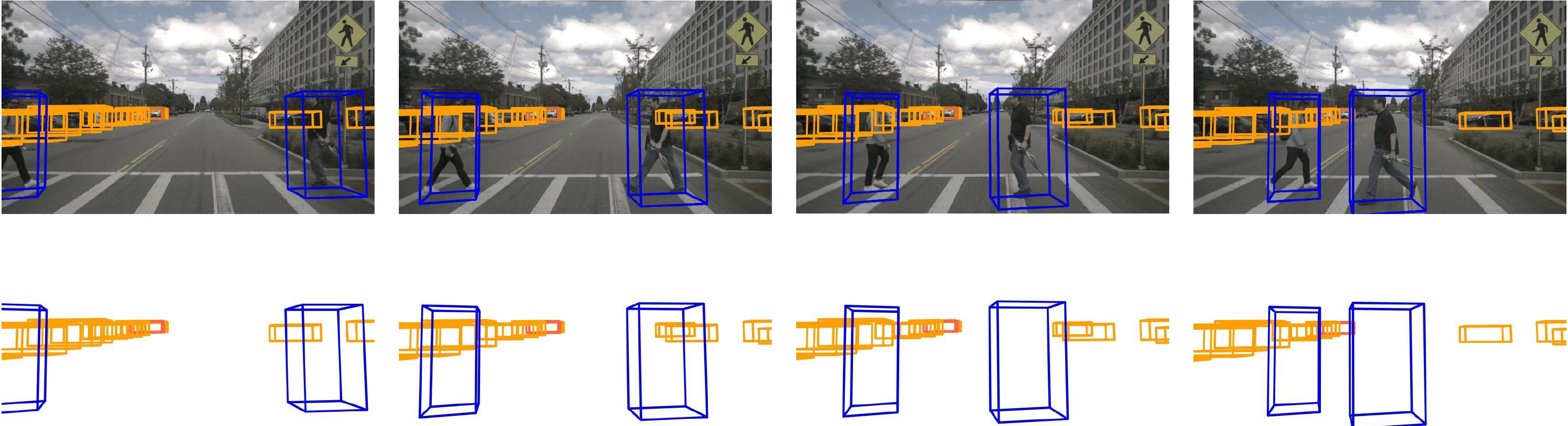}
        \caption{Good example in the \emph{Behavioral Safety} dimension (Score: \textcolor{w_blue}{$10$})}
        \label{fig:human_behavioral_safety_1}
    \end{subfigure}
    \\[2ex]
    \begin{subfigure}[h]{\textwidth}
        \centering
        \includegraphics[width=\linewidth]{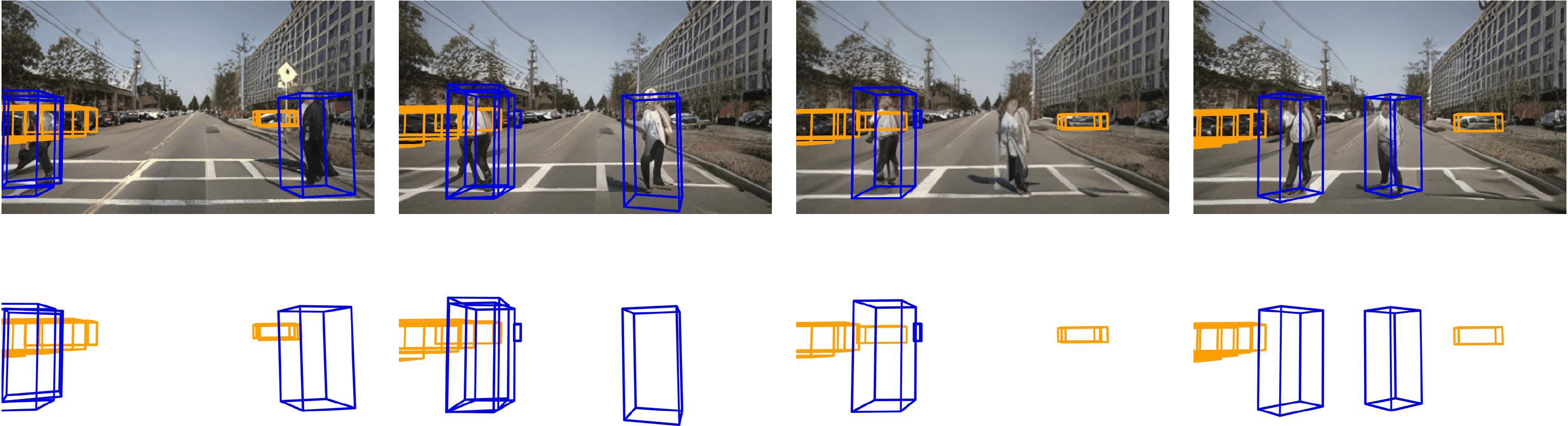}
        \caption{Bad example in the \emph{Behavioral Safety} dimension (Score: \textcolor{red}{$1$})}
        \label{fig:human_behavioral_safety_2}
    \end{subfigure}
    \\[2ex]
    \begin{subfigure}[h]{\textwidth}
        \centering
        \includegraphics[width=\linewidth]{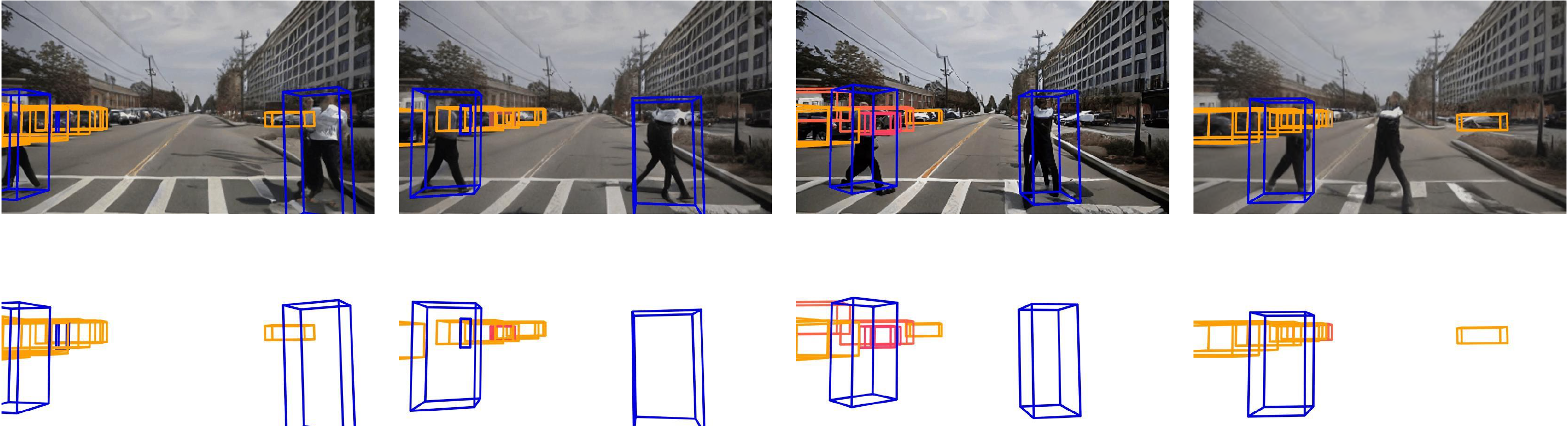}
        \caption{Bad example in the \emph{Behavioral Safety} dimension (Score: \textcolor{red}{$1$})}
        \label{fig:human_behavioral_safety_3}
    \end{subfigure}
    \\[2ex]
    \begin{subfigure}[h]{\textwidth}
        \centering
        \includegraphics[width=\linewidth]{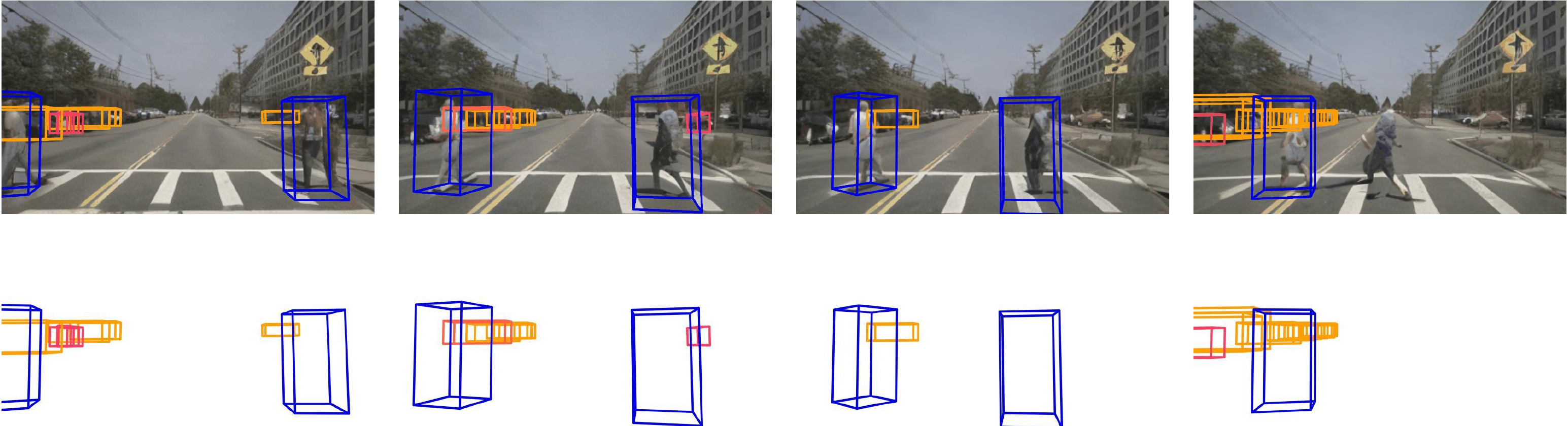}
        \caption{Bad example in the \emph{Behavioral Safety} dimension (Score: \textcolor{red}{$1$})}
        \label{fig:human_behavioral_safety_4}
    \end{subfigure}
    \\
    \vspace{-0.05cm}
    \caption{Examples of ``good'' and ``bad'' human preference alignments in terms of \emph{Behavioral Safety} in WorldLens.}
    \label{fig:human_behavioral_safety}
\end{figure}

Figure~\ref{fig:human_behavioral_safety} provides typical examples of videos with good and bad quality in terms of \emph{Behavioral Safety}.

\subsubsection{~Evaluation \& Analysis}
Table~\ref{tab:supp_human_behavioral_safety} provides the complete results of models in terms of \emph{Behavioral Safety}.

\begin{table*}[h]
    \centering
    \vspace{0.2cm}
    \caption{Complete comparisons of state-of-the-art driving world models in terms of \emph{Behavioral Safety} in WorldLens.}
    \vspace{-0.2cm}
    \label{tab:supp_human_behavioral_safety}
    \resizebox{\linewidth}{!}{
    \begin{tabular}{r|cccccc|c}
        \toprule
        Behavioral& \textbf{MagicDrive} & \textbf{DreamForge}  & \textbf{DriveDreamer-2} & \textbf{OpenDWM} & \textbf{~DiST-4D~} & $\mathcal{X}$\textbf{-Scene} & \textcolor{gray}{\textbf{Empirical}}
        \\
        Safety & \textcolor{gray}{\small[ICLR'24]} & \textcolor{gray}{\small[arXiv'24]} & \textcolor{gray}{\small[AAAI'25]} & \textcolor{gray}{\small[CVPR'25]} & \textcolor{gray}{\small[ICCV'25]} & \textcolor{gray}{\small[NeurIPS'25]} & \textcolor{gray}{\textbf{Max}}
        \\\midrule
        \textsl{min} 
            & $2.0$ & $2.0$ & $2.0$ & $2.0$ & $2.0$ & $2.0$
            & \cellcolor{gray!7}\textcolor{gray}{-} \\
        \textsl{max} 
            & $4.0$ & $4.0$ & $10.0$ & $8.0$ & $6.0$ & $4.0$
            & \cellcolor{gray!7}\textcolor{gray}{-} \\
        \textbf{\textsl{mean}}
            & \cellcolor{w_blue!20}$2.306$
            & \cellcolor{w_blue!20}$2.290$
            & \cellcolor{w_blue!20}$2.533$
            & \cellcolor{w_blue!20}$2.598$
            & \cellcolor{w_blue!20}$2.591$
            & \cellcolor{w_blue!20}$2.318$
            & \cellcolor{gray!7}\textcolor{gray}{$10$} \\
        \textsl{std} 
            & $0.649$ & $0.621$ & $1.184$ & $1.247$ & $1.341$ & $0.686$
            & \cellcolor{gray!7}\textcolor{gray}{-} \\
        \textsl{median} 
            & $2.0$ & $2.0$ & $2.0$ & $2.0$ & $2.0$ & $2.0$
            & \cellcolor{gray!7}\textcolor{gray}{-} \\
        \textsl{q25} 
            & $2.0$ & $2.0$ & $2.0$ & $2.0$ & $2.0$ & $2.0$
            & \cellcolor{gray!7}\textcolor{gray}{-} \\
        \textsl{q75} 
            & $2.0$ & $2.0$ & $2.0$ & $3.0$ & $2.0$ & $2.0$
            & \cellcolor{gray!7}\textcolor{gray}{-} \\
        \bottomrule
    \end{tabular}}
    \vspace{-0.4cm}
\end{table*}

%% file: sections_supp/7_agent.tex
\section{~Evaluation Agent}

In this section, we present additional detail on the proposed \textbf{WorldLens-Agent} model, describing the architecture, prompting scheme, training setup, and providing some qualitative evaluation examples on out-of-distribution driving videos. We observe that evaluating generated worlds often hinges on human-centered criteria (physical plausibility) and subjective preferences (perceived realism) that quantitative metrics inherently miss. Our goal here is to train an auto-evaluation agent that can be utilized in a broader range of generated videos, and, simultaneously, align with the preferences of human annotators.

\subsection{~Agent Architecture}
The \textbf{WorldLens-Agent} is a vision-language critic built on Qwen3-VL-8B \cite{Qwen2.5-VL} and trained to evaluate generated videos along human-centered dimensions, including overall realism, 3D consistency, physical plausibility, and behavioral safety.

\begin{figure}[t]
    \centering
    \includegraphics[width=\linewidth]{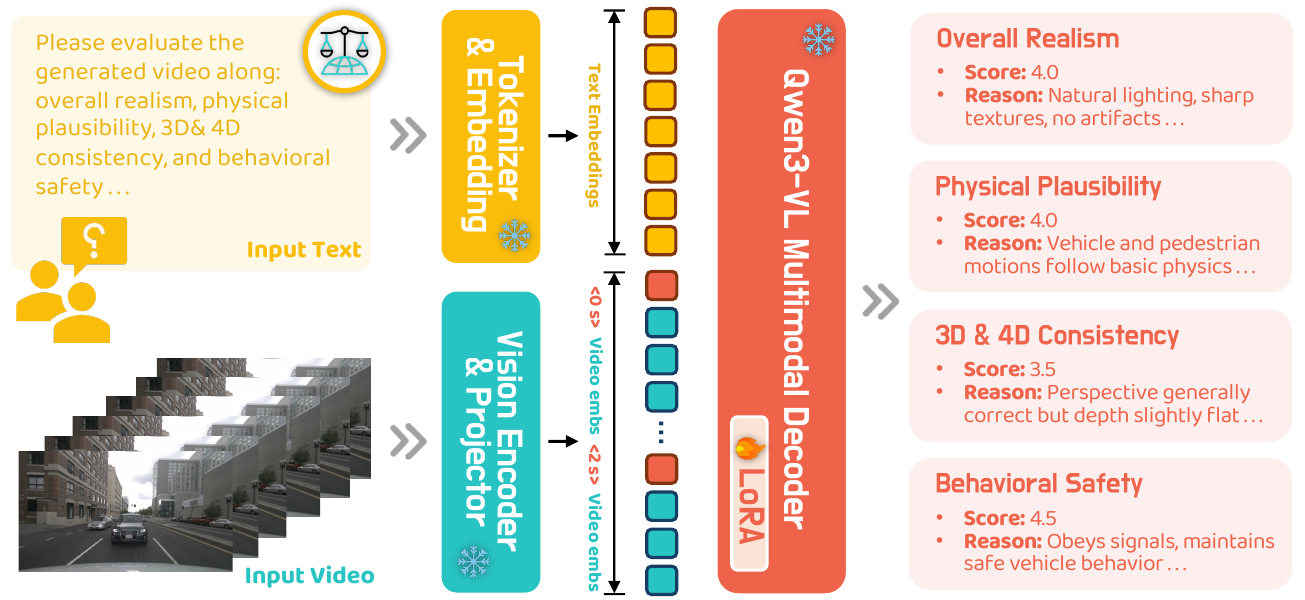}
    \vspace{-0.2cm}
    \caption{The architecture of the proposed \textbf{WorldLens-Agent} for auto-evaluation of generated driving videos.}
    \label{fig:supp_agent_pipeline}
\end{figure}

As shown in Figure~\ref{fig:supp_agent_pipeline}, the agent takes two types of input: an instruction text describing the evaluation criteria, which is processed by the frozen Qwen3 tokenizer, and a video generated by world models, which is encoded by the frozen Qwen3-VL vision encoder. The resulting features are projected into the language token space, forming a unified multimodal token sequence.

This sequence is then passed to the Qwen3-VL decoder, where LoRA adapters are applied only to the attention layers. All other components, including the vision encoder, the projector, the embedding layers, and the MLP blocks, remain frozen. This lightweight adaptation allows the model to incorporate human perceptual and safety-related priors learned from \textbf{WorldLens-26K}, enabling it to capture cues such as lighting realism, depth stability, object dynamics, and safety-critical violations while preserving the general multimodal capability of the base model.

Finally, the agent autoregressively produces a \emph{structured JSON output} that contains a numerical score ($1$-$10$) and a concise rationale for each evaluation dimension. This representation yields a reliable, interpretable, and scalable assessment signal that complements conventional quantitative metrics and serves as a consistent preference oracle for world-model benchmarking and downstream reinforcement learning pipelines.

\subsection{~Prompt Scheme}
The following prompting scheme specifies the instruction protocol for the WorldLens Evaluation Agent. Given a generated driving clip and a dimension-specific human rating rubric, the agent is guided to produce structured, evidence-based scores for multiple aspects of generative video quality.

The prompt enforces strict output formatting, dimension-aware reasoning, and rubric-consistent interpretation, ensuring reliable and reproducible automatic scoring.

\subsection{~Training Setup}
The WorldLens-Agent is fine-tuned from Qwen3-VL-8B through supervised instruction tuning, allowing the model to better align with human evaluation preferences. We adopt LoRA adaptation on all attention modules, using a rank of $16$ and a dropout rate of $0.05$, which provides efficient preference learning while preserving the multimodal reasoning capabilities of the base model.

Training is performed for three epochs with a learning rate of $1$e-$4$, cosine decay scheduling, and a warmup ratio of $0.1$. All experiments are conducted on eight A100 GPUs using \texttt{bfloat16} precision. This configuration ensures stable convergence and effective adaptation, resulting in a vision-language critic that consistently captures realism, geometric consistency, physical plausibility, and safety-related cues in generated videos.

\subsection{~Qualitative Assessment}
\begin{figure}
    \centering
    \includegraphics[width=\linewidth]{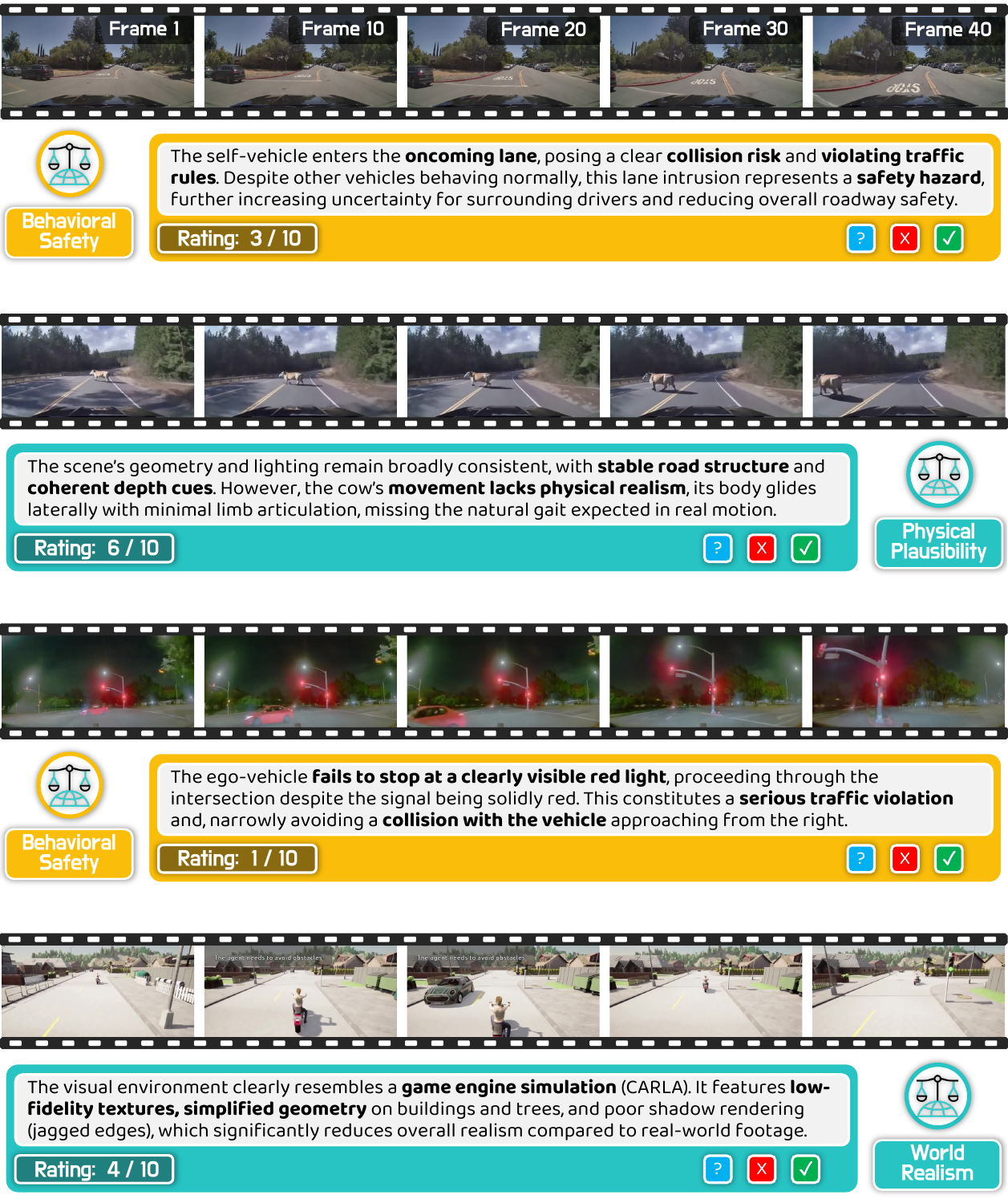}
    \vspace{-0.6cm}
    \caption{Additional qualitative assessments of the \textbf{WorldLens-Agent} evaluation on challenging driving conditions.}
    \label{fig:supp_agent_1}
\end{figure}

\begin{figure}
    \centering
    \includegraphics[width=\linewidth]{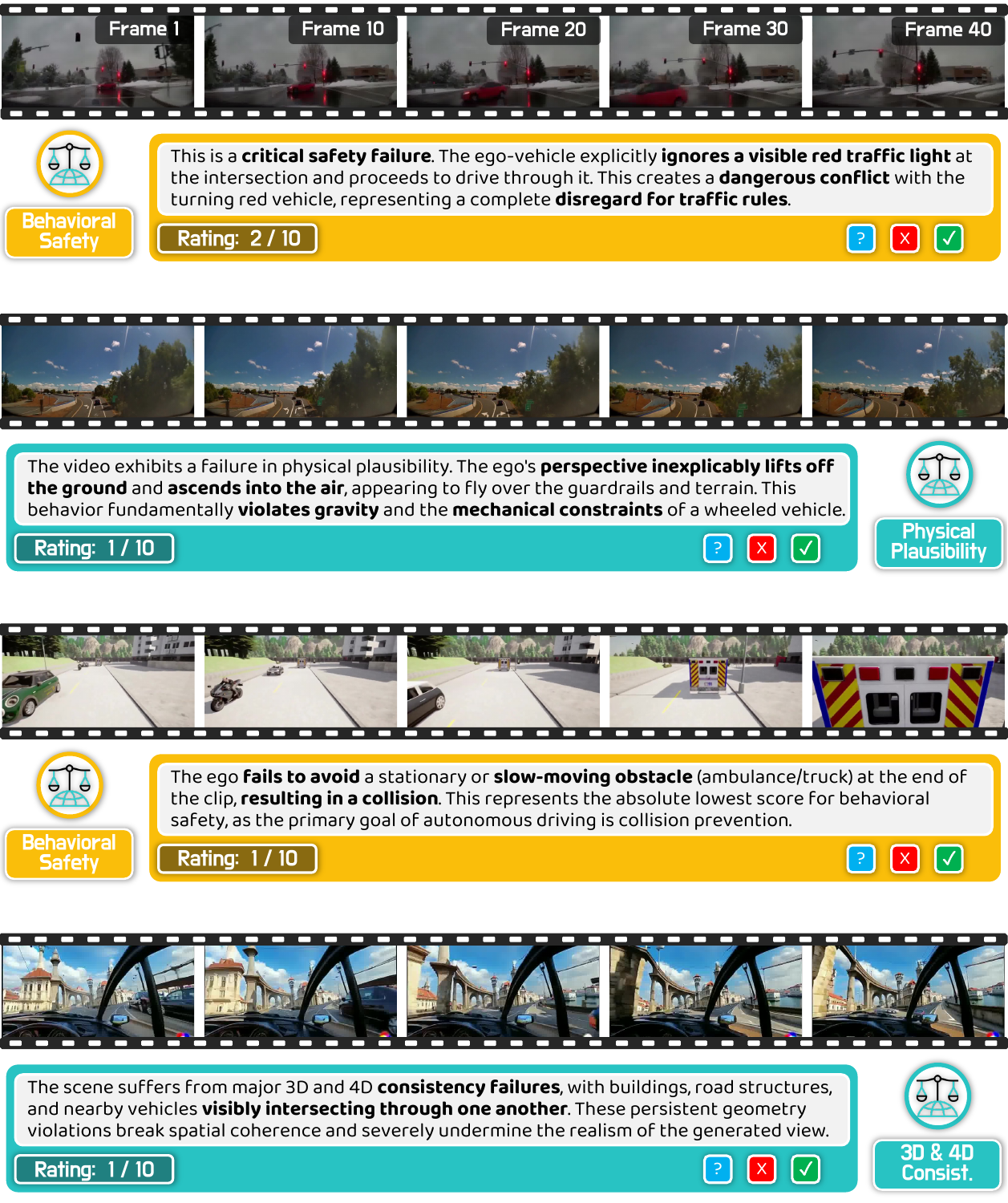}
    \vspace{-0.6cm}
    \caption{Additional qualitative assessments of the \textbf{WorldLens-Agent} evaluation on challenging driving conditions.}
    \label{fig:supp_agent_2}
\end{figure}

Figure~\ref{fig:supp_agent_1} and Figure~\ref{fig:supp_agent_2} present additional qualitative evaluations produced by the \textbf{WorldLens-Agent} on challenging driving scenarios, including \textbf{out-of-distribution} videos rendered or produced by Gen3C~\cite{ren2025gen3c}, Cosmos-Drive~\cite{ren2025cosmos}, and the CARLA~\cite{dosovitskiy2017carla} simulator. These examples illustrate the agent’s ability to generalize beyond its training distribution and maintain consistent, human-aligned judgment across a wide spectrum of visual styles, scene structures, and motion dynamics.

As shown in Figure~\ref{fig:supp_agent_1}, the agent reliably identifies a broad range of safety-critical issues. These include \emph{lane incursions}, \emph{ignoring red lights}, and \emph{near-collision events}, each accompanied by a concise explanation grounded in visible evidence. It also detects failures in physical plausibility, such as \emph{unnatural animal motion} or vehicles exhibiting \emph{incoherent dynamics}, where motion lacks realistic articulation or violates expected mass-gravity relationships. In the realm of realism, the agent highlights artifacts like \emph{low-fidelity textures}, \emph{simplified geometry}, and \emph{game-engine rendering effects}, all of which degrade perceptual authenticity.

Figure~\ref{fig:supp_agent_2} further demonstrates the agent’s sensitivity to more severe and uncommon failure modes. It flags \emph{physically impossible} ego-vehicle trajectories, such as the viewpoint unexpectedly lifting off the ground, as violations of basic mechanical and gravitational constraints. The agent also captures high-impact behavioral safety failures, including \emph{colliding with stationary obstacles} such as ambulances or trucks. Beyond temporal or behavioral issues, it robustly identifies large-scale \emph{3D and 4D consistency violations}, where buildings, vehicles, and road structures visibly intersect or pass through one another, indicating broken geometry and disrupted spatial coherence.

Together, these qualitative cases highlight the strong generalization capability of the proposed \textbf{WorldLens-Agent} and our ability to diagnose diverse, complex failure patterns across unseen generative video domains. The agent not only assigns scalar scores but also provides clear, interpretable rationales, enabling transparent and human-aligned evaluation under significant distribution shift.

\clearpage\clearpage
\begin{tcolorbox}[myboxstyle]

You are an expert \textbf{vision-language evaluator} designed to assess the quality of \textbf{AI-generated driving videos}.  
Given a short video clip together with a human-designed rubric, your task is to assign an objective score and provide a concise, evidence-based explanation.

Your analysis focuses on a \textbf{single target dimension} at a time (\emph{e.g.}, \textit{overall\_realism}, \textit{vehicle\_realism}, \textit{pedestrian\_realism}, \textit{3D\_consistency}, \textit{physical\_plausibility}, \textit{behavioral\_safety}).

\vspace{2ex}
\begin{itemize}[leftmargin=10pt]
  \item {\textbf{\textcolor{w_blue}{Evaluation Target}}: The input specifies which dimension to evaluate, \emph{e.g.}, \texttt{overall\_realism}, \texttt{vehicle\_realism}, etc.}
  \item \textbf{\textcolor{w_blue}{Rubric-Guided Judgment}}: Use the full English rubric provided for the selected dimension, strictly following its definitions, criteria, and scoring scale.
  \item \textbf{\textcolor{w_blue}{Evidence-Based Scoring}}: Judge only what is \textbf{visually observable} in the video, such as 
        \textcolor{teal}{temporal stability}, \textcolor{teal}{geometry}, \textcolor{teal}{textures}, 
        \textcolor{teal}{reflections}, \textcolor{teal}{occlusions}, \textcolor{teal}{physical consistency}, and \textcolor{teal}{artifacts}.
  \item \textbf{\textcolor{w_blue}{Score Range}}: Output a numeric score in $[1, 10]$ with a step of $0.5$ (i.e., $\{1.0, 1.5, \dots, 10.0\}$), rounded to the nearest $0.5$ and clamped to $[1, 10]$.
  \item \textbf{\textcolor{w_blue}{Rationale}}: Provide a short English rationale that cites concrete visual evidence, \emph{e.g.}, 
        \textcolor{teal}{flicker}, \textcolor{teal}{ghosting}, \textcolor{teal}{shape distortions}, \textcolor{teal}{interpenetration}, 
        \textcolor{teal}{lighting jumps}, \textcolor{teal}{unsafe maneuvers}, etc.
\end{itemize}

\vspace{1ex}
\noindent\textbf{\textcolor{purple}{Scoring Rubrics for Each Dimension (Summary):}}
\begin{itemize}[leftmargin=10pt]
\item \textbf{\textcolor{w_blue}{Overall Realism}}  
\begin{itemize}[leftmargin=12pt]
  \item \textbf{1 -- Highly Unrealistic:} Severe artifacts (warping, collapsing geometry, heavy flicker, broken roads, impossible lighting).
  \item \textbf{3 -- Unrealistic:} Structures roughly recognizable but textures blurry; perspective errors; frequent artifacts and instability.
  \item \textbf{5 -- Fair:} Mostly acceptable but clearly synthetic; noticeable unnatural boundaries, lighting jumps, or texture flicker.
  \item \textbf{7 -- Realistic:} Largely natural appearance; coherent lighting and geometry; only minor localized flaws.
  \item \textbf{9 -- Highly Realistic:} Almost indistinguishable from real dashcam footage; stable textures, correct perspective, consistent lighting.
\end{itemize}

\item \textbf{\textcolor{w_blue}{Vehicle Realism}}  
\begin{itemize}[leftmargin=12pt]
  \item \textbf{1 -- Highly Unrealistic:} Strong distortions, split bodies, stretched parts, heavy flicker; vehicles look clearly fake.
  \item \textbf{3 -- Unrealistic:} Coarse material appearance, unstable reflections, poor temporal consistency, visible geometry defects.
  \item \textbf{5 -- Fair:} Car-like but with noticeable flaws in edges, contours, or materials; moderate instability across frames.
  \item \textbf{7 -- Realistic:} Mostly correct geometry, paint, glass, tires, and reflections; minor issues only.
  \item \textbf{9 -- Highly Realistic:} Faithful car shape and materials; natural glass/paint reflections; stable and coherent over time.
\end{itemize}
    
\item \textbf{\textcolor{w_blue}{Pedestrian Realism}}  
\begin{itemize}[leftmargin=12pt]
  \item \textbf{1 -- Highly Unrealistic:} Broken limbs, twisted joints, ghosting, severe flicker, or missing body parts.
  \item \textbf{3 -- Unrealistic:} Human-like but coarse textures, inconsistent appearance, unnatural gait, or floating/sliding.
  \item \textbf{5 -- Fair:} Generally human-shaped with some artifacts or slightly rigid/unnatural motion.
  \item \textbf{7 -- Realistic:} Mostly natural appearance and motion; minor local inconsistencies acceptable.
  \item \textbf{9 -- Highly Realistic:} Convincing human geometry, clothing, motion, and lighting; temporally stable with no major artifacts.
\end{itemize}
    
\item \textbf{\textcolor{w_blue}{3D Consistency}}  
\begin{itemize}[leftmargin=12pt]
  \item \textbf{1 -- Highly Inconsistent:} Strong jitter, scale popping, drift, disappear/reappear, incorrect depth ordering.
  \item \textbf{3 -- Inconsistent:} Overall motion roughly matches but unstable; noticeable mismatches or frequent small jumps.
  \item \textbf{5 -- Fair:} Broadly consistent with occasional jitters, mild drift, or minor alignment errors.
  \item \textbf{7 -- Consistent:} Mostly stable geometry and depth; only small irregularities.
  \item \textbf{9 -- Highly Consistent:} Smooth temporal evolution; stable shapes, positions, and trajectories with near-perfect depth coherence.
\end{itemize}
    
\item \textbf{\textcolor{w_blue}{Physical Plausibility}}  
\begin{itemize}[leftmargin=12pt]
  \item \textbf{1 -- Highly Implausible:} Teleporting, merging/splitting objects, interpenetration, impossible shadows, large occlusion errors.
  \item \textbf{3 -- Implausible:} Noticeable but localized physics violations (\emph{e.g.}, single intersection event, abrupt illumination jump).
  \item \textbf{5 -- Fair:} Mostly plausible but with several visible inconsistencies that do not dominate the clip.
  \item \textbf{7 -- Plausible:} Stable motion, contacts, and occlusions; only minor physical deviations.
  \item \textbf{9 -- Highly Plausible:} Fully consistent with real-world physics; smooth motion, proper occlusions, no interpenetration.
\end{itemize}
    
\item \textbf{\textcolor{w_blue}{Behavioral Safety}}  
\begin{itemize}[leftmargin=12pt]
  \item \textbf{1 -- Highly Unsafe:} Impossible or dangerous behaviors; collisions, severe occlusion failures preventing safety judgment.
  \item \textbf{3 -- Unsafe:} Localized unsafe interactions, obvious violations of signals or right-of-way; unstable visuals harming safety perception.
  \item \textbf{5 -- Generally Safe:} Mostly reasonable behaviors with small issues that do not fundamentally affect safety.
  \item \textbf{7 -- Safe:} Stable, predictable motion; compliant with visible cues; minor cosmetic defects only.
  \item \textbf{9 -- Highly Safe:} Clear, unambiguous, low-risk behavior; natural interaction patterns consistent with real traffic rules.
\end{itemize}

\end{itemize}

\vspace{1ex}
\noindent\textbf{\textcolor{purple}{Instruction:}}
\begin{itemize}[leftmargin=10pt]
  \item Evaluate the clip \textbf{only} from visible evidence; do not hallucinate or infer information not shown in the video.
  \item Follow the corresponding rubric for the given dimension when mapping visual quality to a score in $[1, 10]$.
  \item The output \textbf{must} be a single valid JSON object with exactly two keys: \texttt{"score"} and \texttt{"reason"}.
  \item Do \textbf{not} include any extra keys, text, Markdown, or comments outside the JSON object.
\end{itemize}

\vspace{1ex}
\begin{tcolorbox}[myjsonbox,title=Please format your results as follows:]
\vspace{-4mm}
\begin{lstlisting}[language=json, basicstyle=\scriptsize\ttfamily, frame=none]
{
  "score": <one of 1.0,1.5,...,10.0>
  "reason": "<Brief, evidence-based explanation based only on visible cues.>"
}
\end{lstlisting}
\vspace{-4mm}
\end{tcolorbox}

\vspace{1ex}
\begin{tcolorbox}[myjsonbox,title=Example output:]
\vspace{-4mm}
\begin{lstlisting}[language=json, basicstyle=\scriptsize\ttfamily, frame=none]
{
  "score": 2.5,
  "reason": "Frequent texture flicker on vehicles and unstable shadows reduce realism, but geometry and traffic behaviors remain mostly plausible."
}
\end{lstlisting}
\vspace{-4mm}
\end{tcolorbox}
\end{tcolorbox}
\vspace{-0.2cm}

\clearpage\clearpage

\clearpage\clearpage

%% file: sections_supp/9_impact.tex
\section{~Broader Impact \& Limitations}

In this section, we elaborate on the broader impact, societal influence, and potential limitations of the proposed approach.

\vspace{0.1cm}
\subsection{~Broader Impact}
Our benchmark advances the evaluation of generative world models by establishing a unified, transparent, and reproducible protocol that links perception, geometry, physics, and behavior. By grounding quantitative scores in human perception and physical reasoning, we encourage the development of models that are not only visually convincing but also physically reliable and functionally safe. 

\vspace{0.1cm}
The benchmark, dataset, and agent together promote standardization and comparability in this rapidly evolving domain, helping researchers diagnose weaknesses, track progress, and design more robust embodied simulators. Beyond autonomous driving, the framework can inspire principled evaluation methods for robotics, AR/VR simulation, and broader world-model research.

\vspace{0.1cm}
\subsection{~Societal Influence}
Our benchmark has implications for AI safety, trustworthy simulation, and embodied intelligence. By providing quantitative and human-aligned metrics for realism, physical plausibility, and behavioral safety, our benchmark helps mitigate risks from models that may appear realistic but behave unrealistically when used for planning or training downstream agents. Reliable evaluation of generative simulators could accelerate applications in safe-driving research, synthetic dataset generation, and policy testing under controlled conditions. 

\vspace{0.1cm}
Nonetheless, the framework should be used responsibly, especially when synthetic data influence safety-critical decisions, ensuring transparency in evaluation and avoiding misuse for deceptive content generation.

\vspace{0.1cm}
\subsection{~Potential Limitations}
While our benchmark provides a comprehensive evaluation spectrum, several limitations remain. First, the benchmark currently focuses on driving-world scenarios; extending to indoor, aerial, or humanoid environments requires additional metrics and domain-specific cues. Second, although the human preference dataset (WorldLens-26K) captures rich perceptual reasoning, it may reflect annotator bias toward specific visual styles or regions, which future work could mitigate through more diverse and cross-cultural labeling. Third, the evaluation agent, though effective in zero-shot settings, inherits limitations from its underlying language model and supervision quality. Lastly, physical realism in simulation is inherently open-ended; new metrics may be required as models evolve toward interactive and multimodal 4D reasoning.

%% file: sections_supp/10_ack.tex
\section{~Public Resource Used}
In this section, we acknowledge the use of the following public resources, during the course of this work:

\vspace{0.1cm}
\begin{itemize}
    \item nuScenes\footnote{\url{https://www.nuscenes.org/nuscenes}.} \dotfill CC BY-NC-SA 4.0
    
    \item nuscenes-devkit\footnote{\url{https://github.com/nutonomy/nuscenes-devkit}.} \dotfill Apache License 2.0

    \item KITTI\footnote{\url{http://www.cvlibs.net/datasets/kitti}.} \dotfill Non-Commercial Use Only (Research Purposes)
    
    \item waymo-open-dataset\footnote{\url{https://github.com/waymo-research/waymo-open-dataset}.} \dotfill Apache License 2.0

    \item MagicDrive\footnote{\url{https://github.com/cure-lab/MagicDrive}.} \dotfill Apache License 2.0

    \item DreamForge\footnote{\url{https://github.com/PJLab-ADG/DriveArena}.} \dotfill Apache License 2.0

    \item DriveDreamer-2\footnote{\url{https://github.com/f1yfisher/DriveDreamer2}.} \dotfill Apache License 2.0

    \item OpenDWM\footnote{\url{https://github.com/SenseTime-FVG/OpenDWM}.} \dotfill MIT License

    \item DiST-4D\footnote{\url{https://github.com/royalmelon0505/dist4d}.} \dotfill None

    \item $\mathcal{X}$-Scene\footnote{\url{https://github.com/yuyang-cloud/X-Scene}.} \dotfill None

    \item Panacea\footnote{\url{https://github.com/wenyuqing/panacea}.} \dotfill Apache License 2.0
    
    \item Limsim\footnote{\url{https://github.com/PJLab-ADG/LimSim/tree/LimSim_plus}.} \dotfill None

    \item DriveStudio\footnote{\url{https://github.com/ziyc/drivestudio}.} \dotfill MIT License

    \item DriveArena\footnote{\url{https://github.com/PJLab-ADG/DriveArena}.} \dotfill None

    \item DrivingSphere\footnote{\url{https://github.com/yanty123/DrivingSphere}.} \dotfill Apache License 2.0

    \item MagicDrive-V2\footnote{\url{https://github.com/flymin/MagicDrive-V2}.} \dotfill AGPL-3.0 license

    \item UniAD\footnote{\url{https://github.com/OpenDriveLab/UniAD}.} \dotfill Apache License 2.0
    
    \item Open3D\footnote{\url{http://www.open3d.org}.} \dotfill MIT License 
    
    \item PyTorch\footnote{\url{https://pytorch.org}.} \dotfill BSD License  
    
    \item ROS Humble\footnote{\url{https://docs.ros.org/en/humble}.} \dotfill Apache License 2.0
    
    \item torchsparse\footnote{\url{https://github.com/mit-han-lab/torchsparse}.} \dotfill MIT License
    
    \item VBench\footnote{\url{https://github.com/Vchitect/VBench}.} \dotfill Apache License 2.0
    
    \item SparseOcc\footnote{\url{https://github.com/MCG-NJU/SparseOcc}.} \dotfill Apache License 2.0
    
    \item DINO\footnote{\url{https://github.com/facebookresearch/dino}.} \dotfill Apache License 2.0

    \item DINOv2\footnote{\url{https://github.com/facebookresearch/dinov2}.} \dotfill Apache License 2.0

    \item MMEngine\footnote{\url{https://github.com/open-mmlab/mmengine}.} \dotfill Apache License 2.0
    
    \item MMCV\footnote{\url{https://github.com/open-mmlab/mmcv}.} \dotfill Apache License 2.0
    
    \item MMDetection\footnote{\url{https://github.com/open-mmlab/mmdetection}.} \dotfill Apache License 2.0
    
    \item MMDetection3D\footnote{\url{https://github.com/open-mmlab/mmdetection3d}.} \dotfill Apache License 2.0
    
    \item OpenPCSeg\footnote{\url{https://github.com/PJLab-ADG/OpenPCSeg}.} \dotfill Apache License 2.0
    
    \item OpenPCDet\footnote{\url{https://github.com/open-mmlab/OpenPCDet}.} \dotfill Apache License 2.0

    \item Qwen3-VL\footnote{\url{https://github.com/QwenLM/Qwen3-VL}.} \dotfill Apache License 2.0

    \item LLaMA-Factory\footnote{\url{https://github.com/hiyouga/LLaMA-Factory}.} \dotfill Apache License 2.0
\end{itemize}